\documentclass[10pt,journal,compsoc]{IEEEtran}

\usepackage{amsmath,amsfonts}
\usepackage{array}
\usepackage[caption=false,font=footnotesize,labelfont=sf,textfont=sf]{subfig}
\usepackage{textcomp}
\usepackage{stfloats}
\usepackage{verbatim}
\usepackage{graphicx}
\usepackage[nocompress]{cite}

\usepackage[colorlinks,
linkcolor=blue,
anchorcolor=orange,
citecolor=orange
]{hyperref}
\usepackage[capitalize,noabbrev]{cleveref}

\crefname{figure}{Fig.}{Figs.}
\Crefname{figure}{Fig.}{Figs.}

\crefname{equation}{Eq.}{Eqs.}
\Crefname{equation}{Eq.}{Eqs.}

\crefname{table}{Tab.}{Tabs.}
\Crefname{table}{Tab.}{Tabs.}

\crefname{section}{Sec.}{Secs.}
\Crefname{section}{Sec.}{Secs.}

\crefname{lemma}{Lem.}{Lems.}
\Crefname{lemma}{Lem.}{Lems.}

\crefname{remark}{Remark}{Rem.}
\Crefname{remark}{Remark}{Rems.}

\crefname{proposition}{Proposition}{Prop.}
\Crefname{proposition}{Proposition}{Props.}

\crefname{definition}{Definition}{Def.}
\Crefname{definition}{Definition}{Defs.}

\crefname{appendix}{App.}{App.}

\usepackage{url}

\usepackage{amsthm}
\theoremstyle{plain}
\newtheorem{thm}{Theorem}
\newtheorem{lem}{Lemma}
\newtheorem{prop}{Proposition}
\theoremstyle{definition}
\newtheorem{rem}{Remark}
\newtheorem{defn}{Definition}
\newtheorem{assmp}{Assumption}

\newtheorem*{rthm1}{\textbf{Restate of Theorem 1}}
\newtheorem*{rthm2}{\textbf{Restate of Theorem 2}}
\newtheorem*{rthm3}{\textbf{Restate of Theorem 3}}
\newtheorem*{rthm4}{\textbf{Restate of Theorem 4}}
\newtheorem*{rthm5}{\textbf{Restate of Theorem 5}}
\newtheorem*{rprop1}{\textbf{Restate of Proposition 1}}

\usepackage{algorithmic}
\usepackage[ruled,vlined]{algorithm2e}

\usepackage{tcolorbox}
\tcbuselibrary{skins,breakable}
\newenvironment{qbox}
{\begin{tcolorbox}[enhanced jigsaw, drop shadow=black!50!white,colback=white, width=0.95\linewidth, center, left=2pt,right=2pt,top=1pt,bottom=1pt]}
{\end{tcolorbox}}

\definecolor{softPurple}{HTML}{7D6B7D}
\definecolor{hazePurple}{HTML}{FBF9FB}
\newtcolorbox{simplebox}{
    enhanced,
    breakable,
    frame hidden,
    boxrule=0pt,
    before skip=4pt,
    after skip=5pt,
    left=5pt,
    right=3pt,
    top=3pt,
    bottom=3pt,
    boxsep=0pt,
    borderline west={1.5pt}{0pt}{softPurple},
    colback=hazePurple,
    arc=1mm,
    fuzzy shadow={1mm}{-1mm}{0.5mm}{0.08mm}{black!35}
}

\newtcolorbox{simpleboxbo}{
    enhanced,
    breakable,
    frame hidden,
    boxrule=0pt,
    before skip=4pt,
    after skip=5pt,
    left=5pt,
    right=3pt,
    top=3pt,
    bottom=3pt,
    boxsep=0pt,
    borderline west={1.5pt}{0pt}{softPurple},
    colback=blue!10,
    arc=1mm,
    fuzzy shadow={1mm}{-1mm}{0.5mm}{0.08mm}{black!35}
}

\usepackage{booktabs}
\usepackage{multirow}
\usepackage{colortbl}
\usepackage{tabularx}

\usepackage{pifont}

\usepackage{titletoc}

\usepackage{xcolor}
\definecolor{DisColor}{HTML}{6D339C}
\definecolor{PerColor}{HTML}{285D9B}

\def \x {\mathbf{x}}
\def \z {\mathbf{z}}

\usepackage{fontawesome5} 

\usepackage{ragged2e}

\usepackage[framemethod=TikZ]{mdframed}
\definecolor{myblue}{RGB}{204, 233, 239}
\definecolor{mygreen}{RGB}{234, 240, 225}

\usepackage{enumitem}
\usepackage{amssymb}

\begin{document}

\title{When AUC Meets Generative Reconstruction: \\A Unified Principle for Balanced \\Visual Representation Learning}

\author{
Boyu~Han,
Qianqian~Xu*,~\IEEEmembership{Senior~Member,~IEEE},
Shilong~Bao,~\IEEEmembership{Member,~IEEE}, \\
Zhiyong~Yang,
and~Qingming~Huang*,~\IEEEmembership{Fellow,~IEEE}%
    \IEEEcompsocitemizethanks{%
        \IEEEcompsocthanksitem Boyu Han is with the State Key Laboratory of AI Safety, Institute of Computing Technology, Chinese Academy of Sciences, Beijing 100190, China, and with the School of Computer Science and Technology, University of Chinese Academy of Sciences, Beijing 101408, China (e-mail: \texttt{hanboyu23z@ict.ac.cn}).
        \IEEEcompsocthanksitem Qianqian Xu is with the State Key Laboratory of AI Safety, Institute of Computing Technology, Chinese Academy of Sciences, Beijing 100190, China, and also with Beijing Academy of Artificial Intelligence, Beijing 100084, China (e-mail: \texttt{xuqianqian@ict.ac.cn}).
        \IEEEcompsocthanksitem Shilong Bao is with the State Key Laboratory of AI Safety, Institute of Computing Technology, Chinese Academy of Sciences, Beijing 100190, China (e-mail: \texttt{baoshilong@ict.ac.cn}).
        \IEEEcompsocthanksitem Zhiyong Yang is with the School of Computer Science and Technology, University of Chinese Academy of Sciences, Beijing 101408, China (e-mail: \texttt{yangzhiyong21@ucas.ac.cn}).
        \IEEEcompsocthanksitem Qingming Huang is with the School of Computer Science and Technology,
        University of Chinese Academy of Sciences, Beijing 101408, China, and also with the State Key Laboratory of AI Safety, Institute of Computing Technology, Chinese Academy of Sciences, Beijing 100190, China (e-mail: \texttt{qmhuang@ucas.ac.cn}).
        \IEEEcompsocthanksitem *~Corresponding authors.
    }%
}

\markboth{IEEE TRANSACTIONS ON PATTERN ANALYSIS AND MACHINE INTELLIGENCE}%
{Shell \MakeLowercase{\textit{et al.}}: A Sample Article Using IEEEtran.cls for IEEE Journals}


\IEEEtitleabstractindextext{%
\begin{abstract}
\justifying
Multimodal large language models (MLLMs) require visual encoders that extract fine-grained visual features while constructing semantically disentangled and discriminative representations. Conventional methodologies encounter a fundamental trade-off: contrastive learning enhances inter-class discrimination by suppressing intra-class variability, whereas generative reconstruction emphasizes perceptual fidelity but lacks explicit comparative supervision. The naive aggregation of these objectives via composite loss functions often induces optimization conflicts and instability due to gradient interference within the shared latent space. To resolve this, we introduce a unified representation-learning paradigm that reformulates generative reconstruction as a discriminative ranking problem within the generative output manifold. Our approach maintains perceptual reconstruction as a supervisory signal and integrates a ranking-based discriminative constraint by systematically ordering matched and mismatched reconstruction outputs. This formulation embeds discriminative information seamlessly into the generative process, alleviating the risk of loss competition. Specifically, we propose AUC Generative Reconstruction (AUCGR), which optimizes the area under the ROC curve (AUC) by ensuring higher scores for matched pairs than for mismatched pairs, and Partial AUC Generative Reconstruction (pAUCGR), which addresses supervision bias by focusing on mismatched comparisons within reliable feature subspaces. The proposed objectives are designed to be plug-and-play, requiring no architectural modifications to existing generative reconstruction frameworks. Theoretical analysis establishes rigorous connections to AUC/pAUC optimization and provides generalization guarantees. Extensive experiments across diverse CLIP architectures, more than 30 visual benchmarks, generative backbones, and MLLMs consistently validate that our method advances both discriminative accuracy and fine-grained perceptual representation, establishing a new benchmark for visual representation learning.
\end{abstract}

}

\maketitle
\IEEEdisplaynontitleabstractindextext
\IEEEpeerreviewmaketitle

\section{Introduction}
\label{sec:introduction}
\IEEEPARstart{I}n recent years, Multimodal Large Language Models (MLLMs)~\cite{liu2024improved,zhu2024minigpt,marafioti2025smolvlm} have undergone rapid advancement, with their performance increasingly governed by the quality of multimodal encoders. As a crucial component, vision encoders map images into compact visual representations that facilitate visual understanding and reasoning in downstream tasks~\cite{yao2024tcp,zhou2022conditional,zhang2024overcoming,han2026understanding,hua2024reconboost,weimoka,wang2026clip}. Most encoders leverage contrastive pretraining, benefiting from strong zero-shot generalization capabilities and effective cross-modal alignment~\cite{radford2021learning}. Nonetheless, the self-supervised text-image alignment achieved by these models remains relatively coarse~\cite{zhaosuperclip}. Consequently, the resulting representations are often suboptimal, posing a major bottleneck for broader deployment~\cite{kim2024extending}.

A key challenge in representation learning is that the goal of a good representation is inherently abstract. The seminal review by \textit{Bengio et al.} suggests that effective representations are linked to their capacity to capture the underlying factors of variation within data~\cite{bengio2013representation}. In addition, \textit{LeCun, Bengio, and Hinton's Nature paper} demonstrates that deep neural networks learn hierarchical representations: lower layers encode local structures while higher layers capture object-level abstractions~\cite{lecun2015deep}. This hierarchy exposes a fundamental tension between semantic abstraction and perceptual preservation. Representations must be invariant enough to enable category-level discrimination, while still retaining local details necessary for fine-grained understanding. This phenomenon is also observed in \textit{Kornblith, Shlens, and Quoc V. Le}'s work, which finds that ImageNet-trained features are less general than previously thought, especially for fine-grained tasks~\cite{kornblith2019better}. Building on these insights, recent foundation models increasingly advocate that a robust visual representation should simultaneously exhibit two complementary properties. The first is \textit{Discriminative Ability (\textbf{\textcolor{DisColor}{D-Ability}})}, which assesses whether the representation can identify the primary category or detect the presence of a target concept. The second is \textit{Fine-grained Perceptual Ability (\textbf{\textcolor{PerColor}{P-Ability}})}, which evaluates whether the representation retains pose, orientation, local attributes, and subtle visual differences.

\begin{figure*}[t]
  \centering
   \includegraphics[width=\linewidth]{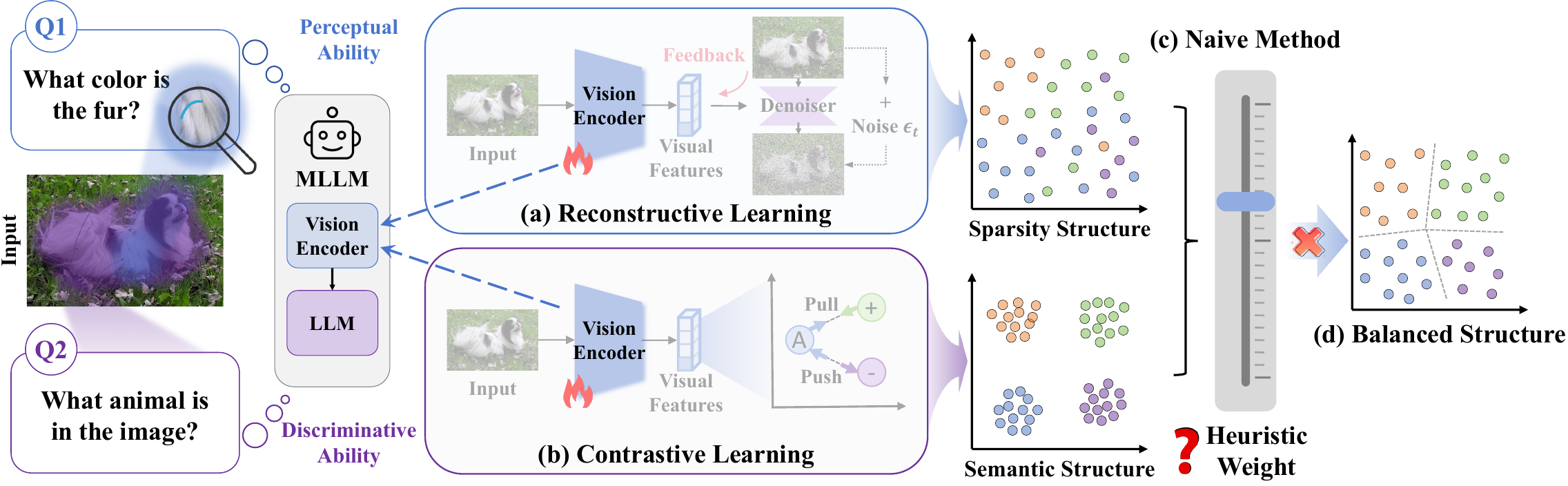}
   \caption{\textbf{Motivation of our method.} (a) Reconstructive learning improves \textbf{\textcolor{PerColor}{P-Ability}} by preserving a sparser representation space. (b) Contrastive learning improves \textbf{\textcolor{DisColor}{D-Ability}} by organizing representations into compact semantic clusters. (c) The naive method searches for a balanced representation through heuristic weight tuning between contrastive learning and reconstructive learning, but the proper weights are difficult to determine. (d) The target balanced representation, which is directly optimized by our method without heuristic weight tuning.}
   \label{fig: intro overview}
\end{figure*}

Existing approaches employ distinct optimization strategies for these two capabilities. (1) Various forms of contrastive learning~\cite{zhou2022learning,wei2023improving,daiexploring,liu2025future,liu2026adaptive} encourage models to suppress local variations and low-level noise, thereby promoting stable semantic representations and strengthening \textbf{\textcolor{DisColor}{D-Ability}}. As depicted in \cref{fig: intro overview}(b), this process tends to cluster learned representations into compact semantic groups. (2) With the emergence of generative models, recent research leverages generative feedback for further enhancement~\cite{wangreconstructive,wangdiffusion,ma2025genhancer,li20252}. In this setting, generative models reconstruct images conditioned on visual features, where accurate reconstruction implies that the visual representation preserves detailed information from the input. This approach defines a reconstruction loss via a mean squared error (MSE) objective, thereby enhancing \textbf{\textcolor{PerColor}{P-Ability}}. As shown in \cref{fig: intro overview}(a), this objective produces a sparser representation space that maintains fine-grained variations across samples.

However, these two approaches typically optimize only one objective at a time, resulting in a trade-off between \textbf{\textcolor{DisColor}{D-Ability}} and \textbf{\textcolor{PerColor}{P-Ability}}. As illustrated in \cref{fig: intro overview}(d), an ideal balanced representation should be sparse while maintaining clear semantic separability. This leads to a natural question: 
\begin{qbox}
    \begin{center}
      \textit{Is there an intuitive optimization strategy that embraces both discriminative and fine-grained perceptual ability?}
    \end{center}
\end{qbox}

In this paper, we answer this question affirmatively. We first motivate our approach by proposing a straightforward solution that linearly combines these two seemingly contradictory objectives within one framework, as depicted in \cref{fig: intro overview}(c). However, both theoretical analysis (\cref{subsubsec: Naive Theoretical Analysis}) and empirical results (\cref{subsubsec: Naive Empirical Analysis}) indicate that this naive solution cannot directly resolve the conflict of the two goals. 

To circumvent the need for explicit balancing, we propose a unified supervision strategy within generative reconstruction. \textbf{The key insight is that reconstruction intrinsically promotes \textcolor{PerColor}{P-Ability}, while comparison-based supervision fundamentally underpins the strong \textcolor{DisColor}{D-Ability} observed in contrastive learning.} Rather than introducing an additional contrastive branch, we seamlessly embed comparison signals into the reconstruction process. Specifically, for each image, we sample a generative timestep alongside its corresponding noisy observation, and compare different one-step denoising results under varying conditioning. (1) When the condition is derived from the \textbf{same image}, such as an augmented view, the reconstruction should remain highly consistent.  These results are treated as \textbf{positive examples}. (2) When the condition originates from \textbf{a different image}, the reconstruction should diverge. These results serve as \textbf{negative examples}. By integrating comparisons directly into the generative process, we avoid directly optimizing two potentially conflicting objectives.

We instantiate this comparison structure using the area under the ROC curve (AUC), as depicted in \cref{fig: overview}(b). Based on the definition of positive/negative examples shown above, optimizing the AUC objective encourages the score of a randomly chosen positive to exceed that of a randomly chosen negative~\cite{hanley1982meaning}. This formulation precisely enforces the desired ordering within the generated image space at an arbitrary scale (since the comparison timestep is randomly chosen each time). On top of this, we introduce \textit{AUC Generative Reconstruction (AUCGR)} as a tractable training objective. Theoretical analysis further demonstrates that this AUC-driven comparison structure supports both \textbf{\textcolor{DisColor}{D-Ability}} and \textbf{\textcolor{PerColor}{P-Ability}}.

AUCGR still encounters a practical challenge in mini-batch training: the pseudo-negative set generated above may include \textbf{noisy} negatives that are semantically similar to the anchor, especially in long-tailed data distributions. Blindly telling apart those noisy examples from positives could introduce additional bias into supervision. To address this, we propose \textit{Partial AUC Generative Reconstruction (pAUCGR)}, yielding a partial-AUC interpretation that preserves uniform generalization guarantees. 

Finally, we conduct comprehensive experiments across diverse benchmark datasets and multimodal large language models to verify the effectiveness of our proposed method.

This work extends our CVPR 2026 paper~\cite{han2026guiding}, where we initially proposed a unified contrastive reconstruction objective to resolve the gradient conflict between contrastive learning and generative reconstruction. In this version, we incorporate systematic enhancements in experimental settings, methodology, and empirical evaluation, resulting in a more general and robust generative reconstruction framework for visual representation learning. In summary, the new contributions are as follows:
\begin{itemize}[leftmargin=*]
    \item \textbf{New Methods.} This version extends the contrastive reconstruction objective into a generic comparison-based theoretical framework (see \cref{sec: Learning with Comparisons in Generative Reconstruction}). As a fundamental principle, it can be instantiated by different comparison strategies. Moreover, it serves as a plug-and-play optimization module that enhances existing generative reconstruction methods and produces more balanced visual representations. We develop an AUC-oriented instantiation of the comparison strategy in the unified paradigm, termed \textit{AUC Generative Reconstruction (AUCGR)}. To mitigate biased supervision arising from noisy negatives, we further introduce \textit{Partial AUC Generative Reconstruction (pAUCGR)}, which restricts negative comparisons to reliable regions.
    \item \textbf{New Theory.} We further provide theoretical analyses for the effectiveness of AUCGR and pAUCGR, establish corresponding generalization bounds (see \cref{sec: Generalization Bound}), and show that the conference version is naturally covered as the logistic-surrogate case of AUCGR.
    \item \textbf{New Empirical Results.} We conduct a broad range of new empirical studies to evaluate the effectiveness of our method. Specifically, we introduce more generative reconstruction backbones, extend the evaluation to over $30$ benchmarks (including zero-shot clustering, zero-shot classification, zero-shot retrieval, fine-grained perception, and MLLM evaluation), and perform more fine-grained ablation studies and visualization analyses.
    \item \textbf{Miscellaneous Contents.} We also improve some existing contents to make our work more complete, including the abstract, introduction, review of prior arts (\cref{sec: relatedwork} and \cref{sec: preliminaries}), methodology and theory (\cref{sec: A Naive Joint Optimization Method}, \cref{sec: Learning with Comparisons in Generative Reconstruction} and \cref{sec: Generalization Bound}), and experiments (\cref{sec: experiments}).
\end{itemize}

\section{Related Work}
\label{sec: relatedwork}

\subsection{Vision Encoders}
Vision encoders serve as the visual backbone of MLLMs~\cite{liu2024improved,liu2023visual}, transforming raw images into high-level feature representations. Among various architectures~\cite{radford2021learning,caron2021emerging,oquab2024dinov2,simeoni2025dinov3,luthra2026self,mehta2026breaking}, CLIP~\cite{radford2021learning} is one of the most widely adopted models. It also provides transferable representations for a wide range of tasks, including detection~\cite{subramanian2022reclip,mu2022slip,yang2024harnessing,lin2025semantically,zhong2026modality}, segmentation~\cite{zhou2022extract,xu2022simple}, video understanding~\cite{xu2021videoclip,tang2021clip4caption}, and generation~\cite{rombach2022high,nichol2022glide,han2025lightfair,li2025hybrid,lione,gao2025devil,gao2025rapo++}. Reliability-aware visual learning has also received increasing attention. SELongVLM~\cite{zhang2026selongvlm} effectively suppresses task-irrelevant visual content through self-corrective clip selection, while \cite{yi2026distilling} explicitly model knowledge uncertainty via Monte Carlo dropout to reduce unreliable supervision in object detection distillation. Overall, recent studies have focused on enhancing vision encoders' understanding capability to improve downstream performance. Some primarily strengthen its discriminative ability~\cite{zhou2022learning,wei2023improving,gong2025boosting}, while others improve its fine-grained perceptual ability~\cite{tong2024eyes,wangreconstructive,wangdiffusion,ma2025genhancer,li20252}. However, an ideal visual encoder should achieve a balance between both aspects. To bridge this gap, this paper aims to develop an enhanced vision encoder with balanced understanding abilities, providing a stronger foundation for MLLMs.

\subsection{Generative Reconstruction Learning}
Generative modeling has recently become a powerful paradigm for improving visual representations. Early studies~\cite{shipard2023diversity,li2024semantic,tian2023stablerep} use generative models as data augmenters to expand training data and boost generalization. More recent works~\cite{wangreconstructive,wangdiffusion,ma2025genhancer,li20252} add self-supervised reconstruction objectives to diffusion frameworks, enabling models to learn richer spatial details and fine-grained semantics. Among them, DIVA~\cite{wangdiffusion} introduces diffusion-based visual feedback conditioned on CLIP embeddings, which refines internal representations through generative reconstruction. GenHancer~\cite{ma2025genhancer} extends this reconstruction process to discrete latent spaces, while un$^2$CLIP~\cite{li20252} resolves the mismatch between CLIP's pretrained feature space and generative model latents. However, these methods often overemphasize fine-grained reconstruction while neglecting discriminative learning, leading to suboptimal representations. In contrast, our work simultaneously strengthens discriminative ability and fine-grained perceptual ability, producing a more effective vision representation.

\subsection{Vision-understanding Benchmarks}
The benchmarks can be broadly divided into two camps. The first focuses on assessing discriminative ability. This ability can be evaluated through zero-shot clustering, zero-shot classification, and zero-shot retrieval. Zero-shot clustering examines whether image representations form separable semantic structures on datasets such as ImageNet~\cite{deng2009imagenet}, Caltech-101~\cite{fei2004learning}, and DTD~\cite{cimpoi2014describing}. Zero-shot classification~\cite{radford2021learning} further tests whether the learned visual features support class-level recognition. Zero-shot retrieval~\cite{young2014image,lin2014microsoft} evaluates whether visual representations can distinguish correct image-text pairs from many candidates. The second group emphasizes fine-grained perceptual ability. MMVP~\cite{tong2024eyes} introduces a challenging benchmark with nine visual patterns, CV-Bench~\cite{tong2024cambrian} extends evaluation to spatial relations, counting, and 3D understanding, and NaturalBench~\cite{li2024naturalbench} adds natural adversarial examples to assess perceptual robustness. Together, these benchmarks provide a comprehensive evaluation of visual representations from both discriminative and fine-grained perceptual perspectives.

\subsection{AUC Optimization}
AUC optimization develops from classical machine learning to modern deep learning. Early work clarifies the distinct motivation for this line of research by showing that maximizing AUC is generally inconsistent with minimizing classification error~\cite{cortes2003auc}. This observation motivates a series of optimization methods for linear models, including logistic regression~\cite{herschtal2004optimising} and support vector machines~\cite{joachims2005support, joachims2006training}. Subsequent studies extend AUC maximization to online and stochastic settings~\cite{zhao2011online,gao2013one,ying2016stochastic,natole2018stochastic}, while generalization and consistency analyses~\cite{agarwal2005generalization, usunier2005data, clemenccon2008ranking,agarwal2014surrogate, gao2012consistency} provide the corresponding theoretical foundations. With the rise of deep models, AUC-oriented training further extends to deep neural networks~\cite{liu2020stochastic}, and a range of variants such as partial AUC, compositional AUC, and long-tailed or class-balanced extensions gradually enriches this line of research~\cite{yang2021all,yang2022auc,yuan2023libauc}. More recently, AUC optimization also extends beyond instance-level prediction to the pixel level~\cite{han2024aucseg,bao2025towards,han2025dual}. In practice, AUC-based optimization demonstrates effectiveness in recommendation systems~\cite{bao2022rethinking,bao2022minority}, disease prediction~\cite{westcott2019chronic,gola2020polygenic}, dense-level segmentation~\cite{han2024aucseg,bao2025towards}, and adversarial training~\cite{hou2022adauc,xiasample,han2016convergence}. In this paper, we introduce AUC optimization into generative reconstruction for visual representation enhancement.

\section{Preliminaries}
\label{sec: preliminaries}
\subsection{Notation}
\label{subsec: Notation}
Let $\epsilon_{\theta}$ denote the generative model, $f_{\phi}$ denote the visual encoder, and $h_{\omega}$ denote the projection module that maps visual features into the generative space. The training dataset is defined as $\mathcal{D}=\{\x^i\in\mathbb{R}^{H\times W\times 3}\}_{i=1}^{n}$, where $H$ and $W$ denote the image height and width. A \textit{variational autoencoder (VAE)} encodes each image $\x$ into a latent $\tilde{\x}$. We denote the visual feature as $\z = f_{\phi}(\x)$, the condition as $c = h_{\omega}(\z)$, and the model prediction as $\hat{\boldsymbol\epsilon}_{\theta} = \epsilon_{\theta}(\tilde{\x}_t, c, t)$ at timestep $t$, with the ground-truth $\boldsymbol\epsilon_t^{\text{gt}}$. For clarity, we include a table of symbol definitions in \cref{sec: supp_Symbol Definitions}.

\subsection{Generative Models}
\label{subsec: Generative Models}
We review two typical generative models: diffusion models and flow-based models.

\noindent \textbf{Diffusion Models.} Diffusion models~\cite{ho2020denoising,rombach2022high} learn a generative process by reversing a gradual noising procedure over a finite number of timesteps. Given a latent representation $\tilde{\x}_0$ encoded by a VAE, the model progressively adds Gaussian noise to obtain $\tilde{\x}_t$ at timestep $t$. A noise prediction network $\epsilon_{\theta}(\tilde{\x}_t, c, t)$ parameterizes the reverse process and predicts the ground-truth noise $\boldsymbol\epsilon_t^{\mathrm{gt}}$ added at timestep $t$. Training diffusion models amounts to minimizing the following denoising objective:
\begin{equation}
\mathcal{L}_{\mathrm{DM}}=\mathbb{E}_{\tilde{\x}_0, \boldsymbol\epsilon_t^{\mathrm{gt}}, t}\Big[\big\|\boldsymbol\epsilon_t^{\mathrm{gt}} - \hat{\boldsymbol\epsilon}_{\theta}\big\|_2^2\Big],
\end{equation}
where $t$ is sampled from a predefined timestep distribution. This objective promotes accurate noise prediction at each diffusion step and implicitly defines the reverse generative dynamics. The model achieves conditional generation by injecting the condition $c$ into the noise prediction network.

\noindent \textbf{Flow-based Models.} Flow-based models~\cite{lipmanflow,liuflow} learn a continuous transformation that maps samples from a base distribution to the target data distribution through deterministic dynamics. In the conditional setting, this transformation is often written as an ordinary differential equation:
\begin{equation}
\frac{d \tilde{\x}_t}{d t} = \epsilon_{\theta}(\tilde{\x}_t, c, t),
\end{equation}
where $\epsilon_{\theta}(\tilde{\x}_t, c, t)$ denotes a time-dependent conditional velocity function. Rectified Flow uses a simpler formulation by constructing linear interpolations between samples from the base distribution and the data distribution. Specifically, let $\tilde{\x}_0 \sim \mathcal{N}(0, \mathbf{I})$ and $\tilde{\x}_1 \sim p_{\mathrm{data}}$, and define $\tilde{\x}_t = t \tilde{\x}_1 + (1 - t) \tilde{\x}_0$. The conditional velocity field is learned by minimizing the regression objective:
\begin{equation}
\mathcal{L}_{\mathrm{FM}}=\mathbb{E}_{t, \tilde{\x}_0, \tilde{\x}_1}\Big[\big\|\boldsymbol\epsilon_t^{\text{gt}} - \hat{\boldsymbol\epsilon}_{\theta}\big\|_2^2\Big],
\end{equation}
where $\boldsymbol\epsilon_t^{\text{gt}} = \tilde{\x}_1 - \tilde{\x}_0$. It directly supervises the instantaneous velocity along straight-line paths conditioned on $c$, defining a deterministic conditional generative process without requiring likelihood estimation or numerical solvers.

Although diffusion and flow-based models define different generative processes, their training objectives share the same basic form. Both supervise the model prediction to match a target signal. They provide a reliable supervision signal for generative reconstruction.

\subsection{AUC Optimization}
\label{subsec: AUC Optimization}
\textit{Area under the Receiver Operating Characteristic Curve (AUC)} is a ranking metric widely used in binary classification~\cite{hanley1982meaning}. It measures the probability that a scoring function assigns a higher score to a positive sample than to a negative sample, namely
\begin{equation}
\mathrm{AUC}(h_{\theta})=\mathbb{P}\big(h_{\theta}(\x^{+}) > h_{\theta}(\x^{-})\mid y^{+}=1,\ y^{-}=0\big),
\end{equation}
where $(\x^{+}, y^{+})$ and $(\x^{-}, y^{-})$ denote positive and negative samples, respectively. A larger AUC indicates a stronger ability to rank positives ahead of negatives.

For a finite dataset, AUC maximization is commonly implemented through an empirical pairwise surrogate objective~\cite{yang2021all,yang2021learning,yang2022optimizing}. Let $\{\x_i^{+}\}_{i=1}^{n^{+}}$ and $\{\x_j^{-}\}_{j=1}^{n^{-}}$ denote the positive and negative sets. The corresponding empirical objective can be written as
\begin{equation}
\widehat{\mathcal{L}}_{\mathrm{AUC}}(h_{\theta})
=
\frac{1}{n^{+}n^{-}}\sum_{i=1}^{n^{+}}\sum_{j=1}^{n^{-}}
\ell\big(h_{\theta}(\x_i^{+}) - h_{\theta}(\x_j^{-})\big),
\label{eq: auc_optimization}
\end{equation}
where $\ell$ denotes a differentiable surrogate loss~\cite{yang2021learning}. Minimizing this objective reduces pairwise ordering violations and therefore improves the induced AUC.

This formulation shows that AUC optimization depends on the relative order between positive and negative scores rather than their absolute calibration. Such a property is particularly valuable under imbalanced or long-tailed data distributions~\cite{fawcett2006introduction, wu2022fast,huang2024adaptive,yuan2021compositional}, where comparison-based objectives can alleviate the bias caused by skewed class frequencies. This perspective provides the foundation for our later AUC instantiation of comparison-based generative reconstruction.

\subsection{The Task of Visual Representation Learning}
\label{subsec:visual_representation_learning}
Visual representation learning aims to learn an encoder $f_{\phi}:\mathcal{X}\!\to\!\mathcal{Z}$ that maps an image $\x$ into a visual representation $\z$. The goal is to obtain a representation with rich understanding capacity, typically characterized by two key properties: strong \textbf{Discriminative Ability (\textcolor{DisColor}{D-Ability})} and \textbf{Fine-grained Perceptual Ability (\textcolor{PerColor}{P-Ability})}~\cite{kim2021hybrid,nozawa2022evaluation}.

\textbf{\textcolor{DisColor}{Property 1 (D-Ability):}} Discriminative ability describes a model's ability to accurately and robustly distinguish between different visual categories. A representation with strong discriminative power can effectively separate distinct classes in the feature space, keeping samples from the same class tightly clustered while pushing samples from different classes farther apart~\cite{li2014fisher,radford2021learning}. Mathematically, for a mini-batch with class set $\mathcal Y$ and feature set $\mathcal Z_y = \{\z_i | y_i = y\}$, we define the class center as $\mu_y = \frac{1}{|\mathcal Z_y|} \sum_{\z \in \mathcal Z_y} \z$. It is formulated as
\begin{equation}
\min S_{\text{inner}},\quad \max S_{\text{inter}},
\label{eq: minmax_s}
\end{equation}
\begin{equation}
S_{\text{inner}}=\frac{1}{|\mathcal Y|}\sum_{y\in\mathcal Y}\frac{1}{|\mathcal Z_y|}\sum_{\z\in\mathcal Z_y}|\z-\mu_y|_2^2,
\label{eq: s_inner}
\end{equation}
\begin{equation}
S_{\text{inter}}=\frac{1}{|\mathcal Y|(|\mathcal Y|-1)}\sum_{y,y'\in\mathcal Y,y\neq y'}|\mu_y-\mu_{y'}|_2^2.
\label{eq: s_inter}
\end{equation}

\textbf{\textcolor{PerColor}{Property 2 (P-Ability):}} Fine-grained perceptual ability reflects the model's capacity to capture fine-grained visual details, enabling a comprehensive understanding of object attributes, spatial relations, and subtle semantic cues within an image. Following previous studies~\cite{wangreconstructive,wangdiffusion,ma2025genhancer,li20252}, this perception is usually evaluated through reconstruction consistency. When a generative model can accurately reconstruct the original image from its learned representation, the representation is considered sufficiently rich. It is formulated as
\begin{equation}
\min \mathbb{E}_{t}\Big[\|\epsilon_{\theta}\big(\tilde{\x}_t,\ h_{\omega}(f_{\phi}(\x)),\ t\big)-\boldsymbol\epsilon_t^{\text{gt}}\|_2^2\Big],
\label{eq: min_rec}
\end{equation}
where $t$ denotes the timestep, $f_{\phi}$ denotes the CLIP visual encoder, and $h_{\omega}$ denotes the projector.

\section{A Naive Joint Optimization Method}
\label{sec: A Naive Joint Optimization Method}
In this section, we propose a naive joint optimization method that aims to pursue the two properties in \cref{subsec:visual_representation_learning}. We focus on self-supervised training, since dense supervision is expensive and does not scale well to broad data sources~\cite{chen2020simple,he2022masked,liu2021self,xia2022moderate,luo2024revive}. We consider a classical and widely used visual representation framework, namely CLIP~\cite{radford2021learning,xudemystifying,zhai2023sigmoid}, as the foundation of our analysis.

\subsection{Linear Combination of Two Objectives}
\label{subsec: Linear Combination of Two Objectives}

We start from two widely adopted self-supervised signals that naturally target the two properties.

\vskip 1 ex
\textcolor{DisColor}{\textbf{D-Ability}}: As shown in \cref{fig: intro overview}(b), mainstream approaches commonly use a contrastive loss that pulls same-class features closer and pushes different-class features apart~\cite{radford2021learning}. Specifically, for a mini-batch $\mathcal{B}$ with cosine similarity $\mathrm{sim}(\z_i,\z_j)=\frac{\z_i^\top \z_j}{\|\z_i\|\,\|\z_j\|}$, temperature $\tau>0$, and the positive set $\mathcal{P}(i)=\{j\in\mathcal{B}\setminus\{i\}\mid y_j=y_i\}$, a InfoNCE-style loss~\cite{oord2018representation} is
\begin{equation}
\mathcal{L}_{\text{con}}=\frac{1}{|\mathcal{B}|}\sum_{i\in\mathcal{B}}-\log\frac{\sum_{j\in\mathcal{P}(i)} \exp\big(\mathrm{sim}(\z_i,\z_j)/\tau\big)}
{\sum_{k\in\mathcal{B}\setminus\{i\}} \exp\big(\mathrm{sim}(\z_i,\z_k)/\tau\big)},
\end{equation}
which decreases $S_{\text{inner}}$ and increases $S_{\text{inter}}$ in \cref{eq: minmax_s}. Here, the class label $y_i$ is introduced only to formalize the class-level discriminative objective and is not used in the actual self-supervised training procedure. In practice, $\mathcal{P}(i)$ is constructed using different augmented views of the same image, while the other samples in the mini-batch are treated as negatives. Therefore, no category annotations are required.

\vskip 1 ex
\textbf{\textcolor{PerColor}{P-Ability}:} As shown in \cref{fig: intro overview}(a), the objective is typically realized by enforcing consistency between the reconstructed image and the original one~\cite{wangdiffusion}, namely
\begin{equation}
\mathcal{L}_{\text{rec}}
=\mathbb{E}_{t}\Big[\big\|\hat{\boldsymbol\epsilon}_{\theta}-\boldsymbol\epsilon_t^{\text{gt}}\big\|_2^2\Big],
\end{equation}
which is exactly \cref{eq: min_rec}.

To obtain richer visual representations, we consider a \textbf{naive method} that combines the two losses described above:
\begin{equation}
\mathcal{L}_{\text{joint}} = \lambda_{\text{con}}\mathcal{L}_{\text{con}} + \lambda_{\text{rec}}\mathcal{L}_{\text{rec}},
\label{eq: joint_loss}
\end{equation}
where $\lambda_{\text{con}}$ and $\lambda_{\text{rec}}$ are hyperparameters that balance the two objectives.

\subsection{Limitation of the Naive Method}
\label{subsec: Limitation of the Naive Method}
\cref{eq: joint_loss} provides a concise way to perform joint optimization. However, directly combining the two objectives does not necessarily produce balanced representations. In practice, the two losses may optimize the shared representation in different directions, leading to interference and suboptimal trade-offs~\cite{sener2018multi,lin2019pareto}. We first analyze this issue theoretically, and then verify it empirically.

\subsubsection{Theoretical Analysis}
\label{subsubsec: Naive Theoretical Analysis}
To analyze the interaction between the two objectives, we first decompose the information contained in the representation. A visual representation includes cues that mainly support semantic discrimination, that mainly support fine-grained perception, and a noise term entangled with both objectives. We model these sources as three independent and orthogonal unit components. Since $Z\in\mathbb{S}^{d-1}$, we have $|Z|_2^2=1$. Under the orthogonal decomposition, the squared coefficient of each component corresponds to its contribution to the total representation energy. Therefore, we interpret the squared coefficients as the energy proportions of different information sources. Based on this observation, we define the following signal decomposition model.

\begin{defn}[Signal Decomposition Model]
Let the feature vector $Z\in\mathbb{S}^{d-1}$ be decomposed into three independent and orthogonal unit components:
\begin{equation}
Z
=
\sqrt{\alpha_D}\,z_D
+\sqrt{\alpha_P}\,z_P
+\sqrt{\alpha_N}\,z_N.
\label{eq: signal_decomposition}
\end{equation}
where $\alpha_D,\alpha_P,\alpha_N\ge0$ and $\alpha_D+\alpha_P+\alpha_N=1$.
\label{def: signal_decomposition}
\end{defn}

Here, $z_D$ denotes the component related to \textbf{\textcolor{DisColor}{D-Ability}}, which contains category-separable and semantic-discriminative information. $z_P$ denotes the component related to \textbf{\textcolor{PerColor}{P-Ability}}, which contains fine-grained perceptual and reconstruction-related information. $z_N$ denotes the noise component that is orthogonal to both signals, which contains task-irrelevant perturbations, sampling noise, or ineffective variations. The coefficients $\alpha_D$, $\alpha_P$, and $\alpha_N$ denote their energy proportions in the representation.

Based on Definition \ref{def: signal_decomposition}, we show that the two objectives favor incompatible allocation regimes.
\begin{simplebox}
\begin{thm}[Objective Conflict in the Naive Method]
Under Assumptions~\ref{ass: naive_orthogonality} and~\ref{ass: naive_recoverability}, consider the representation model in Definition~\ref{def: signal_decomposition}.
There exist constants $c_{\mathrm{con}},c_{\mathrm{rec}}\in\mathbb{R}$ and $\kappa_P,\kappa_N,\kappa_D,\kappa_N'>0$.
With these constants, the achievable envelopes of the two objectives satisfy
\begin{equation}
\begin{aligned}
&\inf \mathcal{L}_{\mathrm{con}}
\ge
c_{\mathrm{con}}+\kappa_P\alpha_P+\kappa_N\alpha_N,\\
&\inf \mathcal{L}_{\mathrm{rec}}
\ge
c_{\mathrm{rec}}+\kappa_D(1-\alpha_P)+\kappa_N'\alpha_N,
\label{eq: lower_envelopes}
\end{aligned}
\end{equation}
where the infimum is taken over all admissible model parameters, optimizing each objective with the shared representation $Z$.
This result implies that the two objectives favor different allocation regimes:
\begin{itemize}
    \item Minimizing $\mathcal{L}_{\mathrm{con}}$ encourages \textbf{smaller} $\alpha_P$.
    \item Minimizing $\mathcal{L}_{\mathrm{rec}}$ encourages \textbf{larger} $\alpha_P$.
\end{itemize}
\textbf{NOTE:} Recall $\alpha_D+\alpha_P+\alpha_N=1$, one only needs to include two of the three parameters in the inequality. 
\label{thm: naive_conflict}
\end{thm}
\end{simplebox}
The proof is provided in \cref{sec: supp_Proof of Theorem 1}. The conflicting inequality in the theorem above suggests that one cannot simultaneously reduce  $\mathcal{L}_{\mathrm{con}}$ and 
$\mathcal{L}_{\mathrm{rec}}$. Therefore, jointly optimizing $\mathcal{L}_{\text{joint}}=\lambda_{\text{con}}\mathcal{L}_{\text{con}} + \lambda_{\text{rec}}\mathcal{L}_{\text{rec}}$ inevitably leads to gradient competition on the shared representation.

\subsubsection{Empirical Analysis}
\label{subsubsec: Naive Empirical Analysis}

Following the \textbf{Tragic Triad} analysis in multi-task learning~\cite{yu2020gradient}, optimization tends to fail to converge properly when three conditions co-occur: \textit{opposite gradient directions}, \textit{low gradient magnitude similarity}, and \textit{high multi-task curvature}. Therefore, we conduct three experiments to verify whether the naive method indeed falls into this regime.

For all experiments, let $\z$ denote the shared visual representation. At training step $t$, the update rule for $\z$ is:
\begin{equation}
\z^{t+1} = \z^{t} - \eta \nabla_{\z}\mathcal{L}_{\text{joint}}
= \z^{t} - \eta \left(
\lambda_{\text{con}}\nabla_{\z}\mathcal{L}_{\text{con}} +
\lambda_{\text{rec}}\nabla_{\z}\mathcal{L}_{\text{rec}}
\right),
\label{eq: joint_update}
\end{equation}
where $\eta$ is the learning rate. We compute the per-step gradients of the two objectives:
\begin{equation}
\mathbf{g}_{\text{con}}=\nabla_{\z}\mathcal{L}_{\text{con}}, \qquad
\mathbf{g}_{\text{rec}}=\nabla_{\z}\mathcal{L}_{\text{rec}},
\end{equation}
Experiments are conducted on OpenAI CLIP ViT-L/14@224~\cite{radford2021learning}.

\vskip 1 ex
\noindent \textbf{Experiment 1: Gradient direction.} We first examine whether the two objectives induce inconsistent descent directions. Specifically, to quantify this interaction, we define the cosine similarity between the two gradient directions:
\begin{equation}
\cos(\mathbf{g}_{\text{con}}, \mathbf{g}_{\text{rec}}) =
\frac{\mathbf{g}_{\text{con}}^\top \mathbf{g}_{\text{rec}}}
{\|\mathbf{g}_{\text{con}}\|_2 \|\mathbf{g}_{\text{rec}}\|_2}.
\label{eq: grad_cos}
\end{equation}
When $\cos(\mathbf{g}_{\text{con}}, \mathbf{g}_{\text{rec}}) > 0$, the two objectives cooperate and reinforce each other. However, when $\cos(\mathbf{g}_{\text{con}}, \mathbf{g}_{\text{rec}}) < 0$, they conflict, leading to destructive interference in the gradient updates. 

\begin{figure}[t]
\centering
\subfloat[Cosine similarity between gradient $\mathbf{g}_{\text{con}}$ and $\mathbf{g}_{\text{rec}}$.]{   
\begin{minipage}{\linewidth}
\includegraphics[width=\linewidth]{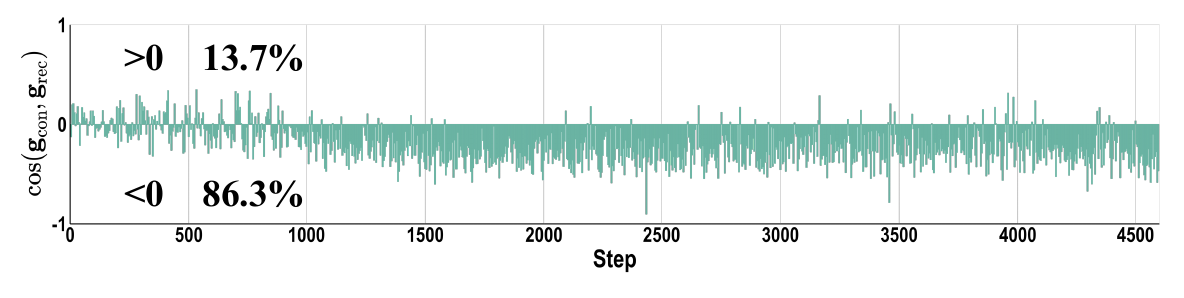}  
\vspace{-20pt}
\label{fig: cos_rec_con}
\end{minipage}
}
\hfil
\subfloat[Contrastive loss $\mathcal{L}_{\text{con}}$.]{
\begin{minipage}{0.47\linewidth}
\includegraphics[width=\linewidth]{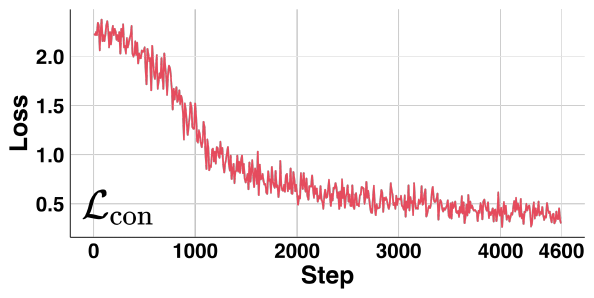} 
\vspace{-10pt}
\label{fig: loss_con}
\end{minipage}
}
\hfil
\subfloat[Reconstruction loss $\mathcal{L}_{\text{rec}}$.]{   
\begin{minipage}{0.47\linewidth}
\includegraphics[width=\linewidth]{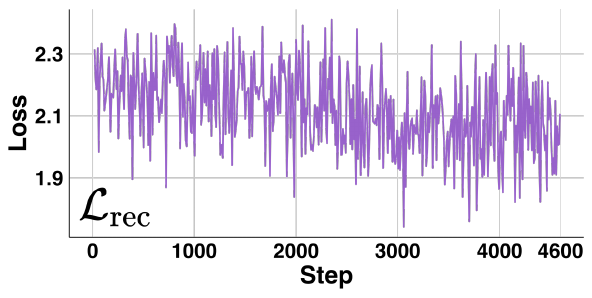}  
\vspace{-10pt}
\label{fig: loss_rec}
\end{minipage}
}
\caption{Training results of linearly combining contrastive learning and reconstructive learning (Naive Method) on the OpenAI CLIP ViT-L@224 backbone, which verify that the naive method indeed falls into the Tragic Triad regime.}
\label{fig: Gradient Conflict}
\end{figure}

As shown in \cref{fig: Gradient Conflict}(a), \underline{\textbf{86.3\%} of the training steps} exhibit a negative cosine similarity $\cos(\mathbf{g}_{\text{con}}, \mathbf{g}_{\text{rec}})$ between the gradients of the two losses, indicating that \textbf{gradient directional conflict is pervasive throughout the optimization process}. Moreover, as training progresses, this value becomes almost entirely negative, suggesting that gradient directional conflicts further intensify in the later stages of optimization.

\vskip 1 ex
\noindent \textbf{Experiment 2: Gradient magnitude similarity.} We next examine whether the optimization is dominated by one task gradient. To quantify this effect, we measure the gradient magnitude similarity
\begin{equation}
\Phi(\mathbf{g}_{\text{con}},\mathbf{g}_{\text{rec}})
=
\frac{2\|\mathbf{g}_{\text{con}}\|_2\|\mathbf{g}_{\text{rec}}\|_2}
{\|\mathbf{g}_{\text{con}}\|_2^2+\|\mathbf{g}_{\text{rec}}\|_2^2},
\end{equation}
where $\Phi\in(0,1]$, and larger values indicate more balanced gradients. When $\Phi$ is small, the joint update is dominated by one objective, while the other objective contributes little to the actual descent direction.

We find that the naive method exhibits \underline{low gradient magnitude similarity in \textbf{51.7\%}} of training steps. It means that even when the two objectives are optimized together, their contributions are highly imbalanced for a large portion of training. As a result, the stronger gradient can override the weaker one, causing the shared encoder to drift toward one objective while sacrificing the other. As shown in \cref{fig: Gradient Conflict}(b)(c), \textbf{$\mathcal{L}_{\text{con}}$ dominates the optimization process, while $\mathcal{L}_{\text{rec}}$ is suppressed and fails to converge.}

\vskip 1 ex
\noindent \textbf{Experiment 3: Multi-task curvature.} We finally examine whether the joint update is taken in a locally sharp region. Following~\cite{yu2020gradient}, we measure the multi-task curvature along the update direction of $\mathcal{L}_{\text{joint}}$. A large curvature implies that even a small step along the joint gradient can deviate substantially from the first-order approximation. In that case, the apparent improvement on the dominant objective may be overestimated, while the degradation on the suppressed objective may be underestimated.

We observe that the naive method reaches a multi-task curvature on the order of \underline{\textbf{$10^6$}} during training, indicating an extremely sharp landscape around the joint descent direction. This result shows that the naive method \textbf{not only suffers from directional conflict and gradient imbalance, but also operates in a regime where these two effects are strongly amplified by curvature}. Consequently, once the update direction is biased toward one objective and misaligned with the other, the optimization error is rapidly magnified, leading to unstable convergence. 

Taken together, Experiments 1--3 verify that the naive method indeed falls into the \textbf{Tragic Triad} regime of multi-task optimization, further demonstrating the existence of severe gradient conflicts between the two objectives.

\section{Learning with Comparisons in Generative Reconstruction}
\label{sec: Learning with Comparisons in Generative Reconstruction}

\begin{figure*}[t]
  \centering
   \includegraphics[width=\linewidth]{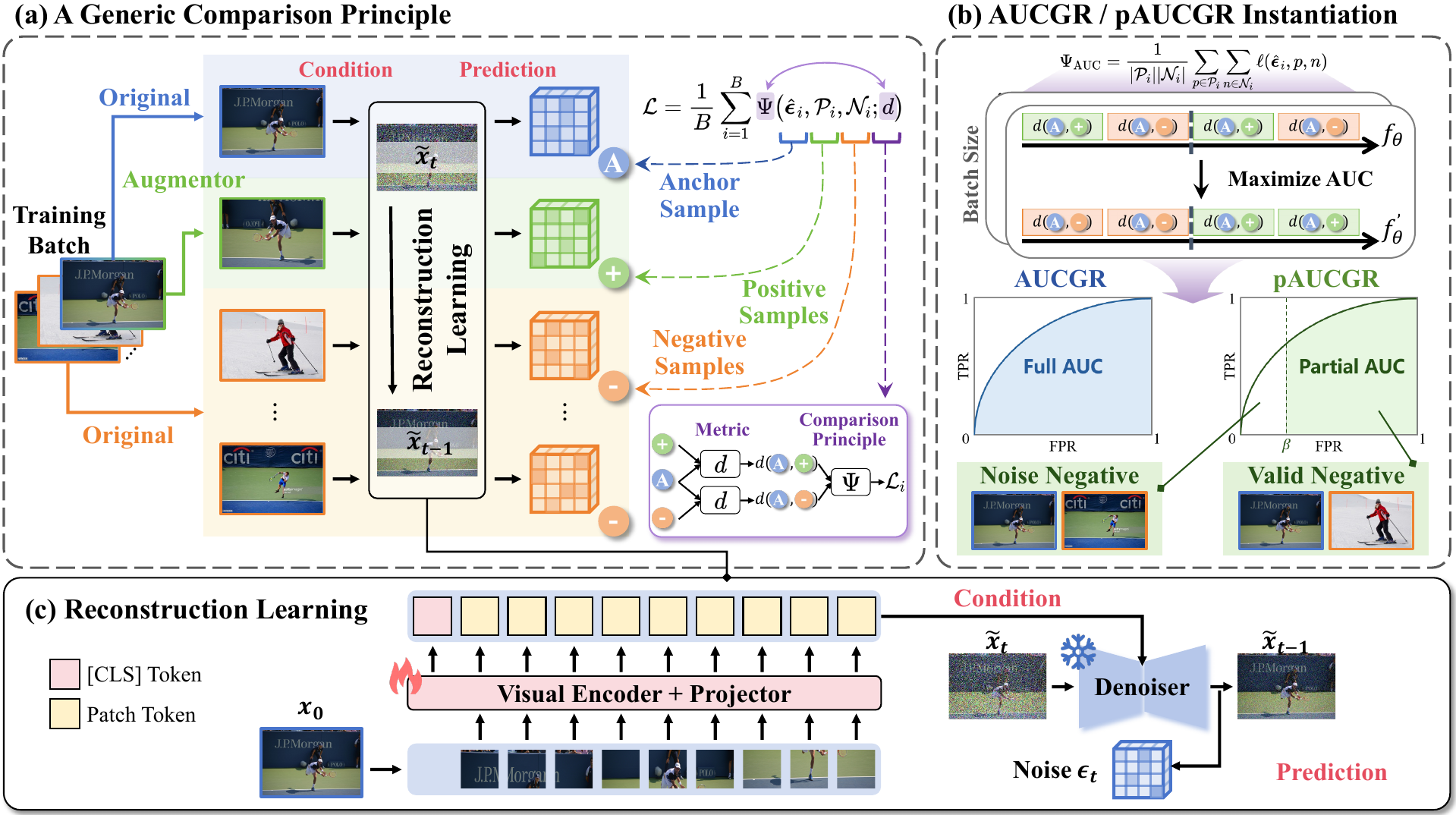}
   \caption{\textbf{An overview of our method.} (a) A generic comparison principle for generative reconstruction. (b) AUC-oriented instantiations, including AUCGR and pAUCGR. (c) Reconstruction learning pipeline. Our method serves as a plug-and-play optimization module that replaces the original optimization strategy in existing reconstruction learning methods, thereby yielding balanced enhanced visual representations.}
   \label{fig: overview}
\end{figure*}

In the previous section, we analyze the limitations of the naive linear weighting strategy. To pursue a unified form of supervision, we inject comparison information into the reconstruction signal, so that generative reconstruction itself can carry relative supervision in the predicted-noise space. A brief overview is provided in \cref{fig: overview}.

\subsection{A Generic Comparison Principle for Generative Reconstruction}
\label{subsec: A Generic Comparison Principle for Generative Reconstruction}

As discussed in \cref{subsec: Linear Combination of Two Objectives}, standard reconstruction learning enhances \textbf{\textcolor{PerColor}{P-Ability}} by enforcing that each image is reconstructed accurately under its own conditioning. However, it does not provide explicit supervision for \textbf{\textcolor{DisColor}{D-Ability}}. \textbf{Importantly, \textbf{\textcolor{DisColor}{D-Ability}} is inherently comparison-driven, as it relies on ensuring that semantically compatible signals are closer than incompatible ones in the representation space.} Rather than externally merging reconstruction and comparison through a weighted sum as in the naive approach, we embed the comparison mechanism directly into the reconstruction learning process. In this manner, both \textbf{\textcolor{PerColor}{P-Ability}} and \textbf{\textcolor{DisColor}{D-Ability}} are jointly supervised under a unified objective.

\textbf{This comparison principle can be operationalized by holding the denoising state fixed in the diffusion process while varying only the reconstruction condition.} As illustrated in \cref{fig: overview}(a), for a reference image $\x_i$, we first sample a diffusion timestep $t$ and generate the corresponding noised image:
\begin{equation}
\tilde{\x}_{t,i}
=
\sqrt{\bar{\alpha}_t}\,\x_i
+
\sqrt{1-\bar{\alpha}_t}\,\boldsymbol{\epsilon},
\quad
\boldsymbol{\epsilon}\sim\mathcal{N}(\mathbf{0},\mathbf{I}),
\end{equation}
where $\bar{\alpha}_t$ denotes the accumulated noise schedule. Given a conditioning vector $\mathbf{c}$, the denoiser predicts the noise component and conducts one-step denoising from $t$ to $t-1$:
\begin{equation}
\tilde{\x}_{t-1,i}^{\mathbf{c}}
=
\frac{1}{\sqrt{\alpha_t}}
\left(
\tilde{\x}_{t,i}
-
\frac{1-\alpha_t}{\sqrt{1-\bar{\alpha}_t}}
\epsilon_{\theta}(\tilde{\x}_{t,i}, \mathbf{c}, t)
\right)
+
\sigma_t \mathbf{z},
\end{equation}
where $\mathbf{z}\sim\mathcal{N}(\mathbf{0},\mathbf{I})$. We keep $\tilde{\x}_{t,i}$ and $t$ fixed, altering only the condition $\mathbf{c}$. The condition derived from $\x_i$ itself establishes the anchor denoising result. A condition from the same image, such as an augmented view, is expected to yield a similar one-step denoising outcome, while a condition from a different image should result in a distinct denoising output. To facilitate meaningful comparison, we note that $\tilde{\x}_{t-1,i}^{\mathbf{c}}$ is inherently stochastic. To mitigate the effects of randomness, we directly compare the corresponding noise predictions $\epsilon_{\theta}(\tilde{\x}_{t,i}, \mathbf{c}, t)$, since:
\begin{align*}
    \mathbb{E}\left[ \tilde{\x}_{t-1,i}^{\mathbf{c}} | \tilde{\x}_{t,i} \right] \propto - \epsilon_{\theta}(\tilde{\x}_{t,i}, \mathbf{c}, t).
\end{align*}
In other words, leveraging the noise prediction enables us to average out random fluctuations, thereby embedding the comparison directly within the diffusion reconstruction process.

We now formalize this mechanism. For an image $\x_i$, let
\begin{equation}
\hat{\boldsymbol\epsilon}_i
=
\epsilon_{\theta}\bigl(\tilde{\x}_{t,i},\,h_{\omega}(f_{\phi}(\x_i)),\,t\bigr)
\label{eq: anchor_predicted_noise}
\end{equation}
denote the anchor predicted noise. We then consider two sets of responses around this anchor in the same predicted-noise space.

The \textbf{positive} set $\mathcal{P}_i$ comprises responses generated by conditions originating from $\x_i$. The \textbf{negative} set $\mathcal{N}_i$ consists of responses induced by conditions from images distinct from $\x_i$. Specifically, $\mathcal{P}_i$ may include the predicted noise from augmented views of $\x_i$ as well as the ground-truth diffusion noise, whereas $\mathcal{N}_i$ is constructed using conditions derived from other images.

With this notation, the comparison structure for $\x_i$ is defined as
\begin{equation}
\mathcal{C}_i
=
(\hat{\boldsymbol\epsilon}_i,\mathcal{P}_i,\mathcal{N}_i,d),
\label{eq: comparison_tuple}
\end{equation}
where $d$ denotes a generic metric in the predicted-noise space. The corresponding generic comparison objective is written as
\begin{equation}
\mathcal{L}=\frac{1}{B}\sum_{i=1}^{B}
\Psi\bigl(\hat{\boldsymbol\epsilon}_i,\mathcal{P}_i,\mathcal{N}_i; d\bigr),
\label{eq: generic_comparison_loss}
\end{equation}
where $B$ is the number of samples and $\Psi(\cdot)$ represents a comparison functional induced by $d$. \cref{eq: generic_comparison_loss} arranges positive and negative responses with respect to the same anchor in the predicted-noise space. Thus, reconstructive and comparative information are unified within a single objective.

This formulation provides a general template for comparison-based generative reconstruction. Variations in the choice of $\mathcal{P}_i$, $\mathcal{N}_i$, $d$, and $\Psi$ yield different realizations, while consistently adhering to the same foundational principle.

\subsection{AUC-oriented Instantiation of the Comparison Principle}
\label{subsec: AUC-oriented Instantiation of the Comparison Principle}

The general principle in \cref{eq: generic_comparison_loss} arranges an anchor and its corresponding positive and negative responses within a unified comparison objective. The remaining challenge is to determine an appropriate metric for quantifying their relative relationships. Building upon \cref{subsec: AUC Optimization}, we instantiate this metric using the Area Under the ROC Curve (AUC). As illustrated in \cref{fig: overview}(b), the AUC criterion naturally models whether positive responses are consistently ranked above negative ones. This comparison-based structure is directly aligned with the objective of enhancing \textbf{\textcolor{DisColor}{D-Ability}}.

In the AUC framework, the score is defined as the similarity between the anchor $i$ and a candidate response $v$.
\begin{equation}
q_i(v)=d(\hat{\boldsymbol\epsilon}_i,v),
\label{eq: anchor_wise_score}
\end{equation}
where $v$ denotes a sample drawn from either $\mathcal{P}_i$ or $\mathcal{N}_i$, and $d$ is the comparison metric introduced in \cref{eq: generic_comparison_loss}. Here, $d$ is instantiated as cosine similarity, such that a higher score indicates greater alignment with the anchor. The corresponding true positive rate (TPR) and false positive rate (FPR) at threshold $\tau$ are given by
\begin{equation}
\begin{aligned}
\mathrm{TPR}(\tau)&=\mathbb{P}_{i,\,p\sim\mathcal{P}_i}\bigl(q_i(p)>\tau\bigr),\\
\mathrm{FPR}(\tau)&=\mathbb{P}_{i,\,n\sim\mathcal{N}_i}\bigl(q_i(n)>\tau\bigr).
\end{aligned}
\label{eq: overall_tpr_fpr}
\end{equation}
Accordingly, the overall AUC is the area under the resulting ROC curve:
\begin{equation}
\mathrm{AUC}
=
\int_{0}^{1}\mathrm{TPR}\bigl(\mathrm{FPR}^{-1}(u)\bigr)\,du.
\label{eq: overall_auc_integral}
\end{equation}
Equivalently, following~\cite{hanley1982meaning}, it can be written as
\begin{equation}
\mathrm{AUC}
=
\mathbb{P}_{i,\,p\sim\mathcal{P}_i,\,n\sim\mathcal{N}_i}
\bigl(q_i(p)>q_i(n)\bigr).
\label{eq: overall_auc}
\end{equation}

However, \cref{eq: overall_auc} is defined through an indicator function and is therefore difficult to optimize directly. By replacing its indicator with a tractable pairwise surrogate, we obtain the loss of \textbf{AUC Generative Reconstruction (AUCGR)} as
\begin{equation}
\mathcal{L}_{\mathrm{AUCGR}}
=
\frac{1}{B}\sum_{i=1}^{B}
\frac{1}{|\mathcal{P}_i||\mathcal{N}_i|}
\sum_{p\in\mathcal{P}_i}\sum_{n\in\mathcal{N}_i}
\ell\bigl(q_i(p)-q_i(n)\bigr),
\label{eq: auc_instantiation_loss}
\end{equation}
where $\ell$ is a pairwise surrogate loss function defined on the ordering gap $q_i(p)-q_i(n)$. The function $\ell$ is \textbf{monotonically decreasing}, as a larger gap indicates that the positive sample is ranked above the negative, thus incurring a smaller penalty. This objective promotes higher scores for positive samples relative to negative samples with respect to the same anchor.

We further provide a theoretical analysis of AUCGR, demonstrating that it is specifically constructed to fulfill the dual objectives discussed in \cref{subsec:visual_representation_learning}. The following two theorems summarize these findings. Proofs are deferred to \cref{sec: supp_Proof of Theorem 2} and \cref{sec: supp_Proof of Theorem 3} for completeness.

\begin{simplebox}
\begin{thm}[AUCGR Supports \textbf{\textcolor{DisColor}{D-Ability}}]
\label{thm: auc_oriented_d_ability}
Fix a diffusion step $t$ and define $T(\z)=\epsilon_{\theta}\bigl(\tilde{\x}_{t},\,h_{\omega}(\z),\,t\bigr)$. Under Assumption~\ref{assmp: bilip_aucgr}, the intra-class scatter and inter-class scatter in the feature space $(S_{\text{inner}},S_{\text{inter}})$ can be bounded by those in the noise space $(S_{\text{inner}}^{(\epsilon)}(t),S_{\text{inter}}^{(\epsilon)}(t))$ as
\begin{equation}
S_{\text{inner}}\le\frac{1}{m^2}S_{\text{inner}}^{(\epsilon)}(t),
\end{equation}
\begin{equation}
S_{\text{inter}}\ge \kappa S_{\text{inter}}^{(\epsilon)}(t)-\eta S_{\text{inner}}^{(\epsilon)}(t),
\end{equation}
where $m$, $\kappa$, and $\eta$ are positive constants depending on the Lipschitz continuity of $T(\cdot)$.
\end{thm}
\end{simplebox}

\begin{simplebox}
\begin{thm}[AUCGR Supports \textbf{\textcolor{PerColor}{P-Ability}}]
\label{thm: auc_oriented_p_ability}
Under the assumptions that negatives in the mini-batch are well separated from the anchor (i.e., $\mathrm{sim}(\hat{\boldsymbol\epsilon},\hat{\boldsymbol\epsilon}_{-}^{j}) \ll \mathrm{sim}(\hat{\boldsymbol\epsilon},\boldsymbol\epsilon_{\mathrm{gt}})$ for all $j\neq i$), predicted-noise norms are bounded away from $0$ and $\infty$, and the surrogate loss is decreasing with bounded slope for the pairwise gaps used by AUCGR, then the AUCGR loss is bounded by a scaled reconstruction loss up to additive constants:
\begin{equation}
\lambda_{\min}R_{\mathrm{rec}}+c_{\min}
\le
\mathcal{L}_{\mathrm{AUCGR}}
\le
\lambda_{\max}R_{\mathrm{rec}}+c_{\max},
\end{equation}
where $R_{\mathrm{rec}}=\frac{1}{B}\sum_{i=1}^{B}\bigl\|\hat{\boldsymbol\epsilon}_i-\boldsymbol\epsilon_{t,i}^{\mathrm{gt}}\bigr\|_2^2$, $\lambda_{\min},\lambda_{\max}>0$, and $c_{\min},c_{\max}$ are constants.
\end{thm}
\end{simplebox}

\begin{rem}
Theorem \ref{thm: auc_oriented_d_ability} establishes that as $\mathcal{L}_{\mathrm{AUCGR}}$ decreases, the intra-class scatter $S_{\text{inner}}$ decreases and the inter-class scatter $S_{\text{inter}}$ increases in the feature space, directly supporting the discriminative objective in \cref{eq: minmax_s}. Theorem \ref{thm: auc_oriented_p_ability} shows that $\mathcal{L}_{\mathrm{AUCGR}}$ is bounded by the reconstruction error up to positive scaling constants, thereby fulfilling the fine-grained perceptual objective in \cref{eq: min_rec}.
\end{rem}

\subsection{Partial AUC-oriented Generative Reconstruction}
\label{subsec: Partial AUC-oriented Generative Reconstruction}

The preceding analysis establishes AUCGR by instantiating $d$ as cosine similarity and $\Psi$ as an AUC-oriented pairwise functional. In this formulation, $\mathcal{P}_i$ and $\mathcal{N}_i$ act as the positive and negative comparison sets sampled from each mini-batch. However, this default construction presents a limitation: \textbf{noisy negatives are intermixed with true negatives and may dominate the optimization process}. This motivates a partial-AUC reformulation, which adheres to the same comparison principle while restricting optimization to reliable negatives.

\subsubsection{Limitation of AUCGR}

Although AUCGR resolves the objective conflict of the naive method, it still inherits a weakness from the default construction of the comparison sets. The AUC objective treats all negatives in $\mathcal{N}_i$ equally. In practice, some in-batch negatives remain semantically compatible with the anchor and should not be treated as equally reliable negatives.

Below, we formalize the issue in the predicted-noise space. Recall that the anchor predicted noise in \cref{eq: anchor_predicted_noise} is $\hat{\boldsymbol\epsilon}_i=\epsilon_{\theta}(\tilde{\x}_{t,i},h_{\omega}(f_{\phi}(\x_i)),t)$.
Under the default AUCGR construction, the in-batch negative set is
\begin{equation}
\mathcal{N}_i
=
\{\hat{\boldsymbol\epsilon}_{i}^{-,k}\}_{k\neq i},
\qquad
\hat{\boldsymbol\epsilon}_{i}^{-,k}
=
\epsilon_{\theta}\bigl(\tilde{\x}_{t,i},\,h_{\omega}(f_{\phi}(\x_k)),\,t\bigr).
\end{equation}
Here, $\tilde{\x}_{t,i}$ is fixed and only the condition changes. A noisy negative corresponds to an index $k\neq i$ whose condition remains compatible with $\x_i$. In the predicted-noise space, this means that
\begin{equation}
q_i\bigl(\hat{\boldsymbol\epsilon}_{i}^{-,k}\bigr)
=
d\bigl(\hat{\boldsymbol\epsilon}_i,\hat{\boldsymbol\epsilon}_{i}^{-,k}\bigr)
\approx 1.
\end{equation}
Since this section instantiates $d$ as cosine similarity, such a negative receives a high anchor-wise similarity score.

We can make its effect explicit from the pairwise AUCGR objective. For anchor $i$, the corresponding loss is
\begin{equation}
\mathcal{L}_{\mathrm{AUCGR}}^{(i)}
=
\frac{1}{|\mathcal{P}_i||\mathcal{N}_i|}
\sum_{p\in\mathcal{P}_i}\sum_{n\in\mathcal{N}_i}
\ell\bigl(q_i(p)-q_i(n)\bigr).
\label{eq: full_aucgr_anchor_loss}
\end{equation}
The contribution of a specific negative $\hat{\boldsymbol\epsilon}_{i}^{-,k}$ to this loss can be written as
\begin{equation}
\rho_{i,k}
=
\frac{1}{|\mathcal{P}_i|}
\sum_{p\in\mathcal{P}_i}
\ell\Bigl(q_i(p)-q_i\bigl(\hat{\boldsymbol\epsilon}_{i}^{-,k}\bigr)\Bigr).
\label{eq: full_aucgr_negative_contribution}
\end{equation}
By the decreasing property of $\ell(\cdot)$ in \cref{eq: auc_instantiation_loss}, a larger negative score $q_i(\hat{\boldsymbol\epsilon}_{i}^{-,k})$ yields a smaller gap and therefore a larger contribution $\rho_{i,k}$. Consequently, optimization is dominated by the highest-score negatives. If these negatives are noisy or ambiguous, the ranking signal becomes systematically biased.

\begin{figure}[t]
\centering
\subfloat[Training data distribution.]{
\begin{minipage}[b]{0.47\linewidth}
\centering
\includegraphics[width=\linewidth]{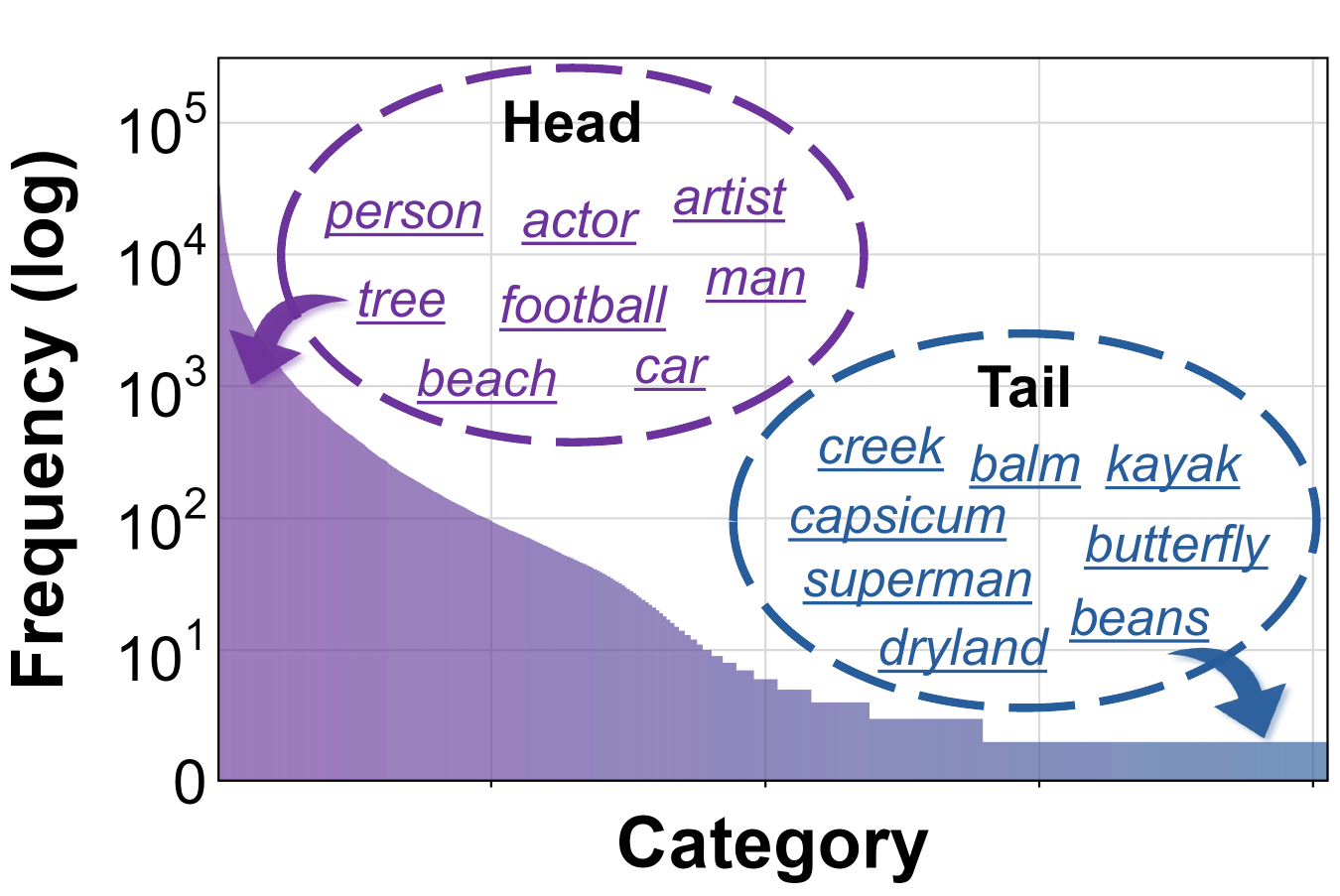}
\label{fig: cc3m_long_tail}
\end{minipage}
}
\hfil
\subfloat[Noisy negatives.]{
\begin{minipage}[b]{0.47\linewidth}
\centering
\includegraphics[width=\linewidth]{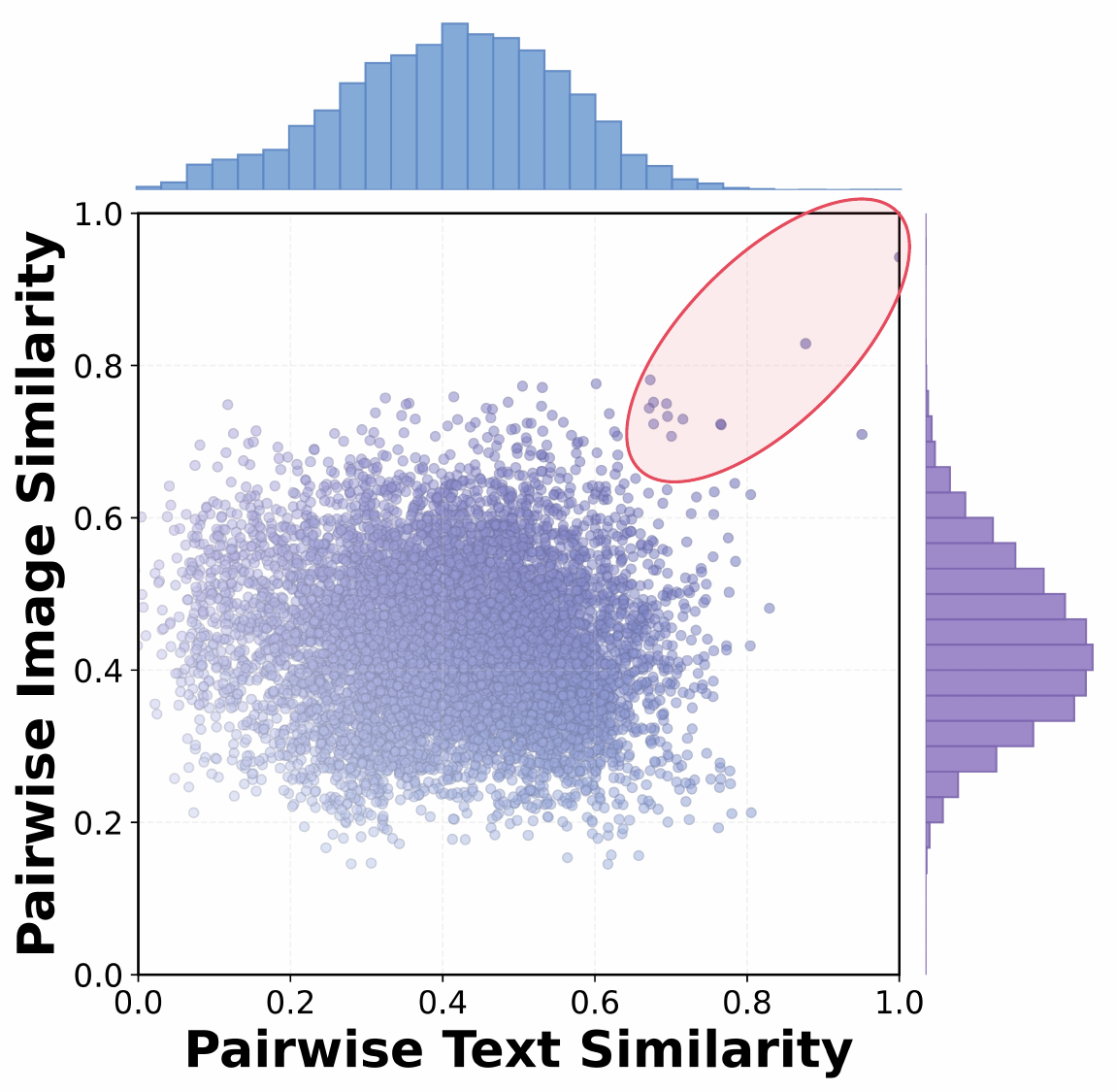}
\label{fig: cc3m_noisy_negatives}
\end{minipage}
}
\caption{\textbf{Empirical verification of noisy negatives.} (a) The training data exhibits a clear long-tail distribution. (b) We randomly sample one mini-batch with batch size $128$ and compute pairwise image similarity and label-text similarity within the batch. The results show that highly similar image pairs indeed exist during training.}
\label{fig: empirical verification of noisy negatives}
\end{figure}

We further verify this argument empirically in \cref{fig: empirical verification of noisy negatives}. Noisy negatives mainly arise from the data distribution and mini-batch sampling. We first examine the sample frequency of semantic categories in the training data. The resulting distribution exhibits a clear long-tail pattern. We then randomly sample one mini-batch and compute the pairwise image similarity and label-text similarity within the batch. The statistics show that highly similar image pairs indeed appear in a normal mini-batch. These highly similar pairs can become noisy negatives under the default in-batch negative construction.

Therefore, the limitation of AUCGR is not the AUC principle itself. The limitation is that AUCGR still allows the least reliable negative comparisons to dominate optimization. This observation motivates a partial-AUC reformulation that keeps all positives while restricting the negative side to the reliable region.

\subsubsection{From AUCGR to pAUCGR}

To suppress this bias, we keep all positives unchanged and truncate only the negative side. Given a truncation ratio $\alpha\in(0,1)$, let $\gamma(\alpha)$ denote the $(1-\alpha)$-quantile of the negative-score distribution $\{q_i(n)\}_{n\in\mathcal{N}_i}$:
\begin{equation}
\mathbb{P}_{n\sim\mathcal{N}_i}\bigl(q_i(n)\le \gamma(\alpha)\bigr)=1-\alpha.
\label{eq: paucgr_quantile_threshold}
\end{equation}
We then retain only the low-score negative subset
\begin{equation}
\mathcal{N}_i^{(\alpha)}
=
\{n\in\mathcal{N}_i:\ q_i(n)\le \gamma(\alpha)\}.
\label{eq: paucgr_retained_negative_set}
\end{equation}
This operation removes the top $\alpha$ fraction of highest-score negatives and keeps the remaining $1-\alpha$ fraction.

The following proposition makes the partial-AUC interpretation explicit.

\begin{simplebox}
\begin{prop}
\label{prop: one_way_partial_auc}
Let $q:\mathcal{X}\to\mathbb{R}$ be a scoring function, and let $x^+\sim P_+$ and $x^-\sim P_-$ denote a positive and a negative sample. Assume that the negative score distribution $F_-(t)=\mathbb{P}(q(x^-)\le t)$ is continuous. For any measurable negative subset $\mathcal{N}$ with positive probability, define the conditional ranking objective
\begin{equation}
\mathrm{AUC}(\mathcal{N})
=
\mathbb{P}\big(q(x^+)>q(x^-)\mid x^-\in\mathcal{N}\big).
\end{equation}
For any $\beta\in(0,1]$, let
\begin{equation}
\mathcal{N}_{\mathrm{low}}^{\beta}
=
\{x^-:F_-(q(x^-))\le \beta\}.
\end{equation}
Then the partial AUC over $[1-\beta,1]$ is defined as
\begin{equation}
\mathrm{pAUC}_{[1-\beta,1]}
:=
\int_{1-\beta}^{1}\mathrm{TPR}\big(\mathrm{FPR}^{-1}(\xi)\big)\,d\xi,
\end{equation}
where $\mathrm{TPR}(t)=\mathbb{P}(q(x^+)>t)$, $\mathrm{FPR}(t)=\mathbb{P}(q(x^-)>t)$, and $\mathrm{FPR}^{-1}(\xi)$ denotes the threshold associated with false positive rate $\xi$.
\end{prop}
\end{simplebox}

The proof is provided in \cref{sec: supp_Proof of Proposition 1}. In our setting, $\mathcal{N}_i^{(\alpha)}$ keeps exactly the lowest-score $1-\alpha$ fraction of negatives. Therefore, by Proposition~\ref{prop: one_way_partial_auc} with $\beta=1-\alpha$, the retained negative set corresponds to the partial AUC over the interval
\begin{equation}
\mathrm{pAUC}_{[\alpha,1]}
=
\int_{\alpha}^{1}\mathrm{TPR}\bigl(\mathrm{FPR}^{-1}(\xi)\bigr)\,d\xi.
\label{eq: paucgr_partial_auc}
\end{equation}
up to the constant factor $1-\alpha$. Since this factor does not depend on the model, the retained negative set does not define a heuristic filter. It restricts optimization to the reliable part of the AUC objective.

Based on \cref{eq: paucgr_retained_negative_set}, we define the final robust objective, termed \textbf{Partial AUC Generative Reconstruction (pAUCGR)}, as
\begin{equation}
\mathcal{L}_{\mathrm{pAUCGR}}
=
\frac{1}{B}\sum_{i=1}^{B}
\frac{1}{|\mathcal{P}_i|\,|\mathcal{N}_i^{(\alpha)}|}
\sum_{p\in\mathcal{P}_i}\sum_{n\in\mathcal{N}_i^{(\alpha)}}
\ell\bigl(q_i(p)-q_i(n)\bigr).
\label{eq: paucgr_loss}
\end{equation}
When $\alpha=0$, \cref{eq: paucgr_loss} collapses to the AUCGR objective in \cref{eq: auc_instantiation_loss}. As $\alpha$ increases, the optimization process incrementally excludes negative comparisons that are most likely to introduce noise. In this manner, pAUCGR maintains the comparison principle outlined in \cref{eq: generic_comparison_loss}, inherits the AUC-based interpretation established previously, and mitigates the dominant bias caused by ambiguous negatives. Consequently, pAUCGR serves as the final optimization framework in our approach.

\begin{rem}
Our goal is to remove highly similar, potentially noisy negatives using a consistent similarity criterion across mini-batches. The ratio $\alpha$ denotes the fraction of high-score negatives in the full training distribution. A mini-batch only provides a local estimate of this distribution. Fixing $\alpha$ in each batch targets the same removed fraction, but its empirical quantile threshold can vary. We instead fix the cosine similarity threshold at $\gamma_0=0.9$ and allow the removed fraction to vary across batches:
\begin{equation}
\widehat{\alpha}_{i,\mathcal{B}}(\gamma_0)
=
\frac{1}{|\mathcal{N}_{i,\mathcal{B}}|}
\sum_{n\in\mathcal{N}_{i,\mathcal{B}}}
\mathbf{1}\!\left[q_i(n)>\gamma_0\right].
\end{equation}
Consequently, each batch removes negatives above a fixed similarity threshold, whose correspondence to the induced population truncation ratio is proved in \cref{sec: alpha_gamma_equivalence}.
\end{rem}

\subsection{Surrogate Loss Instantiation of AUCGR}
\label{subsec: Surrogate Loss Instantiation of AUCGR}

The AUCGR and pAUCGR objectives in \cref{eq: auc_instantiation_loss} and \cref{eq: paucgr_loss} are obtained by replacing the non-differentiable AUC indicator with a surrogate loss $\ell(\cdot)$. Each surrogate converts the original 0-1 ranking event into a tractable pairwise surrogate risk, while preserving the preference that positive samples should be ranked ahead of negative samples. Different choices of $\ell(\cdot)$ correspond to different optimization strategies, which provide AUCGR with a broader family of realizations.

Following the AUC optimization literature~\cite{yang2021learning}, for the pairwise gap $u=q_i(p)-q_i(n)$, the surrogate loss $\ell(\cdot)$ can take the following choices.
\begin{itemize}
    \item Square. $\ell_{\mathrm{square}}(u)=[\lambda-u]_{+}^{2}$.
    \item Hinge. $\ell_{\mathrm{hinge}}(u)=[\lambda-u]_{+}$.
    \item Exponential. $\ell_{\mathrm{exp}}(u)=\exp(-u)$.
    \item Logistic. $\ell_{\mathrm{logistic}}(u)=\log(1+\exp(-u/\tau))$.
\end{itemize}
where $\lambda>0$ denotes a safety margin and $\tau>0$ denotes a temperature.

This view also clarifies the relation to our conference version. The conference version employs a contrastive objective that compares positive and negative reconstruction signals. This objective can be naturally connected to logistic AUCGR through the pairwise ranking formulation. Specifically, the anchor-wise loss $\mathcal{L}_{\mathrm{DCR}}^{(i)}$ is defined as
\begin{equation}
\mathcal{L}_{\mathrm{DCR}}^{(i)}
=
-\frac{1}{|\mathcal{P}_i|}\sum_{p\in\mathcal{P}_i}
\log
\frac{\exp\bigl(q_i(p)/\tau\bigr)}
{\sum_{v\in\mathcal{P}_i\cup\mathcal{N}_i}\exp\bigl(q_i(v)/\tau\bigr)}.
\label{eq: contrastive_instantiation_loss}
\end{equation}
The following theorem establishes this correspondence.

\begin{simplebox}
\begin{thm}
\label{thm: cgr_as_pairwise_ranking_new}
For anchor $i$, let $\mathcal{P}_i$ denote the positive set and let $\mathcal{N}_i$ denote the negative set. Then we have the following result:
\begin{equation}
\mathcal{L}_{\mathrm{DCR}}^{(i)}
\ge
\frac{1}{|\mathcal{P}_i||\mathcal{N}_i|}
\sum_{p\in\mathcal{P}_i}\sum_{n\in\mathcal{N}_i}
\ell\bigl(q_i(p)-q_i(n)\bigr),
\label{eq: cgr_pairwise_ranking_new}
\end{equation}
where $\ell(\cdot)$ is instantiated as the logistic surrogate function
\begin{equation}
\ell_{\mathrm{logistic}}(u)
=
\log\left(1+\exp\left(-\frac{u}{\tau}\right)\right).
\label{eq: cgr_logistic_surrogate_new}
\end{equation}
Moreover, the bound is tight if and only if $|\mathcal{P}_i|=|\mathcal{N}_i|=1$.
\end{thm}
\end{simplebox}

The proof is provided in \cref{sec: supp_Proof of Theorem 4}. Theorem \ref{thm: cgr_as_pairwise_ranking_new} shows that $\mathcal{L}_{\mathrm{DCR}}$ is a special case of AUCGR when $|\mathcal{P}_i|=|\mathcal{N}_i|=1$. Meanwhile, from the ROC perspective, AUCGR supports a broader family of surrogate losses, making it \textbf{a more general formulation with a clearer theoretical interpretation}. We provide a detailed theoretical analysis in the next section.

\section{Generalization Bound}
\label{sec: Generalization Bound}
To further establish that one-way negative truncation does not compromise statistical reliability, we analyze the generalization properties of pAUCGR. Let $f \in \mathcal{F}$ denote a comparison model parameterized by the learnable modules. Each training instance is represented by an anchor tuple
\begin{equation}
z_i=(\hat{\boldsymbol\epsilon}_i, \hat{\boldsymbol\epsilon}^+_i, \hat{\boldsymbol\epsilon}^-_i),  ~~\hat{\boldsymbol\epsilon}^+_i \in \mathcal{P},  \hat{\boldsymbol\epsilon}^-_i\in \mathcal{N}^{-}
\label{eq: paucgr_anchor_tuple}
\end{equation}
where $\mathcal{P}$ follows the same positive-set construction as above, $\mathcal{N}^{-}=\{n_j\}_{j=1}^{m}$ is a candidate negative pool sampled from the  negative condition distribution, and $\hat{\boldsymbol\epsilon}$ is the anchor predicted noise induced by $f$.

For a fixed truncation ratio $\alpha\in(0,1)$, $\gamma(\alpha)$ denote the $(1-\alpha)$-quantile of the negative scores $\{d(\hat{\boldsymbol\epsilon},n):n\in\mathcal{N}^{-}\}$, and define the retained negative set
\begin{equation}
\mathcal{N}_{\alpha}(z)
=
\{n\in\mathcal{N}^{-}:d(\hat{\boldsymbol\epsilon},n)\le \gamma(\alpha)\}.
\label{eq: paucgr_population_retained_negative}
\end{equation}
The corresponding anchor-wise pAUCGR loss is
\begin{equation}
g_f^{(\alpha)}(z)
=
\frac{1}{|\mathcal{P}|\,|\mathcal{N}_{\alpha}(z)|}
\sum_{p\in\mathcal{P}}\sum_{n\in\mathcal{N}_{\alpha}(z)}
\ell\bigl(d(\hat{\boldsymbol\epsilon},p)-d(\hat{\boldsymbol\epsilon},n)\bigr).
\label{eq: paucgr_anchor_loss}
\end{equation}
Accordingly, the population risk and empirical risk are
\begin{equation}
\mathcal{R}_{\alpha}(f)
=
\mathbb{E}_{z\sim\mathcal{Q}}\bigl[g_f^{(\alpha)}(z)\bigr],
\qquad
\hat{\mathcal{R}}_{\alpha}(f)
=
\frac{1}{N}\sum_{i=1}^{N}g_f^{(\alpha)}(z_i),
\label{eq: paucgr_population_empirical_risk}
\end{equation}
where $\mathcal{Q}$ denotes the distribution of such anchor tuples, and $\{z_i\}_{i=1}^{B}$ are i.i.d.\ samples from $\mathcal{Q}$.

\begin{assmp}
\label{ass: paucgr_generalization}
We adopt the following assumptions.

\begin{enumerate}[leftmargin=*]
\item
For every $f\in\mathcal{F}$, the similarity score satisfies $d(\hat{\boldsymbol\epsilon},v)\in[-1,1]$. Hence the pairwise gap in \cref{eq: paucgr_anchor_loss} belongs to $[-2,2]$. The surrogate $\ell$ is continuous on $[-2,2]$, and there exists a constant $M>0$ such that $0\le \ell(u)\le M$ for all $u\in[-2,2]$.

\item
For every $f\in\mathcal{F}$ and every anchor tuple $z$ in the support of $\mathcal{Q}$, the negative-score distribution is continuous. Therefore, $\gamma(\alpha)$ is well defined and the truncation threshold is unambiguous.

\item
Let $\mathcal{G}_{\alpha}=\{g_f^{(\alpha)}:f\in\mathcal{F}\}$. Its covering number under the sup norm satisfies
\begin{equation}
\log\mathcal{N}\bigl(\mathcal{G}_{\alpha},\varepsilon,\|\cdot\|_{\infty}\bigr)
\le
A\log\frac{R}{\varepsilon},
\qquad
\forall \varepsilon\in(0,R],
\label{eq: paucgr_covering_number}
\end{equation}
for some constants $A>0$ and $R>0$.
\end{enumerate}
\end{assmp}

These assumptions are mild and reasonable in our context. The boundedness condition is justified by the use of cosine similarity and a bounded surrogate loss over the relevant interval. The continuity condition simply excludes degenerate score distributions. The covering number requirement is a standard complexity constraint to ensure uniform convergence.

\begin{simplebox}
\begin{thm}[Generalization of pAUCGR]
\label{thm: paucgr_generalization}
Under Assumption \ref{ass: paucgr_generalization}, for any $\delta\in(0,1)$ and any fixed $\varepsilon\in(0,R]$, with probability at least $1-\delta$, the following inequality holds simultaneously for all $f\in\mathcal{F}$:
\begin{equation}
\begin{aligned}
\left|
\mathcal{R}_{\alpha}(f)-\hat{\mathcal{R}}_{\alpha}(f)
\right|
&\le
2\varepsilon \\
&\quad+
M\sqrt{
\frac{
A\log(R/\varepsilon)+\log(2/\delta)
}{
2N
}
}.
\end{aligned}
\label{eq: paucgr_generalization_eps}
\end{equation}
In particular, choosing $\varepsilon=R/N$ yields
\begin{equation}
\begin{aligned}
\left|
\mathcal{R}_{\alpha}(f)-\hat{\mathcal{R}}_{\alpha}(f)
\right|
&\le
\frac{2R}{N} \\
&\quad+
M\sqrt{
\frac{
A\log N+\log(2/\delta)
}{
2N
}
}.
\end{aligned}
\label{eq: paucgr_generalization_final}
\end{equation}
\end{thm}
\end{simplebox}

The proof is presented in \cref{sec: supp_Proof of Theorem 5}. This theorem demonstrates that pAUCGR retains the standard uniform convergence rate of $O(\sqrt{\log N/N})$ even in the presence of one-way truncation, where $N$ denotes the number of samples.

\section{Experiments}
\label{sec: experiments}

\begin{table*}[t!]
\centering
\renewcommand\arraystretch{0.6}
\caption{\textbf{Performance of \textcolor{PerColor}{Fine-grained Perceptual Ability (P-Ability)} on the MMVP-VLM benchmark in terms of \underline{ACC}.} `+DCR' denotes our conference version. The visual patterns are symbolized as: \textbf{\faCompass}: Orientation and Direction, \textbf{\faSearch}: Presence of Specific Features, \textbf{\faSync}: State and Condition, \textbf{\faSortNumericUp}: Quantity and Count, \textbf{\faMapPin}: Positional and Relational Context, \textbf{\faPalette}: Color and Appearance, \textbf{\faCogs}: Structural and Physical Characteristics, \textbf{\faFont}: Texts, \textbf{\faCamera}: Viewpoint and Perspective.}
\resizebox{0.85\linewidth}{!}{
    \begin{tabular}{cccccccccccccc}
    \toprule
    \textbf{CLIP Backbone} & \textbf{Resol.} & \textbf{\#Params} & \textbf{Method} & \textbf{\faCompass} & \textbf{\faSearch} & \textbf{\faSync} & \textbf{\faSortNumericUp} & \textbf{\faMapPin} & \textbf{\faPalette} & \textbf{\faCogs} & \textbf{\faFont} & \textbf{\faCamera} & \textbf{Avg} \\
    \midrule
    \multirow{9}[8]{*}{OpenAI ViT-L-14} & \multirow{9}[8]{*}{224} & \multirow{9}[8]{*}{427.6} & Original &13.3&13.3&20.0&20.0&13.3&53.3&20.0&6.7&13.3&19.2\\
\cmidrule{4-14}
          &       &       & DIVA  &13.3&6.7&13.3&26.7&13.3&46.7&26.7&20.0&26.7&21.5\\
          &       &       & +DCR  & 6.7   & 20.0  & 33.3  & 0.0   & 26.7  & 66.7  & 20.0  & 6.7   & 20.0  & 22.2  \\
          &       &       & \cellcolor[rgb]{ .945,  .973,  .98}+AUCGR & \cellcolor[rgb]{ .945,  .973,  .98}0.0& \cellcolor[rgb]{ .945,  .973,  .98}33.3& \cellcolor[rgb]{ .945,  .973,  .98}26.7& \cellcolor[rgb]{ .945,  .973,  .98}6.7& \cellcolor[rgb]{ .945,  .973,  .98}0.0& \cellcolor[rgb]{ .945,  .973,  .98}73.3& \cellcolor[rgb]{ .945,  .973,  .98}33.3& \cellcolor[rgb]{ .945,  .973,  .98}13.3& \cellcolor[rgb]{ .945,  .973,  .98}13.3& \cellcolor[rgb]{ .945,  .973,  .98}22.2\\
          &       &       & \cellcolor[rgb]{ .945,  .973,  .98}+pAUCGR & \cellcolor[rgb]{ .945,  .973,  .98}6.7& \cellcolor[rgb]{ .945,  .973,  .98}13.3& \cellcolor[rgb]{ .945,  .973,  .98}46.7& \cellcolor[rgb]{ .945,  .973,  .98}6.7& \cellcolor[rgb]{ .945,  .973,  .98}6.7& \cellcolor[rgb]{ .945,  .973,  .98}66.7& \cellcolor[rgb]{ .945,  .973,  .98}20.0& \cellcolor[rgb]{ .945,  .973,  .98}20.0& \cellcolor[rgb]{ .945,  .973,  .98}33.3& \cellcolor[rgb]{ .945,  .973,  .98}\textbf{24.4} \\
\cmidrule{4-14}
          &       &       & GenHancer &13.3&13.3&13.3&13.3&13.3&66.7&40.0&20.0&33.3&25.2\\
          &       &       & +DCR  & 13.3  & 20.0  & 33.3  & 6.7   & 20.0  & 73.3  & 13.3  & 33.3  & 13.3  & 25.2  \\
          &       &       & \cellcolor[rgb]{ .945,  .973,  .98}+AUCGR & \cellcolor[rgb]{ .945,  .973,  .98}13.3& \cellcolor[rgb]{ .945,  .973,  .98}20.0& \cellcolor[rgb]{ .945,  .973,  .98}40.0& \cellcolor[rgb]{ .945,  .973,  .98}0.0& \cellcolor[rgb]{ .945,  .973,  .98}26.7& \cellcolor[rgb]{ .945,  .973,  .98}73.3& \cellcolor[rgb]{ .945,  .973,  .98}26.7& \cellcolor[rgb]{ .945,  .973,  .98}13.3& \cellcolor[rgb]{ .945,  .973,  .98}26.7& \cellcolor[rgb]{ .945,  .973,  .98}26.7\\
          &       &       & \cellcolor[rgb]{ .945,  .973,  .98}+pAUCGR & \cellcolor[rgb]{ .945,  .973,  .98}6.7& \cellcolor[rgb]{ .945,  .973,  .98}40.0& \cellcolor[rgb]{ .945,  .973,  .98}26.7& \cellcolor[rgb]{ .945,  .973,  .98}20.0& \cellcolor[rgb]{ .945,  .973,  .98}6.7& \cellcolor[rgb]{ .945,  .973,  .98}73.3& \cellcolor[rgb]{ .945,  .973,  .98}13.3& \cellcolor[rgb]{ .945,  .973,  .98}33.3& \cellcolor[rgb]{ .945,  .973,  .98}26.7& \cellcolor[rgb]{ .945,  .973,  .98}\textbf{27.4} \\
    \midrule
    \midrule
    \multirow{9}[8]{*}{OpenAI ViT-L-14} & \multirow{9}[8]{*}{336} & \multirow{9}[8]{*}{427.9} & Original &0.0&20.0&40.0&20.0&6.7&20.0&33.3&6.7&33.3&20.0\\
\cmidrule{4-14}
          &       &       & DIVA  &0.0&33.3&26.7&6.7&6.7&73.3&40.0&6.7&20.0&23.7\\
          &       &       & +DCR  & 13.3  & 40.0  & 26.7  & 6.7   & 13.3  & 73.3  & 26.7  & 13.3  & 6.7   & 24.4  \\
          &       &       & \cellcolor[rgb]{ .945,  .973,  .98}+AUCGR & \cellcolor[rgb]{ .945,  .973,  .98}13.3& \cellcolor[rgb]{ .945,  .973,  .98}40.0& \cellcolor[rgb]{ .945,  .973,  .98}26.7& \cellcolor[rgb]{ .945,  .973,  .98}6.7& \cellcolor[rgb]{ .945,  .973,  .98}13.3& \cellcolor[rgb]{ .945,  .973,  .98}73.3& \cellcolor[rgb]{ .945,  .973,  .98}26.7& \cellcolor[rgb]{ .945,  .973,  .98}13.3& \cellcolor[rgb]{ .945,  .973,  .98}6.7& \cellcolor[rgb]{ .945,  .973,  .98}24.4\\
          &       &       & \cellcolor[rgb]{ .945,  .973,  .98}+pAUCGR & \cellcolor[rgb]{ .945,  .973,  .98}33.3& \cellcolor[rgb]{ .945,  .973,  .98}26.7& \cellcolor[rgb]{ .945,  .973,  .98}33.3& \cellcolor[rgb]{ .945,  .973,  .98}20.0& \cellcolor[rgb]{ .945,  .973,  .98}6.7& \cellcolor[rgb]{ .945,  .973,  .98}53.3& \cellcolor[rgb]{ .945,  .973,  .98}40.0& \cellcolor[rgb]{ .945,  .973,  .98}6.7& \cellcolor[rgb]{ .945,  .973,  .98}20.0& \cellcolor[rgb]{ .945,  .973,  .98}\textbf{26.7} \\
\cmidrule{4-14}
          &       &       & GenHancer &13.3&26.7&13.3&13.3&26.7&66.7&20.0&40.0&20.0&26.7\\
          &       &       & +DCR  & 6.7   & 13.3  & 40.0  & 0.0   & 6.7   & 66.7  & 46.7  & 13.3  & 53.3  & 27.4  \\
          &       &       & \cellcolor[rgb]{ .945,  .973,  .98}+AUCGR & \cellcolor[rgb]{ .945,  .973,  .98}6.7& \cellcolor[rgb]{ .945,  .973,  .98}20.0& \cellcolor[rgb]{ .945,  .973,  .98}20.0& \cellcolor[rgb]{ .945,  .973,  .98}20.0& \cellcolor[rgb]{ .945,  .973,  .98}6.7& \cellcolor[rgb]{ .945,  .973,  .98}86.7& \cellcolor[rgb]{ .945,  .973,  .98}20.0& \cellcolor[rgb]{ .945,  .973,  .98}33.3& \cellcolor[rgb]{ .945,  .973,  .98}33.3& \cellcolor[rgb]{ .945,  .973,  .98}27.4\\
          &       &       & \cellcolor[rgb]{ .945,  .973,  .98}+pAUCGR & \cellcolor[rgb]{ .945,  .973,  .98}6.7& \cellcolor[rgb]{ .945,  .973,  .98}20.0& \cellcolor[rgb]{ .945,  .973,  .98}33.3& \cellcolor[rgb]{ .945,  .973,  .98}20.0& \cellcolor[rgb]{ .945,  .973,  .98}6.7& \cellcolor[rgb]{ .945,  .973,  .98}73.3& \cellcolor[rgb]{ .945,  .973,  .98}53.3& \cellcolor[rgb]{ .945,  .973,  .98}26.7& \cellcolor[rgb]{ .945,  .973,  .98}26.7& \cellcolor[rgb]{ .945,  .973,  .98}\textbf{29.6} \\
    \midrule
    \midrule
    \multirow{9}[8]{*}{MetaCLIP ViT-L-14} & \multirow{9}[8]{*}{224} & \multirow{9}[8]{*}{427.6} & Original &13.3&6.7&66.7&6.7&33.3&46.7&20.0&6.7&13.3&23.7\\
\cmidrule{4-14}
          &       &       & DIVA  &6.7&20.0&66.7&6.7&33.3&46.7&20.0&13.3&13.3&25.2\\
          &       &       & +DCR  & 6.7   & 13.3  & 40.0  & 13.3  & 13.3  & 60.0  & 46.7  & 6.7   & 33.3  & 25.9  \\
          &       &       & \cellcolor[rgb]{ .98,  .984,  .957}+AUCGR & \cellcolor[rgb]{ .98,  .984,  .957}6.7& \cellcolor[rgb]{ .98,  .984,  .957}0.0& \cellcolor[rgb]{ .98,  .984,  .957}40.0& \cellcolor[rgb]{ .98,  .984,  .957}0.0& \cellcolor[rgb]{ .98,  .984,  .957}20.0& \cellcolor[rgb]{ .98,  .984,  .957}66.7& \cellcolor[rgb]{ .98,  .984,  .957}46.7& \cellcolor[rgb]{ .98,  .984,  .957}20.0& \cellcolor[rgb]{ .98,  .984,  .957}40.0& \cellcolor[rgb]{ .98,  .984,  .957}26.7\\
          &       &       & \cellcolor[rgb]{ .98,  .984,  .957}+pAUCGR & \cellcolor[rgb]{ .98,  .984,  .957}13.3& \cellcolor[rgb]{ .98,  .984,  .957}20.0& \cellcolor[rgb]{ .98,  .984,  .957}60.0& \cellcolor[rgb]{ .98,  .984,  .957}13.3& \cellcolor[rgb]{ .98,  .984,  .957}13.3& \cellcolor[rgb]{ .98,  .984,  .957}53.3& \cellcolor[rgb]{ .98,  .984,  .957}33.3& \cellcolor[rgb]{ .98,  .984,  .957}26.7& \cellcolor[rgb]{ .98,  .984,  .957}33.3& \cellcolor[rgb]{ .98,  .984,  .957}\textbf{29.6} \\
\cmidrule{4-14}
          &       &       & GenHancer &6.7&13.3&66.7&6.7&26.7&66.7&33.3&6.7&26.7&28.2\\
          &       &       & +DCR  & 6.7   & 20.0  & 53.3  & 13.3  & 13.3  & 73.3  & 20.0  & 13.3  & 53.3  & 29.6  \\
          &       &       & \cellcolor[rgb]{ .98,  .984,  .957}+AUCGR & \cellcolor[rgb]{ .98,  .984,  .957}6.7& \cellcolor[rgb]{ .98,  .984,  .957}20.0& \cellcolor[rgb]{ .98,  .984,  .957}33.3& \cellcolor[rgb]{ .98,  .984,  .957}20.0& \cellcolor[rgb]{ .98,  .984,  .957}6.7& \cellcolor[rgb]{ .98,  .984,  .957}73.3& \cellcolor[rgb]{ .98,  .984,  .957}53.3& \cellcolor[rgb]{ .98,  .984,  .957}26.7& \cellcolor[rgb]{ .98,  .984,  .957}26.7& \cellcolor[rgb]{ .98,  .984,  .957}29.6\\
          &       &       & \cellcolor[rgb]{ .98,  .984,  .957}+pAUCGR & \cellcolor[rgb]{ .98,  .984,  .957}13.3& \cellcolor[rgb]{ .98,  .984,  .957}20.0& \cellcolor[rgb]{ .98,  .984,  .957}53.3& \cellcolor[rgb]{ .98,  .984,  .957}13.3& \cellcolor[rgb]{ .98,  .984,  .957}26.7& \cellcolor[rgb]{ .98,  .984,  .957}80.0& \cellcolor[rgb]{ .98,  .984,  .957}33.3& \cellcolor[rgb]{ .98,  .984,  .957}13.3& \cellcolor[rgb]{ .98,  .984,  .957}33.3& \cellcolor[rgb]{ .98,  .984,  .957}\textbf{31.8} \\
    \midrule
    \midrule
    \multirow{9}[8]{*}{MetaCLIP ViT-H-14} & \multirow{9}[8]{*}{224} & \multirow{9}[8]{*}{986.1} & Original &6.7&13.3&60.0&13.3&6.7&53.3&26.7&13.3&33.3&25.2\\
\cmidrule{4-14}
          &       &       & DIVA  &13.3&20.0&60.0&26.7&20.0&26.7&13.3&20.0&40.0&26.7\\
          &       &       & +DCR  & 13.3  & 6.7   & 53.3  & 26.7  & 20.0  & 53.3  & 33.3  & 13.3  & 33.3  & 28.2  \\
          &       &       & \cellcolor[rgb]{ .98,  .984,  .957}+AUCGR & \cellcolor[rgb]{ .98,  .984,  .957}6.7& \cellcolor[rgb]{ .98,  .984,  .957}6.7& \cellcolor[rgb]{ .98,  .984,  .957}53.3& \cellcolor[rgb]{ .98,  .984,  .957}20.0& \cellcolor[rgb]{ .98,  .984,  .957}26.7& \cellcolor[rgb]{ .98,  .984,  .957}53.3& \cellcolor[rgb]{ .98,  .984,  .957}33.3& \cellcolor[rgb]{ .98,  .984,  .957}13.3& \cellcolor[rgb]{ .98,  .984,  .957}33.3& \cellcolor[rgb]{ .98,  .984,  .957}27.4\\
          &       &       & \cellcolor[rgb]{ .98,  .984,  .957}+pAUCGR & \cellcolor[rgb]{ .98,  .984,  .957}20.0& \cellcolor[rgb]{ .98,  .984,  .957}13.3& \cellcolor[rgb]{ .98,  .984,  .957}53.3& \cellcolor[rgb]{ .98,  .984,  .957}20.0& \cellcolor[rgb]{ .98,  .984,  .957}33.3& \cellcolor[rgb]{ .98,  .984,  .957}53.3& \cellcolor[rgb]{ .98,  .984,  .957}33.3& \cellcolor[rgb]{ .98,  .984,  .957}13.3& \cellcolor[rgb]{ .98,  .984,  .957}33.3& \cellcolor[rgb]{ .98,  .984,  .957}\textbf{30.4} \\
\cmidrule{4-14}
          &       &       & GenHancer &13.3&13.3&53.3&26.7&33.3&66.7&33.3&13.3&40.0&32.6\\
          &       &       & +DCR  & 6.7   & 13.3  & 60.0  & 20.0  & 20.0  & 73.3  & 46.7  & 6.7   & 53.3  & 33.3  \\
          &       &       & \cellcolor[rgb]{ .98,  .984,  .957}+AUCGR & \cellcolor[rgb]{ .98,  .984,  .957}13.3& \cellcolor[rgb]{ .98,  .984,  .957}13.3& \cellcolor[rgb]{ .98,  .984,  .957}60.0& \cellcolor[rgb]{ .98,  .984,  .957}26.7& \cellcolor[rgb]{ .98,  .984,  .957}20.0& \cellcolor[rgb]{ .98,  .984,  .957}66.7& \cellcolor[rgb]{ .98,  .984,  .957}40.0& \cellcolor[rgb]{ .98,  .984,  .957}13.3& \cellcolor[rgb]{ .98,  .984,  .957}53.3& \cellcolor[rgb]{ .98,  .984,  .957}\textbf{34.1} \\
          &       &       & \cellcolor[rgb]{ .98,  .984,  .957}+pAUCGR & \cellcolor[rgb]{ .98,  .984,  .957}13.3& \cellcolor[rgb]{ .98,  .984,  .957}13.3& \cellcolor[rgb]{ .98,  .984,  .957}60.0& \cellcolor[rgb]{ .98,  .984,  .957}26.7& \cellcolor[rgb]{ .98,  .984,  .957}20.0& \cellcolor[rgb]{ .98,  .984,  .957}66.7& \cellcolor[rgb]{ .98,  .984,  .957}40.0& \cellcolor[rgb]{ .98,  .984,  .957}13.3& \cellcolor[rgb]{ .98,  .984,  .957}53.3& \cellcolor[rgb]{ .98,  .984,  .957}\textbf{34.1} \\
    \midrule
    \midrule
    \multirow{9}[8]{*}{SigLIP ViT-SO-14} & \multirow{9}[8]{*}{224} & \multirow{9}[8]{*}{877.4} & Original &26.7&20.0&53.3&40.0&20.0&66.7&40.0&20.0&53.3&37.8\\
\cmidrule{4-14}
          &       &       & DIVA  &13.3&33.3&66.7&33.3&13.3&66.7&46.7&26.7&53.3&39.3\\
          &       &       & +DCR  & 26.7  & 26.7  & 60.0  & 40.0  & 13.3  & 66.7  & 46.7  & 20.0  & 66.7  & 40.7  \\
          &       &       & \cellcolor[rgb]{ .996,  .969,  .945}+AUCGR & \cellcolor[rgb]{ .996,  .969,  .945}6.7& \cellcolor[rgb]{ .996,  .969,  .945}26.7& \cellcolor[rgb]{ .996,  .969,  .945}53.3& \cellcolor[rgb]{ .996,  .969,  .945}40.0& \cellcolor[rgb]{ .996,  .969,  .945}13.3& \cellcolor[rgb]{ .996,  .969,  .945}66.7& \cellcolor[rgb]{ .996,  .969,  .945}60.0& \cellcolor[rgb]{ .996,  .969,  .945}33.3& \cellcolor[rgb]{ .996,  .969,  .945}60.0& \cellcolor[rgb]{ .996,  .969,  .945}40.0\\
          &       &       & \cellcolor[rgb]{ .996,  .969,  .945}+pAUCGR & \cellcolor[rgb]{ .996,  .969,  .945}26.7& \cellcolor[rgb]{ .996,  .969,  .945}33.3& \cellcolor[rgb]{ .996,  .969,  .945}60.0& \cellcolor[rgb]{ .996,  .969,  .945}40.0& \cellcolor[rgb]{ .996,  .969,  .945}20.0& \cellcolor[rgb]{ .996,  .969,  .945}66.7& \cellcolor[rgb]{ .996,  .969,  .945}40.0& \cellcolor[rgb]{ .996,  .969,  .945}33.3& \cellcolor[rgb]{ .996,  .969,  .945}53.3& \cellcolor[rgb]{ .996,  .969,  .945}\textbf{41.5} \\
\cmidrule{4-14}
          &       &       & GenHancer &20.0&26.7&60.0&40.0&13.3&73.3&40.0&26.7&60.0&40.0\\
          &       &       & +DCR  & 26.7  & 20.0  & 53.3  & 40.0  & 20.0  & 73.3  & 40.0  & 26.7  & 60.0  & 40.0  \\
          &       &       & \cellcolor[rgb]{ .996,  .969,  .945}+AUCGR & \cellcolor[rgb]{ .996,  .969,  .945}26.7& \cellcolor[rgb]{ .996,  .969,  .945}20.0& \cellcolor[rgb]{ .996,  .969,  .945}53.3& \cellcolor[rgb]{ .996,  .969,  .945}40.0& \cellcolor[rgb]{ .996,  .969,  .945}20.0& \cellcolor[rgb]{ .996,  .969,  .945}73.3& \cellcolor[rgb]{ .996,  .969,  .945}46.7& \cellcolor[rgb]{ .996,  .969,  .945}26.7& \cellcolor[rgb]{ .996,  .969,  .945}60.0& \cellcolor[rgb]{ .996,  .969,  .945}40.7\\
          &       &       & \cellcolor[rgb]{ .996,  .969,  .945}+pAUCGR & \cellcolor[rgb]{ .996,  .969,  .945}26.7& \cellcolor[rgb]{ .996,  .969,  .945}26.7& \cellcolor[rgb]{ .996,  .969,  .945}53.3& \cellcolor[rgb]{ .996,  .969,  .945}40.0& \cellcolor[rgb]{ .996,  .969,  .945}20.0& \cellcolor[rgb]{ .996,  .969,  .945}73.3& \cellcolor[rgb]{ .996,  .969,  .945}46.7& \cellcolor[rgb]{ .996,  .969,  .945}26.7& \cellcolor[rgb]{ .996,  .969,  .945}60.0& \cellcolor[rgb]{ .996,  .969,  .945}\textbf{41.5} \\
    \midrule
    \midrule
    \multirow{9}[8]{*}{SigLIP ViT-SO-14} & \multirow{9}[8]{*}{384} & \multirow{9}[8]{*}{878.0} & Original &20.0&26.7&60.0&33.3&13.3&66.7&33.3&26.7&53.3&37.0\\
\cmidrule{4-14}
          &       &       & DIVA  &20.0&46.7&53.3&20.0&20.0&80.0&33.3&20.0&53.3&38.5\\
          &       &       & +DCR  & 20.0  & 26.7  & 60.0  & 40.0  & 6.7   & 66.7  & 40.0  & 40.0  & 53.3  & \textbf{39.3} \\
          &       &       & \cellcolor[rgb]{ .996,  .969,  .945}+AUCGR & \cellcolor[rgb]{ .996,  .969,  .945}20.0& \cellcolor[rgb]{ .996,  .969,  .945}26.7& \cellcolor[rgb]{ .996,  .969,  .945}60.0& \cellcolor[rgb]{ .996,  .969,  .945}40.0& \cellcolor[rgb]{ .996,  .969,  .945}6.7& \cellcolor[rgb]{ .996,  .969,  .945}66.7& \cellcolor[rgb]{ .996,  .969,  .945}40.0& \cellcolor[rgb]{ .996,  .969,  .945}33.3& \cellcolor[rgb]{ .996,  .969,  .945}60.0& \cellcolor[rgb]{ .996,  .969,  .945}\textbf{39.3} \\
          &       &       & \cellcolor[rgb]{ .996,  .969,  .945}+pAUCGR & \cellcolor[rgb]{ .996,  .969,  .945}20.0& \cellcolor[rgb]{ .996,  .969,  .945}20.0& \cellcolor[rgb]{ .996,  .969,  .945}60.0& \cellcolor[rgb]{ .996,  .969,  .945}40.0& \cellcolor[rgb]{ .996,  .969,  .945}13.3& \cellcolor[rgb]{ .996,  .969,  .945}66.7& \cellcolor[rgb]{ .996,  .969,  .945}40.0& \cellcolor[rgb]{ .996,  .969,  .945}33.3& \cellcolor[rgb]{ .996,  .969,  .945}60.0& \cellcolor[rgb]{ .996,  .969,  .945}\textbf{39.3} \\
\cmidrule{4-14}
          &       &       & GenHancer &33.3&20.0&53.3&26.7&13.3&80.0&40.0&13.3&60.0&37.8\\
          &       &       & +DCR  & 20.0  & 20.0  & 60.0  & 40.0  & 13.3  & 66.7  & 33.3  & 33.3  & 60.0  & \textbf{38.5}  \\
          &       &       & \cellcolor[rgb]{ .996,  .969,  .945}+AUCGR & \cellcolor[rgb]{ .996,  .969,  .945}20.0& \cellcolor[rgb]{ .996,  .969,  .945}20.0& \cellcolor[rgb]{ .996,  .969,  .945}60.0& \cellcolor[rgb]{ .996,  .969,  .945}40.0& \cellcolor[rgb]{ .996,  .969,  .945}13.3& \cellcolor[rgb]{ .996,  .969,  .945}66.7& \cellcolor[rgb]{ .996,  .969,  .945}33.3& \cellcolor[rgb]{ .996,  .969,  .945}33.3& \cellcolor[rgb]{ .996,  .969,  .945}53.3& \cellcolor[rgb]{ .996,  .969,  .945}37.8\\
          &       &       & \cellcolor[rgb]{ .996,  .969,  .945}+pAUCGR & \cellcolor[rgb]{ .996,  .969,  .945}20.0& \cellcolor[rgb]{ .996,  .969,  .945}26.7& \cellcolor[rgb]{ .996,  .969,  .945}60.0& \cellcolor[rgb]{ .996,  .969,  .945}40.0& \cellcolor[rgb]{ .996,  .969,  .945}13.3& \cellcolor[rgb]{ .996,  .969,  .945}66.7& \cellcolor[rgb]{ .996,  .969,  .945}33.3& \cellcolor[rgb]{ .996,  .969,  .945}33.3& \cellcolor[rgb]{ .996,  .969,  .945}53.3& \cellcolor[rgb]{ .996,  .969,  .945}\textbf{38.5} \\
    \bottomrule
    \end{tabular}%
}
\label{tab: understanding ability results}
\vspace{-10pt}
\end{table*}

In this section, we describe some details of the experiments and present our results. \textbf{Due to space limitations, please refer to \cref{sec: supp_Additional Experiment Settings} and \cref{sec: supp_Additional Experiment Results} for the extended version.}

\subsection{Experimental Setups}
\label{subsec: Experimental Setups}

\begin{table*}[t!]

  \centering
  \renewcommand\arraystretch{1.2}
  \setlength{\tabcolsep}{3pt}
  \caption{\textbf{Performance of \textcolor{DisColor}{Discriminative Ability (D-Ability)} on 26 standard zero-shot clustering benchmarks in terms of \underline{ACC}}. `+DCR' denotes our conference version. O-1, M-1, and S-1 separately represent OpenAI CLIP ViT-L@224, MetaCLIP ViT-L@224, and SigLIP ViT-SO@224. Results in terms of ARI and NMI are provided in \cref{tab: discriminative ability results ARI} and \cref{tab: discriminative ability results NMI}.}
  \resizebox{\linewidth}{!}{
    \begin{tabular}{cc|cccccccccccccccccccccccccc|c}
    \toprule
    \multicolumn{1}{c}{\rotatebox{90}{\textbf{Backbone}}} & \multicolumn{1}{c|}{\rotatebox{90}{\textbf{Method}}} & \rotatebox{90}{\textbf{CIFAR-10}~\cite{krizhevsky2009learning}} & \rotatebox{90}{\textbf{CIFAR-100}~\cite{krizhevsky2009learning}} & \rotatebox{90}{\textbf{MNIST}~\cite{lecun2002gradient}} & \rotatebox{90}{\textbf{Caltech-101}~\cite{fei2004learning}} & \rotatebox{90}{\textbf{SUN397}~\cite{xiao2010sun}} & \rotatebox{90}{\textbf{FGVC Aircraft}~\cite{maji2013fine}} & \rotatebox{90}{\textbf{Country-211}~\cite{radford2021learning}} & \rotatebox{90}{\textbf{Stanford Cars}~\cite{krause20133d}} & \rotatebox{90}{\textbf{DTD}~\cite{cimpoi2014describing}} & \rotatebox{90}{\textbf{Eurosat}~\cite{helber2019eurosat}} & \rotatebox{90}{\textbf{FER2013}~\cite{goodfellow2013challenges}} & \rotatebox{90}{\textbf{Flowers-102}~\cite{nilsback2008automated}} & \rotatebox{90}{\textbf{Food-101}~\cite{bossard2014food}} & \rotatebox{90}{\textbf{GTSRB}~\cite{stallkamp2012man}} & \rotatebox{90}{\textbf{PCam}~\cite{veeling2018rotation}} & \rotatebox{90}{\textbf{Pets}~\cite{parkhi2012cats}} & \rotatebox{90}{\textbf{Rendered SST2}~\cite{radford2021learning}} & \rotatebox{90}{\textbf{Resisc45}~\cite{cheng2017remote}} & \rotatebox{90}{\textbf{STL10}~\cite{coates2011analysis}} & \rotatebox{90}{\textbf{VOC2007}~\cite{everingham2010pascal}} & \rotatebox{90}{\textbf{ImageNet-1K}~\cite{deng2009imagenet}} & \rotatebox{90}{\textbf{ImageNet-V2}~\cite{recht2019imagenet}} & \rotatebox{90}{\textbf{ImageNet-Adv.}~\cite{hendrycks2021natural}} & \rotatebox{90}{\textbf{ImageNet-Ren.}~\cite{hendrycks2021many}} & \rotatebox{90}{\textbf{ImageNet-OOD}~\cite{hendrycks2021natural}} & \rotatebox{90}{\textbf{ObjectNet}~\cite{barbu2019objectnet}} & \rotatebox{90}{\textbf{Avg. ACC}} \\
    \midrule
    \multirow{9}[6]{*}{O-1} & Original & 83.57  & 51.32  & 67.95  & 49.63  & 50.03  & 35.70  & 13.19  & 54.94  & 49.41  & 69.30  & 25.62  & 82.14  & 85.01  & 46.69  & 62.92  & 64.90  & 52.22  & 70.71  & 95.74  & 41.03  & 53.92  & 42.39  & 35.77  & 38.84  & 27.85  & 44.24  & 53.66  \\
\cmidrule{2-29}          & DIVA  & 80.57  & 41.80  & 62.71  & 45.57  & 47.21  & 22.77  & 7.83  & 20.07  & 48.46  & 65.02  & 28.45  & 68.58  & 68.82  & 36.64  & 65.03  & 41.37  & 52.11  & 67.56  & 97.83  & 37.47  & 36.80  & 31.38  & 17.31  & 17.29  & 28.70  & 26.20  & 44.75  \\
          & +DCR  & 75.47  & 42.18  & 58.57  & 42.94  & 49.80  & 30.60  & 12.05  & 41.43  & 42.13  & 53.22  & 28.53  & 74.96  & 77.55  & 35.31  & 61.42  & 51.05  & 50.69  & 68.71  & 95.25  & 39.77  & 45.35  & 35.85  & 32.47  & 14.01  & 26.90  & 41.45  & 47.22  \\
          & \cellcolor[rgb]{ .945,  .973,  .98}+AUCGR & \cellcolor[rgb]{ .945,  .973,  .98}86.14  & \cellcolor[rgb]{ .945,  .973,  .98}43.49  & \cellcolor[rgb]{ .945,  .973,  .98}65.82  & \cellcolor[rgb]{ .945,  .973,  .98}47.05  & \cellcolor[rgb]{ .945,  .973,  .98}48.14  & \cellcolor[rgb]{ .945,  .973,  .98}28.11  & \cellcolor[rgb]{ .945,  .973,  .98}10.53  & \cellcolor[rgb]{ .945,  .973,  .98}37.68  & \cellcolor[rgb]{ .945,  .973,  .98}48.99  & \cellcolor[rgb]{ .945,  .973,  .98}77.93  & \cellcolor[rgb]{ .945,  .973,  .98}26.02  & \cellcolor[rgb]{ .945,  .973,  .98}77.04  & \cellcolor[rgb]{ .945,  .973,  .98}76.32  & \cellcolor[rgb]{ .945,  .973,  .98}43.59  & \cellcolor[rgb]{ .945,  .973,  .98}64.87  & \cellcolor[rgb]{ .945,  .973,  .98}49.74  & \cellcolor[rgb]{ .945,  .973,  .98}54.20  & \cellcolor[rgb]{ .945,  .973,  .98}75.38  & \cellcolor[rgb]{ .945,  .973,  .98}94.39  & \cellcolor[rgb]{ .945,  .973,  .98}39.82  & \cellcolor[rgb]{ .945,  .973,  .98}43.71  & \cellcolor[rgb]{ .945,  .973,  .98}36.43  & \cellcolor[rgb]{ .945,  .973,  .98}25.37  & \cellcolor[rgb]{ .945,  .973,  .98}21.12  & \cellcolor[rgb]{ .945,  .973,  .98}27.65  & \cellcolor[rgb]{ .945,  .973,  .98}39.77  & \cellcolor[rgb]{ .945,  .973,  .98}49.59  \\
          & \cellcolor[rgb]{ .945,  .973,  .98}+pAUCGR & \cellcolor[rgb]{ .945,  .973,  .98}83.47  & \cellcolor[rgb]{ .945,  .973,  .98}51.72  & \cellcolor[rgb]{ .945,  .973,  .98}66.43  & \cellcolor[rgb]{ .945,  .973,  .98}49.88  & \cellcolor[rgb]{ .945,  .973,  .98}50.71  & \cellcolor[rgb]{ .945,  .973,  .98}34.56  & \cellcolor[rgb]{ .945,  .973,  .98}12.78  & \cellcolor[rgb]{ .945,  .973,  .98}55.74  & \cellcolor[rgb]{ .945,  .973,  .98}51.81  & \cellcolor[rgb]{ .945,  .973,  .98}69.57  & \cellcolor[rgb]{ .945,  .973,  .98}25.61  & \cellcolor[rgb]{ .945,  .973,  .98}78.89  & \cellcolor[rgb]{ .945,  .973,  .98}83.98  & \cellcolor[rgb]{ .945,  .973,  .98}47.20  & \cellcolor[rgb]{ .945,  .973,  .98}62.69  & \cellcolor[rgb]{ .945,  .973,  .98}61.41  & \cellcolor[rgb]{ .945,  .973,  .98}52.11  & \cellcolor[rgb]{ .945,  .973,  .98}73.71  & \cellcolor[rgb]{ .945,  .973,  .98}95.76  & \cellcolor[rgb]{ .945,  .973,  .98}46.65  & \cellcolor[rgb]{ .945,  .973,  .98}53.89  & \cellcolor[rgb]{ .945,  .973,  .98}41.57  & \cellcolor[rgb]{ .945,  .973,  .98}37.97  & \cellcolor[rgb]{ .945,  .973,  .98}39.55  & \cellcolor[rgb]{ .945,  .973,  .98}27.80  & \cellcolor[rgb]{ .945,  .973,  .98}45.46  & \cellcolor[rgb]{ .945,  .973,  .98}\textbf{53.88} \\
\cmidrule{2-29}          & GenHancer & 44.25  & 19.39  & 60.34  & 33.54  & 30.20  & 14.46  & 7.17  & 12.29  & 30.96  & 65.44  & 30.59  & 36.67  & 32.04  & 26.88  & 64.03  & 21.10  & 51.78  & 56.13  & 55.83  & 26.70  & 18.20  & 20.87  & 11.31  & 7.43  & 24.00  & 7.85  & 31.13  \\
          & +DCR  & 72.29  & 35.46  & 59.99  & 43.32  & 42.02  & 16.50  & 7.09  & 12.22  & 44.15  & 58.33  & 28.34  & 57.60  & 51.35  & 30.57  & 64.87  & 24.78  & 52.33  & 58.68  & 84.64  & 39.37  & 27.47  & 24.99  & 14.21  & 11.41  & 28.65  & 18.19  & 38.80  \\
          & \cellcolor[rgb]{ .945,  .973,  .98}+AUCGR & \cellcolor[rgb]{ .945,  .973,  .98}76.05  & \cellcolor[rgb]{ .945,  .973,  .98}34.91  & \cellcolor[rgb]{ .945,  .973,  .98}72.69  & \cellcolor[rgb]{ .945,  .973,  .98}44.85  & \cellcolor[rgb]{ .945,  .973,  .98}43.49  & \cellcolor[rgb]{ .945,  .973,  .98}16.44  & \cellcolor[rgb]{ .945,  .973,  .98}8.42  & \cellcolor[rgb]{ .945,  .973,  .98}20.27  & \cellcolor[rgb]{ .945,  .973,  .98}46.12  & \cellcolor[rgb]{ .945,  .973,  .98}70.80  & \cellcolor[rgb]{ .945,  .973,  .98}30.61  & \cellcolor[rgb]{ .945,  .973,  .98}54.87  & \cellcolor[rgb]{ .945,  .973,  .98}28.16  & \cellcolor[rgb]{ .945,  .973,  .98}36.05  & \cellcolor[rgb]{ .945,  .973,  .98}62.64  & \cellcolor[rgb]{ .945,  .973,  .98}42.14  & \cellcolor[rgb]{ .945,  .973,  .98}53.05  & \cellcolor[rgb]{ .945,  .973,  .98}49.79  & \cellcolor[rgb]{ .945,  .973,  .98}91.19  & \cellcolor[rgb]{ .945,  .973,  .98}31.94  & \cellcolor[rgb]{ .945,  .973,  .98}29.21  & \cellcolor[rgb]{ .945,  .973,  .98}24.62  & \cellcolor[rgb]{ .945,  .973,  .98}13.19  & \cellcolor[rgb]{ .945,  .973,  .98}15.76  & \cellcolor[rgb]{ .945,  .973,  .98}30.75  & \cellcolor[rgb]{ .945,  .973,  .98}12.29  & \cellcolor[rgb]{ .945,  .973,  .98}40.01  \\
          & \cellcolor[rgb]{ .945,  .973,  .98}+pAUCGR & \cellcolor[rgb]{ .945,  .973,  .98}73.99  & \cellcolor[rgb]{ .945,  .973,  .98}36.62  & \cellcolor[rgb]{ .945,  .973,  .98}58.50  & \cellcolor[rgb]{ .945,  .973,  .98}42.04  & \cellcolor[rgb]{ .945,  .973,  .98}45.50  & \cellcolor[rgb]{ .945,  .973,  .98}17.16  & \cellcolor[rgb]{ .945,  .973,  .98}7.32  & \cellcolor[rgb]{ .945,  .973,  .98}13.49  & \cellcolor[rgb]{ .945,  .973,  .98}44.63  & \cellcolor[rgb]{ .945,  .973,  .98}61.70  & \cellcolor[rgb]{ .945,  .973,  .98}27.49  & \cellcolor[rgb]{ .945,  .973,  .98}61.60  & \cellcolor[rgb]{ .945,  .973,  .98}56.87  & \cellcolor[rgb]{ .945,  .973,  .98}33.91  & \cellcolor[rgb]{ .945,  .973,  .98}65.00  & \cellcolor[rgb]{ .945,  .973,  .98}28.02  & \cellcolor[rgb]{ .945,  .973,  .98}52.33  & \cellcolor[rgb]{ .945,  .973,  .98}62.33  & \cellcolor[rgb]{ .945,  .973,  .98}86.45  & \cellcolor[rgb]{ .945,  .973,  .98}37.20  & \cellcolor[rgb]{ .945,  .973,  .98}30.51  & \cellcolor[rgb]{ .945,  .973,  .98}27.08  & \cellcolor[rgb]{ .945,  .973,  .98}15.12  & \cellcolor[rgb]{ .945,  .973,  .98}13.09  & \cellcolor[rgb]{ .945,  .973,  .98}28.05  & \cellcolor[rgb]{ .945,  .973,  .98}20.54  & \cellcolor[rgb]{ .945,  .973,  .98}\textbf{40.25} \\
    \midrule
    \midrule
    \multirow{9}[6]{*}{M-1} & Original & 88.06  & 53.03  & 67.24  & 50.55  & 51.15  & 45.36  & 13.96  & 61.39  & 49.31  & 66.52  & 27.82  & 83.02  & 81.09  & 44.67  & 62.02  & 69.42  & 55.08  & 68.52  & 95.69  & 36.83  & 56.53  & 43.51  & 37.07  & 35.44  & 26.65  & 50.87  & 54.65  \\
\cmidrule{2-29}          & DIVA  & 70.56  & 44.00  & 71.82  & 47.00  & 43.33  & 21.72  & 8.30  & 31.28  & 43.46  & 59.94  & 25.61  & 72.32  & 72.15  & 33.03  & 64.67  & 40.58  & 53.98  & 68.83  & 95.01  & 33.17  & 39.05  & 32.88  & 19.69  & 15.99  & 27.50  & 39.35  & 45.20  \\
          & +DCR  & 85.02  & 47.18  & 67.31  & 49.78  & 43.34  & 27.60  & 10.25  & 33.08  & 46.97  & 66.31  & 26.64  & 76.22  & 74.58  & 33.92  & 62.64  & 40.88  & 51.84  & 64.06  & 95.45  & 34.05  & 42.70  & 35.43  & 26.57  & 16.00  & 25.25  & 44.30  & 47.21  \\
          & \cellcolor[rgb]{ .98,  .984,  .957}+AUCGR & \cellcolor[rgb]{ .98,  .984,  .957}84.30  & \cellcolor[rgb]{ .98,  .984,  .957}48.01  & \cellcolor[rgb]{ .98,  .984,  .957}68.04  & \cellcolor[rgb]{ .98,  .984,  .957}48.28  & \cellcolor[rgb]{ .98,  .984,  .957}43.88  & \cellcolor[rgb]{ .98,  .984,  .957}31.83  & \cellcolor[rgb]{ .98,  .984,  .957}10.76  & \cellcolor[rgb]{ .98,  .984,  .957}37.94  & \cellcolor[rgb]{ .98,  .984,  .957}47.98  & \cellcolor[rgb]{ .98,  .984,  .957}63.06  & \cellcolor[rgb]{ .98,  .984,  .957}26.83  & \cellcolor[rgb]{ .98,  .984,  .957}76.87  & \cellcolor[rgb]{ .98,  .984,  .957}73.66  & \cellcolor[rgb]{ .98,  .984,  .957}40.57  & \cellcolor[rgb]{ .98,  .984,  .957}61.22  & \cellcolor[rgb]{ .98,  .984,  .957}45.00  & \cellcolor[rgb]{ .98,  .984,  .957}52.72  & \cellcolor[rgb]{ .98,  .984,  .957}63.14  & \cellcolor[rgb]{ .98,  .984,  .957}95.59  & \cellcolor[rgb]{ .98,  .984,  .957}28.71  & \cellcolor[rgb]{ .98,  .984,  .957}46.27  & \cellcolor[rgb]{ .98,  .984,  .957}38.29  & \cellcolor[rgb]{ .98,  .984,  .957}29.59  & \cellcolor[rgb]{ .98,  .984,  .957}18.34  & \cellcolor[rgb]{ .98,  .984,  .957}25.75  & \cellcolor[rgb]{ .98,  .984,  .957}49.50  & \cellcolor[rgb]{ .98,  .984,  .957}48.31  \\
          & \cellcolor[rgb]{ .98,  .984,  .957}+pAUCGR & \cellcolor[rgb]{ .98,  .984,  .957}88.39  & \cellcolor[rgb]{ .98,  .984,  .957}55.63  & \cellcolor[rgb]{ .98,  .984,  .957}67.17  & \cellcolor[rgb]{ .98,  .984,  .957}48.58  & \cellcolor[rgb]{ .98,  .984,  .957}50.59  & \cellcolor[rgb]{ .98,  .984,  .957}42.90  & \cellcolor[rgb]{ .98,  .984,  .957}13.62  & \cellcolor[rgb]{ .98,  .984,  .957}57.22  & \cellcolor[rgb]{ .98,  .984,  .957}49.84  & \cellcolor[rgb]{ .98,  .984,  .957}68.83  & \cellcolor[rgb]{ .98,  .984,  .957}27.50  & \cellcolor[rgb]{ .98,  .984,  .957}79.22  & \cellcolor[rgb]{ .98,  .984,  .957}82.45  & \cellcolor[rgb]{ .98,  .984,  .957}42.00  & \cellcolor[rgb]{ .98,  .984,  .957}62.00  & \cellcolor[rgb]{ .98,  .984,  .957}78.11  & \cellcolor[rgb]{ .98,  .984,  .957}52.83  & \cellcolor[rgb]{ .98,  .984,  .957}69.98  & \cellcolor[rgb]{ .98,  .984,  .957}85.66  & \cellcolor[rgb]{ .98,  .984,  .957}33.73  & \cellcolor[rgb]{ .98,  .984,  .957}56.30  & \cellcolor[rgb]{ .98,  .984,  .957}43.84  & \cellcolor[rgb]{ .98,  .984,  .957}37.24  & \cellcolor[rgb]{ .98,  .984,  .957}34.39  & \cellcolor[rgb]{ .98,  .984,  .957}26.05  & \cellcolor[rgb]{ .98,  .984,  .957}50.67  & \cellcolor[rgb]{ .98,  .984,  .957}\textbf{54.03} \\
\cmidrule{2-29}          & GenHancer & 58.07  & 30.95  & 64.67  & 36.25  & 21.46  & 11.07  & 6.09  & 8.06  & 35.85  & 70.15  & 27.29  & 49.54  & 12.95  & 27.02  & 62.68  & 19.46  & 51.51  & 41.62  & 54.71  & 22.72  & 14.04  & 17.36  & 9.89  & 6.77  & 24.70  & 5.13  & 30.39  \\
          & +DCR  & 82.54  & 50.12  & 63.23  & 47.76  & 41.03  & 27.24  & 10.35  & 35.12  & 48.09  & 64.37  & 27.90  & 60.01  & 61.50  & 32.01  & 62.38  & 37.04  & 52.72  & 67.16  & 84.48  & 27.18  & 40.93  & 32.01  & 18.89  & 13.15  & 26.05  & 39.59  & 44.34  \\
          & \cellcolor[rgb]{ .98,  .984,  .957}+AUCGR & \cellcolor[rgb]{ .98,  .984,  .957}87.62  & \cellcolor[rgb]{ .98,  .984,  .957}49.70  & \cellcolor[rgb]{ .98,  .984,  .957}66.05  & \cellcolor[rgb]{ .98,  .984,  .957}47.44  & \cellcolor[rgb]{ .98,  .984,  .957}39.10  & \cellcolor[rgb]{ .98,  .984,  .957}27.03  & \cellcolor[rgb]{ .98,  .984,  .957}9.93  & \cellcolor[rgb]{ .98,  .984,  .957}33.14  & \cellcolor[rgb]{ .98,  .984,  .957}46.65  & \cellcolor[rgb]{ .98,  .984,  .957}63.43  & \cellcolor[rgb]{ .98,  .984,  .957}27.89  & \cellcolor[rgb]{ .98,  .984,  .957}60.71  & \cellcolor[rgb]{ .98,  .984,  .957}56.05  & \cellcolor[rgb]{ .98,  .984,  .957}31.11  & \cellcolor[rgb]{ .98,  .984,  .957}62.73  & \cellcolor[rgb]{ .98,  .984,  .957}38.08  & \cellcolor[rgb]{ .98,  .984,  .957}53.05  & \cellcolor[rgb]{ .98,  .984,  .957}71.84  & \cellcolor[rgb]{ .98,  .984,  .957}82.45  & \cellcolor[rgb]{ .98,  .984,  .957}39.40  & \cellcolor[rgb]{ .98,  .984,  .957}39.00  & \cellcolor[rgb]{ .98,  .984,  .957}31.97  & \cellcolor[rgb]{ .98,  .984,  .957}19.55  & \cellcolor[rgb]{ .98,  .984,  .957}12.73  & \cellcolor[rgb]{ .98,  .984,  .957}27.75  & \cellcolor[rgb]{ .98,  .984,  .957}35.36  & \cellcolor[rgb]{ .98,  .984,  .957}44.61  \\
          & \cellcolor[rgb]{ .98,  .984,  .957}+pAUCGR & \cellcolor[rgb]{ .98,  .984,  .957}87.98  & \cellcolor[rgb]{ .98,  .984,  .957}52.40  & \cellcolor[rgb]{ .98,  .984,  .957}63.69  & \cellcolor[rgb]{ .98,  .984,  .957}54.46  & \cellcolor[rgb]{ .98,  .984,  .957}48.77  & \cellcolor[rgb]{ .98,  .984,  .957}42.33  & \cellcolor[rgb]{ .98,  .984,  .957}13.22  & \cellcolor[rgb]{ .98,  .984,  .957}59.27  & \cellcolor[rgb]{ .98,  .984,  .957}51.17  & \cellcolor[rgb]{ .98,  .984,  .957}66.78  & \cellcolor[rgb]{ .98,  .984,  .957}26.58  & \cellcolor[rgb]{ .98,  .984,  .957}77.39  & \cellcolor[rgb]{ .98,  .984,  .957}82.81  & \cellcolor[rgb]{ .98,  .984,  .957}39.23  & \cellcolor[rgb]{ .98,  .984,  .957}61.96  & \cellcolor[rgb]{ .98,  .984,  .957}54.16  & \cellcolor[rgb]{ .98,  .984,  .957}53.87  & \cellcolor[rgb]{ .98,  .984,  .957}70.54  & \cellcolor[rgb]{ .98,  .984,  .957}95.73  & \cellcolor[rgb]{ .98,  .984,  .957}32.57  & \cellcolor[rgb]{ .98,  .984,  .957}53.20  & \cellcolor[rgb]{ .98,  .984,  .957}40.60  & \cellcolor[rgb]{ .98,  .984,  .957}34.71  & \cellcolor[rgb]{ .98,  .984,  .957}29.78  & \cellcolor[rgb]{ .98,  .984,  .957}26.80  & \cellcolor[rgb]{ .98,  .984,  .957}52.16  & \cellcolor[rgb]{ .98,  .984,  .957}\textbf{52.78} \\
    \midrule
    \midrule
    \multirow{9}[6]{*}{S-1} & Original & 93.79  & 66.83  & 75.77  & 47.54  & 54.92  & 59.50  & 12.92  & 82.25  & 54.36  & 67.78  & 33.80  & 85.27  & 84.16  & 52.62  & 57.66  & 82.77  & 53.27  & 74.63  & 94.99  & 48.48  & 62.48  & 48.36  & 38.72  & 39.89  & 26.95  & 45.95  & 59.45  \\
\cmidrule{2-29}          & DIVA  & 53.16  & 33.91  & 73.71  & 28.09  & 10.44  & 11.07  & 5.36  & 6.91  & 20.27  & 48.02  & 23.11  & 20.15  & 4.95  & 34.43  & 52.51  & 14.75  & 51.95  & 48.00  & 85.89  & 14.88  & 11.56  & 13.30  & 9.36  & 4.41  & 13.75  & 4.93  & 26.88  \\
          & +DCR  & 97.17  & 67.07  & 76.09  & 46.72  & 54.28  & 59.86  & 12.59  & 81.73  & 55.53  & 66.80  & 33.95  & 85.67  & 86.05  & 51.35  & 57.56  & 85.91  & 52.99  & 73.62  & 95.00  & 42.88  & 61.93  & 48.13  & 36.96  & 38.48  & 27.95  & 32.45  & 58.80  \\
          & \cellcolor[rgb]{ .996,  .969,  .945}+AUCGR & \cellcolor[rgb]{ .996,  .969,  .945}84.18  & \cellcolor[rgb]{ .996,  .969,  .945}60.05  & \cellcolor[rgb]{ .996,  .969,  .945}72.56  & \cellcolor[rgb]{ .996,  .969,  .945}47.26  & \cellcolor[rgb]{ .996,  .969,  .945}53.18  & \cellcolor[rgb]{ .996,  .969,  .945}56.92  & \cellcolor[rgb]{ .996,  .969,  .945}11.07  & \cellcolor[rgb]{ .996,  .969,  .945}79.68  & \cellcolor[rgb]{ .996,  .969,  .945}54.63  & \cellcolor[rgb]{ .996,  .969,  .945}60.96  & \cellcolor[rgb]{ .996,  .969,  .945}32.22  & \cellcolor[rgb]{ .996,  .969,  .945}82.06  & \cellcolor[rgb]{ .996,  .969,  .945}84.65  & \cellcolor[rgb]{ .996,  .969,  .945}34.20  & \cellcolor[rgb]{ .996,  .969,  .945}60.76  & \cellcolor[rgb]{ .996,  .969,  .945}81.22  & \cellcolor[rgb]{ .996,  .969,  .945}54.04  & \cellcolor[rgb]{ .996,  .969,  .945}73.43  & \cellcolor[rgb]{ .996,  .969,  .945}99.28  & \cellcolor[rgb]{ .996,  .969,  .945}41.76  & \cellcolor[rgb]{ .996,  .969,  .945}58.15  & \cellcolor[rgb]{ .996,  .969,  .945}44.08  & \cellcolor[rgb]{ .996,  .969,  .945}28.47  & \cellcolor[rgb]{ .996,  .969,  .945}31.43  & \cellcolor[rgb]{ .996,  .969,  .945}27.65  & \cellcolor[rgb]{ .996,  .969,  .945}37.08  & \cellcolor[rgb]{ .996,  .969,  .945}55.81  \\
          & \cellcolor[rgb]{ .996,  .969,  .945}+pAUCGR & \cellcolor[rgb]{ .996,  .969,  .945}97.11  & \cellcolor[rgb]{ .996,  .969,  .945}66.01  & \cellcolor[rgb]{ .996,  .969,  .945}76.41  & \cellcolor[rgb]{ .996,  .969,  .945}50.71  & \cellcolor[rgb]{ .996,  .969,  .945}54.18  & \cellcolor[rgb]{ .996,  .969,  .945}60.49  & \cellcolor[rgb]{ .996,  .969,  .945}12.68  & \cellcolor[rgb]{ .996,  .969,  .945}81.92  & \cellcolor[rgb]{ .996,  .969,  .945}53.88  & \cellcolor[rgb]{ .996,  .969,  .945}66.00  & \cellcolor[rgb]{ .996,  .969,  .945}33.52  & \cellcolor[rgb]{ .996,  .969,  .945}83.98  & \cellcolor[rgb]{ .996,  .969,  .945}86.93  & \cellcolor[rgb]{ .996,  .969,  .945}50.89  & \cellcolor[rgb]{ .996,  .969,  .945}57.56  & \cellcolor[rgb]{ .996,  .969,  .945}81.82  & \cellcolor[rgb]{ .996,  .969,  .945}53.10  & \cellcolor[rgb]{ .996,  .969,  .945}70.92  & \cellcolor[rgb]{ .996,  .969,  .945}94.98  & \cellcolor[rgb]{ .996,  .969,  .945}45.74  & \cellcolor[rgb]{ .996,  .969,  .945}61.65  & \cellcolor[rgb]{ .996,  .969,  .945}48.19  & \cellcolor[rgb]{ .996,  .969,  .945}37.09  & \cellcolor[rgb]{ .996,  .969,  .945}41.05  & \cellcolor[rgb]{ .996,  .969,  .945}27.50  & \cellcolor[rgb]{ .996,  .969,  .945}47.42  & \cellcolor[rgb]{ .996,  .969,  .945}\textbf{59.30} \\
\cmidrule{2-29}          & GenHancer & 97.06  & 67.13  & 76.97  & 48.18  & 54.78  & 61.09  & 12.38  & 84.21  & 53.03  & 66.35  & 34.24  & 82.84  & 85.05  & 52.60  & 57.91  & 80.54  & 52.11  & 71.03  & 95.13  & 46.39  & 62.15  & 47.96  & 34.47  & 36.62  & 27.40  & 44.07  & 58.91  \\
          & +DCR  & 96.96  & 65.76  & 73.18  & 49.01  & 53.14  & 60.94  & 12.64  & 81.76  & 52.82  & 68.87  & 28.73  & 86.24  & 88.23  & 49.48  & 57.15  & 81.38  & 54.20  & 78.56  & 94.88  & 43.69  & 60.95  & 46.34  & 36.63  & 38.58  & 28.15  & 46.12  & 59.01  \\
          & \cellcolor[rgb]{ .996,  .969,  .945}+AUCGR & \cellcolor[rgb]{ .996,  .969,  .945}97.20  & \cellcolor[rgb]{ .996,  .969,  .945}65.34  & \cellcolor[rgb]{ .996,  .969,  .945}73.42  & \cellcolor[rgb]{ .996,  .969,  .945}49.25  & \cellcolor[rgb]{ .996,  .969,  .945}54.39  & \cellcolor[rgb]{ .996,  .969,  .945}62.20  & \cellcolor[rgb]{ .996,  .969,  .945}12.77  & \cellcolor[rgb]{ .996,  .969,  .945}81.10  & \cellcolor[rgb]{ .996,  .969,  .945}55.53  & \cellcolor[rgb]{ .996,  .969,  .945}68.65  & \cellcolor[rgb]{ .996,  .969,  .945}33.71  & \cellcolor[rgb]{ .996,  .969,  .945}85.87  & \cellcolor[rgb]{ .996,  .969,  .945}85.16  & \cellcolor[rgb]{ .996,  .969,  .945}54.46  & \cellcolor[rgb]{ .996,  .969,  .945}57.64  & \cellcolor[rgb]{ .996,  .969,  .945}82.47  & \cellcolor[rgb]{ .996,  .969,  .945}53.10  & \cellcolor[rgb]{ .996,  .969,  .945}72.17  & \cellcolor[rgb]{ .996,  .969,  .945}84.06  & \cellcolor[rgb]{ .996,  .969,  .945}46.75  & \cellcolor[rgb]{ .996,  .969,  .945}62.32  & \cellcolor[rgb]{ .996,  .969,  .945}47.71  & \cellcolor[rgb]{ .996,  .969,  .945}36.08  & \cellcolor[rgb]{ .996,  .969,  .945}37.83  & \cellcolor[rgb]{ .996,  .969,  .945}27.55  & \cellcolor[rgb]{ .996,  .969,  .945}46.93  & \cellcolor[rgb]{ .996,  .969,  .945}58.99  \\
          & \cellcolor[rgb]{ .996,  .969,  .945}+pAUCGR & \cellcolor[rgb]{ .996,  .969,  .945}97.22  & \cellcolor[rgb]{ .996,  .969,  .945}65.34  & \cellcolor[rgb]{ .996,  .969,  .945}76.20  & \cellcolor[rgb]{ .996,  .969,  .945}49.73  & \cellcolor[rgb]{ .996,  .969,  .945}53.99  & \cellcolor[rgb]{ .996,  .969,  .945}61.75  & \cellcolor[rgb]{ .996,  .969,  .945}12.64  & \cellcolor[rgb]{ .996,  .969,  .945}79.31  & \cellcolor[rgb]{ .996,  .969,  .945}55.37  & \cellcolor[rgb]{ .996,  .969,  .945}67.50  & \cellcolor[rgb]{ .996,  .969,  .945}33.67  & \cellcolor[rgb]{ .996,  .969,  .945}83.07  & \cellcolor[rgb]{ .996,  .969,  .945}85.71  & \cellcolor[rgb]{ .996,  .969,  .945}51.84  & \cellcolor[rgb]{ .996,  .969,  .945}57.55  & \cellcolor[rgb]{ .996,  .969,  .945}81.77  & \cellcolor[rgb]{ .996,  .969,  .945}53.16  & \cellcolor[rgb]{ .996,  .969,  .945}71.14  & \cellcolor[rgb]{ .996,  .969,  .945}94.99  & \cellcolor[rgb]{ .996,  .969,  .945}43.68  & \cellcolor[rgb]{ .996,  .969,  .945}62.33  & \cellcolor[rgb]{ .996,  .969,  .945}48.32  & \cellcolor[rgb]{ .996,  .969,  .945}37.23  & \cellcolor[rgb]{ .996,  .969,  .945}39.98  & \cellcolor[rgb]{ .996,  .969,  .945}27.75  & \cellcolor[rgb]{ .996,  .969,  .945}47.38  & \cellcolor[rgb]{ .996,  .969,  .945}\textbf{59.18} \\
    \bottomrule
    \end{tabular}%
   }
\label{tab: discriminative ability results ACC}

\vspace{-10pt}
\end{table*}

\begin{figure*}[t]
  \centering
   \includegraphics[width=0.95\linewidth]{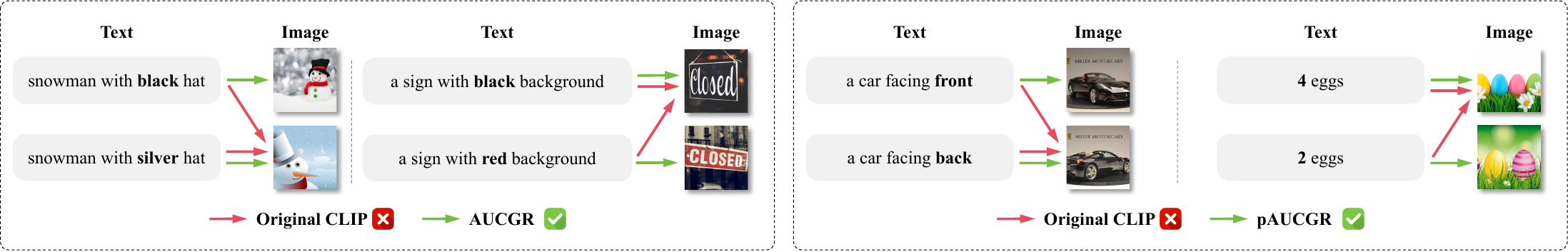}
      \caption{\textbf{Qualitative results of \textcolor{PerColor}{P-Ability} on the MMVP-VLM benchmark.} The predictions from the original CLIP and our improved version are indicated by red and green arrows, respectively. The improved CLIP effectively addresses the original model's limitations in capturing fine-grained visual details. More images are provided in \cref{fig: expanded_mmvp_vlm} and \cref{fig: expanded_mmvp_vlm2}.}
      \label{fig: mmvp vlm qualitative results}
      \vspace{-10pt}
\end{figure*}

\noindent \textbf{Implementation Details.} We implement our method as plug-and-play optimization modules for existing generative reconstruction methods. In the main experiments, we apply them to the following two base methods:
\begin{itemize}
    \item \textbf{DIVA}~\cite{wangdiffusion} introduces diffusion-based visual feedback to refine CLIP features. It performs reconstruction conditioned on CLIP vision embeddings, enabling the model to recover more detailed visual information.
    \item \textbf{GenHancer}~\cite{ma2025genhancer} systematically investigates how generative models can enhance CLIP by refining conditioning design, denoising strategies, and generation paradigms. It shows that using global conditions and lightweight denoisers leads to more stable reconstruction-based representation learning. In addition, it extends the reconstruction process to flow-based generative models.
\end{itemize}
For both base methods, we replace the original training objective with our objective, while keeping the network architecture, training data, and other optimization settings unchanged whenever possible for fair comparison. We denote the enhanced versions as \textit{+AUCGR} and \textit{+pAUCGR}. For AUCGR, we use the hinge surrogate listed above. For pAUCGR, we set $\gamma(\alpha)=0.9$. This protocol directly evaluates whether the proposed AUC-oriented optimization improves existing generative reconstruction pipelines. We retrain all models using public code on eight NVIDIA A100 80GB GPUs. 

\vskip 1 ex
\noindent \textbf{Evaluation Protocol.} Following \cite{wangdiffusion,ma2025genhancer}, we enhance $6$ CLIP backbones: OpenAI CLIP ViT-L@224/@336~\cite{radford2021learning}, MetaCLIP ViT-L/H@224~\cite{xudemystifying}, and SigLIP ViT-SO@224/@384~\cite{zhai2023sigmoid}. To evaluate the \textbf{\textcolor{PerColor}{P-Ability}}, we use MMVP-VLM~\cite{tong2024eyes}. To evaluate \textbf{\textcolor{DisColor}{D-Ability}}, we conduct zero-shot clustering and zero-shot classification on $26$ standard datasets. For clustering, we report \textit{Normalized Mutual Information (NMI)}, \textit{Accuracy (ACC)}, and \textit{Adjusted Rand Index (ARI)}. For classification, we report \textit{Top-1 Accuracy (ACC@1)}, \textit{Top-5 Accuracy (ACC@5)}, and \textit{Mean Recall (mRecall)}. We also evaluate zero-shot retrieval on Flickr30k~\cite{young2014image} and COCO~\cite{lin2014microsoft}, covering both image-to-text and text-to-image retrieval. Finally, we train MLLMs with our enhanced CLIP encoders using the official LLaVA-1.5~\cite{liu2024improved} recipe. We then evaluate these MLLMs on the same benchmarks~\cite{tong2024eyes,tong2024cambrian,li2024naturalbench,li2023evaluating,lu2022learn,guan2024hallusionbench}.


\subsection{Overall Performance}
\label{subsec: Overall Performance}

\begin{figure*}[t]
\centering
\subfloat[DIVA on S-1]{
\begin{minipage}{0.15\linewidth}
\includegraphics[width=\linewidth]{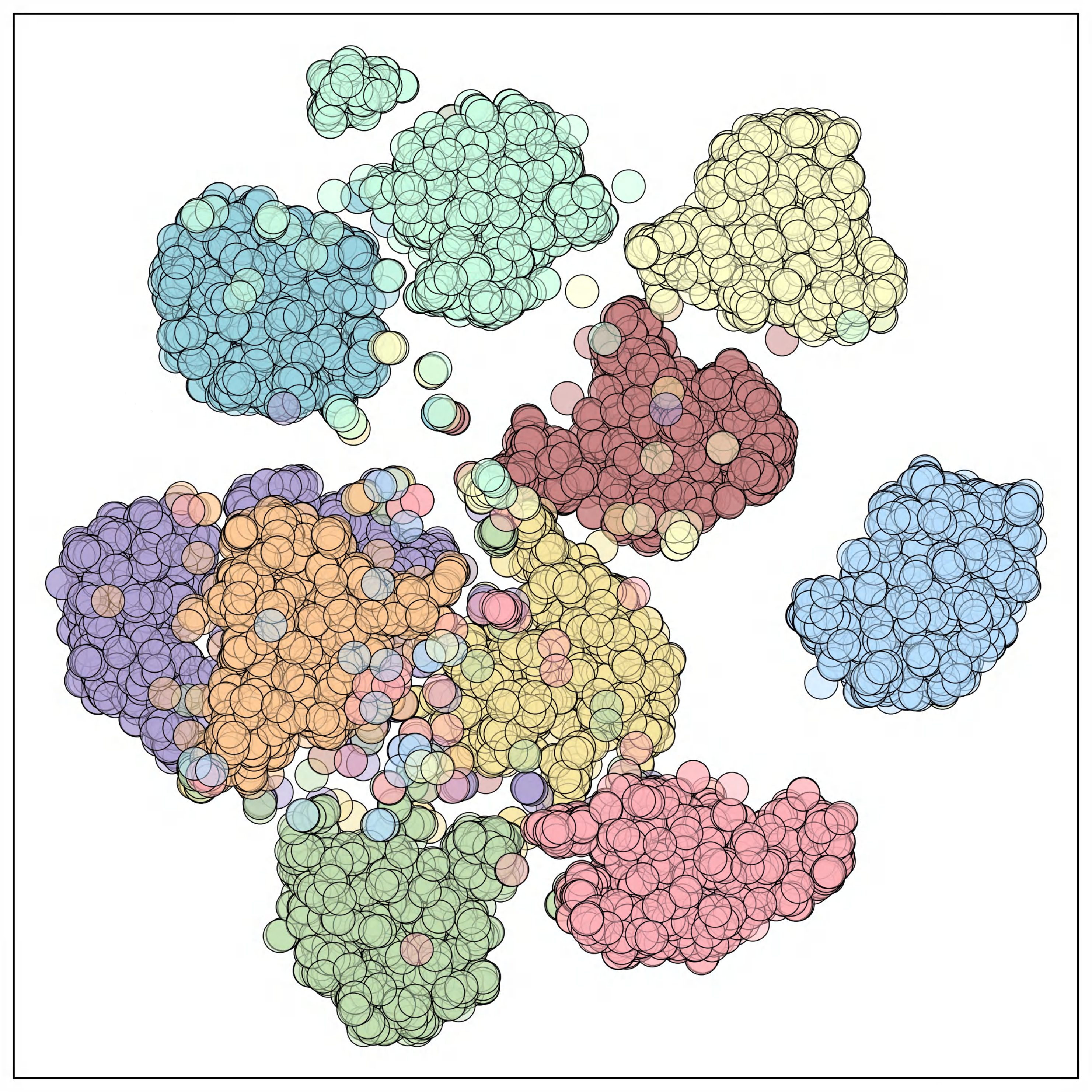}  
\end{minipage}
}
\hfil
\subfloat[+AUCGR on S-1]{   
\begin{minipage}{0.15\linewidth}
\includegraphics[width=\linewidth]{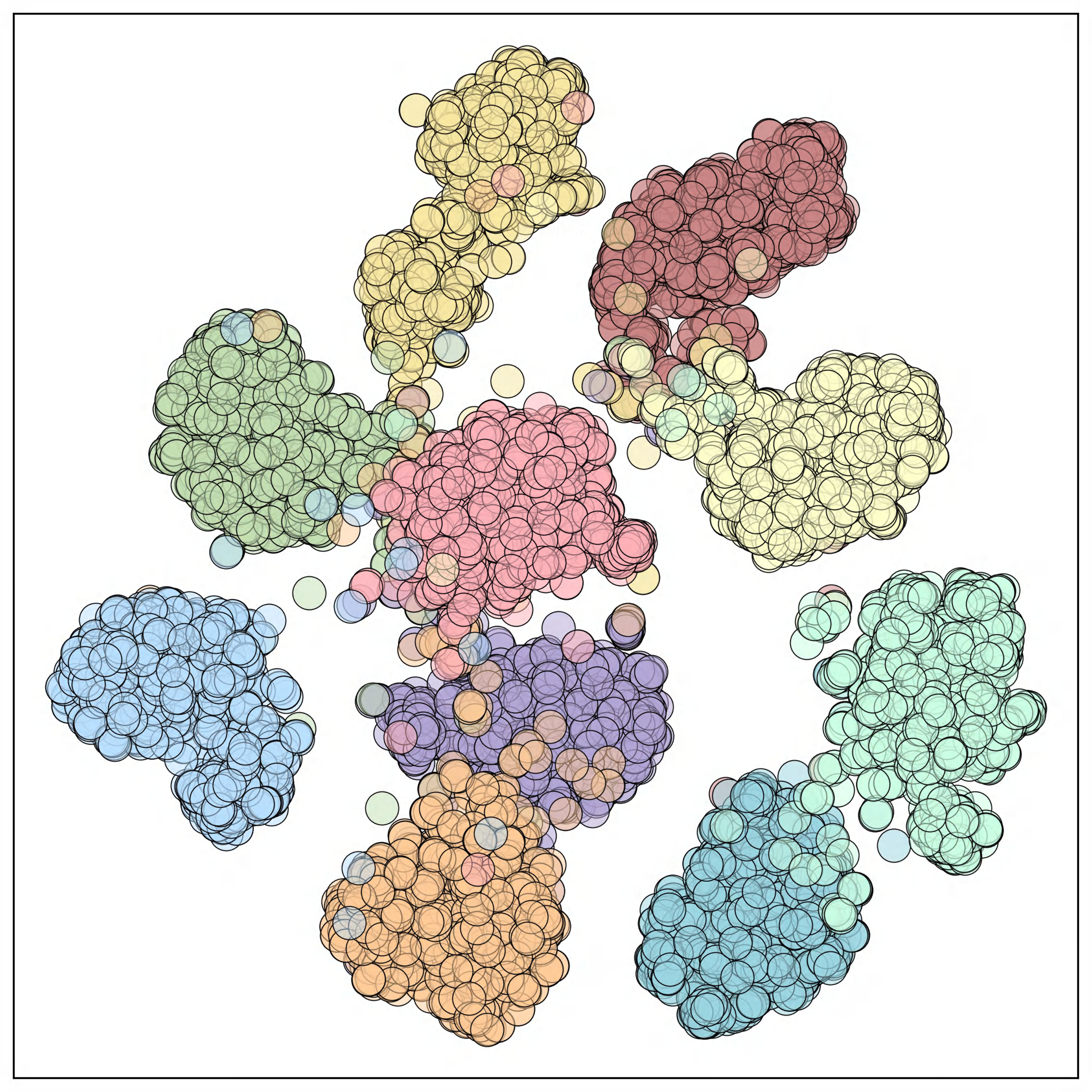} 
\end{minipage}
}
\hfil
\subfloat[+pAUCGR on S-1]{   
\begin{minipage}{0.15\linewidth}
\includegraphics[width=\linewidth]{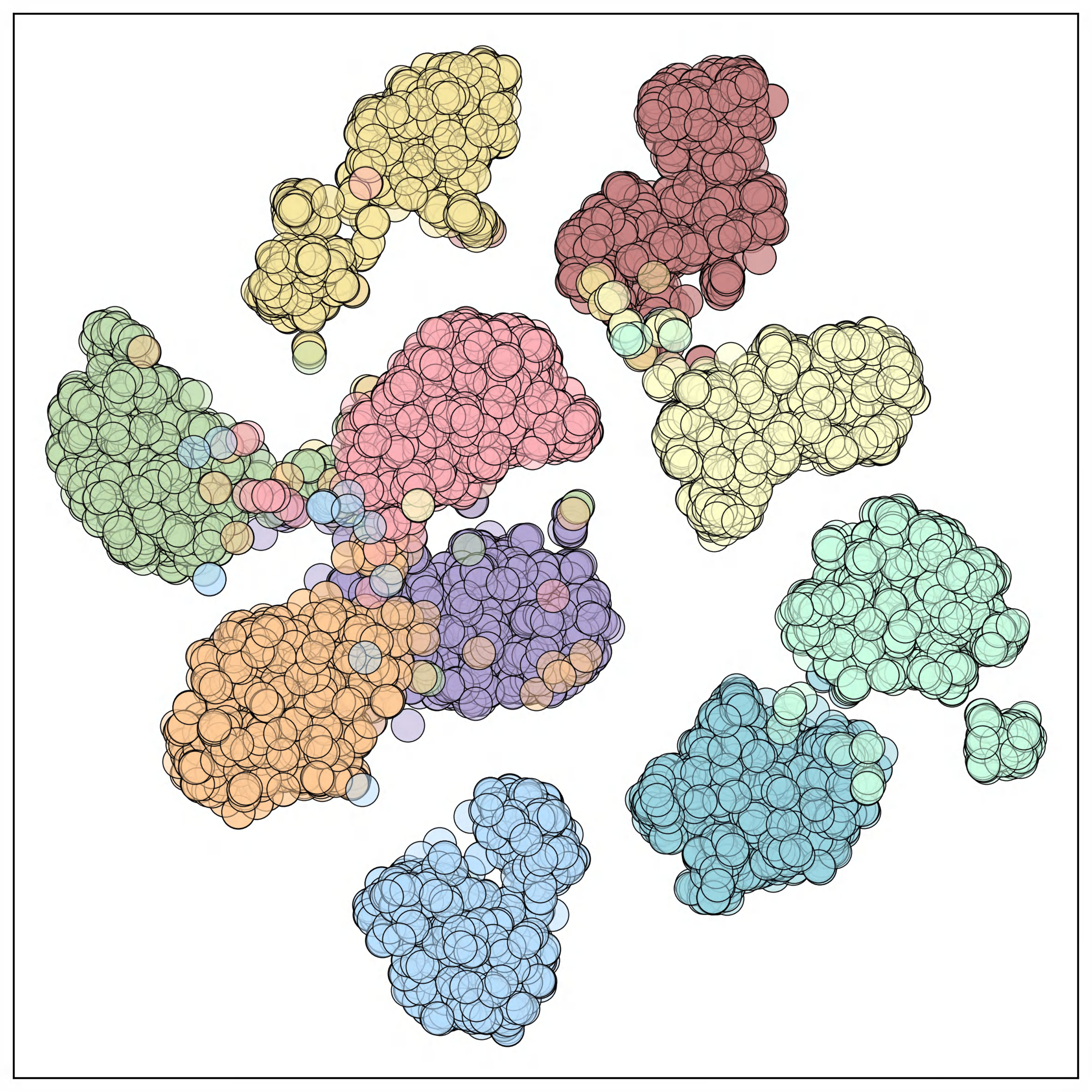}
\end{minipage}
}
\hfil
\subfloat[GenHancer on M-1]{
\begin{minipage}{0.15\linewidth}
\includegraphics[width=\linewidth]{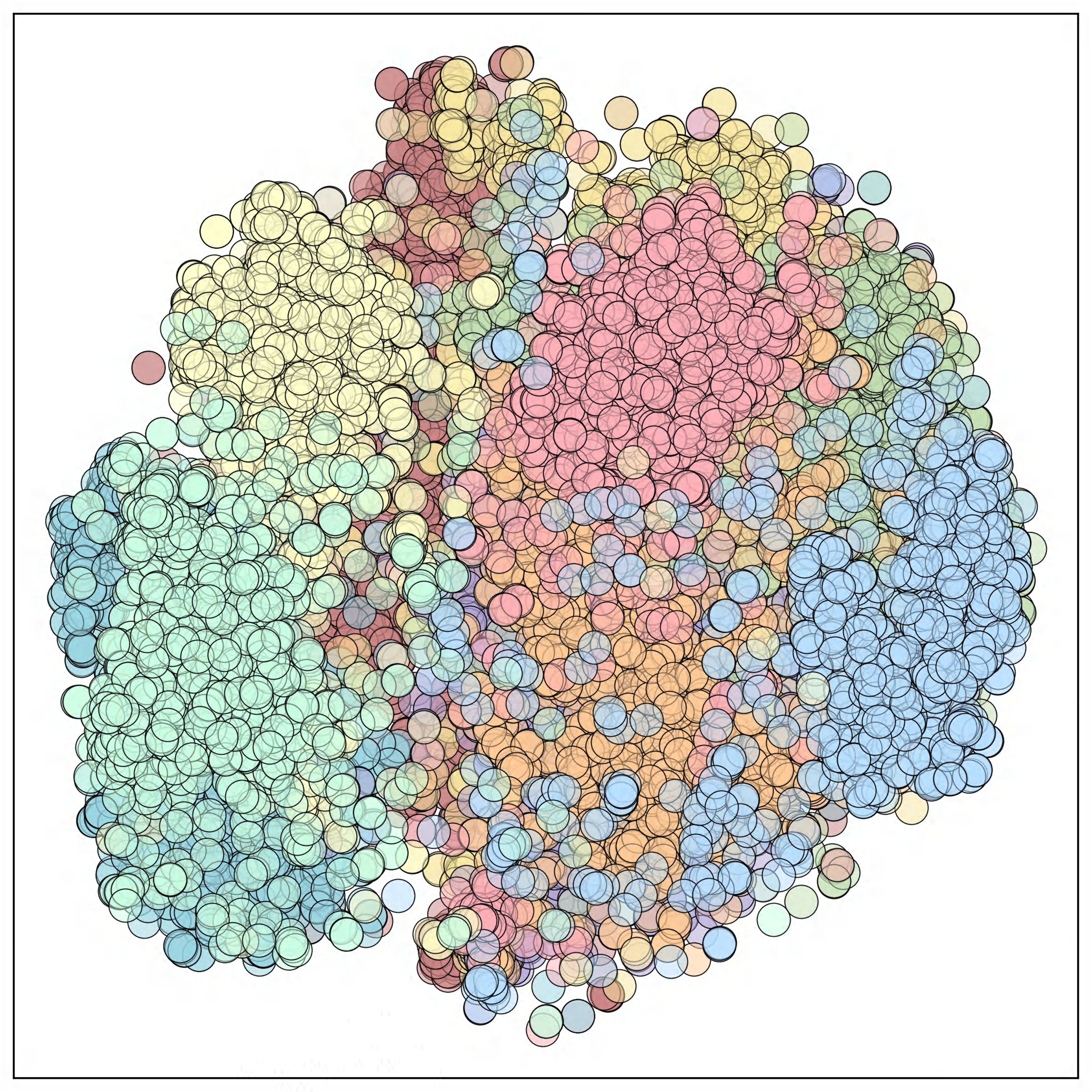}  
\end{minipage}
}
\hfil
\subfloat[+AUCGR on M-1]{   
\begin{minipage}{0.15\linewidth}
\includegraphics[width=\linewidth]{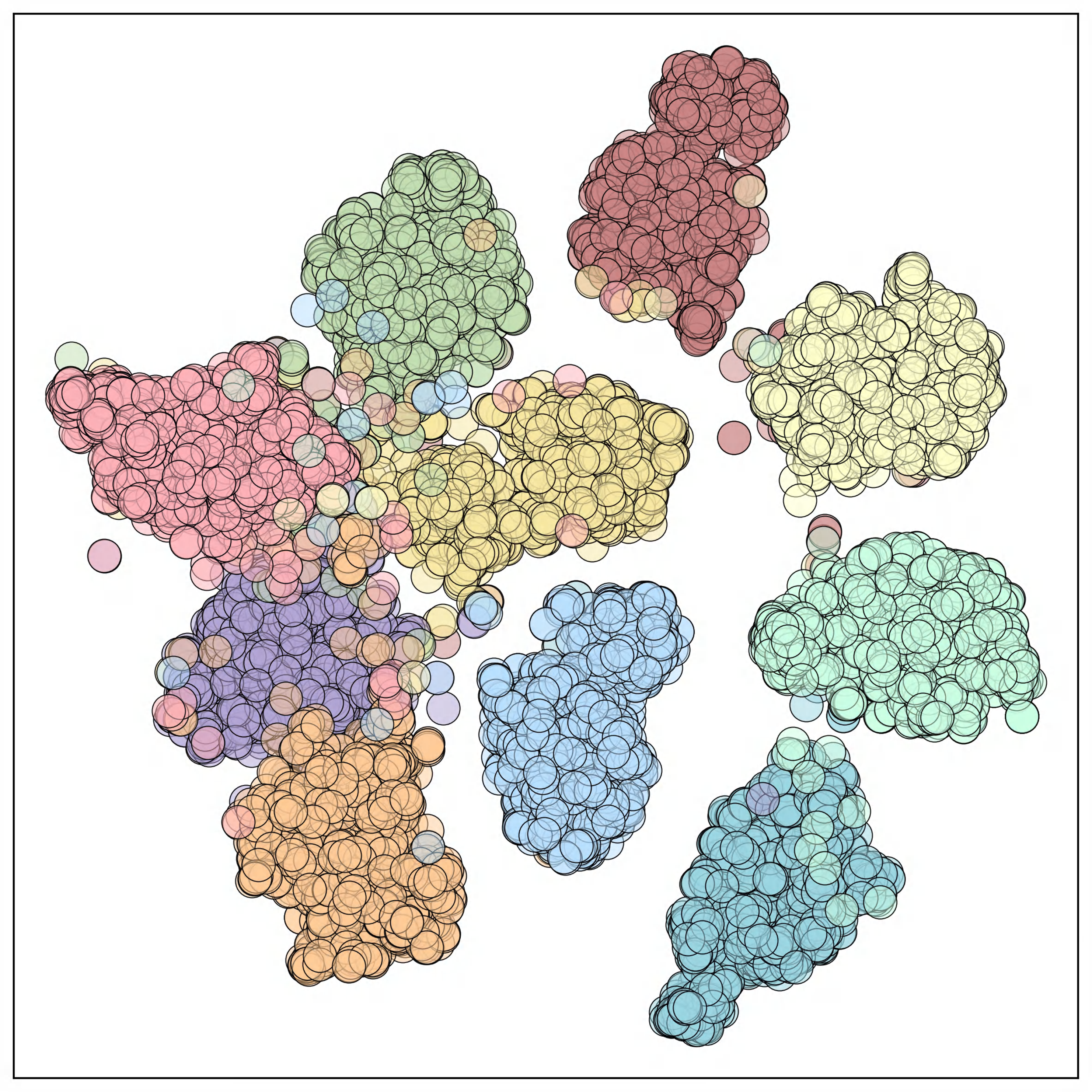} 
\end{minipage}
}
\hfil
\subfloat[+pAUCGR on M-1]{   
\begin{minipage}{0.15\linewidth}
\includegraphics[width=\linewidth]{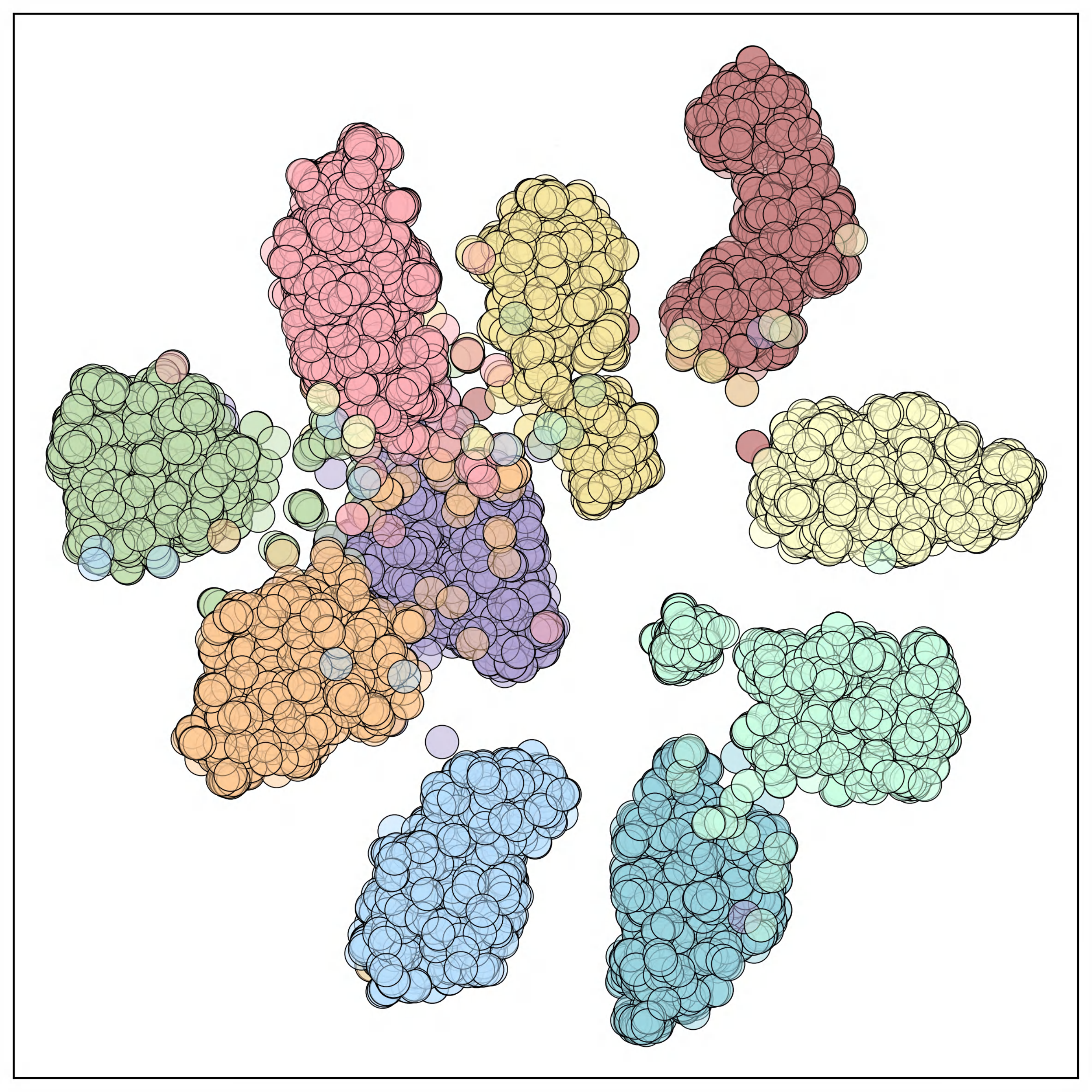}
\end{minipage}
}
\caption{\textbf{Qualitative results of \textcolor{DisColor}{D-Ability} on the CIFAR-10 zero-shot clustering benchmark by using the t-SNE method.} The first three panels use DIVA with SigLIP ViT-SO@224 (S-1), and the last three panels use GenHancer with MetaCLIP ViT-L@224 (M-1). The improved CLIP achieves better class separability.}
\label{fig: discriminative ability qualitative results}
\vspace{-10pt}
\end{figure*}

\begin{figure*}[t]
  \centering
  \subfloat[Image@5 - Flickr30k\label{fig: O1_Image_Flickr30k}]{
    \includegraphics[width=0.23\linewidth]{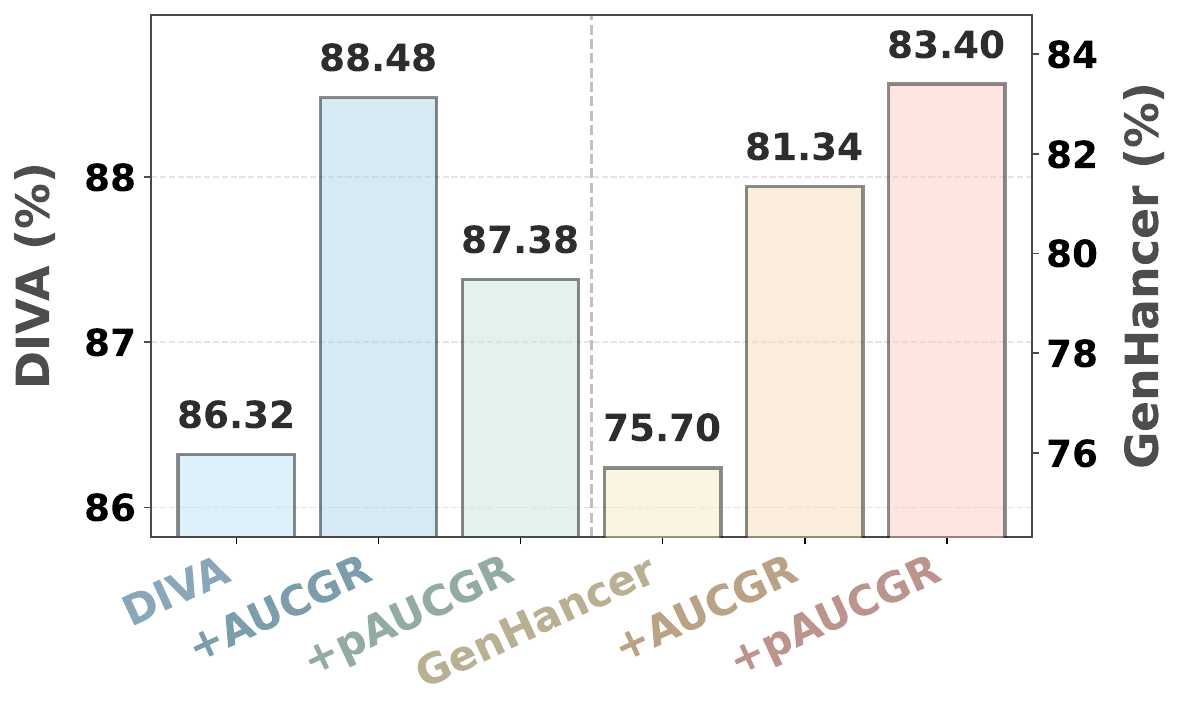}}
  \hfill
  \subfloat[Image@5 - COCO\label{fig: O1_Image_COCO}]{
    \includegraphics[width=0.23\linewidth]{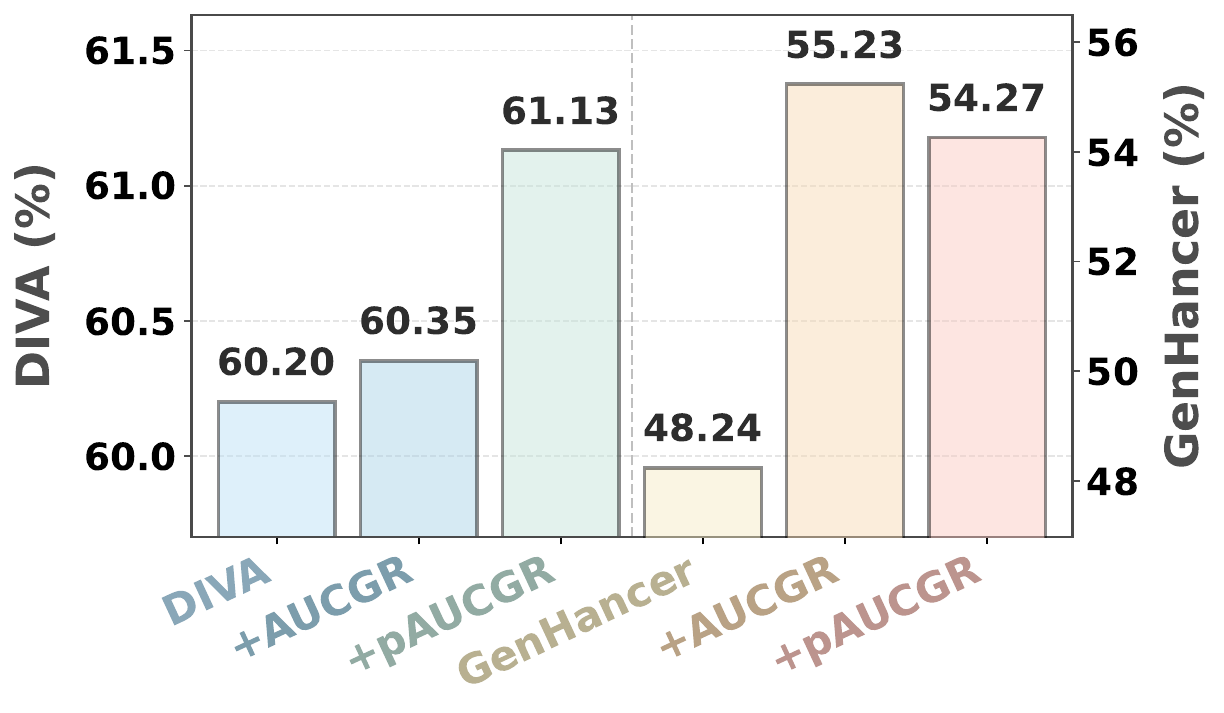}}
  \hfill
  \subfloat[Text@5 - Flickr30k\label{fig: O1_Text_Flickr30k}]{
    \includegraphics[width=0.23\linewidth]{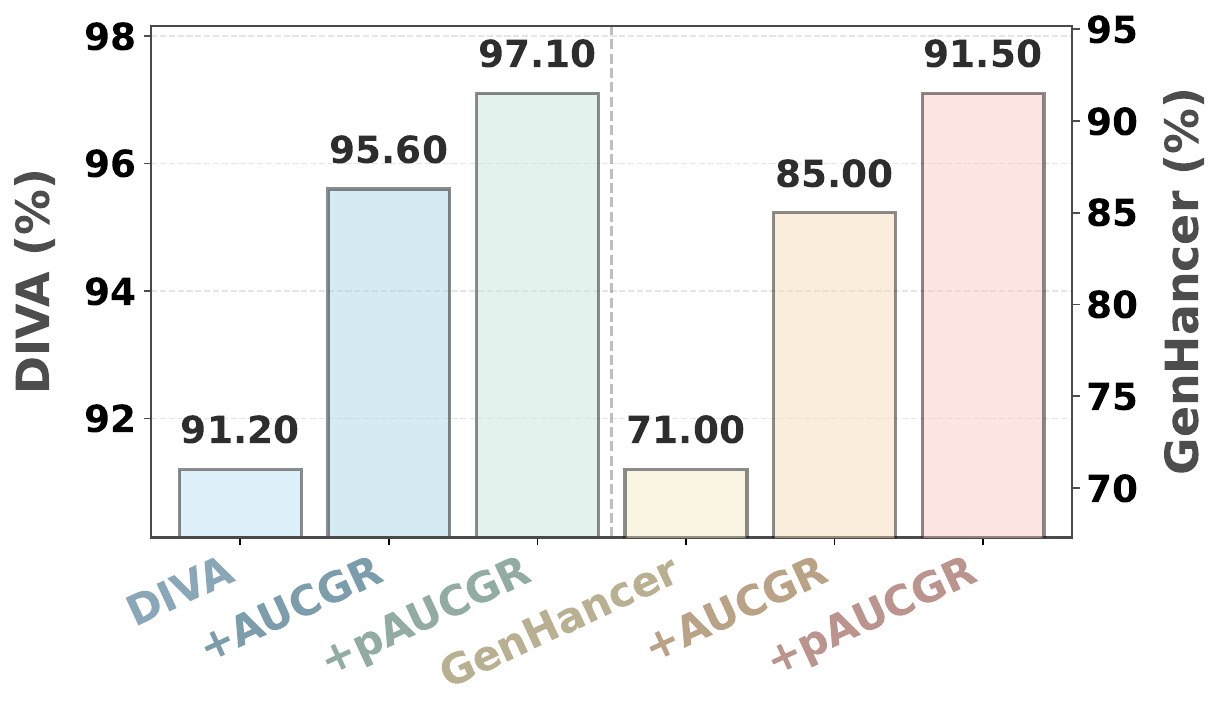}}
  \hfill
  \subfloat[Text@5 - COCO\label{fig: O1_Text_COCO}]{
    \includegraphics[width=0.23\linewidth]{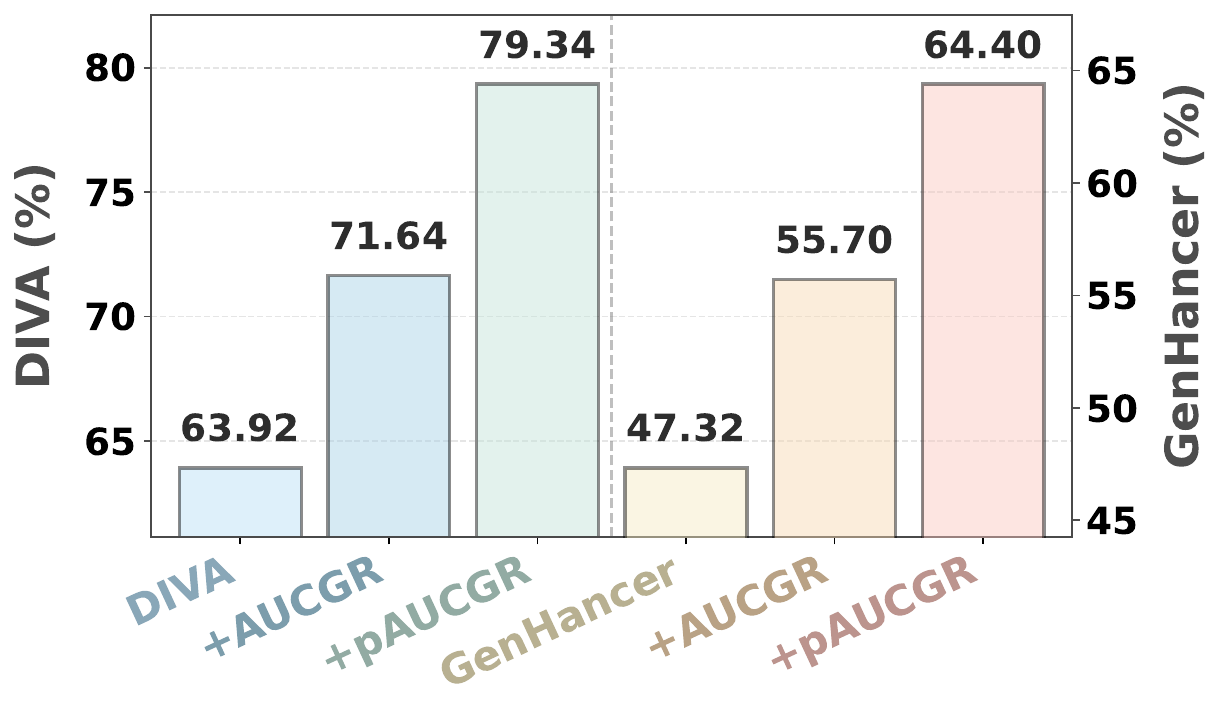}}
  \caption{\textbf{Performance on zero-shot retrieval on the OpenAI CLIP ViT-L@224.} Compared with the original DIVA/GenHancer model, our method consistently improves both image-to-text and text-to-image retrieval on Flickr30k and COCO. The results on the MetaCLIP ViT-L@224 and SigLIP ViT-SO@224 are provided in \cref{fig: M1_retrieval} and \cref{fig: S1_retrieval}.}
  \label{fig: O1_retrieval}
  \vspace{-10pt}
\end{figure*}

\begin{figure}[t]
\centering
\subfloat[DIVA]{
\begin{minipage}{0.45\linewidth}
\includegraphics[width=\linewidth]{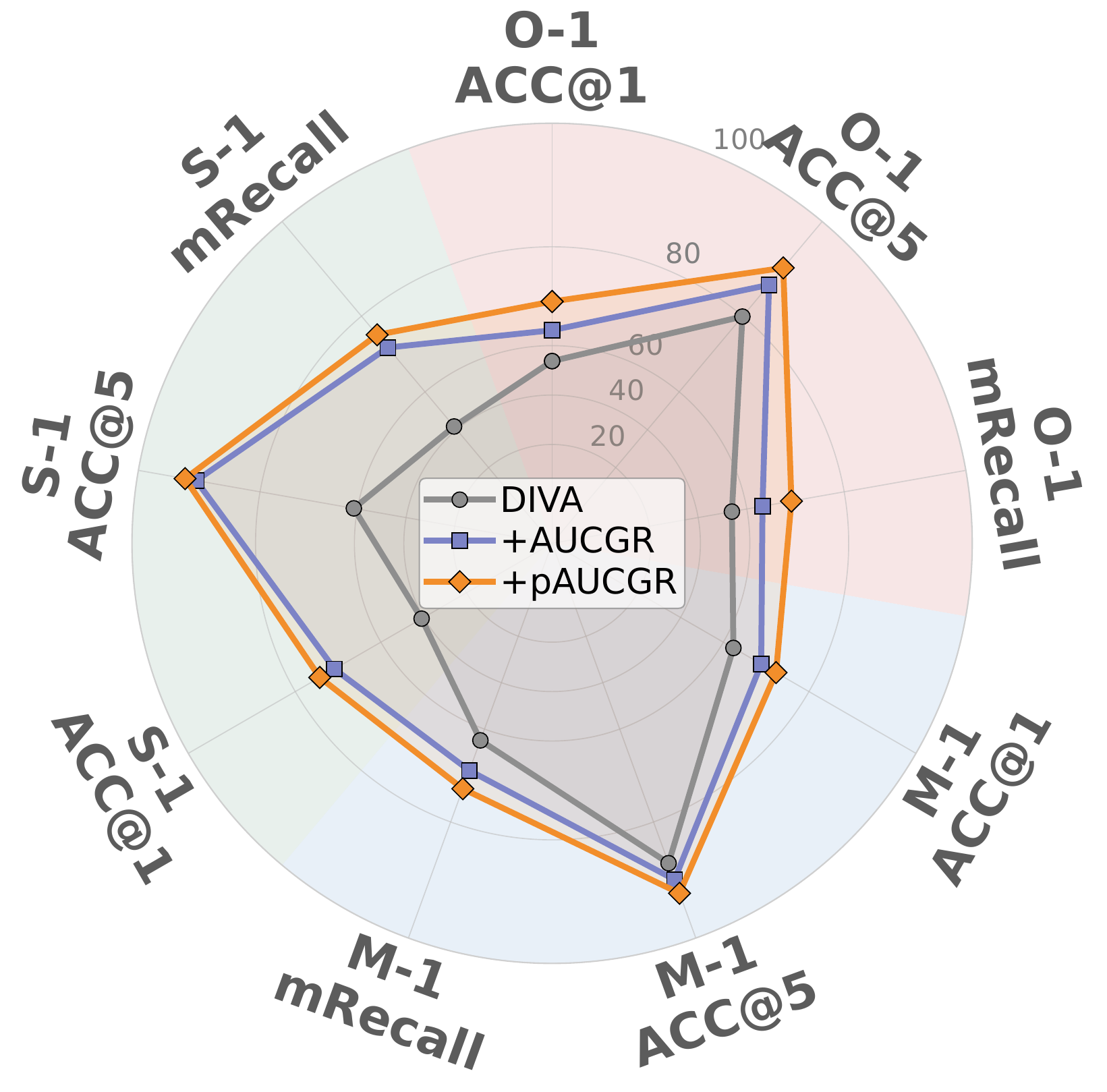}  
\end{minipage}
}
\hfil
\subfloat[GenHancer]{   
\begin{minipage}{0.45\linewidth}
\includegraphics[width=\linewidth]{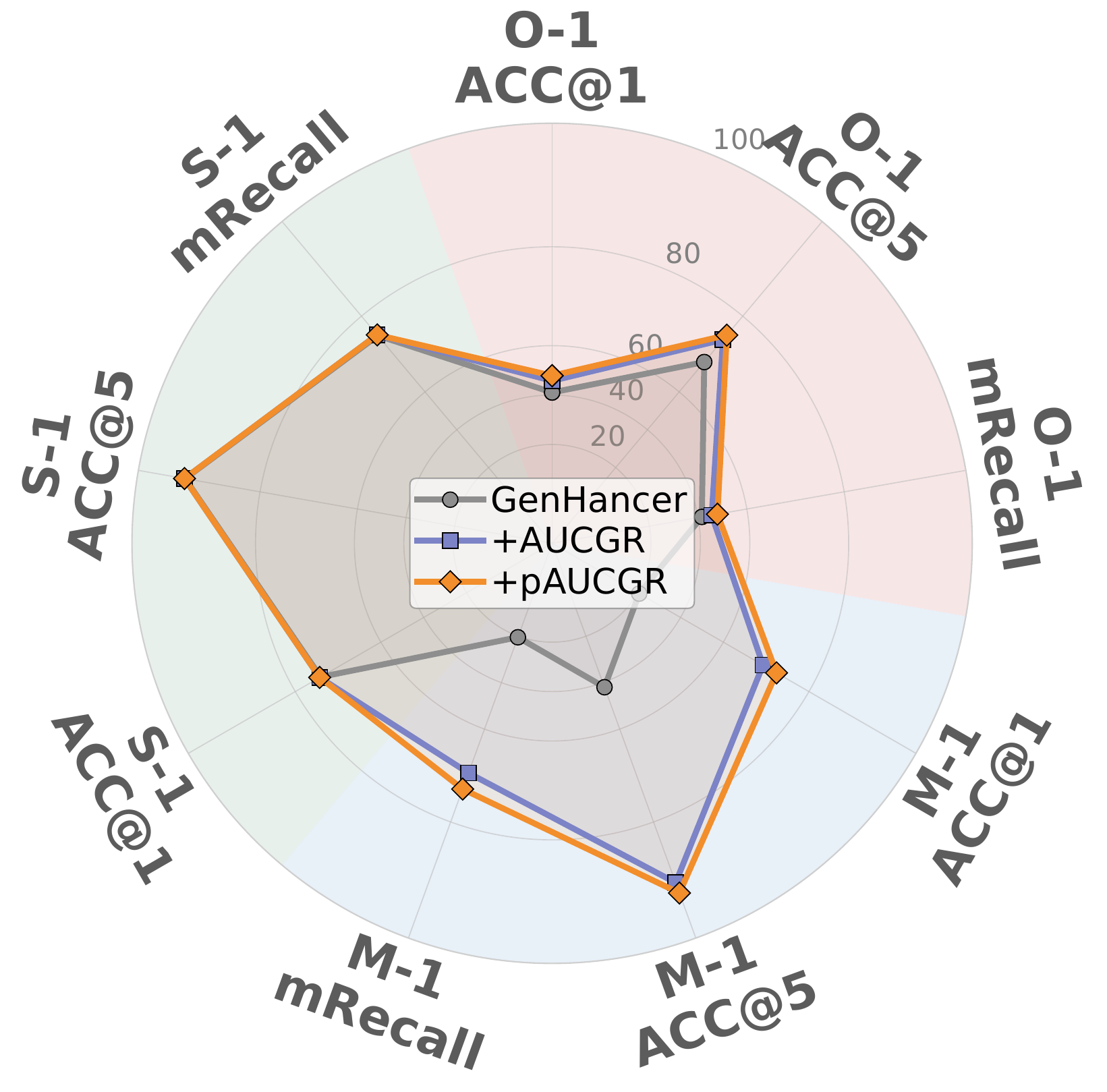} 
\end{minipage}
}
\caption{\textbf{Average performance on 26 standard zero-shot classification benchmarks.} O-1, M-1, and S-1 separately represent OpenAI CLIP ViT-L@224, MetaCLIP ViT-L@224, and SigLIP ViT-SO@224. Detailed results are provided in \cref{tab: classification acc1}, \cref{tab: classification acc5} and \cref{tab: classification mrecall}.}
\label{fig: classification results}
\vspace{-10pt}
\end{figure}

\subsubsection{Performance of P-Ability}

\noindent \noindent \textbf{Quantitative results.} As shown in \cref{tab: understanding ability results}, AUCGR consistently improves the original generative reconstruction baselines on MMVP-VLM. For OpenAI CLIP ViT-L@224, AUCGR improves GenHancer from $25.2\%$ to $26.7\%$, and raises the color-and-appearance score from $66.7\%$ to $73.3\%$. These gains show that AUCGR improves fine-grained perceptual ability across visual patterns such as color, state, structure, and viewpoint. Our conference version, DCR, also yields clear improvements over the corresponding baselines. pAUCGR further improves the results by filtering ambiguous negatives with a suitable similarity threshold. On MetaCLIP ViT-L@224, pAUCGR improves DIVA from $25.2\%$ to $29.6\%$ and GenHancer from $28.2\%$ to $31.8\%$. Overall, AUCGR brings consistent gains over generative reconstruction baselines, and pAUCGR achieves stronger performance by reducing unreliable negative comparisons.

\vskip 1 ex
\noindent \textbf{Qualitative results.} \cref{fig: mmvp vlm qualitative results} further shows representative examples from MMVP-VLM. The original CLIP often fails when the question depends on local visual evidence, such as color, viewpoint, and subtle object appearance. After applying our plug-in optimization, the enhanced CLIP gives more accurate predictions on these detail-sensitive cases. The qualitative results are consistent with the quantitative gains and indicate that our method improves fine-grained perceptual ability without changing the model architecture. Additional qualitative results are provided in \cref{subsec: supp_Expanded Qualitative Results on P-Ability Performance}.

\subsubsection{Performance of D-Ability}

We evaluate \textbf{\textcolor{DisColor}{D-Ability}} from three complementary perspectives, including zero-shot clustering, zero-shot classification, and zero-shot retrieval.

\vskip 1 ex
\noindent \textbf{Zero-shot clustering.} We first evaluate whether image representations form separable semantic structures in image clustering tasks. As shown in \cref{tab: discriminative ability results ACC}, AUCGR improves the two generative reconstruction baselines on $26$ standard datasets. Specifically, on SigLIP ViT-SO@224, AUCGR improves DIVA from $26.88\%$ to $55.81\%$ Avg ACC. On MetaCLIP ViT-L@224, AUCGR improves GenHancer from $30.39\%$ to $44.61\%$ Avg ACC. pAUCGR further improves the results by filtering ambiguous negatives. For example, it raises the DIVA result on SigLIP ViT-SO@224 to $59.30\%$ Avg ACC and the GenHancer result on MetaCLIP ViT-L@224 to $52.78\%$ Avg ACC. Compared with the original pretrained models, pAUCGR largely restores their discriminative ability in most settings and even achieves improvements on some datasets. The ARI and NMI results in \cref{tab: discriminative ability results ARI} and \cref{tab: discriminative ability results NMI} show the same trend. \textbf{This indicates that the comparison-based supervision encourages samples with related semantics to be grouped more compactly and separates samples with different semantics more clearly.} The t-SNE visualization in \cref{fig: discriminative ability qualitative results} further supports this observation.

\vskip 1 ex
\noindent \textbf{Zero-shot classification.} We next examine the performance on zero-shot classification. As shown in \cref{fig: classification results}, AUCGR improves the average classification accuracy of both DIVA and GenHancer, showing that comparison-based supervision helps improve the discriminative ability weakened by plain generative reconstruction supervision. Specifically, on MetaCLIP ViT-L@224, AUCGR improves GenHancer from $20.69\%$ to $69.26\%$ Avg ACC@1. On SigLIP ViT-SO@224, AUCGR improves DIVA from $40.97\%$ to $70.86\%$ Avg ACC@1. pAUCGR further improves the results by filtering ambiguous negatives with the similarity threshold, reaching $72.42\%$ and $74.28\%$ Avg ACC@1, respectively. These results show that \textbf{the proposed comparison-based optimization improves the quality of visual representations and remains effective in the more demanding classification setting.} The complete results are provided in \cref{tab: classification acc1}, \cref{tab: classification acc5} and \cref{tab: classification mrecall}.

\vskip 1 ex
\noindent \textbf{Zero-shot retrieval.} Finally, we evaluate cross-modal alignment on Flickr30k and COCO. As shown in \cref{fig: O1_retrieval}, our method improves both image-to-text and text-to-image retrieval on OpenAI CLIP ViT-L@224. Specifically, for Text@5 on COCO, AUCGR improves the DIVA baseline from $63.92$ to $71.64$, and pAUCGR further reaches $79.34$. For Image@5 on Flickr30k, AUCGR improves the GenHancer baseline from $75.70$ to $81.34$, and pAUCGR further improves it to $83.40$. The results on MetaCLIP ViT-L@224 and SigLIP ViT-SO@224 are provided in \cref{fig: M1_retrieval} and \cref{fig: S1_retrieval}, where the same trend is observed. Since retrieval requires distinguishing correct image-text pairs from many candidates, these results further show that the enhanced visual representations have stronger discriminative ability.

\subsection{Multimodal Large Language Model Evaluation}
\label{subsec: Multimodal Large Language Model Evaluation}
We further examine whether the enhanced visual representations learned by our method can be effectively transferred to \textit{Multimodal Large Language Models (MLLMs)}. Specifically, we substitute the original CLIP vision encoder in the LLaVA-1.5~\cite{liu2024improved} framework with our improved CLIP while keeping all training configurations unchanged to ensure a fair comparison. The evaluation is performed on a series of vision-centric and general multimodal benchmarks, covering both vision-centric and general tasks. As shown in \cref{tab: mllm results}, the integration of our enhanced CLIP substantially boosts MLLM performance, particularly in benchmarks that require detailed visual understanding. These results demonstrate that improving the granularity of visual representations not only benefits standalone vision models but also strengthens the visual grounding and reasoning ability of multimodal systems. Some qualitative results are provided in \cref{subsec: supp_Expanded Qualitative Results on MLLM Performance}.

\subsection{Ablation Studies}
\label{subsec: Ablation Studies}

\begin{figure*}[t]
\vspace{-20pt}
\centering
\subfloat[O-1/DIVA]{
\begin{minipage}{0.15\linewidth}
\includegraphics[width=\linewidth]{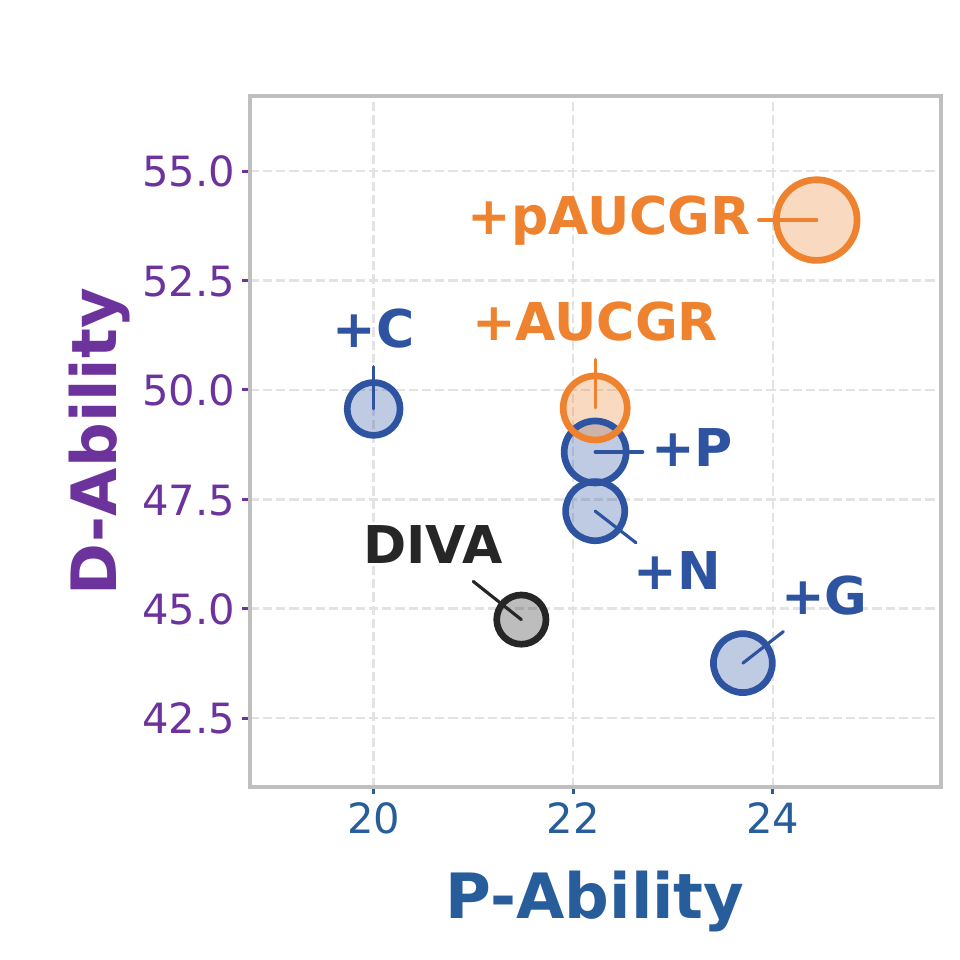}  
\end{minipage}
}
\hfil
\subfloat[O-1/GenHancer]{   
\begin{minipage}{0.15\linewidth}
\includegraphics[width=\linewidth]{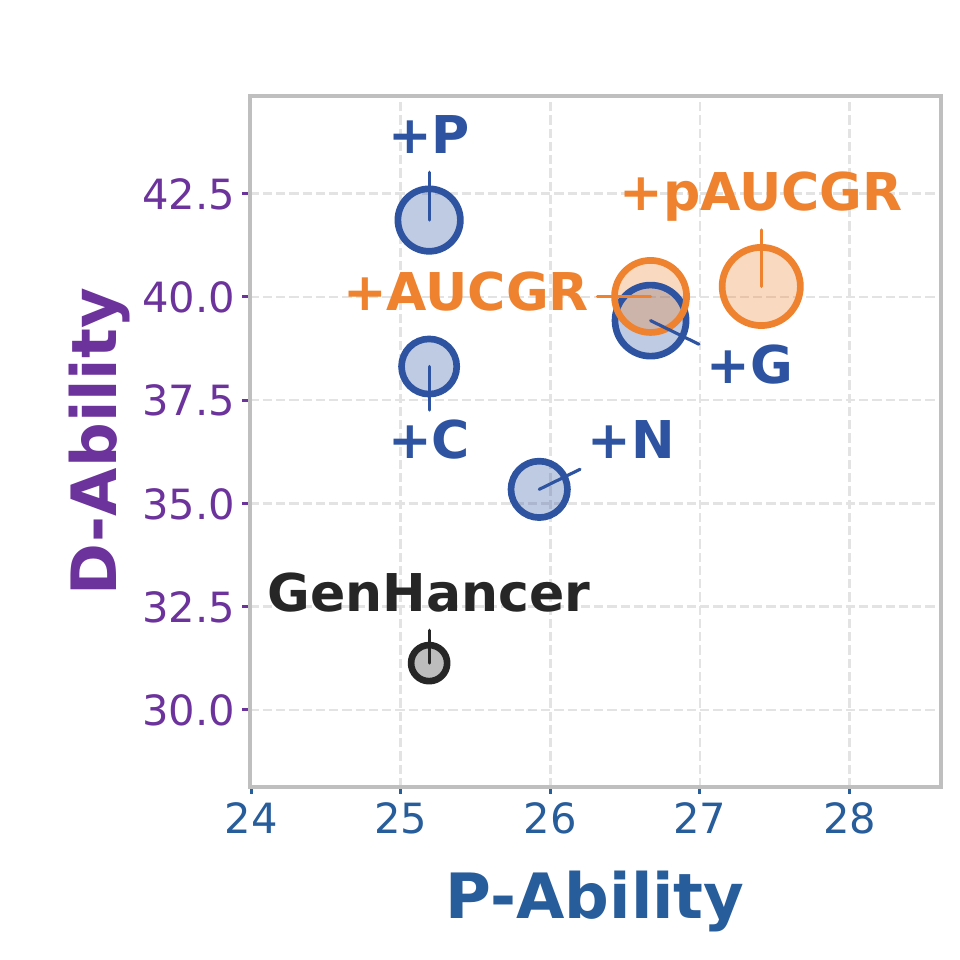} 
\end{minipage}
}
\hfil
\subfloat[M-1/DIVA]{   
\begin{minipage}{0.15\linewidth}
\includegraphics[width=\linewidth]{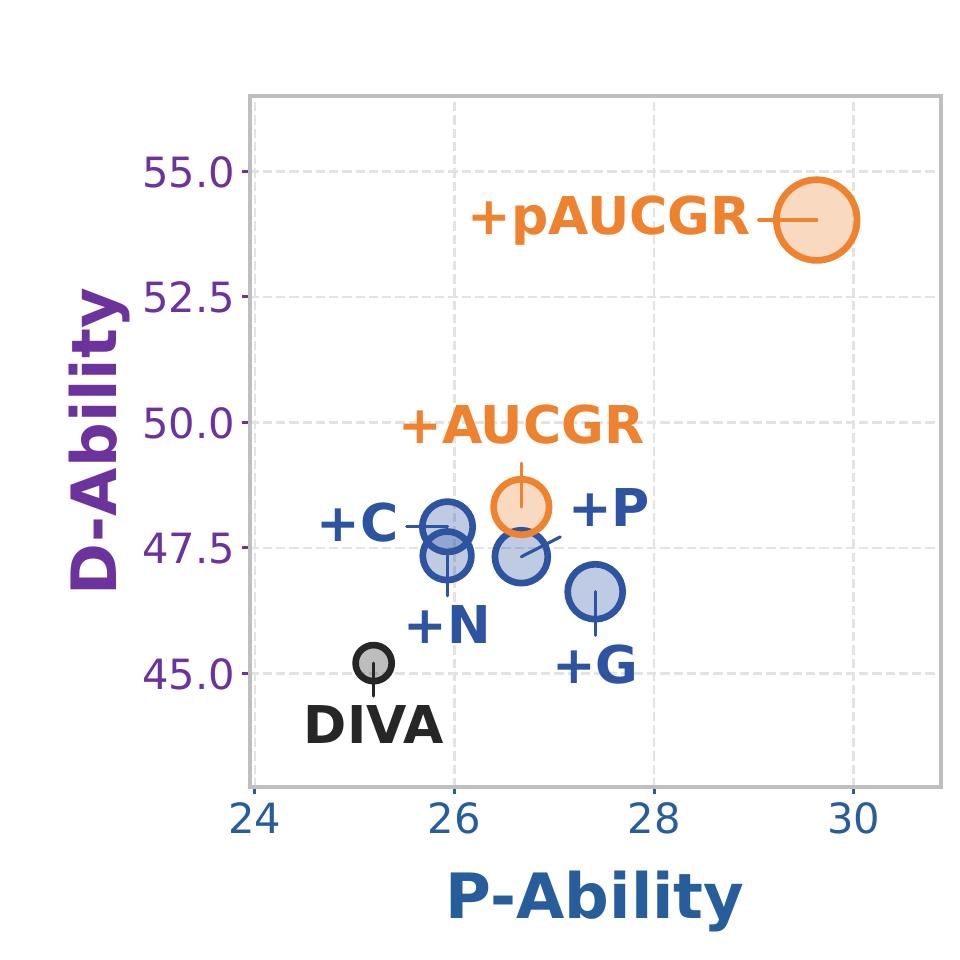}
\end{minipage}
}
\hfil
\subfloat[M-1/GenHancer]{
\begin{minipage}{0.15\linewidth}
\includegraphics[width=\linewidth]{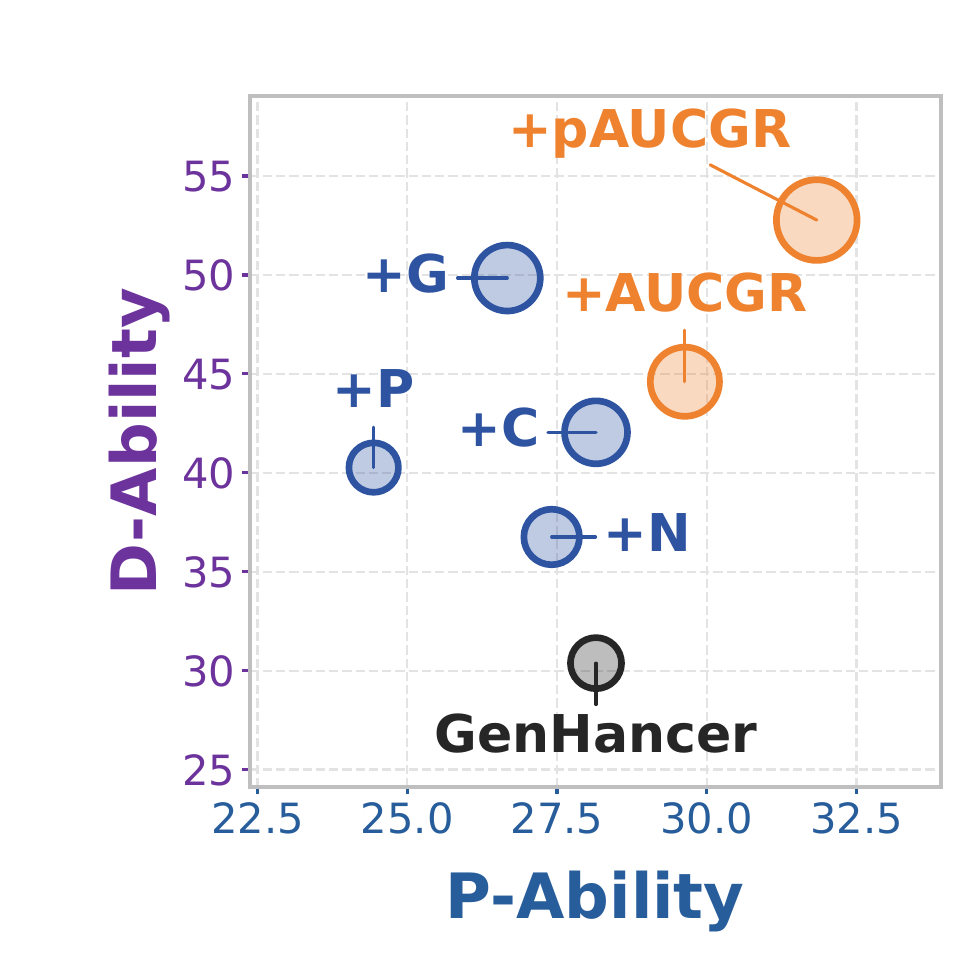}  
\end{minipage}
}
\hfil
\subfloat[S-1/DIVA]{   
\begin{minipage}{0.15\linewidth}
\includegraphics[width=\linewidth]{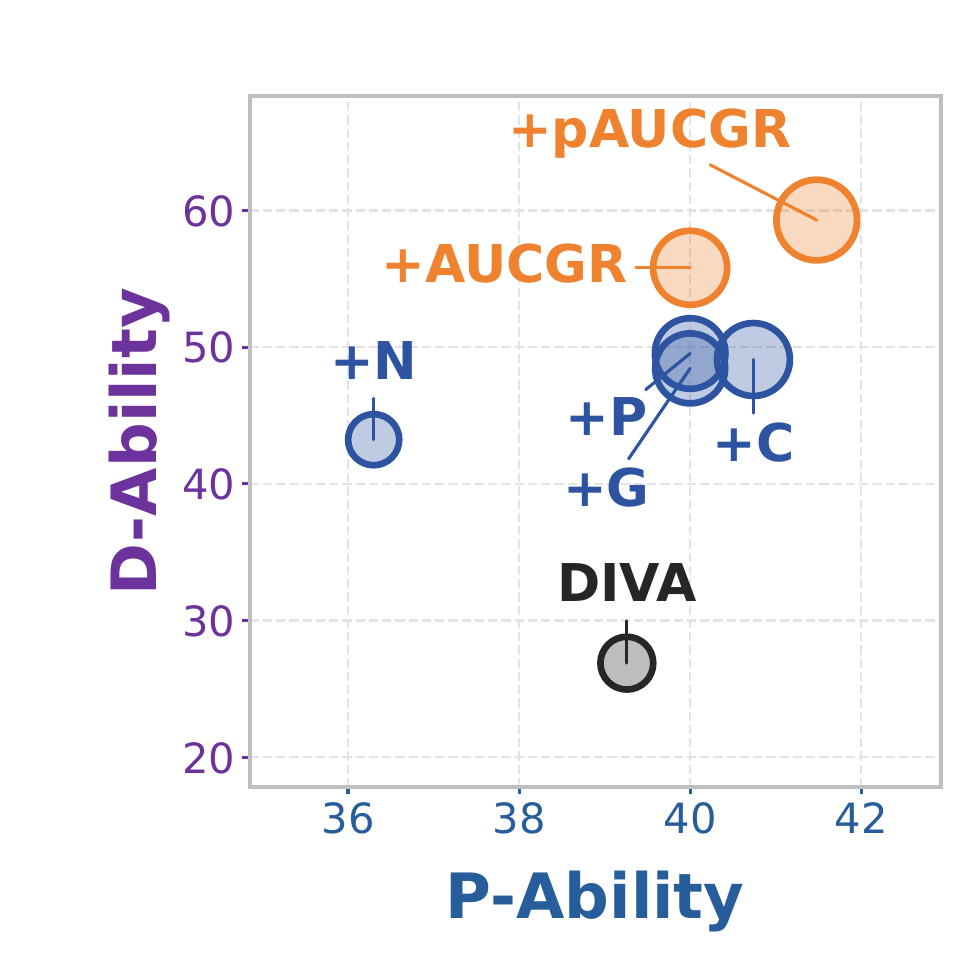} 
\end{minipage}
}
\hfil
\subfloat[S-1/GenHancer]{   
\begin{minipage}{0.15\linewidth}
\includegraphics[width=\linewidth]{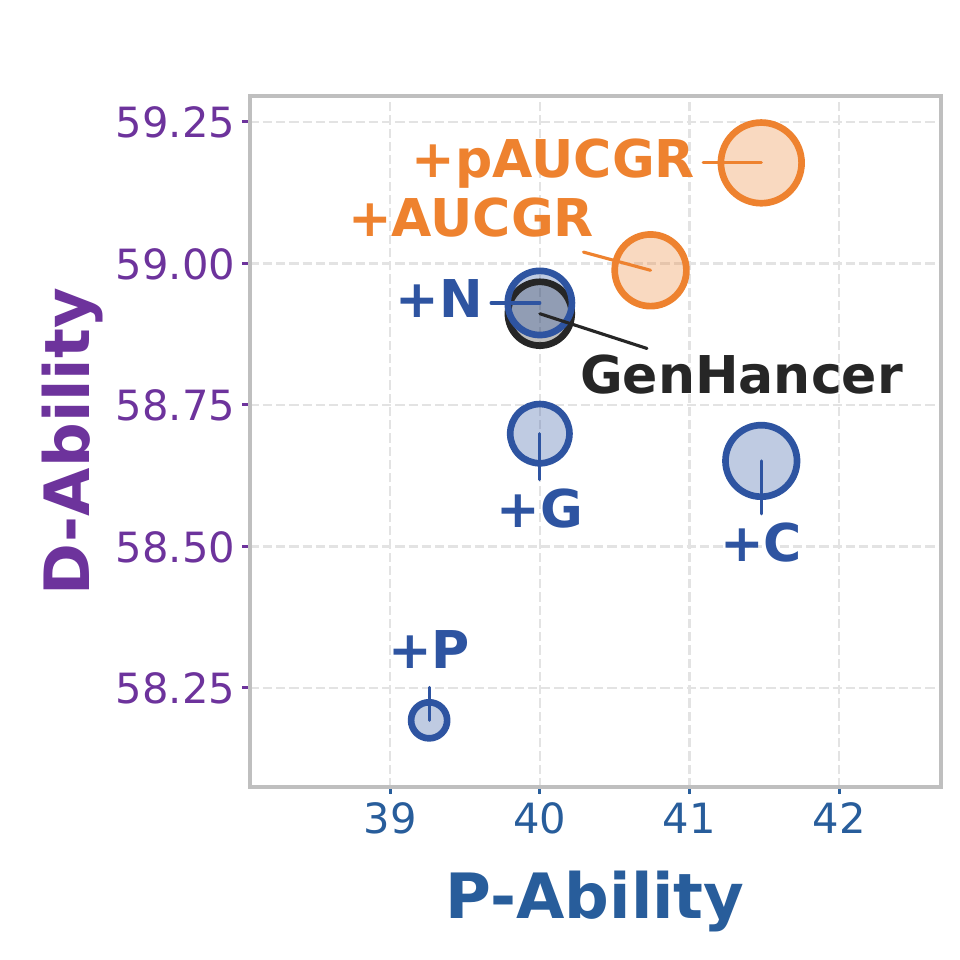}
\end{minipage}
}
\caption{\textbf{Ablation comparison with naive linear weighting and gradient-balancing methods.} We evaluate the trade-off between \textbf{\textcolor{PerColor}{P-Ability}} and \textbf{\textcolor{DisColor}{D-Ability}} on two generative reconstruction baselines, DIVA and GenHancer, across OpenAI CLIP ViT-L@224 (O-1), MetaCLIP ViT-L@224 (M-1), and SigLIP ViT-SO@224 (S-1). Here, \textbf{\textcolor{DisColor}{D-Ability}} is measured by ACC on zero-shot clustering benchmarks, while +N, +P, +C, and +G denote +Naive, +PCGrad, +CAGrad, and +GradNorm, respectively. Compared with both naive linear weighting and gradient-balancing methods, AUCGR and pAUCGR achieve a better balance between the two abilities.}
\label{fig: Naive Method}
\end{figure*}

\vskip 1 ex
\noindent \textbf{Results on Naive and Gradient-Balancing Methods.} \cref{fig: Naive Method} compares our AUCGR with the naive linear weighting method and three representative gradient-balancing methods, i.e., PCGrad~\cite{yu2020gradient}, CAGrad~\cite{liu2021conflict}, and GradNorm~\cite{chen2018gradnorm}. Although these methods can improve \textbf{\textcolor{DisColor}{D-Ability}} or \textbf{\textcolor{PerColor}{P-Ability}} in some cases, their gains are generally limited or inconsistent across different baselines and backbones. This suggests that the conflict between the two optimization objectives remains insufficiently resolved, making it difficult to approach the Pareto frontier. In contrast, AUCGR and pAUCGR consistently achieve a better balance between the two abilities through a unified optimization objective. These results further demonstrate the effectiveness of the proposed AUCGR framework.

\begin{table*}[t!]
    \centering
    \renewcommand\arraystretch{0.9}
    \caption{\textbf{Performance of MLLMs (LLaVA-1.5~\cite{liu2024improved}) on various benchmarks.} The champion is highlighted in \textbf{bold}.}
    \vspace{-10pt}
    \resizebox{\linewidth}{!}{
    \begin{tabular}{@{}clccccccccccccc@{}}
    \toprule
    \multirow{3}{*}{LLM} & \multirow{3}{*}{CLIP} & \multicolumn{8}{c}{Vision-Centric Benchmarks} & \multicolumn{5}{c}{Conventional MLLM Benchmarks} \\ \cmidrule(l){3-10} \cmidrule(l){11-15} 
     &  & \multirow{2}{*}{\begin{tabular}[c]{@{}c@{}}MMVP-\\ MLLM~\cite{tong2024eyes}\end{tabular}} & \multicolumn{4}{c}{NaturalBench~\cite{li2024naturalbench}} & \multicolumn{2}{c}{CV-Bench 2D~\cite{tong2024cambrian}} & \multirow{2}{*}{\begin{tabular}[c]{@{}c@{}}CV-Bench\\ 3D~\cite{tong2024cambrian}\end{tabular}} & \multicolumn{3}{c}{POPE~\cite{li2023evaluating}} & \multirow{2}{*}{\begin{tabular}[c]{@{}c@{}}SciQA-\\ IMG~\cite{lu2022learn}\end{tabular}} & \multirow{2}{*}{\begin{tabular}[c]{@{}c@{}}Hallusion\\ Avg.~\cite{guan2024hallusionbench}\end{tabular}} \\ \cmidrule(lr){4-7} \cmidrule(lr){8-9} \cmidrule(lr){11-13}
     &  &  & Acc & Q-Acc & I-Acc & G-Acc & ADE20K & COCO &  & rand & pop & adv &  &  \\ \midrule
    \multirow{5}{*}{Vicuna-7B} & Original & 24.7 & 76.4 & 53.6 & 56.4 & 17.6 & 49.6 & 60.9 & 58.7 & 87.3 & 86.1 & 84.2 & 66.8 & 27.6 \\
    \cmidrule(l){2-15}
     & DIVA & \textbf{31.3} & 75.3 & 51.7 & \textbf{56.1} & 22.3 & 51.3 & 63.4 & 60.2 & \textbf{87.9} & 87.0 & 84.6 & 66.3 & \textbf{28.6} \\
     & \cellcolor[rgb]{ .851,  .882,  .957}+pAUCGR & \cellcolor[rgb]{ .851,  .882,  .957}\textbf{31.3} & \cellcolor[rgb]{ .851,  .882,  .957}\textbf{76.3} & \cellcolor[rgb]{ .851,  .882,  .957}\textbf{53.9} & \cellcolor[rgb]{ .851,  .882,  .957}\textbf{56.1} & \cellcolor[rgb]{ .851,  .882,  .957}\textbf{25.5} & \cellcolor[rgb]{ .851,  .882,  .957}\textbf{54.1} & \cellcolor[rgb]{ .851,  .882,  .957}\textbf{64.7} & \cellcolor[rgb]{ .851,  .882,  .957}\textbf{62.6} & \cellcolor[rgb]{ .851,  .882,  .957}87.3 & \cellcolor[rgb]{ .851,  .882,  .957}\textbf{87.2} & \cellcolor[rgb]{ .851,  .882,  .957}\textbf{84.9} & \cellcolor[rgb]{ .851,  .882,  .957}\textbf{66.4} & \cellcolor[rgb]{ .851,  .882,  .957}\textbf{28.6} \\ 
     \cmidrule(l){2-15}
     & GenHancer & 30.7 & 77.3 & 55.6 & 59.1 & 24.4 & 52.9 & 63.6 & 63.2 & 88.1 & 86.7 & 84.6 & 66.5 & 28.4 \\ 
     & \cellcolor[rgb]{ .851,  .882,  .957}+pAUCGR & \cellcolor[rgb]{ .851,  .882,  .957}\textbf{31.3} & \cellcolor[rgb]{ .851,  .882,  .957}\textbf{79.9} & \cellcolor[rgb]{ .851,  .882,  .957}\textbf{56.1} & \cellcolor[rgb]{ .851,  .882,  .957}\textbf{61.2} & \cellcolor[rgb]{ .851,  .882,  .957}\textbf{26.5} & \cellcolor[rgb]{ .851,  .882,  .957}\textbf{54.9} & \cellcolor[rgb]{ .851,  .882,  .957}\textbf{65.2} & \cellcolor[rgb]{ .851,  .882,  .957}\textbf{63.5} & \cellcolor[rgb]{ .851,  .882,  .957}\textbf{88.7} & \cellcolor[rgb]{ .851,  .882,  .957}\textbf{87.5} & \cellcolor[rgb]{ .851,  .882,  .957}\textbf{85.2} & \cellcolor[rgb]{ .851,  .882,  .957}\textbf{67.1} & \cellcolor[rgb]{ .851,  .882,  .957}\textbf{29.1} \\ 
     \bottomrule
    \end{tabular}
    }
    \label{tab: mllm results}
\end{table*}

\begin{figure*}[t]
\vspace{-10pt}
\centering
\subfloat[Stable Diffusion v1.5]{
\begin{minipage}{0.25\linewidth}
\includegraphics[width=\linewidth]{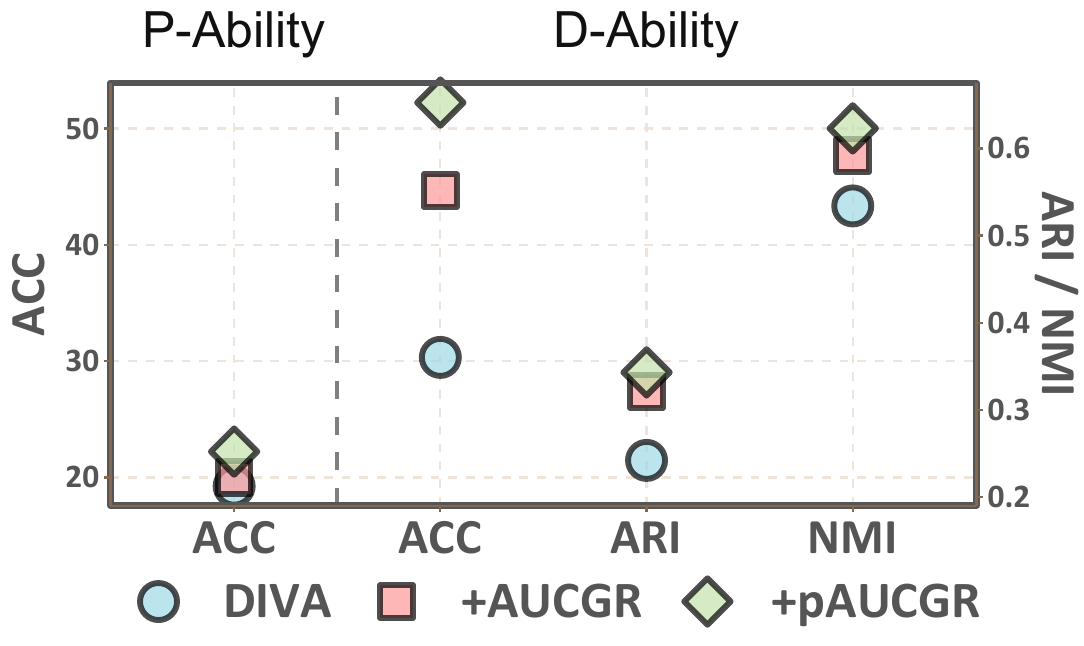}  
\end{minipage}
}
\hfil
\subfloat[Stable Diffusion v2.1]{   
\begin{minipage}{0.25\linewidth}
\includegraphics[width=\linewidth]{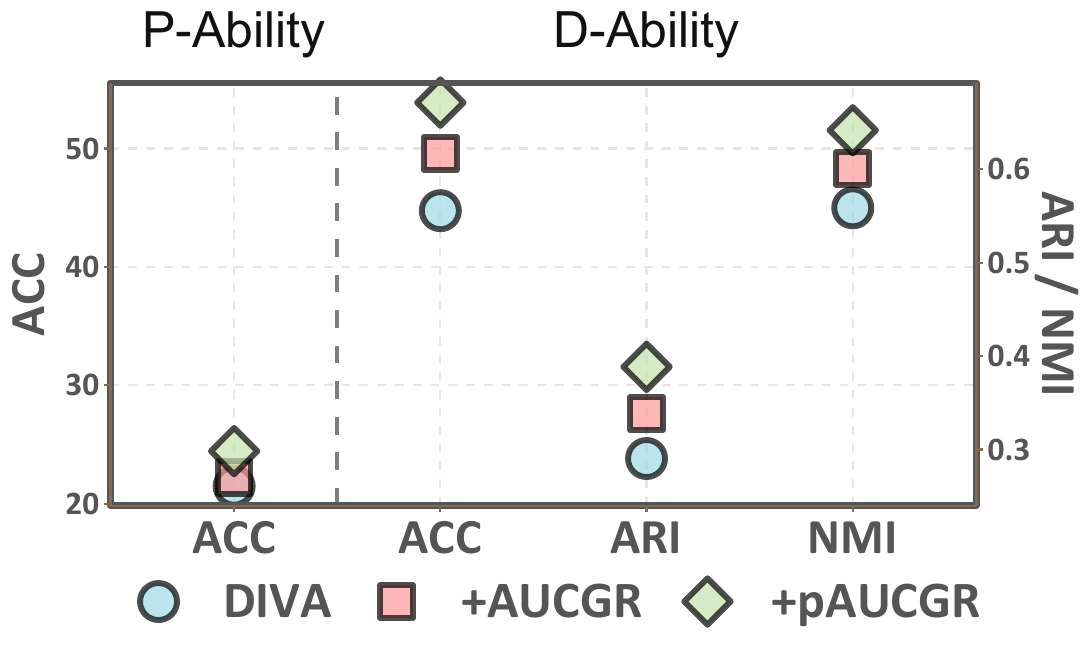} 
\end{minipage}
}
\hfil
\subfloat[Rectified Flow]{   
\begin{minipage}{0.25\linewidth}
\includegraphics[width=\linewidth]{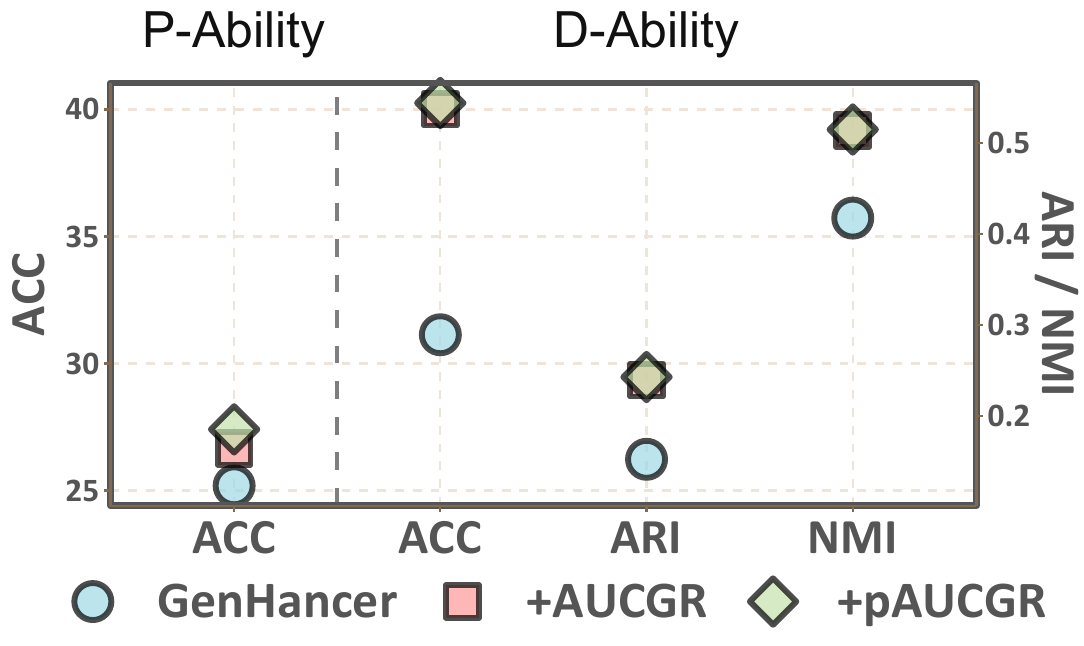}
\end{minipage}
}
\caption{\textbf{Ablation study on different generative model structures.} We evaluate our method with Stable Diffusion v1.5, Stable Diffusion v2.1, and Rectified Flow as the generative reconstruction backbone. All experiments use OpenAI CLIP ViT-L@224 as the visual backbone. The results show that our framework is not tied to a specific generative architecture and consistently improves both \textbf{\textcolor{PerColor}{P-Ability}} and \textbf{\textcolor{DisColor}{D-Ability}}.}
\label{fig: ablation_studies_generative_model}
\vspace{-15pt}
\end{figure*}

\begin{figure}[t]
\vspace{-10pt}
\centering
\subfloat[DIVA]{
\begin{minipage}{0.35\linewidth}
\includegraphics[width=\linewidth]{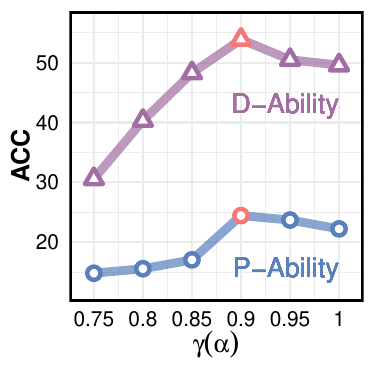}  
\end{minipage}
}
\hfil
\subfloat[GenHancer]{   
\begin{minipage}{0.35\linewidth}
\includegraphics[width=\linewidth]{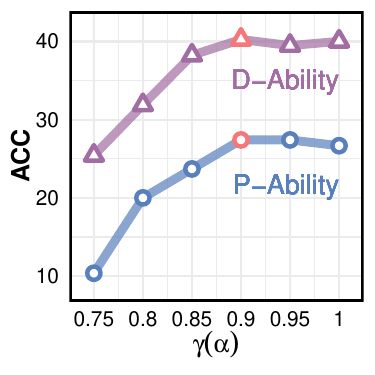} 
\end{minipage}
}
\caption{\textbf{Ablation study on the pAUCGR parameter.} We vary the retained-negative threshold $\gamma(\alpha)$. All experiments use OpenAI CLIP ViT-L@224 as the visual backbone.}
\label{fig: ablation_studies_gamma_alpha}
\end{figure}

\vskip 1 ex
\noindent \textbf{Results on Different Generative Model Structures.} \cref{fig: ablation_studies_generative_model} evaluates our method with different generative reconstruction backbones, including diffusion models and flow-based models. Specifically, we test Stable Diffusion v1.5, Stable Diffusion v2.1, and Rectified Flow. The results show that our plug-in consistently improves both \textbf{\textcolor{DisColor}{D-Ability}} and \textbf{\textcolor{PerColor}{P-Ability}} across these generative structures. This indicates that the proposed optimization is not tied to a specific architecture, but can be applied to different generative reconstruction models in a plug-and-play manner.

\vskip 1 ex
\noindent \textbf{Ablation Study on Hyper-Parameter $\gamma(\alpha)$ in pAUCGR.} The hyper-parameter $\gamma(\alpha)$ controls which negatives are filtered in pAUCGR. A larger value keeps more high-similarity negatives, but it may fail to remove noisy negatives that share similar visual or semantic cues with the anchor. A smaller value removes more negatives, but it may also discard useful mismatched samples that provide valid comparison signals. As shown in \cref{fig: ablation_studies_gamma_alpha}, both extremes lead to suboptimal performance. In our experiments, $\gamma(\alpha)=0.9$ provides a better balance between removing noisy negatives and retaining reliable negative comparisons. We therefore use this setting as the default choice.

\vskip 1 ex
Due to space limitations, experiments validating negative filtering and robustness to surrogate choices and random seeds are presented in \cref{subsec: supp_Expanded Ablation Studies}. The computational cost analysis is provided in \cref{subsec: supp_Computational Costs}.

\section{Conclusion}
\label{sec: conclusion}
In this paper, we focus on balanced visual representation learning by jointly considering semantic discrimination and fine-grained perception. Our analysis shows that the balance cannot be reliably achieved by a naive weighted combination of contrastive learning and generative reconstruction, because the resulting optimization is structurally conflicted. To address this issue, we formulate generative reconstruction as a comparison-based optimization problem and instantiate it with AUCGR. We further introduce pAUCGR to reduce biased supervision from noisy negatives through one-way partial AUC. This formulation provides a plug-and-play optimization module for existing generative reconstruction methods. Theoretical results and empirical studies consistently support the proposed framework and demonstrate its effectiveness across CLIP backbones, visual benchmarks, and multimodal large language models.

\section*{Acknowledgments}
This work was supported in part by National Natural Science Foundation of China: 62525212, U23B2051, 62236008, 62441232, 62521007, U21B2038, 62502496, and 62576332, in part by Youth Innovation Promotion Association CAS, in part by the Strategic Priority Research Program of the Chinese Academy of Sciences, Grant No. XDB0680201, in part by the project ZR2025ZD01 supported by Shandong Provincial Natural Science Foundation, in part by the Young Elite Scientists Sponsorship Program of the Beijing High Innovation Plan No. 20250687, and in part by Beijing Natural Science Foundation-Xiaomi Innovation Joint Fund (Grant No. 26L030072).

\bibliographystyle{IEEEtran}
\bibliography{main}

\clearpage
\onecolumn
\appendices
\definecolor{app_blue}{RGB}{0,20,115}
\textcolor{white}{dasdsa}

\section*{\textcolor{app_blue}{\Large{Contents}}}

\titlecontents{section}
[0em]
{\color{app_blue}\bfseries}
{\thecontentslabel. }
{}
{\hfill\contentspage}
[\vspace{0.6em}]

\titlecontents{subsection}
[1.5em]
{\color{app_blue}}
{\thecontentslabel. }
{}
{\titlerule*[0.75em]{.}\contentspage}
[\vspace{0.35em}]

\startcontents[sections]

\printcontents[sections]{1}{1}{\setcounter{tocdepth}{3}}

\newpage

\section{Symbol Definitions}
\label{sec: supp_Symbol Definitions}

In this section, \Cref{tab: symbols} summarizes the key notations used in the paper.

\begin{table}[htbp]
    \centering
    \renewcommand{\arraystretch}{1.2}
    \caption{A summary of key notations and descriptions in the paper.}
    \resizebox{\linewidth}{!}{
    \begin{tabular}{p{0.28\linewidth}p{0.66\linewidth}}
        \toprule
        \textbf{Notation} & \textbf{Description} \\
        \midrule
        $\x\in\mathbb{R}^{H\times W\times 3}$ & Input image with height $H$ and width $W$. \\
        $\mathcal{D}=\{\x_i\}_{i=1}^{n}$ & Training dataset. \\
        $\tilde{\x}$, $\tilde{\x}_t$ & VAE latent of $\x$ and its noisy latent at step $t$. \\
        $t$ & Diffusion or flow timestep. \\
        $f_{\phi}$ & visual encoder. \\
        $\z=f_{\phi}(\x)$ & Visual representation extracted by $f_{\phi}$. \\
        $h_{\omega}$ & Projection module that maps visual features into the generative condition space. \\
        $c=h_{\omega}(\z)$ & Conditional input for generative reconstruction. \\
        $\epsilon_{\theta}$ & Conditional generative prediction network. \\
        $\hat{\boldsymbol\epsilon}_{\theta}=\epsilon_{\theta}(\tilde{\x}_t,c,t)$ & Predicted noise or velocity response produced by the generative model. \\
        $\boldsymbol\epsilon_t^{\mathrm{gt}}$ & Ground-truth noise or velocity target at step $t$. \\
        $\mathcal{L}_{\mathrm{con}}$, $\mathcal{L}_{\mathrm{rec}}$, $\mathcal{L}_{\mathrm{joint}}$ & Contrastive loss, reconstruction loss, and naive weighted joint objective. \\
        $Z$ & Normalized representation used in the signal decomposition analysis. \\
        $z_D$, $z_P$, $z_N$ & Discriminative component, fine-grained perceptual component, and noise component. \\
        $\alpha_D$, $\alpha_P$, $\alpha_N$ & Energy proportions of $z_D$, $z_P$, and $z_N$ in $Z$. \\
        $S_{\text{inner}}$, $S_{\text{inter}}$ & Intra-class and inter-class scatters in the feature space. \\
        $S_{\text{inner}}^{(\epsilon)}(t)$, $S_{\text{inter}}^{(\epsilon)}(t)$ & Intra-class and inter-class scatters in the predicted-noise space at step $t$. \\
        $\hat{\boldsymbol\epsilon}_i$ & Anchor predicted noise for image $\x_i$. \\
        $\mathcal{P}_i$, $\mathcal{N}_i$ & Matched and mismatched comparison sets for anchor $i$. \\
        $d(\cdot,\cdot)$ & Comparison measure between predicted-noise responses. \\
        $q_i(v)=d(\hat{\boldsymbol\epsilon}_i,v)$ & Anchor-wise comparison score for $v\in\mathcal{P}_i\cup\mathcal{N}_i$. \\
        $\ell(\cdot)$ & Decreasing pairwise surrogate loss for the ordering gap $q_i(p)-q_i(n)$. \\
        $\Psi$ & Anchor-wise comparison functional in the generic comparison objective. \\
        $\mathcal{L}_{\mathrm{AUCGR}}$ & AUC Generative Reconstruction loss. \\
        $\alpha$ & Truncation ratio used in pAUCGR. \\
        $\gamma(\alpha)$ & Negative-score quantile threshold. \\
        $\mathcal{N}_i^{(\alpha)}$ & Retained negative set after one-way negative truncation. \\
        $\mathcal{L}_{\mathrm{pAUCGR}}$ & Partial AUC Generative Reconstruction loss. \\
        $\mathcal{L}_{\mathrm{DCR}}$ & Contrastive generative reconstruction loss used in the conference version. \\
        $f\in\mathcal{F}$ & Comparison model induced by the learnable modules in the generalization analysis. \\
        $z=(\hat{\boldsymbol\epsilon},\mathcal{P},\mathcal{N}^{-})$ & Anchor tuple for the pAUCGR population risk. \\
        $g_f^{(\alpha)}(z)$ & Anchor-wise pAUCGR loss for model $f$ and tuple $z$. \\
        $\mathcal{R}_{\alpha}(f)$, $\hat{\mathcal{R}}_{\alpha}(f)$ & Population risk and empirical risk of pAUCGR. \\
        $B$ & Mini-batch size. \\
        $N$ & The number of samples. \\
        \bottomrule
    \end{tabular}
    } \label{tab: symbols}
\end{table}

\newpage

\section{Algorithms of AUCGR and pAUCGR}
\label{sec: supp_Algorithms of AUCGR and pAUCGR}

For clarity, we provide the plug-and-play training procedures of AUCGR and pAUCGR in Algorithms~\ref{alg: aucgr} and \ref{alg: paucgr}. Both algorithms only require an existing generative reconstruction framework to produce an anchor response, a matched set, and a mismatched set. Therefore, they can be inserted into different reconstruction-based enhancement pipelines without assuming a specific generative architecture.

\begin{algorithm}[htbp]
   \caption{AUCGR Algorithm.}
   \label{alg: aucgr}
   \begin{algorithmic}[1]
   \REQUIRE Dataset $\mathcal{D}$; existing generative reconstruction framework with trainable modules $\Theta$; comparison measure $d$; decreasing surrogate loss $\ell$; batch size $B$; training iterations $K$.
   \ENSURE Enhanced visual encoder in the base framework.
   \FOR{$k = 1$ \textbf{to} $K$}
      \STATE Sample a mini-batch $\{\x_i\}_{i=1}^{B}$ from $\mathcal{D}$.
      \FOR{$i = 1$ \textbf{to} $B$}
         \STATE Use the base framework to obtain the anchor response $\hat{\boldsymbol\epsilon}_i$.
         \STATE Build the positive set $\mathcal{P}_i$ and negative set $\mathcal{N}_i$ using the base framework.
         \FOR{each $v\in\mathcal{P}_i\cup\mathcal{N}_i$}
            \STATE Compute the anchor-wise score $q_i(v)\gets d(\hat{\boldsymbol\epsilon}_i,v)$.
         \ENDFOR
         \STATE Compute the AUCGR anchor loss
         \[
         \mathcal{L}_{\mathrm{AUCGR}}^{(i)}
         \gets
         \frac{1}{|\mathcal{P}_i||\mathcal{N}_i|}
         \sum_{p\in\mathcal{P}_i}\sum_{n\in\mathcal{N}_i}
         \ell\bigl(q_i(p)-q_i(n)\bigr).
         \]
      \ENDFOR
      \STATE $\mathcal{L}_{\mathrm{AUCGR}}\gets\frac{1}{B}\sum_{i=1}^{B}\mathcal{L}_{\mathrm{AUCGR}}^{(i)}$.
      \STATE Update the trainable modules $\Theta$.
   \ENDFOR
   \STATE \textbf{return} the enhanced visual encoder.
   \end{algorithmic}
\end{algorithm}

\begin{algorithm}[htbp]
   \caption{pAUCGR Algorithm.}
   \label{alg: paucgr}
   \begin{algorithmic}[1]
   \REQUIRE Dataset $\mathcal{D}$; existing generative reconstruction framework with trainable modules $\Theta$; comparison measure $d$; decreasing surrogate loss $\ell$; threshold $\gamma(\alpha)$; batch size $B$; training iterations $K$.
   \ENSURE Enhanced visual encoder in the base framework.
   \FOR{$k = 1$ \textbf{to} $K$}
      \STATE Sample a mini-batch $\{\x_i\}_{i=1}^{B}$ from $\mathcal{D}$.
      \FOR{$i = 1$ \textbf{to} $B$}
         \STATE Use the base framework to obtain the anchor response $\hat{\boldsymbol\epsilon}_i$.
         \STATE Build the positive set $\mathcal{P}_i$ and candidate negative set $\mathcal{N}_i$ using the base framework.
         \FOR{each $v\in\mathcal{P}_i\cup\mathcal{N}_i$}
            \STATE Compute the anchor-wise score $q_i(v)\gets d(\hat{\boldsymbol\epsilon}_i,v)$.
         \ENDFOR
         \STATE Retain reliable negatives $\mathcal{N}_i^{(\alpha)}\gets\{n\in\mathcal{N}_i:q_i(n)\le\gamma(\alpha)\}$.
         \STATE Compute the pAUCGR anchor loss
         \[
         \mathcal{L}_{\mathrm{pAUCGR}}^{(i)}
         \gets
         \frac{1}{|\mathcal{P}_i|\,|\mathcal{N}_i^{(\alpha)}|}
         \sum_{p\in\mathcal{P}_i}\sum_{n\in\mathcal{N}_i^{(\alpha)}}
         \ell\bigl(q_i(p)-q_i(n)\bigr).
         \]
      \ENDFOR
      \STATE $\mathcal{L}_{\mathrm{pAUCGR}}\gets\frac{1}{B}\sum_{i=1}^{B}\mathcal{L}_{\mathrm{pAUCGR}}^{(i)}$.
      \STATE Update the trainable modules.
   \ENDFOR
   \STATE \textbf{return} the enhanced visual encoder.
   \end{algorithmic}
\end{algorithm}

\newpage
\section{Proof of Theorem 1}
\label{sec: supp_Proof of Theorem 1}

\begin{assmp}[Three-Component Orthogonality]
Let
\begin{equation}
Z
=
\sqrt{\alpha_D}\,z_D
+\sqrt{\alpha_P}\,z_P
+\sqrt{\alpha_N}\,z_N,
\qquad
Z^{+}
=
\sqrt{\alpha_D}\,z_D
+\sqrt{\alpha_P}\,z_P^{+}
+\sqrt{\alpha_N}\,z_N^{+},
\label{eq: naive_positive_pair}
\end{equation}
where $z_D,z_P,z_N,z_P^{+},z_N^{+}\in\mathbb{S}^{d-1}$, $\alpha_D,\alpha_P,\alpha_N\ge0$, and $\alpha_D+\alpha_P+\alpha_N=1$. We impose the following conditions.
\begin{enumerate}[label=(\roman*)]
\item $z_D$, $z_P$, and $z_N$ are mutually independent and orthogonal.
\item The positive-view perceptual and noise components are conditionally centered in the sense that
\begin{equation}
\mathbb{E}\bigl[z_P^{+}\mid z_D,z_P,z_N\bigr]=0,
\qquad
\mathbb{E}\bigl[z_N^{+}\mid z_D,z_P,z_N\bigr]=0.
\label{eq: detail_zero_mean}
\end{equation}
\item Conditioned on $Z$, the component of $z_P$ orthogonal to $Z$ is isotropic in $Z^{\perp}$.
\end{enumerate}
\label{ass: naive_orthogonality}
\end{assmp}

\begin{assmp}[Recoverability of the Reconstruction Target]
For each diffusion step $t$, there exist constants $\lambda,\eta>0$ and a measurable function $r_t(\cdot)$ such that
\begin{equation}
\boldsymbol\epsilon_t^{\mathrm{gt}}
=
r_t(x_t,z_D,t)+\lambda z_P+\eta\sqrt{\alpha_N}\,z_N.
\label{eq: naive_noise_decomposition}
\end{equation}
Let $c=h_{\omega}(Z)$ and $U=(x_t,c,t)$. We assume that
\begin{equation}
\mathbb{E}\bigl[z_P\mid U\bigr]=\mathbb{E}\bigl[z_P\mid Z\bigr],
\label{eq: detail_sufficiency}
\end{equation}
and that the cross-noise residual is non-degenerate:
\begin{equation}
\mathbb{E}\Bigl\|z_N-\mathbb{E}\bigl[z_N\mid U\bigr]\Bigr\|_2^2\ge m_N
\label{eq: noise_non_degenerate}
\end{equation}
for some constant $m_N>0$. Define
\begin{equation}
\begin{aligned}
R_D&=r_t(x_t,z_D,t)-\mathbb{E}\bigl[r_t(x_t,z_D,t)\mid U\bigr],\\
R_P&=z_P-\mathbb{E}\bigl[z_P\mid U\bigr],\\
R_N&=z_N-\mathbb{E}\bigl[z_N\mid U\bigr].
\end{aligned}
\label{eq: naive_residual_terms}
\end{equation}
We assume that these three residuals are conditionally pairwise orthogonal:
\begin{equation}
\mathbb{E}\bigl[\langle R_a,R_b\rangle\mid U\bigr]=0,
\qquad
a\ne b,\quad a,b\in\{D,P,N\}.
\label{eq: residual_orthogonality}
\end{equation}
\label{ass: naive_recoverability}
\end{assmp}

\begin{lem}[Conditional Mean Minimizes Mean-Squared Error]
Let $Y$ be a random vector and let $U$ be an information variable. Then, for every $U$-measurable predictor $\widehat{Y}(U)$,
\begin{equation}
\mathbb{E}\bigl\|Y-\widehat{Y}(U)\bigr\|_2^2
\ge
\mathbb{E}\Bigl\|Y-\mathbb{E}\bigl[Y\mid U\bigr]\Bigr\|_2^2.
\label{eq: conditional_mean_optimality}
\end{equation}
\label{lem: conditional_mean_optimality}
\end{lem}

\begin{proof}
Let $m(U)=\mathbb{E}\bigl[Y\mid U\bigr]$. Then
\begin{equation}
Y-\widehat{Y}(U)
=
\bigl(Y-m(U)\bigr)
+
\bigl(m(U)-\widehat{Y}(U)\bigr).
\end{equation}
Taking squared norms gives
\begin{equation}
\begin{aligned}
\bigl\|Y-\widehat{Y}(U)\bigr\|_2^2
=&
\bigl\|Y-m(U)\bigr\|_2^2
+
\bigl\|m(U)-\widehat{Y}(U)\bigr\|_2^2 \\
&+
2\Bigl\langle
Y-m(U),\,
m(U)-\widehat{Y}(U)
\Bigr\rangle.
\end{aligned}
\end{equation}
We next take the conditional expectation with respect to $U$. Since $m(U)-\widehat{Y}(U)$ is $U$-measurable, we have
\begin{equation}
\begin{aligned}
\mathbb{E}\Bigl[
\Bigl\langle
Y-m(U),\,
m(U)-\widehat{Y}(U)
\Bigr\rangle
\Bigm| U
\Bigr]
=&
\Bigl\langle
\mathbb{E}\bigl[Y-m(U)\mid U\bigr],\,
m(U)-\widehat{Y}(U)
\Bigr\rangle \\
=&
\Bigl\langle
\mathbb{E}\bigl[Y\mid U\bigr]-m(U),\,
m(U)-\widehat{Y}(U)
\Bigr\rangle \\
=& 0.
\end{aligned}
\end{equation}
Therefore,
\begin{equation}
\mathbb{E}\Bigl[\bigl\|Y-\widehat{Y}(U)\bigr\|_2^2\Bigm| U\Bigr]
=
\mathbb{E}\Bigl[\bigl\|Y-m(U)\bigr\|_2^2\Bigm| U\Bigr]
+
\bigl\|m(U)-\widehat{Y}(U)\bigr\|_2^2.
\end{equation}
Taking the full expectation yields
\begin{equation}
\mathbb{E}\bigl\|Y-\widehat{Y}(U)\bigr\|_2^2
=
\mathbb{E}\bigl\|Y-m(U)\bigr\|_2^2
+
\mathbb{E}\bigl\|m(U)-\widehat{Y}(U)\bigr\|_2^2
\ge
\mathbb{E}\bigl\|Y-m(U)\bigr\|_2^2.
\end{equation}
This completes the proof.
\end{proof}

\begin{lem}[Perceptual Residual Under the Signal Decomposition]
Under Assumption~\ref{ass: naive_orthogonality}, the perceptual component admits the decomposition
\begin{equation}
z_P
=
\sqrt{\alpha_P}Z+z_{P,\perp},
\qquad
z_{P,\perp}\perp Z,
\qquad
\bigl\|z_{P,\perp}\bigr\|_2^2=1-\alpha_P.
\label{eq: f_orthogonal_decomposition}
\end{equation}
Moreover,
\begin{equation}
\mathbb{E}\bigl[z_P\mid Z\bigr]=\sqrt{\alpha_P}Z,
\qquad
\mathbb{E}\Bigl\|z_P-\mathbb{E}\bigl[z_P\mid Z\bigr]\Bigr\|_2^2=1-\alpha_P.
\label{eq: detail_residual_identity}
\end{equation}
\label{lem: naive_detail_residual}
\end{lem}

\begin{proof}
By \cref{eq: signal_decomposition} and Assumption~\ref{ass: naive_orthogonality},
\begin{equation}
\begin{aligned}
\langle z_P, Z\rangle
=&
\left\langle
z_P,\,
\sqrt{\alpha_D}\,z_D+\sqrt{\alpha_P}\,z_P+\sqrt{\alpha_N}\,z_N
\right\rangle \\
=&
\sqrt{\alpha_D}\,\langle z_P,z_D\rangle
+
\sqrt{\alpha_P}\langle z_P,z_P\rangle
+
\sqrt{\alpha_N}\langle z_P,z_N\rangle \\
=& \sqrt{\alpha_P}.
\end{aligned}
\end{equation}
Therefore, the projection of $z_P$ onto the direction $Z$ is $\sqrt{\alpha_P}Z$. Define
\begin{equation}
z_{P,\perp}=z_P-\sqrt{\alpha_P}Z.
\end{equation}
Then $z_{P,\perp}\perp Z$, and
\begin{equation}
z_P=\sqrt{\alpha_P}Z+z_{P,\perp}.
\end{equation}
Since $\|Z\|_2=\|z_P\|_2=1$ and $z_{P,\perp}\perp Z$, the Pythagorean identity gives
\begin{equation}
1=\|z_P\|_2^2=\|\sqrt{\alpha_P}Z\|_2^2+\|z_{P,\perp}\|_2^2=\alpha_P+\|z_{P,\perp}\|_2^2,
\end{equation}
which implies
\begin{equation}
\|z_{P,\perp}\|_2^2=1-\alpha_P.
\end{equation}

By item (iii) of Assumption~\ref{ass: naive_orthogonality}, conditioned on $Z$, the vector $z_{P,\perp}$ is isotropic in $Z^{\perp}$. Hence its conditional mean vanishes:
\begin{equation}
\mathbb{E}\bigl[z_{P,\perp}\mid Z\bigr]=0.
\end{equation}
Combining this relation with \cref{eq: f_orthogonal_decomposition} gives
\begin{equation}
\mathbb{E}\bigl[z_P\mid Z\bigr]
=
\mathbb{E}\bigl[\sqrt{\alpha_P}Z+z_{P,\perp}\mid Z\bigr]
=
\sqrt{\alpha_P}Z.
\end{equation}
Finally,
\begin{equation}
z_P-\mathbb{E}\bigl[z_P\mid Z\bigr]
=
z_P-\sqrt{\alpha_P}Z
=
z_{P,\perp},
\end{equation}
so
\begin{equation}
\mathbb{E}\Bigl\|z_P-\mathbb{E}\bigl[z_P\mid Z\bigr]\Bigr\|_2^2
=
\mathbb{E}\|z_{P,\perp}\|_2^2
=
1-\alpha_P.
\end{equation}
This completes the proof.
\end{proof}

\begin{lem}[Lower Envelope of the Contrastive Objective]
Under Assumption~\ref{ass: naive_orthogonality}, there exist constants $c_{\mathrm{con}}\in\mathbb{R}$ and $\kappa_P,\kappa_N>0$, which depend only on the temperature $\tau$ and the number $K$ of negatives, such that
\begin{equation}
\mathcal{L}_{\mathrm{con}}
\ge
c_{\mathrm{con}}+\kappa_P\alpha_P+\kappa_N\alpha_N.
\label{eq: contrastive_lower_envelope_appendix}
\end{equation}
\label{lem: naive_contrastive_lower}
\end{lem}

\begin{proof}
Consider one anchor together with one positive and $K$ negatives. Let
\begin{equation}
s^{+}=\mathrm{sim}(Z,Z^{+}),
\qquad
s_k^{-}=\mathrm{sim}(Z,Z_k^{-}),
\qquad
k=1,\dots,K.
\end{equation}
The corresponding InfoNCE term is
\begin{equation}
\ell_{\mathrm{con}}
=
-\log
\frac{\exp\bigl(s^{+}/\tau\bigr)}
{\exp\bigl(s^{+}/\tau\bigr)+\sum_{k=1}^{K}\exp\bigl(s_k^{-}/\tau\bigr)}.
\label{eq: anchor_contrastive_loss}
\end{equation}
Factoring out $\exp\bigl(s^{+}/\tau\bigr)$ from the denominator yields
\begin{equation}
\ell_{\mathrm{con}}
=
\log\left(
1+\sum_{k=1}^{K}
\exp\left(\frac{s_k^{-}-s^{+}}{\tau}\right)
\right).
\label{eq: contrastive_factorized}
\end{equation}
Since all similarities lie in $[-1,1]$, we have $s_k^{-}\ge -1$ for every $k$. Therefore,
\begin{equation}
\ell_{\mathrm{con}}
\ge
\log\left(
1+
K\exp\left(\frac{-1-s^{+}}{\tau}\right)
\right).
\label{eq: contrastive_first_lower}
\end{equation}

We next compute the positive similarity. By \cref{eq: naive_positive_pair},
\begin{equation}
\begin{aligned}
s^{+}
=&
\left\langle
\sqrt{\alpha_D}\,z_D+\sqrt{\alpha_P}\,z_P+\sqrt{\alpha_N}\,z_N,\,
\sqrt{\alpha_D}\,z_D+\sqrt{\alpha_P}\,z_P^{+}+\sqrt{\alpha_N}\,z_N^{+}
\right\rangle \\
=&
\alpha_D\langle z_D,z_D\rangle
+
\sqrt{\alpha_P\alpha_D}
\bigl(
\langle z_D,z_P^{+}\rangle
+
\langle z_P,z_D\rangle
\bigr)
+
\sqrt{\alpha_N\alpha_D}
\bigl(
\langle z_D,z_N^{+}\rangle
+
\langle z_N,z_D\rangle
\bigr) \\
&+
\alpha_P\langle z_P,z_P^{+}\rangle
+
\alpha_N\langle z_N,z_N^{+}\rangle \\
&+
\sqrt{\alpha_P\alpha_N}
\bigl(
\langle z_P,z_N^{+}\rangle
+
\langle z_N,z_P^{+}\rangle
\bigr).
\end{aligned}
\label{eq: positive_similarity_decomposition}
\end{equation}
Taking the conditional expectation with respect to $z_D,z_P,z_N$, and using \cref{eq: detail_zero_mean}, we obtain
\begin{equation}
\mathbb{E}\bigl[s^{+}\mid z_D,z_P,z_N\bigr]
=
\alpha_D
=
1-\alpha_P-\alpha_N.
\label{eq: positive_similarity_mean}
\end{equation}

Define
\begin{equation}
g(u)=\log\left(1+K\exp\left(\frac{-1-u}{\tau}\right)\right).
\end{equation}
Then \cref{eq: contrastive_first_lower} becomes $\ell_{\mathrm{con}}\ge g(s^{+})$. A direct calculation gives
\begin{equation}
g''(u)
=
\frac{K\exp\left(\frac{-1-u}{\tau}\right)}
{\tau^2\left(1+K\exp\left(\frac{-1-u}{\tau}\right)\right)^2}
>0,
\end{equation}
so $g$ is convex. Applying Jensen's inequality together with \cref{eq: positive_similarity_mean} gives
\begin{equation}
\begin{aligned}
\mathbb{E}\bigl[\ell_{\mathrm{con}}\mid z_D,z_P,z_N\bigr]
\ge&
\mathbb{E}\bigl[g(s^{+})\mid z_D,z_P,z_N\bigr] \\
\ge&
g\Bigl(\mathbb{E}\bigl[s^{+}\mid z_D,z_P,z_N\bigr]\Bigr) \\
=&
\log\left(
1+K\exp\left(\frac{-2+\alpha_P+\alpha_N}{\tau}\right)
\right).
\end{aligned}
\label{eq: contrastive_jensen}
\end{equation}
Set
\begin{equation}
C=K\exp\left(-\frac{2}{\tau}\right),
\qquad
h(x)=\log\bigl(1+Ce^x\bigr).
\end{equation}
Then \cref{eq: contrastive_jensen} becomes
\begin{equation}
\mathbb{E}\bigl[\ell_{\mathrm{con}}\mid z_D,z_P,z_N\bigr]\ge h\left(\frac{\alpha_P+\alpha_N}{\tau}\right).
\end{equation}

Since
\begin{equation}
h'(x)=\frac{Ce^x}{1+Ce^x},
\end{equation}
we have
\begin{equation}
h'(x)\ge h'(0)=\frac{C}{1+C}
\qquad
\text{for all } x\ge 0.
\end{equation}

Therefore,
\begin{equation}
\begin{aligned}
h(x)
=&
h(0)+\int_{0}^{x} h'(u)\,du \\
\ge&
h(0)+\frac{C}{1+C}x.
\end{aligned}
\end{equation}

Substituting $x=(\alpha_P+\alpha_N)/\tau$ gives
\begin{equation}
\mathbb{E}\bigl[\ell_{\mathrm{con}}\mid z_D,z_P,z_N\bigr]
\ge
\log(1+C)+\frac{C}{\tau(1+C)}(\alpha_P+\alpha_N).
\end{equation}

Finally, taking the full expectation over anchors yields
\begin{equation}
\mathcal{L}_{\mathrm{con}}
\ge
\log(1+C)+\frac{C}{\tau(1+C)}(\alpha_P+\alpha_N).
\end{equation}

Hence \cref{eq: contrastive_lower_envelope_appendix} holds with
\begin{equation}
c_{\mathrm{con}}=\log(1+C),
\qquad
\kappa_P=\kappa_N=\frac{C}{\tau(1+C)}.
\end{equation}

This completes the proof.
\end{proof}

\begin{lem}[Lower Envelope of the Reconstruction Objective]
Under Assumptions~\ref{ass: naive_orthogonality} and \ref{ass: naive_recoverability}, there exist constants $c_{\mathrm{rec}}\in\mathbb{R}$ and $\kappa_D,\kappa_N'>0$ such that
\begin{equation}
\mathcal{L}_{\mathrm{rec}}
\ge
c_{\mathrm{rec}}+\kappa_D(1-\alpha_P)+\kappa_N'\alpha_N.
\label{eq: reconstruction_lower_envelope_appendix}
\end{equation}
\label{lem: naive_reconstruction_lower}
\end{lem}

\begin{proof}
Recall from the main text that
\begin{equation}
\mathcal{L}_{\mathrm{rec}}
=
\mathbb{E}_{t}\Bigl[
\bigl\|\hat{\boldsymbol\epsilon}_{\theta}-\boldsymbol\epsilon_t^{\mathrm{gt}}\bigr\|_2^2
\Bigr].
\end{equation}
Let
\begin{equation}
U=(x_t,c,t),
\qquad
Y=\boldsymbol\epsilon_t^{\mathrm{gt}},
\qquad
\widehat{Y}(U)=\hat{\boldsymbol\epsilon}_{\theta}.
\end{equation}
By Lemma~\ref{lem: conditional_mean_optimality},
\begin{equation}
\mathcal{L}_{\mathrm{rec}}
\ge
\mathbb{E}\Bigl\|
\boldsymbol\epsilon_t^{\mathrm{gt}}
-
\mathbb{E}\bigl[\boldsymbol\epsilon_t^{\mathrm{gt}}\mid U\bigr]
\Bigr\|_2^2.
\label{eq: reconstruction_conditional_variance}
\end{equation}

We next expand the conditional residual. By \cref{eq: naive_noise_decomposition},
\begin{equation}
\boldsymbol\epsilon_t^{\mathrm{gt}}
-
\mathbb{E}\bigl[\boldsymbol\epsilon_t^{\mathrm{gt}}\mid U\bigr]
=
\Bigl(
r_t(x_t,z_D,t)-\mathbb{E}\bigl[r_t(x_t,z_D,t)\mid U\bigr]
\Bigr)
+
\lambda\Bigl(
z_P-\mathbb{E}\bigl[z_P\mid U\bigr]
\Bigr)
+
\eta\sqrt{\alpha_N}
\Bigl(
z_N-\mathbb{E}\bigl[z_N\mid U\bigr]
\Bigr).
\label{eq: reconstruction_residual_decomposition}
\end{equation}
Taking squared norms and then conditioning on $U$ gives
\begin{equation}
\begin{aligned}
&\mathbb{E}\Bigl[
\Bigl\|
\boldsymbol\epsilon_t^{\mathrm{gt}}
-
\mathbb{E}\bigl[\boldsymbol\epsilon_t^{\mathrm{gt}}\mid U\bigr]
\Bigr\|_2^2
\Bigm| U
\Bigr] \\
=&
\mathbb{E}\Bigl[
\Bigl\|
r_t(x_t,z_D,t)-\mathbb{E}\bigl[r_t(x_t,z_D,t)\mid U\bigr]
\Bigr\|_2^2
\Bigm| U
\Bigr] \\
&+
\lambda^2
\mathbb{E}\Bigl[
\Bigl\|
z_P-\mathbb{E}\bigl[z_P\mid U\bigr]
\Bigr\|_2^2
\Bigm| U
\Bigr] \\
&+
\eta^2\alpha_N
\mathbb{E}\Bigl[
\Bigl\|
z_N-\mathbb{E}\bigl[z_N\mid U\bigr]
\Bigr\|_2^2
\Bigm| U
\Bigr]
\end{aligned}
\end{equation}
where all cross terms vanish by \cref{eq: residual_orthogonality}. Hence
\begin{equation}
\mathbb{E}\Bigl[
\Bigl\|
\boldsymbol\epsilon_t^{\mathrm{gt}}
-
\mathbb{E}\bigl[\boldsymbol\epsilon_t^{\mathrm{gt}}\mid U\bigr]
\Bigr\|_2^2
\Bigm| U
\Bigr]
\ge
\lambda^2
\mathbb{E}\Bigl[
\Bigl\|
z_P-\mathbb{E}\bigl[z_P\mid U\bigr]
\Bigr\|_2^2
\Bigm| U
\Bigr]
+
\eta^2\alpha_N
\mathbb{E}\Bigl[
\Bigl\|
z_N-\mathbb{E}\bigl[z_N\mid U\bigr]
\Bigr\|_2^2
\Bigm| U
\Bigr].
\end{equation}
Taking the full expectation and using \cref{eq: detail_sufficiency} yields
\begin{equation}
\mathcal{L}_{\mathrm{rec}}
\ge
\lambda^2
\mathbb{E}\Bigl\|
z_P-\mathbb{E}\bigl[z_P\mid Z\bigr]
\Bigr\|_2^2
+
\eta^2\alpha_N
\mathbb{E}\Bigl\|
z_N-\mathbb{E}\bigl[z_N\mid U\bigr]
\Bigr\|_2^2.
\end{equation}
Combining Lemma~\ref{lem: naive_detail_residual} with \cref{eq: noise_non_degenerate} then implies
\begin{equation}
\mathcal{L}_{\mathrm{rec}}
\ge
\lambda^2(1-\alpha_P)+\eta^2m_N\alpha_N.
\end{equation}
Therefore, \cref{eq: reconstruction_lower_envelope_appendix} holds with
\begin{equation}
c_{\mathrm{rec}}=0,
\qquad
\kappa_D=\lambda^2,
\qquad
\kappa_N'=\eta^2m_N.
\end{equation}
This completes the proof.
\end{proof}

\begin{rthm1}
Under Assumptions~\ref{ass: naive_orthogonality} and~\ref{ass: naive_recoverability}, consider the representation model in Definition~\ref{def: signal_decomposition}.
There exist constants $c_{\mathrm{con}},c_{\mathrm{rec}}\in\mathbb{R}$ and $\kappa_P,\kappa_N,\kappa_D,\kappa_N'>0$.
With these constants, the achievable envelopes of the two objectives satisfy
\begin{equation}
\begin{aligned}
&\inf \mathcal{L}_{\mathrm{con}}
\ge
c_{\mathrm{con}}+\kappa_P\alpha_P+\kappa_N\alpha_N,\\
&\inf \mathcal{L}_{\mathrm{rec}}
\ge
c_{\mathrm{rec}}+\kappa_D(1-\alpha_P)+\kappa_N'\alpha_N,
\end{aligned}
\end{equation}
where the infimum is taken over all admissible model parameters, optimizing each objective with the shared representation $Z$.
This result implies that the two objectives favor different allocation regimes:
\begin{itemize}
    \item Minimizing $\mathcal{L}_{\mathrm{con}}$ encourages \textbf{smaller} $\alpha_P$.
    \item Minimizing $\mathcal{L}_{\mathrm{rec}}$ encourages \textbf{larger} $\alpha_P$.
\end{itemize}
\end{rthm1}

\begin{proof}
Lemma~\ref{lem: naive_contrastive_lower} shows that every admissible choice of model parameters satisfies
\begin{equation}
\mathcal{L}_{\mathrm{con}}
\ge
c_{\mathrm{con}}+\kappa_P\alpha_P+\kappa_N\alpha_N
\end{equation}
for some constants $c_{\mathrm{con}}\in\mathbb{R}$ and $\kappa_P,\kappa_N>0$.
These constants do not depend on the particular parameter choice.
Taking the infimum over all admissible parameters preserves this lower bound, and therefore
\begin{equation}
\inf \mathcal{L}_{\mathrm{con}}
\ge
c_{\mathrm{con}}+\kappa_P\alpha_P+\kappa_N\alpha_N.
\end{equation}

Lemma~\ref{lem: naive_reconstruction_lower} similarly gives
\begin{equation}
\mathcal{L}_{\mathrm{rec}}
\ge
c_{\mathrm{rec}}+\kappa_D(1-\alpha_P)+\kappa_N'\alpha_N
\end{equation}
for constants $c_{\mathrm{rec}}\in\mathbb{R}$ and $\kappa_D,\kappa_N'>0$ that are independent of the admissible parameter choice. Taking the infimum again yields
\begin{equation}
\inf \mathcal{L}_{\mathrm{rec}}
\ge
c_{\mathrm{rec}}+\kappa_D(1-\alpha_P)+\kappa_N'\alpha_N.
\end{equation}

The first lower envelope is an increasing function of $\alpha_P$.
Therefore, minimizing $\mathcal{L}_{\mathrm{con}}$ favors smaller $\alpha_P$.
The second lower envelope is a decreasing function of $\alpha_P$.
Therefore, minimizing $\mathcal{L}_{\mathrm{rec}}$ favors larger $\alpha_P$.
The two objectives thus prefer incompatible allocations of the shared representation.
This is the structural source of the objective conflict in the naive method.

This completes the proof.
\end{proof}

\section{Proof of Theorem 2}
\label{sec: supp_Proof of Theorem 2}

\begin{defn}[Bi-Lipschitz Mapping~\cite{david2016bi}]
Let $T:\mathcal{X} \to \mathcal{Y}$ be a mapping between two metric spaces $(\mathcal{X}, \|\cdot\|_{\mathcal{X}})$ and $(\mathcal{Y}, \|\cdot\|_{\mathcal{Y}})$. We say that $T$ is bi-Lipschitz on a subset $\mathcal{M} \subseteq \mathcal{X}$ if there exist constants $0 < m \le L < \infty$ such that, for all $z_1, z_2 \in \mathcal{M}$,
\begin{equation}
m\|z_1 - z_2\|_{\mathcal{X}} \le \bigl\|T(z_1) - T(z_2)\bigr\|_{\mathcal{Y}} \le L \|z_1 - z_2\|_{\mathcal{X}}.
\end{equation}
In particular, a bi-Lipschitz map preserves pairwise distances up to constant factors and is injective on $\mathcal{M}$.
\label{defn: bilip_aucgr}
\end{defn}

\begin{assmp}
Fix a diffusion step $t$ and define $T(\z)=\epsilon_{\theta}\bigl(\tilde{\x}_{t},\,h_{\omega}(\z),\,t\bigr)$. According to Definition \ref{defn: bilip_aucgr}, we assume that $T(\z)$ satisfies the bi-Lipschitz property on a subset that contains all feature samples and their class means.
\label{assmp: bilip_aucgr}
\end{assmp}

\begin{lem}[Equivalent Variance Identity~\cite{zhang2012some}]
Let $e_1,\dots,e_n \in \mathbb{R}^d$ be arbitrary vectors and denote their mean by $\bar{e} = \frac{1}{n}\sum_{i=1}^n e_i$. Then the empirical variance can be written as the average of all pairwise squared distances:
\begin{equation}
\sum_{i=1}^n \|e_i - \bar{e}\|_2^2 = \frac{1}{2n}\sum_{i=1}^n \sum_{j=1}^n \|e_i - e_j\|_2^2.
\label{eq: var_pair_aucgr}
\end{equation}
\label{lem: equivalent_variance_identity_aucgr}
\end{lem}

\begin{proof}
\textbf{For the left-hand side}, using $\bar{e} = \frac{1}{n}\sum_{k=1}^n e_k$, we have:
\begin{equation}
\begin{aligned}
\sum_{i=1}^n \|e_i - \bar{e}\|_2^2
&= \sum_{i=1}^n \langle e_i - \bar{e}, e_i - \bar{e} \rangle \\
&= \sum_{i=1}^n \bigl(\|e_i\|_2^2 + \|\bar{e}\|_2^2 - 2\langle e_i, \bar{e} \rangle \bigr) \\
&= \sum_{i=1}^n \|e_i\|_2^2 + n\|\bar{e}\|_2^2 - 2\left\langle \sum_{i=1}^n e_i, \bar{e} \right\rangle.
\end{aligned}
\end{equation}

Since $\sum_{i=1}^n e_i = n\bar{e}$, it follows that
\begin{equation}
\left\langle \sum_{i=1}^n e_i, \bar{e} \right\rangle = \langle n\bar{e}, \bar{e} \rangle = n\|\bar{e}\|_2^2.
\end{equation}

Therefore,
\begin{equation}
\sum_{i=1}^n \|e_i - \bar{e}\|_2^2 = \sum_{i=1}^n \|e_i\|_2^2 - n\|\bar{e}\|_2^2.
\label{eq: var_left_aucgr}
\end{equation}

\textbf{For the right-hand side}, using $\|e_i - e_j\|_2^2 = \|e_i\|_2^2 + \|e_j\|_2^2 - 2\langle e_i, e_j\rangle$, we obtain
\begin{equation}
\begin{aligned}
&\sum_{i=1}^n \sum_{j=1}^n \|e_i - e_j\|_2^2 \\
=& \sum_{i,j} \|e_i\|_2^2 + \sum_{i,j} \|e_j\|_2^2 - 2\sum_{i,j} \langle e_i, e_j\rangle \\
=& n \sum_{i=1}^n \|e_i\|_2^2 + n \sum_{j=1}^n \|e_j\|_2^2 - 2 \left\langle \sum_{i=1}^n e_i,\ \sum_{j=1}^n e_j \right\rangle \\
=& 2n \sum_{i=1}^n \|e_i\|_2^2 - 2 \left\|\sum_{i=1}^n e_i\right\|_2^2.
\end{aligned}
\end{equation}

Again using $\sum_{i=1}^n e_i = n\bar{e}$, we obtain
\begin{equation}
\left\|\sum_{i=1}^n e_i\right\|_2^2 = \|n\bar{e}\|_2^2 = n^2\|\bar{e}\|_2^2,
\end{equation}
so
\begin{equation}
\sum_{i=1}^n \sum_{j=1}^n \|e_i - e_j\|_2^2 = 2n \sum_{i=1}^n \|e_i\|_2^2 - 2n^2\|\bar{e}\|_2^2.
\label{eq: var_right_aucgr}
\end{equation}

Dividing \cref{eq: var_right_aucgr} by $2n$ yields
\begin{equation}
\frac{1}{2n} \sum_{i=1}^n \sum_{j=1}^n \|e_i - e_j\|_2^2
= \sum_{i=1}^n \|e_i\|_2^2 - n\|\bar{e}\|_2^2.
\label{eq: var_right_2n_aucgr}
\end{equation}

Therefore, by combining \cref{eq: var_left_aucgr} and \cref{eq: var_right_2n_aucgr}, we can obtain:
\begin{equation}
\sum_{i=1}^n \|e_i - \bar{e}\|_2^2 = \frac{1}{2n}\sum_{i=1}^n \sum_{j=1}^n \|e_i - e_j\|_2^2.
\end{equation}

This completes the proof.
\end{proof}

\begin{rthm2}[AUCGR Supports \textbf{\textcolor{DisColor}{D-Ability}}]
Fix a diffusion step $t$ and define $T(\z)=\epsilon_{\theta}\bigl(\tilde{\x}_{t},\,h_{\omega}(\z),\,t\bigr)$. Under Assumption~\ref{assmp: bilip_aucgr}, the intra-class scatter and inter-class scatter in the feature space $(S_{\text{inner}},S_{\text{inter}})$ can be bounded by those in the noise space $(S_{\text{inner}}^{(\epsilon)}(t),S_{\text{inter}}^{(\epsilon)}(t))$ as
\begin{equation*}
S_{\text{inner}}\le\frac{1}{m^2}S_{\text{inner}}^{(\epsilon)}(t),
\end{equation*}
\begin{equation*}
S_{\text{inter}}\ge \kappa S_{\text{inter}}^{(\epsilon)}(t)-\eta S_{\text{inner}}^{(\epsilon)}(t),
\end{equation*}
where $m$, $\kappa$, and $\eta$ are positive constants depending on the Lipschitz continuity of $T(\cdot)$.
\end{rthm2}

\begin{proof}
We proceed in three steps. Step 1 identifies the pairwise ordering target induced by AUCGR. Steps 2 and 3 establish the two scatter bounds stated in the theorem. Throughout, fix a diffusion step $t$. We also fix a mini-batch with class set $\mathcal Y$.

For class $y\in\mathcal Y$, let the feature set be $\mathcal Z_y=\{\z_i\}_{i=1}^{n_y}$ with $\z_i=f_{\phi}(\x_i)$ and the noise set be $\mathcal E_{y,t}=\{\hat{\boldsymbol\epsilon}_i\}_{i=1}^{n_y}$ with $\hat{\boldsymbol\epsilon}_i=\epsilon_{\theta}(\tilde{\x}_{t},h_{\omega}(\z_i),t)$. Denote the class means by $\mu_y=\frac{1}{n_y}\sum_{i=1}^{n_y}\z_i$ and $\nu_{y,t}=\frac{1}{n_y}\sum_{i=1}^{n_y}\hat{\boldsymbol\epsilon}_i$. As shown in \cref{eq: s_inner,eq: s_inter} in the main text, we define the intra-/inter-class scatters in the CLIP feature space by
\begin{equation}
S_{\text{inner}}=\frac{1}{|\mathcal Y|}\sum_{y\in\mathcal Y}\frac{1}{n_y}\sum_{i=1}^{n_y}\|\z_i-\mu_y\|_2^2,
\label{eq: feature_inner_aucgr}
\end{equation}
\begin{equation}
S_{\text{inter}}=\frac{1}{|\mathcal Y|(|\mathcal Y|-1)}\sum_{\substack{y,y'\in\mathcal Y\\y\ne y'}}\|\mu_y-\mu_{y'}\|_2^2.
\label{eq: feature_inter_aucgr}
\end{equation}
Similarly, we define the intra-/inter-class scatters in the diffusion noise space by
\begin{equation}
S_{\text{inner}}^{(\epsilon)}(t)=\frac{1}{|\mathcal Y|}\sum_{y\in\mathcal Y}\frac{1}{n_y}\sum_{i=1}^{n_y}\|\hat{\boldsymbol\epsilon}_i-\nu_{y,t}\|_2^2,
\label{eq: noise_inner_aucgr}
\end{equation}
\begin{equation}
S_{\text{inter}}^{(\epsilon)}(t)=\frac{1}{|\mathcal Y|(|\mathcal Y|-1)}\sum_{\substack{y,y'\in\mathcal Y\\y\ne y'}}\|\nu_{y,t}-\nu_{y',t}\|_2^2.
\label{eq: noise_inter_aucgr}
\end{equation}

\vskip 1 ex
\noindent \textbf{Step 1: AUCGR induces an ordering target in the predicted-noise space.}
\vskip 1 ex

Recall the anchor-wise AUCGR term
\begin{equation}
\mathcal{L}_{\mathrm{AUCGR}}^{(i)}
=
\frac{1}{|\mathcal{P}_i||\mathcal{N}_i|}
\sum_{p\in\mathcal{P}_i}\sum_{n\in\mathcal{N}_i}
\ell\bigl(q_i(p)-q_i(n)\bigr),
\label{eq: aucgr_anchor_loss_appendix}
\end{equation}
where $q_i(v)=\mathrm{sim}(\hat{\boldsymbol\epsilon}_i,v)$ and $\mathrm{sim}(\cdot,\cdot)$ denotes cosine similarity. For a fixed anchor $i$, define the pairwise margin
\begin{equation}
\Delta_{ipn}:=q_i(p)-q_i(n),
\qquad
\forall p\in\mathcal P_i,\ \forall n\in\mathcal N_i.
\label{eq: aucgr_margin_appendix}
\end{equation}
Then \cref{eq: aucgr_anchor_loss_appendix} becomes
\begin{equation}
\mathcal{L}_{\mathrm{AUCGR}}^{(i)}
=
\frac{1}{|\mathcal{P}_i||\mathcal{N}_i|}
\sum_{p\in\mathcal{P}_i}\sum_{n\in\mathcal{N}_i}
\ell\bigl(\Delta_{ipn}\bigr).
\label{eq: aucgr_anchor_loss_margin_appendix}
\end{equation}
Since $\ell(\cdot)$ is decreasing, a smaller AUCGR loss is obtained by increasing the ordering gaps $\Delta_{ipn}$ in the active comparison pairs. Thus, an ordering-consistent solution satisfies
\begin{equation}
q_i(p)\ge q_i(n),
\qquad
\forall p\in\mathcal P_i,\ \forall n\in\mathcal N_i.
\label{eq: aucgr_ordering_appendix}
\end{equation}
Equivalently, AUCGR promotes the expected ordering
\begin{equation}
\mathbb{E}_{p\sim\mathcal P_i}\bigl[q_i(p)\bigr]
\ge
\mathbb{E}_{n\sim\mathcal N_i}\bigl[q_i(n)\bigr],
\label{eq: aucgr_expected_ordering_appendix}
\end{equation}
for the active comparisons optimized by the surrogate.

Since cosine similarity compares normalized vectors, for any nonzero $u,v$,
\begin{equation}
\mathrm{sim}(u,v)=\langle \bar u,\bar v\rangle,
\qquad
\bar u=\frac{u}{\|u\|_2},
\qquad
\bar v=\frac{v}{\|v\|_2},
\end{equation}
and hence
\begin{equation}
\|\bar u-\bar v\|_2^2 = 2\bigl(1-\mathrm{sim}(u,v)\bigr).
\label{eq: cosine_distance_identity_aucgr}
\end{equation}
Applying \cref{eq: cosine_distance_identity_aucgr} to anchor-positive and anchor-negative pairs yields
\begin{equation}
\|\bar{\hat{\boldsymbol\epsilon}}_i-\bar p\|_2^2
=
2\bigl(1-q_i(p)\bigr),
\qquad
\|\bar{\hat{\boldsymbol\epsilon}}_i-\bar n\|_2^2
=
2\bigl(1-q_i(n)\bigr).
\label{eq: aucgr_pair_distance_appendix}
\end{equation}
Hence the ordering relation in \cref{eq: aucgr_ordering_appendix} is equivalent to
\begin{equation}
\|\bar{\hat{\boldsymbol\epsilon}}_i-\bar p\|_2^2
\le
\|\bar{\hat{\boldsymbol\epsilon}}_i-\bar n\|_2^2,
\qquad
\forall p\in\mathcal P_i,\ \forall n\in\mathcal N_i.
\label{eq: aucgr_expected_pairwise_shift_appendix}
\end{equation}
In expectation, this gives
\begin{equation}
\mathbb{E}_{i,p}\bigl[\|\bar{\hat{\boldsymbol\epsilon}}_i-\bar p\|_2^2\bigr]
\le
\mathbb{E}_{i,n}\bigl[\|\bar{\hat{\boldsymbol\epsilon}}_i-\bar n\|_2^2\bigr].
\label{eq: aucgr_expected_pairwise_shift_batch_appendix}
\end{equation}
Thus, AUCGR explicitly encodes the desired ordering between matched and mismatched responses in the predicted-noise space. We next use the bi-Lipschitz property of $T(\cdot)$ to transfer the relevant scatter quantities to the feature space.

\vskip 1 ex
\noindent \textbf{Step 2: Bounding feature-space intra-class scatter.}
\vskip 1 ex

Fix a class $y\in\mathcal Y$, and write its feature set as $\mathcal Z_y=\{\z_i\}_{i=1}^{n_y}$ and its noise set as $\mathcal E_{y,t}=\{\hat{\boldsymbol\epsilon}_i\}_{i=1}^{n_y}$ with $\hat{\boldsymbol\epsilon}_i = T(\z_i)$.

By Lemma \ref{lem: equivalent_variance_identity_aucgr}, the intra-class variance in the feature space can be written as
\begin{equation}
\frac{1}{n_y}\sum_{i=1}^{n_y}\|\z_i-\mu_y\|_2^2 = \frac{1}{2n_y^2}\sum_{i=1}^{n_y}\sum_{j=1}^{n_y}\|\z_i-\z_j\|_2^2.
\label{eq: feature_var_pair_aucgr}
\end{equation}
Similarly, the intra-class variance in the noise space is
\begin{equation}
\frac{1}{n_y}\sum_{i=1}^{n_y}\|\hat{\boldsymbol\epsilon}_i-\nu_{y,t}\|_2^2
= \frac{1}{2n_y^2}\sum_{i=1}^{n_y}\sum_{j=1}^{n_y}
\|\hat{\boldsymbol\epsilon}_i-\hat{\boldsymbol\epsilon}_j\|_2^2.
\label{eq: noise_var_pair_aucgr}
\end{equation}

By Assumption \ref{assmp: bilip_aucgr}, there exists $m>0$ such that, for all $\z_1,\z_2\in\mathcal M$,
\begin{equation}
m\|\z_1-\z_2\| \le \|T(\z_1)-T(\z_2)\|.
\end{equation}

Equivalently,
\begin{equation}
\|\z_1-\z_2\| \le \frac{1}{m}\|T(\z_1)-T(\z_2)\|.
\end{equation}

Applying this inequality to each pair $(\z_i,\z_j)$ gives
\begin{equation}
\|\z_i-\z_j\|^2 \le \frac{1}{m^2}\|T(\z_i)-T(\z_j)\|^2 = \frac{1}{m^2}\|\hat{\boldsymbol\epsilon}_i-\hat{\boldsymbol\epsilon}_j\|^2.
\label{eq: feature_noise_aucgr}
\end{equation}

Substituting \cref{eq: feature_noise_aucgr} into \cref{eq: feature_var_pair_aucgr} and comparing with \cref{eq: noise_var_pair_aucgr}, we obtain
\begin{equation}
\begin{aligned}
\frac{1}{n_y}\sum_{i=1}^{n_y}\|\z_i-\mu_y\|_2^2 &\le \frac{1}{2n_y^2}\sum_{i,j=1}^{n_y} \frac{1}{m^2}\|\hat{\boldsymbol\epsilon}_i-\hat{\boldsymbol\epsilon}_j\|_2^2 \\
&= \frac{1}{m^2}\cdot\frac{1}{n_y}\sum_{i=1}^{n_y}\|\hat{\boldsymbol\epsilon}_i-\nu_{y,t}\|_2^2.
\end{aligned}
\end{equation}

Averaging over all classes $y\in\mathcal Y$ yields
\begin{equation}
\begin{aligned}
S_{\text{inner}} &= \frac{1}{|\mathcal Y|}\sum_{y\in\mathcal Y} \frac{1}{n_y}\sum_{i=1}^{n_y}\|\z_i-\mu_y\|_2^2 \\
&\le \frac{1}{m^2}\cdot \frac{1}{|\mathcal Y|}\sum_{y\in\mathcal Y} \frac{1}{n_y}\sum_{i=1}^{n_y}\|\hat{\boldsymbol\epsilon}_i-\nu_{y,t}\|_2^2 \\
&= \frac{1}{m^2} S_{\text{inner}}^{(\epsilon)}(t),
\end{aligned}
\end{equation}
which proves the first inequality in Theorem \ref{thm: auc_oriented_d_ability}.

\vskip 1 ex
\noindent \textbf{Step 3: Bounding feature-space inter-class scatter.}
\vskip 1 ex

By Assumption \ref{assmp: bilip_aucgr}, there exists $L>0$ such that, for all $\z_1,\z_2\in\mathcal M$,
\begin{equation}
\|T(\z_1)-T(\z_2)\| \le L\|\z_1-\z_2\|.
\end{equation}
Equivalently,
\begin{equation}
\|\z_1-\z_2\| \ge \frac{1}{L}\|T(\z_1)-T(\z_2)\|.
\label{eq: lower_bound_T_aucgr}
\end{equation}

Fix two distinct classes $y,y'\in\mathcal Y$. Applying \cref{eq: lower_bound_T_aucgr} with $\z_1=\mu_y$, $\z_2=\mu_{y'}$ yields
\begin{equation}
\|\mu_y-\mu_{y'}\| \ge \frac{1}{L}\|T(\mu_y)-T(\mu_{y'})\|.
\label{eq: mu_lower_bound_aucgr}
\end{equation}

We next relate $T(\mu_y)$ to $\nu_{y,t}$. We have
\begin{equation}
\nu_{y,t} = \frac{1}{n_y}\sum_{i=1}^{n_y}\hat{\boldsymbol\epsilon}_i = \frac{1}{n_y}\sum_{i=1}^{n_y} T(\z_i).
\end{equation}
Then
\begin{equation}
\begin{aligned}
\|T(\mu_y)-\nu_{y,t}\|_2
&= \left\|T(\mu_y)-\frac{1}{n_y}\sum_{i=1}^{n_y}T(\z_i)\right\|_2 \\
&= \left\|\frac{1}{n_y}\sum_{i=1}^{n_y}\bigl(T(\mu_y)-T(\z_i)\bigr)\right\|_2 \\
&\le \frac{1}{n_y}\sum_{i=1}^{n_y}\|T(\mu_y)-T(\z_i)\|_2 \\
&\le \frac{L}{n_y}\sum_{i=1}^{n_y}\|\mu_y-\z_i\|_2,
\end{aligned}
\end{equation}
where we used the triangle inequality and then the Lipschitz property of $T$. By the Cauchy-Schwarz inequality,
\begin{equation}
\sum_{i=1}^{n_y}\|\mu_y-\z_i\|_2 \le \sqrt{n_y}\left(\sum_{i=1}^{n_y}\|\mu_y-\z_i\|_2^2\right)^{1/2}.
\end{equation}
Hence,
\begin{equation}
\|T(\mu_y)-\nu_{y,t}\|_2 \le \frac{L}{\sqrt{n_y}} \left(\sum_{i=1}^{n_y}\|\mu_y-\z_i\|_2^2\right)^{1/2}.
\label{eq: Tmu_nu_bound_aucgr}
\end{equation}

Similarly, for class $y'$, we have:
\begin{equation}
\|T(\mu_{y'})-\nu_{y',t}\|_2
\le \frac{L}{\sqrt{n_{y'}}}
\left(\sum_{i=1}^{n_{y'}}\|\mu_{y'}-\z_i\|_2^2\right)^{1/2}.
\label{eq: Tmuprime_nuprime_bound_aucgr}
\end{equation}

Using the triangle inequality, we get
\begin{equation}
\begin{aligned}
&\|T(\mu_y)-T(\mu_{y'})\|_2 \\
\ge& \|\nu_{y,t}-\nu_{y',t}\|_2
   - \|T(\mu_y)-\nu_{y,t}\|_2 - \|T(\mu_{y'})-\nu_{y',t}\|_2.
\end{aligned}
\end{equation}

Combining this with \cref{eq: Tmu_nu_bound_aucgr,eq: Tmuprime_nuprime_bound_aucgr}, we obtain
\begin{equation}
\|T(\mu_y)-T(\mu_{y'})\|_2 \ge \|\nu_{y,t}-\nu_{y',t}\|_2
 - \frac{L}{\sqrt{n_y}}
   \left(\sum_{i=1}^{n_y}\|\mu_y-\z_i\|_2^2\right)^{1/2} - \frac{L}{\sqrt{n_{y'}}}
   \left(\sum_{i=1}^{n_{y'}}\|\mu_{y'}-\z_i\|_2^2\right)^{1/2}.
\label{eq: Tmu_diff_bound_aucgr}
\end{equation}

Substituting \cref{eq: Tmu_diff_bound_aucgr} into \cref{eq: mu_lower_bound_aucgr}, we obtain
\begin{equation}
\|\mu_y-\mu_{y'}\|_2 \ge \frac{1}{L}\|\nu_{y,t}-\nu_{y',t}\|_2 - \frac{1}{\sqrt{n_y}}
   \left(\sum_{i=1}^{n_y}\|\mu_y-\z_i\|_2^2\right)^{1/2} - \frac{1}{\sqrt{n_{y'}}}
   \left(\sum_{i=1}^{n_{y'}}\|\mu_{y'}-\z_i\|_2^2\right)^{1/2}.
\end{equation}

Let
\begin{equation}
a=\frac{1}{L}\|\nu_{y,t}-\nu_{y',t}\|_2,\quad
b=\frac{1}{\sqrt{n_y}}
\left(\sum_{i=1}^{n_y}\|\mu_y-\z_i\|_2^2\right)^{1/2},\quad
c=\frac{1}{\sqrt{n_{y'}}}
\left(\sum_{i=1}^{n_{y'}}\|\mu_{y'}-\z_i\|_2^2\right)^{1/2}.
\end{equation}
If $a-b-c<0$, then $\frac{1}{2}a^2-2(b^2+c^2)\le0$ and the following lower bound is trivial. If $a-b-c\ge0$, then we square the preceding inequality and use $(a-b-c)^2 \ge \frac{1}{2}a^2-2(b^2+c^2)$. In both cases, we obtain
\begin{equation}
\|\mu_y-\mu_{y'}\|_2^2 \ge \frac{1}{2L^2}\|\nu_{y,t}-\nu_{y',t}\|_2^2 - 2\left(\frac{1}{n_y}\sum_{i=1}^{n_y}\|\mu_y-\z_i\|_2^2\right) - 2\left(\frac{1}{n_{y'}}\sum_{i=1}^{n_{y'}}\|\mu_{y'}-\z_i\|_2^2\right).
\end{equation}

Averaging the above inequality over all ordered pairs $(y,y')$ with $y\neq y'$,
and recalling the definition of $S_{\text{inter}}$ and $S_{\text{inter}}^{(\epsilon)}(t)$, we obtain
\begin{equation}
\begin{aligned}
S_{\text{inter}}
&= \frac{1}{|\mathcal Y|(|\mathcal Y|-1)}
   \sum_{\substack{y,y'\in\mathcal Y\\y\ne y'}}
   \|\mu_y-\mu_{y'}\|_2^2 \\
&\ge \frac{1}{|\mathcal Y|(|\mathcal Y|-1)}
   \sum_{\substack{y,y'\in\mathcal Y\\y\ne y'}}
   \left[
      \frac{1}{2L^2}\|\nu_{y,t}-\nu_{y',t}\|_2^2 
      - 2 B_{y,y'}
   \right] \\
&= \frac{1}{2L^2} S_{\text{inter}}^{(\epsilon)}(t)
   - \frac{2}{|\mathcal Y|(|\mathcal Y|-1)}
     \sum_{\substack{y,y'\in\mathcal Y\\y\ne y'}}
     B_{y,y'},
\end{aligned}
\end{equation}
where $B_{y,y'} = \frac{1}{n_y}\sum_{i\in I_y}\|\mu_y-\z_i\|_2^2 + \frac{1}{n_{y'}}\sum_{i\in I_{y'}}\|\mu_{y'}-\z_i\|_2^2$.

For the second term, note that for each fixed $y$, the quantity $A_y = \frac{1}{n_y}\sum_{i\in I_y}\|\mu_y-\z_i\|_2^2$ appears exactly $(|\mathcal Y|-1)$ times as the first index and $(|\mathcal Y|-1)$ times as the second index when summing over all ordered pairs $(y,y')$ with $y\neq y'$. Hence
\begin{equation}
\sum_{\substack{y,y'\in\mathcal Y\\y\ne y'}} 
\left(A_y + A_{y'}\right)
= 2(|\mathcal Y|-1)\sum_{y\in\mathcal Y} A_y.
\end{equation}

Therefore,
\begin{equation}
\begin{aligned}
&\frac{2}{|\mathcal Y|(|\mathcal Y|-1)}
\sum_{\substack{y,y'\in\mathcal Y\\y\ne y'}}
\left(A_y + A_{y'}\right) \\
=& \frac{2}{|\mathcal Y|(|\mathcal Y|-1)}\,
   2(|\mathcal Y|-1)\sum_{y\in\mathcal Y} A_y \\
=& \frac{4}{|\mathcal Y|}\sum_{y\in\mathcal Y} A_y \\
 =& 4 S_{\text{inner}}.
\end{aligned}
\end{equation}

Substituting this back, we obtain the explicit bound
\begin{equation}
S_{\text{inter}}
\;\ge\; \frac{1}{2L^2} S_{\text{inter}}^{(\epsilon)}(t)
      \;-\; 4 S_{\text{inner}}.
\label{eq:Sinter_feature_noise_aucgr}
\end{equation}

Finally, using the bound from Step 2, $S_{\text{inner}} \le \frac{1}{m^2} S_{\text{inner}}^{(\epsilon)}(t)$, we arrive at
\begin{equation}
S_{\text{inter}} \ge \frac{1}{2L^2} S_{\text{inter}}^{(\epsilon)}(t) - \frac{4}{m^2} S_{\text{inner}}^{(\epsilon)}(t).
\end{equation}
Thus, we have:
\begin{equation}
S_{\text{inter}}\ge\kappa S_{\text{inter}}^{(\epsilon)}(t)-\eta S_{\text{inner}}^{(\epsilon)}(t),
\end{equation}
where $\kappa = \frac{1}{2L^2}$ and $\eta = \frac{4}{m^2}$. This proves the second inequality in Theorem \ref{thm: auc_oriented_d_ability}.

This completes the proof.
\end{proof}

\newpage
\section{Proof of Theorem 3}
\label{sec: supp_Proof of Theorem 3}

\begin{rthm3}[AUCGR Supports \textbf{\textcolor{PerColor}{P-Ability}}]
Under the assumptions that negatives in the mini-batch are well separated from the anchor (i.e., $\mathrm{sim}(\hat{\boldsymbol\epsilon},\hat{\boldsymbol\epsilon}_{-}^{j}) \ll \mathrm{sim}(\hat{\boldsymbol\epsilon},\boldsymbol\epsilon_{\mathrm{gt}})$ for all $j\neq i$), predicted-noise norms are bounded away from $0$ and $\infty$, and the surrogate loss is decreasing with bounded slope for the pairwise gaps used by AUCGR, then the AUCGR loss is bounded by a scaled reconstruction loss up to additive constants:
\begin{equation*}
\lambda_{\min}R_{\mathrm{rec}}+c_{\min}
\le
\mathcal{L}_{\mathrm{AUCGR}}
\le
\lambda_{\max}R_{\mathrm{rec}}+c_{\max},
\end{equation*}
where $R_{\mathrm{rec}}=\frac{1}{B}\sum_{i=1}^{B}\bigl\|\hat{\boldsymbol\epsilon}_i-\boldsymbol\epsilon_{t,i}^{\mathrm{gt}}\bigr\|_2^2$, $\lambda_{\min},\lambda_{\max}>0$, and $c_{\min},c_{\max}$ are constants.
\end{rthm3}

\begin{proof}
We consider a single anchor sample and drop the sample index for brevity.

Recall the anchor predicted noise
\begin{equation}
\hat{\boldsymbol\epsilon}
=
\epsilon_{\theta}\bigl(\tilde{\x}_{t},\,h_{\omega}(f_{\phi}(\x)),\,t\bigr).
\end{equation}
Let $\boldsymbol\epsilon_{\text{gt}}$ denote the ground-truth predicted-noise target contained in the positive set. Let $\mathcal{P}_{\mathrm{aux}}$ denote the remaining positives, let $\mathcal P=\{\boldsymbol\epsilon_{\text{gt}}\}\cup\mathcal P_{\mathrm{aux}}$, and let $\mathcal N=\{\hat{\boldsymbol\epsilon}^{j}_{-}\}_{j}$ denote the set of negatives in the mini-batch. The corresponding anchor-wise AUCGR loss can be decomposed as
\begin{equation}
\mathcal{L}_{\mathrm{AUCGR}}^{(i)}
=
\mathcal{L}_{\mathrm{gt}}+\mathcal{L}_{\mathrm{aux}},
\label{eq: aucgr_anchor_reconstruction}
\end{equation}
where
\begin{equation}
u=\mathrm{sim}(\hat{\boldsymbol\epsilon},\boldsymbol\epsilon_{\text{gt}}),
\qquad
r_j=\mathrm{sim}(\hat{\boldsymbol\epsilon},\hat{\boldsymbol\epsilon}_{-}^{j}),
\qquad
q(p)=\mathrm{sim}(\hat{\boldsymbol\epsilon},p).
\end{equation}
\begin{equation}
\mathcal{L}_{\mathrm{gt}}
=
\frac{1}{|\mathcal P||N|}\sum_{j}\ell\bigl(u-r_j\bigr),
\qquad
\mathcal{L}_{\mathrm{aux}}
=
\frac{1}{|\mathcal P||N|}
\sum_{p\in\mathcal P_{\mathrm{aux}}}\sum_j
\ell\bigl(q(p)-r_j\bigr).
\label{eq: aucgr_positive_split}
\end{equation}

Since cosine similarity lies in $[-1,1]$ and the surrogate is bounded on compact intervals, the auxiliary term $\mathcal{L}_{\mathrm{aux}}$ is bounded. Therefore, there exist constants $C_{\mathrm{aux}}^{\min},C_{\mathrm{aux}}^{\max}\in\mathbb{R}$ such that
\begin{equation}
C_{\mathrm{aux}}^{\min}\le \mathcal{L}_{\mathrm{aux}}\le C_{\mathrm{aux}}^{\max}.
\label{eq: positive_term_bound_aucgr}
\end{equation}

We now analyze the ground-truth term. By the separation assumption, there exists a constant $\Delta>0$ such that
\begin{equation}
r_j\le u-\Delta,
\qquad
\forall j.
\label{eq: aucgr_negative_separation}
\end{equation}
Hence, for each $j$, the score gap $u-r_j$ belongs to the pairwise gaps used by AUCGR. We assume that the surrogate loss is decreasing with bounded slope on these gaps. That is, there exist constants $0<\mu_{\ell}\le L_{\ell}<\infty$ such that, for every such gap $s$,
\begin{equation}
\mu_{\ell}
\le
-\ell'(s)
\le
L_{\ell}.
\label{eq: surrogate_slope_bound_aucgr}
\end{equation}
Next, define the normalized vectors
\begin{equation}
\hat{e} = \frac{\hat{\boldsymbol\epsilon}}{\|\hat{\boldsymbol\epsilon}\|_2},
\qquad
g = \frac{\boldsymbol\epsilon_{\text{gt}}}{\|\boldsymbol\epsilon_{\text{gt}}\|_2}.
\end{equation}
Let
\begin{equation}
d_{\text{gt}}^2 = \|\hat{e}-g\|_2^2.
\end{equation}
Since $\hat{e}$ and $g$ are unit vectors, we have
\begin{equation}
u = \langle \hat{e},g\rangle = 1-\frac{1}{2}d_{\text{gt}}^2,
\qquad
0\le d_{\text{gt}}^2\le 4.
\label{eq: similarity_distance_aucgr}
\end{equation}

Substituting \cref{eq: similarity_distance_aucgr} into each term of $\mathcal{L}_{\mathrm{gt}}$ gives
\begin{equation}
\ell\bigl(u-r_j\bigr)
=
\ell\left(1-\frac{1}{2}d_{\text{gt}}^2-r_j\right).
\label{eq: gt_term_distance_aucgr}
\end{equation}
By the mean value theorem and \cref{eq: surrogate_slope_bound_aucgr}, the right-hand side of \cref{eq: gt_term_distance_aucgr} is bounded above and below by affine functions of $d_{\text{gt}}^2$. Therefore, there exist positive constants $a_j,b_j$ and constants $c_j,d_j$ such that
\begin{equation}
a_j d_{\text{gt}}^2 + c_j
\le
\ell\bigl(u-r_j\bigr)
\le
b_j d_{\text{gt}}^2 + d_j.
\label{eq: gt_affine_bound_aucgr}
\end{equation}
Averaging over all $j$ in \cref{eq: gt_affine_bound_aucgr}, there exist positive constants $\tilde{\lambda}_{\min},\tilde{\lambda}_{\max}$ and constants $\tilde{c}_{\min},\tilde{c}_{\max}$ such that
\begin{equation}
\tilde{\lambda}_{\min} d_{\text{gt}}^2 + \tilde{c}_{\min}
\le
\mathcal{L}_{\mathrm{gt}}
\le
\tilde{\lambda}_{\max} d_{\text{gt}}^2 + \tilde{c}_{\max}.
\label{eq: gt_average_affine_aucgr}
\end{equation}

Combining \cref{eq: positive_term_bound_aucgr} and \cref{eq: gt_average_affine_aucgr}, we obtain
\begin{equation}
\bar{\lambda}_{\min} d_{\text{gt}}^2 + \bar{c}_{\min}
\le
\mathcal{L}_{\mathrm{AUCGR}}^{(i)}
\le
\bar{\lambda}_{\max} d_{\text{gt}}^2 + \bar{c}_{\max},
\label{eq: aucgr_dgt_bounds}
\end{equation}
for some positive constants $\bar{\lambda}_{\min},\bar{\lambda}_{\max}$ and constants $\bar{c}_{\min},\bar{c}_{\max}$.

We now relate $d_{\text{gt}}^2$ to the true reconstruction error $\|\hat{\boldsymbol\epsilon}-\boldsymbol\epsilon_{\text{gt}}\|_2^2$. For any nonzero vectors $u,v$, the law of cosines gives
\begin{equation}
\|u-v\|_2^2 = \|u\|_2^2 + \|v\|_2^2 - 2\|u\|_2\|v\|_2\,\mathrm{sim}(u,v).
\end{equation}

With $u=\hat{\boldsymbol\epsilon}$, $v=\boldsymbol\epsilon_{\text{gt}}$ and $\mathrm{sim}(u,v)=1-d_{\text{gt}}^2/2$, we obtain
\begin{equation}
\begin{aligned}
\bigl\|\hat{\boldsymbol\epsilon}-\boldsymbol\epsilon_{\text{gt}}\bigr\|_2^2
&= \|\hat{\boldsymbol\epsilon}\|_2^2 + \|\boldsymbol\epsilon_{\text{gt}}\|_2^2 - 2\|\hat{\boldsymbol\epsilon}\|_2\|\boldsymbol\epsilon_{\text{gt}}\|_2 \left(1-\frac{d_{\text{gt}}^2}{2}\right) \\
&= \bigl(\|\hat{\boldsymbol\epsilon}\|_2 - \|\boldsymbol\epsilon_{\text{gt}}\|_2\bigr)^2 + \|\hat{\boldsymbol\epsilon}\|_2\|\boldsymbol\epsilon_{\text{gt}}\|_2\,d_{\text{gt}}^2.
\end{aligned}
\label{eq: MSE_exact_aucgr}
\end{equation}

By assumption, predicted-noise norms are bounded away from $0$ and $\infty$. Thus, there exist constants $0<\alpha\le\beta<\infty$ such that
\begin{equation}
\alpha \le \|\hat{\boldsymbol\epsilon}\|_2,\ \|\boldsymbol\epsilon_{\text{gt}}\|_2 \le \beta.
\label{eq: norm_bounds_aucgr}
\end{equation}

Using \cref{eq: MSE_exact_aucgr} and the fact that $\bigl(\|\hat{\boldsymbol\epsilon}\|_2 - \|\boldsymbol\epsilon_{\text{gt}}\|_2\bigr)^2\ge 0$, we have
\begin{equation}
\bigl\|\hat{\boldsymbol\epsilon}-\boldsymbol\epsilon_{\text{gt}}\bigr\|_2^2 \ge \|\hat{\boldsymbol\epsilon}\|_2\|\boldsymbol\epsilon_{\text{gt}}\|_2\,d_{\text{gt}}^2 \ge \alpha^2 d_{\text{gt}}^2.
\end{equation}

On the other hand, using $\|\hat{\boldsymbol\epsilon}\|_2^2 + \|\boldsymbol\epsilon_{\text{gt}}\|_2^2 \le 2\beta^2$ and $\|\hat{\boldsymbol\epsilon}\|_2\|\boldsymbol\epsilon_{\text{gt}}\|_2 \le \beta^2$, we obtain
\begin{equation}
\bigl\|\hat{\boldsymbol\epsilon}-\boldsymbol\epsilon_{\text{gt}}\bigr\|_2^2 \le 2\beta^2 + \beta^2 d_{\text{gt}}^2.
\end{equation}

Rearranging these inequalities yields
\begin{equation}
\frac{1}{\beta^2} \left(\bigl\|\hat{\boldsymbol\epsilon}-\boldsymbol\epsilon_{\text{gt}}\bigr\|_2^2-2\beta^2\right) \le d_{\text{gt}}^2 \le \frac{1}{\alpha^2} \bigl\|\hat{\boldsymbol\epsilon}-\boldsymbol\epsilon_{\text{gt}}\bigr\|_2^2.
\label{eq: dgt_MSE_aucgr}
\end{equation}

Combining \cref{eq: aucgr_dgt_bounds} and \cref{eq: dgt_MSE_aucgr}, we obtain affine bounds on the anchor-wise AUCGR loss in terms of the reconstruction error. Therefore, there exist positive constants $\lambda_{\min}^{(i)},\lambda_{\max}^{(i)}$ and constants $c_{\min}^{(i)},c_{\max}^{(i)}$ such that
\begin{equation}
\lambda_{\min}^{(i)}\bigl\|\hat{\boldsymbol\epsilon}-\boldsymbol\epsilon_{\text{gt}}\bigr\|_2^2+c_{\min}^{(i)}
\le
\mathcal{L}_{\mathrm{AUCGR}}^{(i)}
\le
\lambda_{\max}^{(i)}\bigl\|\hat{\boldsymbol\epsilon}-\boldsymbol\epsilon_{\text{gt}}\bigr\|_2^2+c_{\max}^{(i)}.
\label{eq: anchor_reconstruction_bounds_aucgr}
\end{equation}

Averaging \cref{eq: anchor_reconstruction_bounds_aucgr} over all anchors in the mini-batch and using the smallest lower coefficient and the largest upper coefficient yields
\begin{equation}
\lambda_{\min}\frac{1}{B}\sum_{i=1}^{B}
\bigl\|\hat{\boldsymbol\epsilon}_i-\boldsymbol\epsilon_{t,i}^{\mathrm{gt}}\bigr\|_2^2+c_{\min}
\le
\mathcal{L}_{\mathrm{AUCGR}}
\le
\lambda_{\max}\frac{1}{B}\sum_{i=1}^{B}
\bigl\|\hat{\boldsymbol\epsilon}_i-\boldsymbol\epsilon_{t,i}^{\mathrm{gt}}\bigr\|_2^2+c_{\max},
\end{equation}
for some positive constants $\lambda_{\min},\lambda_{\max}$ and constants $c_{\min},c_{\max}$. This proves the claimed affine control by the reconstruction error.

This completes the proof.
\end{proof}

\newpage

\section{Proof of Proposition 1}
\label{sec: supp_Proof of Proposition 1}

\begin{rprop1}
Let $q:\mathcal{X}\to\mathbb{R}$ be a scoring function, and let $x^+\sim P_+$ and $x^-\sim P_-$ denote a positive and a negative sample. Assume that the negative score distribution $F_-(t)=\mathbb{P}(q(x^-)\le t)$ is continuous. For any measurable negative subset $\mathcal{N}$ with positive probability, define the conditional ranking objective
\begin{equation*}
\mathrm{AUC}(\mathcal{N})
=
\mathbb{P}\big(q(x^+)>q(x^-)\mid x^-\in\mathcal{N}\big).
\end{equation*}
For any $\beta\in(0,1]$, let
\begin{equation*}
\mathcal{N}_{\mathrm{low}}^{\beta}
=
\{x^-:F_-(q(x^-))\le \beta\}.
\end{equation*}
Then the partial AUC over $[1-\beta,1]$ is defined as
\begin{equation*}
\mathrm{pAUC}_{[1-\beta,1]}
:=
\int_{1-\beta}^{1}\mathrm{TPR}\big(\mathrm{FPR}^{-1}(\xi)\big)\,d\xi,
\end{equation*}
where $\mathrm{TPR}(t)=\mathbb{P}(q(x^+)>t)$, $\mathrm{FPR}(t)=\mathbb{P}(q(x^-)>t)$, and $\mathrm{FPR}^{-1}(\xi)$ denotes the threshold associated with false positive rate $\xi$.
\end{rprop1}

\begin{proof}
Let
\begin{equation}
Y^- = q(x^-),
\qquad
Y^+ = q(x^+).
\end{equation}
Then the negative-score cumulative distribution function is
\begin{equation}
F_-(t)=\mathbb{P}(Y^-\le t).
\end{equation}
We also define the false positive rate and the true positive rate at threshold $t$ as
\begin{equation}
\mathrm{FPR}(t)=\mathbb{P}(Y^->t)=1-F_-(t),
\qquad
\mathrm{TPR}(t)=\mathbb{P}(Y^+>t).
\end{equation}
Since $F_-$ is continuous, the transformed variable
\begin{equation}
U=F_-(Y^-)
\end{equation}
is uniformly distributed on $[0,1]$. We also define the quantile function
\begin{equation}
Q_-(u)=\inf\{t\in\mathbb{R}:F_-(t)\ge u\},
\qquad u\in[0,1].
\end{equation}
For any score level $z$, define
\begin{equation}
G(z)=\mathbb{P}(Y^+>z).
\end{equation}
The quantity $G(z)$ is exactly $\mathrm{TPR}(z)$.

\vskip 1 ex
\noindent \textbf{Step 1. Conditioning on low-score negatives.}
\vskip 1 ex

Consider
\begin{equation}
\mathcal{N}_{\mathrm{low}}^{\beta}
=
\{x^-:F_-(Y^-)\le \beta\}
=
\{x^-:U\le \beta\}.
\end{equation}
Because $U\sim\mathrm{Unif}[0,1]$, we have
\begin{equation}
\mathbb{P}(U\le \beta)=\beta.
\end{equation}
Therefore, conditioned on the event $U\le \beta$, the variable $U$ is uniformly distributed on $[0,\beta]$ with density $1/\beta$. Hence
\begin{equation}
\begin{aligned}
\mathrm{AUC}(\mathcal{N}_{\mathrm{low}}^{\beta})
&=\mathbb{P}(Y^+>Y^-\mid U\le \beta)\\
&=\frac{1}{\beta}\int_0^{\beta} G\big(Q_-(u)\big)\,du.
\end{aligned}
\end{equation}
This identity means that, after restricting negatives to the lowest-score $\beta$ fraction, we average the ranking probability over the quantile range $u\in[0,\beta]$.

Now make the change of variable
\begin{equation}
\xi = 1-u.
\end{equation}
When $u=0$, we have $\xi=1$. When $u=\beta$, we have $\xi=1-\beta$. Since $du=-d\xi$, reversing the integration limits removes the minus sign. Therefore,
\begin{equation}
\mathrm{AUC}(\mathcal{N}_{\mathrm{low}}^{\beta})
=
\frac{1}{\beta}\int_{1-\beta}^{1} G\big(Q_-(1-\xi)\big)\,d\xi.
\label{eq: pauc_low_intermediate}
\end{equation}

\vskip 1 ex
\noindent \textbf{Step 2. Connecting the conditional objective to partial AUC.}
\vskip 1 ex

For any $\xi\in[0,1]$, define the generalized inverse of the false positive rate as
\begin{equation}
\mathrm{FPR}^{-1}(\xi)=Q_-(1-\xi).
\end{equation}
We now show that $\mathrm{FPR}^{-1}(\xi)$ is exactly the score threshold whose false positive rate is $\xi$. By definition,
\begin{equation}
\mathrm{FPR}\big(\mathrm{FPR}^{-1}(\xi)\big)
=
\mathbb{P}\big(Y^->Q_-(1-\xi)\big)
=
1-\mathbb{P}\big(Y^-\le Q_-(1-\xi)\big).
\end{equation}
Since $F_-$ is continuous, we have
\begin{equation}
\mathbb{P}\big(Y^-\le Q_-(1-\xi)\big)=F_-\big(Q_-(1-\xi)\big)=1-\xi.
\end{equation}
Substituting this identity above yields
\begin{equation}
\mathrm{FPR}\big(\mathrm{FPR}^{-1}(\xi)\big)=\xi.
\end{equation}
Therefore, $\mathrm{FPR}^{-1}(\xi)$ is the threshold associated with false positive rate $\xi$. The corresponding true positive rate is
\begin{equation}
\mathrm{TPR}\big(\mathrm{FPR}^{-1}(\xi)\big)
=
\mathbb{P}\big(Y^+>\mathrm{FPR}^{-1}(\xi)\big)
=
\mathbb{P}\big(Y^+>Q_-(1-\xi)\big)
=
G\big(Q_-(1-\xi)\big).
\label{eq: tpr_quantile_identity}
\end{equation}
Substituting \cref{eq: tpr_quantile_identity} into \cref{eq: pauc_low_intermediate}, we obtain
\begin{equation}
\mathrm{AUC}(\mathcal{N}_{\mathrm{low}}^{\beta})
=
\frac{1}{\beta}\int_{1-\beta}^{1}\mathrm{TPR}\big(\mathrm{FPR}^{-1}(\xi)\big)\,d\xi.
\end{equation}
Multiplying both sides by $\beta$ gives
\begin{equation}
\mathrm{pAUC}_{[1-\beta,1]}
:=
\int_{1-\beta}^{1}\mathrm{TPR}\big(\mathrm{FPR}^{-1}(\xi)\big)\,d\xi
=
\beta\,\mathrm{AUC}(\mathcal{N}_{\mathrm{low}}^{\beta}).
\end{equation}

This completes the proof.
\end{proof}

\newpage

\section{Proof of Theorem 4}
\label{sec: supp_Proof of Theorem 4}

\begin{rthm4}
For anchor $i$, let $\mathcal{P}_i$ denote the positive set and let $\mathcal{N}_i$ denote the negative set. Then we have the following result:
\begin{equation*}
\mathcal{L}_{\mathrm{DCR}}^{(i)}
\ge
\frac{1}{|\mathcal{P}_i||\mathcal{N}_i|}
\sum_{p\in\mathcal{P}_i}\sum_{n\in\mathcal{N}_i}
\ell\bigl(q_i(p)-q_i(n)\bigr).
\end{equation*}
where $\ell(\cdot)$ is instantiated as the logistic surrogate function
\begin{equation*}
\ell_{\mathrm{logistic}}(u)
=
\log\left(1+\exp\left(-\frac{u}{\tau}\right)\right).
\end{equation*}
Moreover, the bound is tight if and only if $|\mathcal{P}_i|=|\mathcal{N}_i|=1$.
\end{rthm4}

\begin{proof}
Let $\mathcal{C}_i=\mathcal{P}_i\cup\mathcal{N}_i$ denote the comparison set. The anchor-wise conference loss takes the form
\begin{equation}
\mathcal{L}_{\mathrm{DCR}}^{(i)}
=
-\frac{1}{|\mathcal{P}_i|}
\sum_{p\in\mathcal{P}_i}
\log
\frac{\exp\bigl(q_i(p)/\tau\bigr)}
{\sum_{c\in\mathcal{C}_i}\exp\bigl(q_i(c)/\tau\bigr)}.
\end{equation}
Fix a positive $p\in\mathcal{P}_i$. The corresponding summand in $\mathcal{L}_{\mathrm{DCR}}^{(i)}$ is
\begin{equation}
T_{i,p}
:=
-\log
\frac{\exp\bigl(q_i(p)/\tau\bigr)}
{\sum_{c\in\mathcal{C}_i}\exp\bigl(q_i(c)/\tau\bigr)}.
\end{equation}
By direct expansion,
\begin{equation}
\begin{aligned}
T_{i,p}
&= -\log \frac{\exp\bigl(q_i(p)/\tau\bigr)}{\sum_{c\in\mathcal{P}_i\cup\mathcal{N}_i}\exp\bigl(q_i(c)/\tau\bigr)} \\
&= \log\Bigg( 1+ \sum_{p'\in\mathcal{P}_i\setminus\{p\}} \exp\left(\frac{q_i(p')-q_i(p)}{\tau}\right) + \sum_{n\in\mathcal{N}_i} \exp\left(\frac{q_i(n)-q_i(p)}{\tau}\right) \Bigg).
\end{aligned}
\end{equation}
Since the terms contributed by $\mathcal{P}_i\setminus\{p\}$ are nonnegative, we obtain
\begin{equation}
\begin{aligned}
T_{i,p}
\ge
\log\left(
1+
\sum_{n\in\mathcal{N}_i}
\exp\left(\frac{q_i(n)-q_i(p)}{\tau}\right)
\right).
\end{aligned}
\end{equation}
By the definition of the pairwise logistic surrogate, we have
\begin{equation}
\ell\bigl(q_i(p)-q_i(n)\bigr)
=
\log\left(
1+
\exp\left(\frac{q_i(n)-q_i(p)}{\tau}\right)
\right).
\end{equation}
Averaging over $n\in\mathcal{N}_i$ and applying Jensen's inequality to the concave function $\log(\cdot)$ yields
\begin{equation}
\begin{aligned}
\frac{1}{|\mathcal{N}_i|}
\sum_{n\in\mathcal{N}_i}
\ell\bigl(q_i(p)-q_i(n)\bigr)
&=
\frac{1}{|\mathcal{N}_i|}
\sum_{n\in\mathcal{N}_i}
\log\left(
1+
\exp\left(\frac{q_i(n)-q_i(p)}{\tau}\right)
\right) \\
&\le
\log\left(
1+
\frac{1}{|\mathcal{N}_i|}
\sum_{n\in\mathcal{N}_i}
\exp\left(\frac{q_i(n)-q_i(p)}{\tau}\right)
\right).
\end{aligned}
\end{equation}
Because $|\mathcal{N}_i|\ge 1$, we further have
\begin{equation}
\begin{aligned}
\log\left(
1+
\frac{1}{|\mathcal{N}_i|}
\sum_{n\in\mathcal{N}_i}
\exp\left(\frac{q_i(n)-q_i(p)}{\tau}\right)
\right)
&\le
\log\left(
1+
\sum_{n\in\mathcal{N}_i}
\exp\left(\frac{q_i(n)-q_i(p)}{\tau}\right)
\right)
\le T_{i,p}.
\end{aligned}
\end{equation}
Therefore,
\begin{equation}
\frac{1}{|\mathcal{N}_i|}
\sum_{n\in\mathcal{N}_i}
\ell\bigl(q_i(p)-q_i(n)\bigr)
\le T_{i,p}.
\end{equation}
Averaging the above inequality over $p\in\mathcal{P}_i$ gives
\begin{equation}
\frac{1}{|\mathcal{P}_i||\mathcal{N}_i|}
\sum_{p\in\mathcal{P}_i}
\sum_{n\in\mathcal{N}_i}
\ell\bigl(q_i(p)-q_i(n)\bigr)
\le
\mathcal{L}_{\mathrm{DCR}}^{(i)}.
\end{equation}

We finally characterize when the bound is tight. Assume $\tau>0$ and all scores are finite, so every exponential term is strictly positive.

For a fixed $p\in\mathcal{P}_i$, the inequality
\begin{equation}
T_{i,p}
\ge
\log\left(
1+
\sum_{n\in\mathcal{N}_i}
\exp\left(\frac{q_i(n)-q_i(p)}{\tau}\right)
\right)
\end{equation}
is an equality if and only if
\begin{equation}
\mathcal{P}_i\setminus\{p\}=\varnothing,
\end{equation}
i.e., $|\mathcal{P}_i|=1$.

Moreover, since
\begin{equation}
\log\left(
1+
\frac{1}{|\mathcal{N}_i|}
\sum_{n\in\mathcal{N}_i}
\exp\left(\frac{q_i(n)-q_i(p)}{\tau}\right)
\right)
\le
\log\left(
1+
\sum_{n\in\mathcal{N}_i}
\exp\left(\frac{q_i(n)-q_i(p)}{\tau}\right)
\right),
\end{equation}
and the summed exponential term is strictly positive, equality holds if and only if $|\mathcal{N}_i|=1$.

Therefore, the bound is tight if and only if
\begin{equation}
|\mathcal{P}_i|=|\mathcal{N}_i|=1.
\end{equation}
In this case, letting $\mathcal{P}_i=\{p\}$ and $\mathcal{N}_i=\{n\}$,
\begin{equation}
\mathcal{L}_{\mathrm{DCR}}^{(i)}
=
\log\left(
1+\exp\left(\frac{q_i(n)-q_i(p)}{\tau}\right)
\right)
=
\ell\bigl(q_i(p)-q_i(n)\bigr).
\end{equation}
Hence, DCR and logistic AUCGR are exactly equivalent when each anchor has one positive and one negative; otherwise, the upper bound is strict.

This completes the proof.
\end{proof}

\section{Proof of Theorem 5}
\label{sec: supp_Proof of Theorem 5}

\begin{lem}[Hoeffding Inequality~\cite{hoeffding1963probability}]
\label{lem: hoeffding_aucgr}
Let $X_1,\dots,X_N$ be independent random variables such that
\begin{equation}
X_i \in [a_i,b_i]
\qquad
\text{for all } i=1,\dots,N.
\end{equation}
Then, for any $t>0$,
\begin{equation}
\mathbb{P}\left(
\left|
\frac{1}{N}\sum_{i=1}^{N} X_i
-
\frac{1}{N}\sum_{i=1}^{N} \mathbb{E}[X_i]
\right|
\ge t
\right)
\le
2\exp\left(
-\frac{2N^2t^2}{\sum_{i=1}^{N}(b_i-a_i)^2}
\right).
\label{eq: hoeffding_aucgr}
\end{equation}
\end{lem}

\begin{lem}[Uniform Convergence over a Finite Function Class]
\label{lem: finite_class_concentration_aucgr}
Let $\{g_1,\dots,g_{N_{\mathcal{G}}}\}$ be a finite collection of measurable functions on $\mathcal{Z}$. Assume that, for every $k$ and every $z$,
\begin{equation}
0 \le g_k(z) \le M.
\end{equation}
Then, for any $\delta \in (0,1)$, with probability at least $1-\delta$,
\begin{equation}
\sup_{1 \le k \le N_{\mathcal{G}}}
\left|\mathbb{E}[g_k(z)]-\hat{\mathbb{E}}_{N}[g_k]\right|
\le
M\sqrt{
\frac{
\log(2N_{\mathcal{G}}/\delta)
}{
2N
}
}.
\end{equation}
\end{lem}

\begin{proof}
Fix one function $g_k$. Since the anchor tuples $z_1,\dots,z_N$ are i.i.d., define
\begin{equation}
X_i := g_k(z_i),
\qquad
i=1,\dots,N.
\end{equation}

Then $X_1,\dots,X_N$ are independent random variables, and they have the same expectation:
\begin{equation}
\mathbb{E}[X_i] = \mathbb{E}[g_k(z)].
\end{equation}

Moreover, the assumption $0 \le g_k(z) \le M$ implies
\begin{equation}
0 \le X_i \le M
\qquad
\text{for all } i=1,\dots,N.
\end{equation}

Therefore, Lemma~\ref{lem: hoeffding_aucgr} applies with
\begin{equation}
a_i = 0,
\qquad
b_i = M
\qquad
\text{for all } i=1,\dots,N.
\end{equation}

Under this choice,
\begin{equation}
b_i-a_i = M
\qquad
\text{for all } i=1,\dots,N,
\end{equation}
and hence
\begin{equation}
\sum_{i=1}^{N}(b_i-a_i)^2
=
\sum_{i=1}^{N} M^2
=
NM^2.
\end{equation}

Substituting these identities into \cref{eq: hoeffding_aucgr} yields
\begin{equation}
\mathbb{P}\left(
\left|
\frac{1}{N}\sum_{i=1}^{N} X_i
-
\frac{1}{N}\sum_{i=1}^{N}\mathbb{E}[X_i]
\right|
> t
\right)
\le
2\exp\left(-\frac{2Nt^2}{M^2}\right).
\end{equation}

Since
\begin{equation}
\frac{1}{N}\sum_{i=1}^{N} X_i
=
\frac{1}{N}\sum_{i=1}^{N} g_k(z_i)
=
\hat{\mathbb{E}}_{N}[g_k]
\end{equation}
and
\begin{equation}
\frac{1}{N}\sum_{i=1}^{N}\mathbb{E}[X_i]
=
\mathbb{E}[g_k(z)],
\end{equation}
we obtain
\begin{equation}
\mathbb{P}\left(
\left|\mathbb{E}[g_k(z)]-\hat{\mathbb{E}}_{N}[g_k]\right|>t
\right)
\le
2\exp\left(-\frac{2Nt^2}{M^2}\right).
\label{eq: single_function_concentration_aucgr}
\end{equation}

This bound holds for every index $k$. We now require the same event to hold simultaneously for all $k=1,\dots,N_{\mathcal{G}}$. Using the union bound,
\begin{equation}
\mathbb{P}\left(
\sup_{1 \le k \le N_{\mathcal{G}}}
\left|\mathbb{E}[g_k(z)]-\hat{\mathbb{E}}_{N}[g_k]\right|>t
\right)
\le
2N_{\mathcal{G}}
\exp\left(-\frac{2Nt^2}{M^2}\right).
\end{equation}

Indeed, the event
\begin{equation}
\sup_{1 \le k \le N_{\mathcal{G}}}
\left|\mathbb{E}[g_k(z)]-\hat{\mathbb{E}}_{N}[g_k]\right|>t
\end{equation}
means that there exists at least one index $k$ such that
\begin{equation}
\left|\mathbb{E}[g_k(z)]-\hat{\mathbb{E}}_{N}[g_k]\right|>t.
\end{equation}

Therefore,
\begin{equation}
\mathbb{P}\left(
\sup_{1 \le k \le N_{\mathcal{G}}}
\left|\mathbb{E}[g_k(z)]-\hat{\mathbb{E}}_{N}[g_k]\right|>t
\right)
\le
\sum_{k=1}^{N_{\mathcal{G}}}
\mathbb{P}\left(
\left|\mathbb{E}[g_k(z)]-\hat{\mathbb{E}}_{N}[g_k]\right|>t
\right),
\end{equation}
and each summand is bounded by \cref{eq: single_function_concentration_aucgr}.

To make the right hand side of the union bound no larger than $\delta$, it is sufficient to require
\begin{equation}
2N_{\mathcal{G}}
\exp\left(-\frac{2Nt^2}{M^2}\right)
\le
\delta.
\end{equation}
Since $\delta \in (0,1)$ and $N_{\mathcal{G}}\ge 1$, this requirement is equivalent to
\begin{equation}
\exp\left(-\frac{2Nt^2}{M^2}\right)
\le
\frac{\delta}{2N_{\mathcal{G}}},
\end{equation}
which gives
\begin{equation}
-\frac{2Nt^2}{M^2}
\le
\log\frac{\delta}{2N_{\mathcal{G}}}
=
-\log\frac{2N_{\mathcal{G}}}{\delta}.
\end{equation}
Equivalently,
\begin{equation}
\frac{2Nt^2}{M^2}
\ge
\log\frac{2N_{\mathcal{G}}}{\delta}.
\end{equation}
Solving for the nonnegative threshold yields
\begin{equation}
t
\ge
M\sqrt{
\frac{
\log(2N_{\mathcal{G}}/\delta)
}{
2N
}
}.
\end{equation}
Thus the smallest threshold that ensures a failure probability at most $\delta$ is
\begin{equation}
t =
M\sqrt{
\frac{
\log(2N_{\mathcal{G}}/\delta)
}{
2N
}
}.
\end{equation}

Hence, with probability at least $1-\delta$,
\begin{equation}
\sup_{1 \le k \le N_{\mathcal{G}}}
\left|\mathbb{E}[g_k(z)]-\hat{\mathbb{E}}_{N}[g_k]\right|
\le
M\sqrt{
\frac{
\log(2N_{\mathcal{G}}/\delta)
}{
2N
}
}.
\end{equation}

This completes the proof.
\end{proof}

\begin{rthm5}
Under Assumption \ref{ass: paucgr_generalization}, for any $\delta\in(0,1)$ and any fixed $\varepsilon\in(0,R]$, with probability at least $1-\delta$, the following inequality holds simultaneously for all $f\in\mathcal{F}$:
\begin{equation}
\begin{aligned}
\left|
\mathcal{R}_{\alpha}(f)-\hat{\mathcal{R}}_{\alpha}(f)
\right|
&\le
2\varepsilon \\
&\quad+
M\sqrt{
\frac{
A\log(R/\varepsilon)+\log(2/\delta)
}{
2N
}
}.
\end{aligned}
\end{equation}

In particular, choosing $\varepsilon=R/N$ yields
\begin{equation}
\begin{aligned}
\left|
\mathcal{R}_{\alpha}(f)-\hat{\mathcal{R}}_{\alpha}(f)
\right|
&\le
\frac{2R}{N} \\
&\quad+
M\sqrt{
\frac{
A\log N+\log(2/\delta)
}{
2N
}
}.
\end{aligned}
\end{equation}
\end{rthm5}

\begin{proof}
Fix any $\delta \in (0,1)$ and any $\varepsilon \in (0,R]$.

For convenience, define
\begin{equation}
\hat{\mathbb{E}}_{N}[g]
=
\frac{1}{N}\sum_{i=1}^{N} g(z_i).
\end{equation}

Write $g_f(z)=g_f^{(\alpha)}(z)$. Then
\begin{equation}
\mathcal{R}_{\alpha}(f)=\mathbb{E}[g_f(z)],
\qquad
\hat{\mathcal{R}}_{\alpha}(f)=\hat{\mathbb{E}}_{N}[g_f].
\end{equation}

For every anchor tuple $z$ and every retained pair $(p,n)$, define
\begin{equation}
\Delta_f(z;p,n)
=
d(\hat{\boldsymbol\epsilon},p)-d(\hat{\boldsymbol\epsilon},n).
\end{equation}

Assumption~\ref{ass: paucgr_generalization} implies
\begin{equation}
-2 \le \Delta_f(z;p,n) \le 2,
\qquad
0\le \ell\bigl(\Delta_f(z;p,n)\bigr)\le M.
\end{equation}

Since $g_f(z)$ is the average of these surrogate terms over all retained positive-negative pairs, it follows that
\begin{equation}
0 \le g_f(z) \le M.
\label{eq: aucgr_boundedness_appendix}
\end{equation}

Let
\begin{equation}
\{g_1,\dots,g_{N_{\mathcal{G}}}\}
\subset
\mathcal{G}_{\alpha}
\end{equation}
be an $\varepsilon$-cover of $\mathcal{G}_{\alpha}$ under the sup norm, where
\begin{equation}
N_{\mathcal{G}}
=
\mathcal{N}(\mathcal{G}_{\alpha}, \varepsilon, \|\cdot\|_{\infty}).
\end{equation}

Applying Lemma~\ref{lem: finite_class_concentration_aucgr} to the bound in \cref{eq: aucgr_boundedness_appendix}, we obtain that with probability at least $1-\delta$,
\begin{equation}
\sup_{1 \le k \le N_{\mathcal{G}}}
\left|\mathbb{E}[g_k(z)]-\hat{\mathbb{E}}_{N}[g_k]\right|
\le
M\sqrt{
\frac{
\log(2N_{\mathcal{G}}/\delta)
}{
2N
}
}.
\label{eq: finite_cover_bound_aucgr}
\end{equation}

Now take any $g \in \mathcal{G}_{\alpha}$. Since $\{g_1,\dots,g_{N_{\mathcal{G}}}\}$ is an $\varepsilon$-cover, there exists some $g_k$ such that
\begin{equation}
\|g-g_k\|_{\infty} \le \varepsilon.
\end{equation}

This means that, for every anchor tuple $z$,
\begin{equation}
|g(z)-g_k(z)| \le \varepsilon.
\end{equation}

Taking expectation on both sides gives
\begin{equation}
\left|\mathbb{E}[g(z)]-\mathbb{E}[g_k(z)]\right|
\le
\mathbb{E}\left[|g(z)-g_k(z)|\right]
\le
\varepsilon.
\end{equation}

For the empirical averages, we have
\begin{equation}
\begin{aligned}
\left|\hat{\mathbb{E}}_{N}[g]-\hat{\mathbb{E}}_{N}[g_k]\right|
&=
\left|
\frac{1}{N}\sum_{i=1}^{N}
\bigl(g(z_i)-g_k(z_i)\bigr)
\right| \\
&\le
\frac{1}{N}\sum_{i=1}^{N}
|g(z_i)-g_k(z_i)| \\
&\le
\frac{1}{N}\sum_{i=1}^{N}
\varepsilon \\
&=
\varepsilon.
\end{aligned}
\end{equation}

We now insert $g_k$ between the population and empirical terms for $g$. By adding and subtracting $\mathbb{E}[g_k(z)]$ and $\hat{\mathbb{E}}_{N}[g_k]$, we obtain the exact identity
\begin{equation}
\begin{aligned}
\mathbb{E}[g(z)]-\hat{\mathbb{E}}_{N}[g]
=&
\bigl(\mathbb{E}[g(z)]-\mathbb{E}[g_k(z)]\bigr) \\
&+
\bigl(\mathbb{E}[g_k(z)]-\hat{\mathbb{E}}_{N}[g_k]\bigr) \\
&+
\bigl(\hat{\mathbb{E}}_{N}[g_k]-\hat{\mathbb{E}}_{N}[g]\bigr).
\end{aligned}
\end{equation}

Taking absolute values on both sides and applying the triangle inequality gives
\begin{equation}
\begin{aligned}
\left|
\mathbb{E}[g(z)]-\hat{\mathbb{E}}_{N}[g]
\right|
\le&
\left|
\mathbb{E}[g(z)]-\mathbb{E}[g_k(z)]
\right| \\
&+
\left|
\mathbb{E}[g_k(z)]-\hat{\mathbb{E}}_{N}[g_k]
\right| \\
&+
\left|
\hat{\mathbb{E}}_{N}[g_k]-\hat{\mathbb{E}}_{N}[g]
\right|.
\end{aligned}
\label{eq: cover_bridge_decomposition_aucgr}
\end{equation}

The first term is bounded by $\varepsilon$ from the expectation estimate above. The third term is also bounded by $\varepsilon$ from the empirical estimate above. For the middle term, \cref{eq: finite_cover_bound_aucgr} gives
\begin{equation}
\left|
\mathbb{E}[g_k(z)]-\hat{\mathbb{E}}_{N}[g_k]
\right|
\le
M\sqrt{
\frac{
\log(2N_{\mathcal{G}}/\delta)
}{
2N
}
}.
\end{equation}

Substituting these three bounds into \cref{eq: cover_bridge_decomposition_aucgr} yields
\begin{equation}
\left|
\mathbb{E}[g(z)]-\hat{\mathbb{E}}_{N}[g]
\right|
\le
2\varepsilon+
M\sqrt{
\frac{
\log(2N_{\mathcal{G}}/\delta)
}{
2N
}
}.
\end{equation}

Because this bound holds for every $g \in \mathcal{G}_{\alpha}$, we may take the supremum over $\mathcal{G}_{\alpha}$:
\begin{equation}
\sup_{g \in \mathcal{G}_{\alpha}}
\left|
\mathbb{E}[g(z)]-\hat{\mathbb{E}}_{N}[g]
\right|
\le
2\varepsilon+
M\sqrt{
\frac{
\log(2N_{\mathcal{G}}/\delta)
}{
2N
}
}.
\end{equation}

Every element of $\mathcal{G}_{\alpha}$ has the form $g_f$ for some $f \in \mathcal{F}$. Therefore,
\begin{equation}
\sup_{f \in \mathcal{F}}
\left|
\mathcal{R}_{\alpha}(f)-\hat{\mathcal{R}}_{\alpha}(f)
\right|
\le
2\varepsilon+
M\sqrt{
\frac{
\log(2N_{\mathcal{G}}/\delta)
}{
2N
}
}.
\end{equation}

By Assumption~\ref{ass: paucgr_generalization},
\begin{equation}
\log N_{\mathcal{G}}
\le
A\log(R/\varepsilon).
\end{equation}

Hence,
\begin{equation}
\log(2N_{\mathcal{G}}/\delta)
=
\log N_{\mathcal{G}}
+
\log(2/\delta)
\le
A\log(R/\varepsilon)
+
\log(2/\delta).
\end{equation}

Substituting this estimate into the previous display yields
\begin{equation}
\sup_{f \in \mathcal{F}}
\left|
\mathcal{R}_{\alpha}(f)-\hat{\mathcal{R}}_{\alpha}(f)
\right|
\le
2\varepsilon+
M\sqrt{
\frac{
A\log(R/\varepsilon)+\log(2/\delta)
}{
2N
}
}.
\end{equation}

The above argument is valid for any fixed $\varepsilon \in (0,R]$. We now take
\begin{equation}
\varepsilon = \frac{R}{N}.
\end{equation}

Since $N \ge 1$, this choice belongs to $(0,R]$, and the supremum bound implies that for every $f\in\mathcal{F}$,
\begin{equation}
\left|
\mathcal{R}_{\alpha}(f)-\hat{\mathcal{R}}_{\alpha}(f)
\right|
\le
\frac{2R}{N}
+
M\sqrt{
\frac{
A\log N+\log(2/\delta)
}{
2N
}
}.
\end{equation}

This completes the proof.
\end{proof}

\newpage

\section{Additional Experimental Settings}
\label{sec: supp_Additional Experiment Settings}
In this section, we make a supplementation to \cref{subsec: Experimental Setups}.

\subsection{CLIP Backbones}
\label{subsec: supp_CLIP Backbones}
Here, we provide a detailed summary of the $6$ types of CLIP backbones used in our experiments. They fall into $3$ categories:
\begin{itemize}
    \item \textbf{OpenAI CLIP ViT-L@224 and OpenAI CLIP ViT-L@336}~\cite{radford2021learning} establish the canonical formulation of contrastive language-image pretraining, defining the feature space that later CLIP variants build upon. Their large-scale but noisy training data gives them strong open-world generalization, while their ability to perceive fine-grained visual details remains limited. The 336-resolution variant uniquely extends CLIP's receptive field without altering the architecture, making it a standard choice for studying high-resolution alignment.
    \item \textbf{MetaCLIP ViT-L@224 and MetaCLIP ViT-H@224}~\cite{xudemystifying} distinguishes itself by reconstructing a high-quality, well-aligned training corpus using a principled filtering pipeline rather than relying on raw web data. This dataset-centric redesign leads to representations that are more stable, less noisy, and better calibrated than those of OpenAI CLIP. The ViT-H model further pushes scaling laws within the CLIP paradigm.
    \item \textbf{SigLIP ViT-SO@224 and SigLIP ViT-SO@384}~\cite{zhai2023sigmoid} departs from the classic softmax contrastive loss by introducing a sigmoid-based objective, fundamentally altering how positive and negative pairs influence training. This probabilistic formulation enables more fine-grained pairwise alignment and reduces overconfidence artifacts commonly observed in CLIP-like models. Its higher-resolution variant leverages the smoother objective to maintain stable training.
\end{itemize}

\subsection{Evaluation Protocol for P-Ability}
\label{subsec: supp_Evaluation Protocol for P-Ability}
\noindent \textbf{Datasets.} Following \cite{wangdiffusion,ma2025genhancer,li20252}, we use MMVP-VLM~\cite{tong2024eyes} to evaluate the \textbf{\textcolor{PerColor}{P-Ability}}. It is a benchmark designed to evaluate fine-grained visual perception by testing VLMs on a wide range of visual patterns. It contains human-designed symbols, shapes, transformations, and composites that isolate specific perceptual abilities such as symmetry, color matching, and geometric relations. The dataset includes tens of thousands of diverse pattern-label pairs, providing a controlled environment for assessing detailed visual understanding.

\noindent \textbf{Evaluation metrics.} Following \cite{tong2024eyes}, we report \textit{Accuracy (ACC)} as the evaluation metric. ACC measures the proportion of samples for which the model correctly recognizes the underlying visual pattern, serving as a direct indicator of fine-grained visual perception.

\subsection{Evaluation Protocol for D-Ability}
\label{subsec: supp_Evaluation Protocol for D-Ability}
We evaluate \textbf{\textcolor{DisColor}{D-Ability}} from three complementary perspectives. Zero-shot clustering examines whether image representations form separable semantic structures without using textual class names. Zero-shot classification evaluates whether image features can be aligned with class-name prompts. Zero-shot retrieval further tests cross-modal alignment between images and natural language descriptions.

\subsubsection{Zero-shot Clustering}
\noindent \textbf{Datasets.} Following \cite{radford2021learning}, we perform zero-shot clustering on $26$ standard datasets to evaluate whether the learned visual representations form separable semantic structures. These datasets cover handwritten digits, generic objects, fine-grained categories, textures, scenes, remote sensing images, medical images, and distribution-shifted recognition benchmarks.
\begin{itemize}
    \item \textbf{CIFAR-10}~\cite{krizhevsky2009learning} is a natural image dataset designed for general object recognition across $10$ categories. Its low-resolution $32\times32$ images contain large appearance variations, making it a compact benchmark for evaluating semantic separability.
    \item \textbf{CIFAR-100}~\cite{krizhevsky2009learning} extends CIFAR-10 to $100$ fine-grained categories. Compared with CIFAR-10, it provides a more challenging test of whether visual features can distinguish visually similar classes.
    \item \textbf{MNIST}~\cite{lecun2002gradient} contains grayscale handwritten digits from $0$ to $9$, covering $10$ categories. It is a standard benchmark for testing whether representations preserve simple but clearly separable visual categories.
    \item \textbf{Caltech-101}~\cite{fei2004learning} contains images from $101$ object categories and a background class, covering $102$ classes in total. The dataset includes substantial variation in object scale, pose, and background, making it useful for evaluating general object-level discrimination.
    \item \textbf{SUN397}~\cite{xiao2010sun} is a large-scale scene recognition dataset covering $397$ scene categories. It evaluates whether visual representations can separate scene-level semantics rather than only object-centric categories.
    \item \textbf{FGVC Aircraft}~\cite{maji2013fine} focuses on fine-grained aircraft recognition over $100$ aircraft variants. The categories are visually similar, so the dataset tests whether representations can capture subtle appearance cues.
    \item \textbf{Country-211}~\cite{radford2021learning} contains geolocation-related visual categories from $211$ countries and regions. It evaluates whether representations encode high-level geographic and cultural visual cues.
    \item \textbf{Stanford Cars}~\cite{krause20133d} is a fine-grained vehicle recognition dataset with $196$ car categories. It requires distinguishing car models with small differences in shape, texture, and local details.
    \item \textbf{Describable Textures (DTD)}~\cite{cimpoi2014describing} focuses on texture classification over $47$ human-interpretable texture categories. It is useful for testing whether representations capture local visual patterns.
    \item \textbf{Eurosat}~\cite{helber2019eurosat} is a satellite image dataset for land-use and land-cover recognition. It contains $10$ diverse remote-sensing categories such as forests, rivers, and residential areas.
    \item \textbf{Facial Emotion Recognition 2013 (FER2013)}~\cite{goodfellow2013challenges} is a facial expression recognition dataset with $7$ expression categories. It evaluates whether representations can separate subtle facial states under variations in identity, illumination, and pose.
    \item \textbf{Oxford Flowers 102 (Flowers-102)}~\cite{nilsback2008automated} is a fine-grained flower classification dataset with $102$ categories. The dataset emphasizes color, shape, and petal-level differences.
    \item \textbf{Food-101}~\cite{bossard2014food} contains food images from $101$ categories. It tests recognition under large intra-class appearance variation caused by cooking style, viewpoint, and presentation.
    \item \textbf{GTSRB}~\cite{stallkamp2012man} is a traffic sign recognition dataset with $43$ traffic sign categories. It evaluates the ability to distinguish symbolic visual categories with strong shape and color regularities.
    \item \textbf{PatchCamelyon (PCam)}~\cite{veeling2018rotation} is a medical image dataset for patch-level histopathology recognition. It provides a binary classification setting with $2$ classes and subtle texture and tissue patterns.
    \item \textbf{Oxford-IIIT Pets (Pets)}~\cite{parkhi2012cats} contains $37$ cat and dog breed categories. It evaluates fine-grained animal recognition with variations in pose, fur pattern, and background.
    \item \textbf{Rendered SST2}~\cite{radford2021learning} converts binary sentiment classification examples into rendered text images, covering $2$ classes. It tests whether visual representations can handle text-like visual inputs.
    \item \textbf{RESISC45}~\cite{cheng2017remote} is a remote sensing scene classification dataset with $45$ categories. It evaluates scene-level discrimination from aerial imagery.
    \item \textbf{STL10}~\cite{coates2011analysis} is an object recognition dataset with $10$ categories and higher-resolution images than CIFAR. It provides another compact benchmark for semantic clustering.
    \item \textbf{Pascal VOC 2007 (VOC2007)}~\cite{everingham2010pascal} is a multi-object recognition dataset containing $20$ object categories in natural scenes. It tests representation quality under cluttered backgrounds and multi-label visual content.
    \item \textbf{ImageNet-1K}~\cite{deng2009imagenet} is a large-scale recognition benchmark with $1000$ categories. It provides a comprehensive test of general visual discrimination.
    \item \textbf{ImageNet-V2}~\cite{recht2019imagenet} is a distribution-shifted ImageNet test set collected with the same $1000$-class label space. It evaluates whether separability is preserved under natural dataset shift.
    \item \textbf{ImageNet-Adversarial (ImageNet-Adv.)}~\cite{hendrycks2021natural} contains naturally occurring images from $200$ ImageNet categories that are difficult for standard ImageNet classifiers. It tests robustness to challenging visual examples.
    \item \textbf{ImageNet-Rendition (ImageNet-Ren.)}~\cite{hendrycks2021many} contains artistic renditions from $200$ ImageNet categories. It evaluates whether representations capture semantic identity beyond natural photo statistics.
    \item \textbf{ImageNet-OOD}~\cite{hendrycks2021natural} evaluates out-of-distribution recognition over $200$ ImageNet-related categories. It measures robustness when image statistics deviate from the original ImageNet distribution.
    \item \textbf{ObjectNet}~\cite{barbu2019objectnet} is an object recognition dataset with $313$ object categories, designed to stress test models under changes in viewpoint, rotation, and background. It evaluates robust semantic discrimination in realistic settings.
\end{itemize}

\noindent \textbf{Evaluation metrics.} Following \cite{li2024image}, we adopt $3$ widely-used metrics to evaluate clustering performance.
\begin{itemize}
    \item \textbf{Normalized Mutual Information (NMI)} measures the mutual dependence between predicted clusters and ground-truth labels, normalized to $[0,1]$. Higher values indicate better alignment between the clustering structure and the true class partition.
    \item \textbf{Accuracy (ACC)} evaluates the best-matched assignment between predicted clusters and ground-truth classes. It computes the fraction of correctly assigned samples after optimal label permutation using the Hungarian algorithm.
    \item \textbf{Adjusted Rand Index (ARI)} quantifies the similarity between two partitions by counting pairwise agreements, adjusted for random chance. An ARI of $0$ corresponds to random clustering, while higher values indicate more consistent partitioning.
\end{itemize}

\subsubsection{Zero-shot Classification}
\noindent \textbf{Datasets.} We use the same $26$ datasets introduced in the zero-shot clustering protocol. This shared dataset suite allows us to evaluate the same visual representations from two complementary perspectives. Clustering measures the intrinsic separability of image features without using text labels during grouping, while classification measures whether image features can be aligned with textual class names in a zero-shot manner.

\noindent \textbf{Evaluation metrics.} We adopt $3$ metrics to evaluate zero-shot classification performance.
\begin{itemize}
    \item \textbf{Top-1 Accuracy (ACC@1)} measures the percentage of samples whose top-ranked predicted class matches the ground-truth label. It directly reflects whether the strongest image-text match is correct.
    \item \textbf{Top-5 Accuracy (ACC@5)} measures whether the ground-truth label appears among the top $5$ predicted classes. This metric is useful for large-category datasets where several classes may be visually or semantically close. For datasets with fewer than $5$ classes, ACC@5 is not applicable.
    \item \textbf{Mean Recall (mRecall)} computes recall for each class and then averages over classes. It reduces the dominance of frequent classes and better reflects class-balanced recognition performance.
\end{itemize}

\subsubsection{Zero-shot Retrieval}
\noindent \textbf{Datasets.} We evaluate cross-modal retrieval on Flickr30k and COCO, which are widely used image-text retrieval benchmarks.
\begin{itemize}
    \item \textbf{Flickr30k}~\cite{young2014image} contains images collected from Flickr, each paired with multiple human-written captions. It evaluates whether visual representations can align images with natural language descriptions in open-domain scenes.
    \item \textbf{COCO}~\cite{lin2014microsoft} contains complex natural images with rich object interactions and multiple captions per image. It provides a more diverse retrieval benchmark with dense object and scene content.
\end{itemize}

\noindent \textbf{Evaluation metrics.} We evaluate both image-to-text and text-to-image retrieval.
\begin{itemize}
    \item \textbf{Image-to-Text Recall@5 (Image@5)} measures whether at least one correct caption of a query image appears in the top $5$ retrieved texts. Higher values indicate stronger image-to-language alignment.
    \item \textbf{Text-to-Image Recall@5 (Text@5)} measures whether the correct image of a query caption appears in the top $5$ retrieved images. Higher values indicate stronger language-to-image alignment.
\end{itemize}

\subsection{Evaluation Protocol for MLLMs}
\label{subsec: supp_Evaluation Protocol for MLLMs}
Following \cite{tong2024eyes,ma2025genhancer,li20252}, we evaluate MLLMs using two complementary groups of benchmarks: \textbf{Vision-Centric Benchmarks}, which focus on fine-grained visual perception and scene understanding, and \textbf{Conventional MLLM Benchmarks}, which measure robustness, hallucination resistance, and multimodal reasoning. Together, they provide a comprehensive view of an MLLM's visual competence, reliability, and general multimodal capability.

\noindent \textbf{Vision-Centric Benchmarks.} We evaluate on $4$ datasets:
\begin{itemize}
    \item \textbf{MMVP-MLLM}~\cite{tong2024eyes} evaluates the model's fine-grained visual perception by testing its ability to identify structured visual patterns. We report classification accuracy (ACC).
    \item \textbf{NaturalBench}~\cite{li2024naturalbench} examines multimodal reasoning across natural images using four metrics: overall accuracy (Acc), question-type accuracy (Q-Acc), instance-level accuracy (I-Acc), and group-level accuracy (G-Acc).
    \item \textbf{CV-Bench 2D}~\cite{tong2024cambrian} measures perception quality on 2D dense prediction tasks. We report the accuracy (ACC) of ADE20K~\cite{zhou2017scene} and COCO~\cite{caesar2018coco}.
    \item \textbf{CV-Bench 3D}~\cite{tong2024cambrian} assesses 3D spatial understanding through structured 3D reasoning queries. We report overall accuracy (ACC).
\end{itemize}

\noindent \textbf{Conventional MLLM Benchmarks.} We evaluate on $3$ datasets:
\begin{itemize}
    \item \textbf{POPE}~\cite{li2023evaluating} evaluates object hallucination robustness under three settings, including random (rand), popularity-based (pop), and adversarial (adv). Higher scores indicate stronger resistance to hallucination.
    \item \textbf{SciQA-IMG}~\cite{lu2022learn} assesses scientific visual question answering involving diagrams, plots, and structured visual cues. We report accuracy (ACC).
    \item \textbf{Hallusion}~\cite{guan2024hallusionbench} measures multimodal robustness by testing whether the model avoids visually induced false inferences. We report the average accuracy across all query types.
\end{itemize}

\section{Additional Experimental Results}
\label{sec: supp_Additional Experiment Results}

\subsection{Expanded Qualitative Results on P-Ability Performance}
\label{subsec: supp_Expanded Qualitative Results on P-Ability Performance}

\begin{figure}[htbp]
  \centering
   \includegraphics[width=0.95\linewidth]{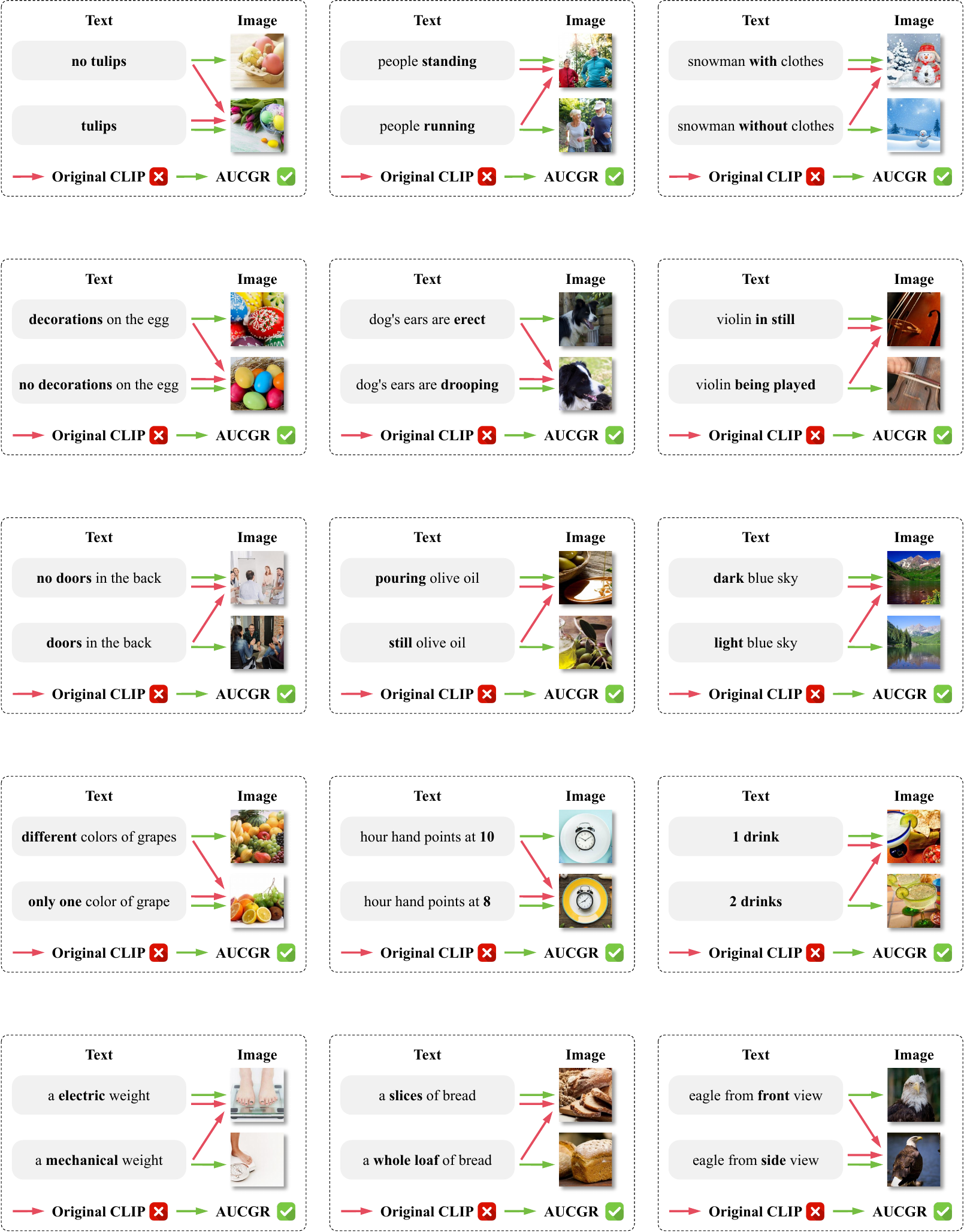}
   \caption{\textbf{Expanded qualitative results of AUCGR on the MMVP-VLM benchmark.} The predictions from the original CLIP and the AUCGR-enhanced CLIP are indicated by red and green arrows, respectively. AUCGR helps the model capture fine-grained visual details that are missed by the original CLIP.}
   \label{fig: expanded_mmvp_vlm}
\end{figure}

\begin{figure}[htbp]
  \centering
   \includegraphics[width=0.95\linewidth]{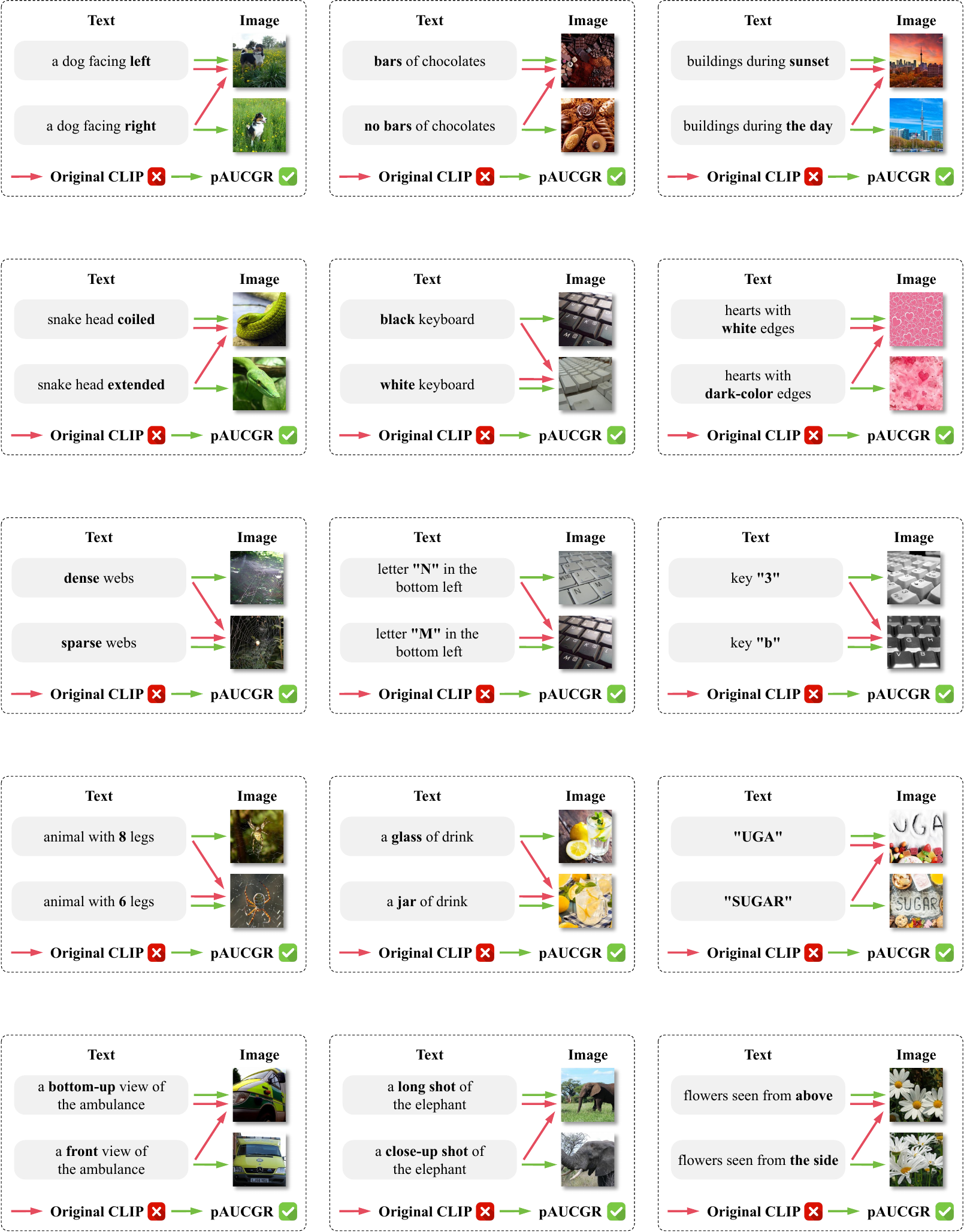}
   \caption{\textbf{Expanded qualitative results of pAUCGR on the MMVP-VLM benchmark.} The predictions from the original CLIP and the pAUCGR-enhanced CLIP are indicated by red and green arrows, respectively. pAUCGR further improves fine-grained visual perception by reducing misleading negative comparisons during optimization.}
   \label{fig: expanded_mmvp_vlm2}
\end{figure}

\cref{fig: expanded_mmvp_vlm} and \cref{fig: expanded_mmvp_vlm2} provide expanded qualitative results on the MMVP-VLM benchmark. \cref{fig: expanded_mmvp_vlm} visualizes the results obtained by AUCGR, while \cref{fig: expanded_mmvp_vlm2} visualizes the results obtained by pAUCGR. These examples cover visual questions that depend on local and fine-grained evidence, such as object state, color, spatial relation, structure, and viewpoint. The original CLIP often gives incorrect answers when the required evidence is subtle or appears only in a local region. After applying AUCGR, the enhanced representation better preserves such detail-sensitive cues and produces more accurate predictions. pAUCGR further improves the qualitative behavior by reducing misleading negative comparisons during training. The two groups of visualizations therefore provide complementary evidence that the proposed optimization strengthens \textbf{\textcolor{PerColor}{P-Ability}} without changing the visual encoder architecture.

\clearpage

\clearpage

\clearpage

\subsection{Expanded Quantitative Results on D-Ability Performance}
\label{subsec: supp_Expanded Quantitative Results on D-Ability Performance}

In this section, we provide a more comprehensive version of the quantitative evaluation for \textbf{\textcolor{DisColor}{D-Ability}}, covering zero-shot clustering, zero-shot classification, and zero-shot retrieval.

\subsubsection{Zero-shot Clustering}
\label{subsubsec: supp_Zero-shot Clustering}
The main paper reports zero-shot clustering results in terms of ACC in \cref{tab: discriminative ability results ACC}. To provide a more complete evaluation, \cref{tab: discriminative ability results ARI} and \cref{tab: discriminative ability results NMI} further report ARI and NMI on the same $26$ datasets. ARI evaluates whether the predicted clusters agree with the ground-truth semantic partitions after correcting for chance. NMI measures the information overlap between predicted clusters and ground-truth labels. \textbf{Across these complementary metrics, AUCGR and pAUCGR generally improve the DIVA and GenHancer baselines.} The consistent gains indicate that the improvement is not tied to a single clustering metric. Instead, the comparison-based optimization encourages more coherent semantic grouping in the visual feature space.

\begin{table}[htbp]

  \centering
  \renewcommand\arraystretch{2}
  \setlength{\tabcolsep}{3pt}
  \caption{\textbf{Performance of \textcolor{DisColor}{Discriminative Ability (D-Ability)} on 26 standard zero-shot clustering benchmarks in terms of \underline{ARI}.} `+DCR' denotes our conference. This table provides supplementary results for \cref{tab: discriminative ability results ACC}.}
  \resizebox{\linewidth}{!}{
    \begin{tabular}{c|c|cccc|cccc|c|cccc|cccc|c|cccc|cccc}
    \toprule
    \multirow[c]{2}{*}{\textbf{Dataset}} & \multicolumn{9}{c|}{\textbf{OpenAI CLIP ViT-L@224 (O-1)}} & \multicolumn{9}{c|}{\textbf{MetaCLIP ViT-L@224 (M-1)}} & \multicolumn{9}{c}{\textbf{SigLIP ViT-SO@224 (S-1)}} \\
    & \rotatebox[origin=c]{90}{\textbf{Original}} & \rotatebox[origin=c]{90}{\textbf{DIVA}} & \rotatebox[origin=c]{90}{\textbf{+DCR}} & \rotatebox[origin=c]{90}{\textbf{+AUCGR}} & \rotatebox[origin=c]{90}{\textbf{+pAUCGR}} & \rotatebox[origin=c]{90}{\textbf{GenHancer}} & \rotatebox[origin=c]{90}{\textbf{+DCR}} & \rotatebox[origin=c]{90}{\textbf{+AUCGR}} & \rotatebox[origin=c]{90}{\textbf{+pAUCGR}} & \rotatebox[origin=c]{90}{\textbf{Original}} & \rotatebox[origin=c]{90}{\textbf{DIVA}} & \rotatebox[origin=c]{90}{\textbf{+DCR}} & \rotatebox[origin=c]{90}{\textbf{+AUCGR}} & \rotatebox[origin=c]{90}{\textbf{+pAUCGR}} & \rotatebox[origin=c]{90}{\textbf{GenHancer}} & \rotatebox[origin=c]{90}{\textbf{+DCR}} & \rotatebox[origin=c]{90}{\textbf{+AUCGR}} & \rotatebox[origin=c]{90}{\textbf{+pAUCGR}} & \rotatebox[origin=c]{90}{\textbf{Original}} & \rotatebox[origin=c]{90}{\textbf{DIVA}} & \rotatebox[origin=c]{90}{\textbf{+DCR}} & \rotatebox[origin=c]{90}{\textbf{+AUCGR}} & \rotatebox[origin=c]{90}{\textbf{+pAUCGR}} & \rotatebox[origin=c]{90}{\textbf{GenHancer}} & \rotatebox[origin=c]{90}{\textbf{+DCR}} & \rotatebox[origin=c]{90}{\textbf{+AUCGR}} & \rotatebox[origin=c]{90}{\textbf{+pAUCGR}} \\
    \midrule
    CIFAR-10 & 0.786 & 0.675 & 0.616 & \cellcolor[rgb]{ .945,  .973,  .98}0.759 & \cellcolor[rgb]{ .945,  .973,  .98}0.784 & 0.252 & 0.538 & \cellcolor[rgb]{ .945,  .973,  .98}0.573 & \cellcolor[rgb]{ .945,  .973,  .98}0.565 & 0.791 & 0.630 & 0.774 & \cellcolor[rgb]{ .98,  .984,  .957}0.754 & \cellcolor[rgb]{ .98,  .984,  .957}0.801 & 0.399 & 0.751 & \cellcolor[rgb]{ .98,  .984,  .957}0.789 & \cellcolor[rgb]{ .98,  .984,  .957}0.793 & 0.878 & 0.404 & 0.740 & \cellcolor[rgb]{ .996,  .969,  .945}0.801 & \cellcolor[rgb]{ .996,  .969,  .945}0.937 & 0.936 & 0.938 & \cellcolor[rgb]{ .996,  .969,  .945}0.939 & \cellcolor[rgb]{ .996,  .969,  .945}0.939 \\
    CIFAR-100 & 0.384 & 0.286 & 0.274 & \cellcolor[rgb]{ .945,  .973,  .98}0.303 & \cellcolor[rgb]{ .945,  .973,  .98}0.382 & 0.088 & 0.209 & \cellcolor[rgb]{ .945,  .973,  .98}0.214 & \cellcolor[rgb]{ .945,  .973,  .98}0.226 & 0.396 & 0.309 & 0.340 & \cellcolor[rgb]{ .98,  .984,  .957}0.330 & \cellcolor[rgb]{ .98,  .984,  .957}0.431 & 0.182 & 0.366 & \cellcolor[rgb]{ .98,  .984,  .957}0.364 & \cellcolor[rgb]{ .98,  .984,  .957}0.403 & 0.551 & 0.206 & 0.537 & \cellcolor[rgb]{ .996,  .969,  .945}0.465 & \cellcolor[rgb]{ .996,  .969,  .945}0.538 & 0.548 & 0.549 & \cellcolor[rgb]{ .996,  .969,  .945}0.537 & \cellcolor[rgb]{ .996,  .969,  .945}0.534 \\
    MNIST & 0.449 & 0.407 & 0.333 & \cellcolor[rgb]{ .945,  .973,  .98}0.463 & \cellcolor[rgb]{ .945,  .973,  .98}0.501 & 0.365 & 0.422 & \cellcolor[rgb]{ .945,  .973,  .98}0.585 & \cellcolor[rgb]{ .945,  .973,  .98}0.406 & 0.496 & 0.513 & 0.476 & \cellcolor[rgb]{ .98,  .984,  .957}0.492 & \cellcolor[rgb]{ .98,  .984,  .957}0.472 & 0.524 & 0.449 & \cellcolor[rgb]{ .98,  .984,  .957}0.475 & \cellcolor[rgb]{ .98,  .984,  .957}0.475 & 0.555 & 0.544 & 0.605 & \cellcolor[rgb]{ .996,  .969,  .945}0.551 & \cellcolor[rgb]{ .996,  .969,  .945}0.562 & 0.571 & 0.559 & \cellcolor[rgb]{ .996,  .969,  .945}0.579 & \cellcolor[rgb]{ .996,  .969,  .945}0.561 \\
    Caltech-101 & 0.302 & 0.299 & 0.250 & \cellcolor[rgb]{ .945,  .973,  .98}0.283 & \cellcolor[rgb]{ .945,  .973,  .98}0.306 & 0.215 & 0.282 & \cellcolor[rgb]{ .945,  .973,  .98}0.276 & \cellcolor[rgb]{ .945,  .973,  .98}0.258 & 0.302 & 0.298 & 0.312 & \cellcolor[rgb]{ .98,  .984,  .957}0.309 & \cellcolor[rgb]{ .98,  .984,  .957}0.298 & 0.235 & 0.294 & \cellcolor[rgb]{ .98,  .984,  .957}0.305 & \cellcolor[rgb]{ .98,  .984,  .957}0.309 & 0.268 & 0.207 & 0.287 & \cellcolor[rgb]{ .996,  .969,  .945}0.268 & \cellcolor[rgb]{ .996,  .969,  .945}0.289 & 0.281 & 0.276 & \cellcolor[rgb]{ .996,  .969,  .945}0.294 & \cellcolor[rgb]{ .996,  .969,  .945}0.289 \\
    SUN397 & 0.379 & 0.363 & 0.371 & \cellcolor[rgb]{ .945,  .973,  .98}0.355 & \cellcolor[rgb]{ .945,  .973,  .98}0.380 & 0.186 & 0.306 & \cellcolor[rgb]{ .945,  .973,  .98}0.304 & \cellcolor[rgb]{ .945,  .973,  .98}0.336 & 0.386 & 0.326 & 0.321 & \cellcolor[rgb]{ .98,  .984,  .957}0.320 & \cellcolor[rgb]{ .98,  .984,  .957}0.387 & 0.113 & 0.302 & \cellcolor[rgb]{ .98,  .984,  .957}0.292 & \cellcolor[rgb]{ .98,  .984,  .957}0.369 & 0.432 & 0.052 & 0.418 & \cellcolor[rgb]{ .996,  .969,  .945}0.404 & \cellcolor[rgb]{ .996,  .969,  .945}0.430 & 0.430 & 0.431 & \cellcolor[rgb]{ .996,  .969,  .945}0.428 & \cellcolor[rgb]{ .996,  .969,  .945}0.431 \\
    FGVC Aircraft & 0.270 & 0.119 & 0.195 & \cellcolor[rgb]{ .945,  .973,  .98}0.179 & \cellcolor[rgb]{ .945,  .973,  .98}0.252 & 0.044 & 0.062 & \cellcolor[rgb]{ .945,  .973,  .98}0.064 & \cellcolor[rgb]{ .945,  .973,  .98}0.078 & 0.379 & 0.114 & 0.159 & \cellcolor[rgb]{ .98,  .984,  .957}0.203 & \cellcolor[rgb]{ .98,  .984,  .957}0.348 & 0.023 & 0.170 & \cellcolor[rgb]{ .98,  .984,  .957}0.157 & \cellcolor[rgb]{ .98,  .984,  .957}0.344 & 0.550 & 0.021 & 0.560 & \cellcolor[rgb]{ .996,  .969,  .945}0.526 & \cellcolor[rgb]{ .996,  .969,  .945}0.569 & 0.551 & 0.566 & \cellcolor[rgb]{ .996,  .969,  .945}0.572 & \cellcolor[rgb]{ .996,  .969,  .945}0.562 \\
    Country-211 & 0.052 & 0.021 & 0.044 & \cellcolor[rgb]{ .945,  .973,  .98}0.037 & \cellcolor[rgb]{ .945,  .973,  .98}0.050 & 0.017 & 0.017 & \cellcolor[rgb]{ .945,  .973,  .98}0.022 & \cellcolor[rgb]{ .945,  .973,  .98}0.018 & 0.055 & 0.024 & 0.035 & \cellcolor[rgb]{ .98,  .984,  .957}0.038 & \cellcolor[rgb]{ .98,  .984,  .957}0.055 & 0.012 & 0.035 & \cellcolor[rgb]{ .98,  .984,  .957}0.031 & \cellcolor[rgb]{ .98,  .984,  .957}0.053 & 0.048 & 0.010 & 0.048 & \cellcolor[rgb]{ .996,  .969,  .945}0.039 & \cellcolor[rgb]{ .996,  .969,  .945}0.049 & 0.048 & 0.047 & \cellcolor[rgb]{ .996,  .969,  .945}0.049 & \cellcolor[rgb]{ .996,  .969,  .945}0.048 \\
    Stanford Cars & 0.474 & 0.099 & 0.318 & \cellcolor[rgb]{ .945,  .973,  .98}0.260 & \cellcolor[rgb]{ .945,  .973,  .98}0.474 & 0.035 & 0.038 & \cellcolor[rgb]{ .945,  .973,  .98}0.100 & \cellcolor[rgb]{ .945,  .973,  .98}0.049 & 0.552 & 0.203 & 0.227 & \cellcolor[rgb]{ .98,  .984,  .957}0.275 & \cellcolor[rgb]{ .98,  .984,  .957}0.514 & 0.013 & 0.234 & \cellcolor[rgb]{ .98,  .984,  .957}0.211 & \cellcolor[rgb]{ .98,  .984,  .957}0.530 & 0.799 & 0.008 & 0.616 & \cellcolor[rgb]{ .996,  .969,  .945}0.768 & \cellcolor[rgb]{ .996,  .969,  .945}0.803 & 0.821 & 0.788 & \cellcolor[rgb]{ .996,  .969,  .945}0.787 & \cellcolor[rgb]{ .996,  .969,  .945}0.754 \\
    DTD & 0.333 & 0.317 & 0.255 & \cellcolor[rgb]{ .945,  .973,  .98}0.313 & \cellcolor[rgb]{ .945,  .973,  .98}0.355 & 0.147 & 0.275 & \cellcolor[rgb]{ .945,  .973,  .98}0.302 & \cellcolor[rgb]{ .945,  .973,  .98}0.291 & 0.338 & 0.282 & 0.319 & \cellcolor[rgb]{ .98,  .984,  .957}0.311 & \cellcolor[rgb]{ .98,  .984,  .957}0.347 & 0.201 & 0.322 & \cellcolor[rgb]{ .98,  .984,  .957}0.304 & \cellcolor[rgb]{ .98,  .984,  .957}0.348 & 0.404 & 0.070 & 0.388 & \cellcolor[rgb]{ .996,  .969,  .945}0.391 & \cellcolor[rgb]{ .996,  .969,  .945}0.386 & 0.386 & 0.413 & \cellcolor[rgb]{ .996,  .969,  .945}0.404 & \cellcolor[rgb]{ .996,  .969,  .945}0.410 \\
    Eurosat & 0.569 & 0.491 & 0.344 & \cellcolor[rgb]{ .945,  .973,  .98}0.649 & \cellcolor[rgb]{ .945,  .973,  .98}0.570 & 0.503 & 0.438 & \cellcolor[rgb]{ .945,  .973,  .98}0.578 & \cellcolor[rgb]{ .945,  .973,  .98}0.471 & 0.556 & 0.466 & 0.548 & \cellcolor[rgb]{ .98,  .984,  .957}0.515 & \cellcolor[rgb]{ .98,  .984,  .957}0.576 & 0.608 & 0.446 & \cellcolor[rgb]{ .98,  .984,  .957}0.430 & \cellcolor[rgb]{ .98,  .984,  .957}0.528 & 0.523 & 0.280 & 0.531 & \cellcolor[rgb]{ .996,  .969,  .945}0.430 & \cellcolor[rgb]{ .996,  .969,  .945}0.500 & 0.506 & 0.514 & \cellcolor[rgb]{ .996,  .969,  .945}0.534 & \cellcolor[rgb]{ .996,  .969,  .945}0.521 \\
    FER2013 & 0.054 & 0.064 & 0.064 & \cellcolor[rgb]{ .945,  .973,  .98}0.043 & \cellcolor[rgb]{ .945,  .973,  .98}0.054 & 0.100 & 0.061 & \cellcolor[rgb]{ .945,  .973,  .98}0.085 & \cellcolor[rgb]{ .945,  .973,  .98}0.057 & 0.057 & 0.036 & 0.046 & \cellcolor[rgb]{ .98,  .984,  .957}0.044 & \cellcolor[rgb]{ .98,  .984,  .957}0.057 & 0.043 & 0.046 & \cellcolor[rgb]{ .98,  .984,  .957}0.049 & \cellcolor[rgb]{ .98,  .984,  .957}0.050 & 0.105 & 0.024 & 0.069 & \cellcolor[rgb]{ .996,  .969,  .945}0.094 & \cellcolor[rgb]{ .996,  .969,  .945}0.099 & 0.112 & 0.105 & \cellcolor[rgb]{ .996,  .969,  .945}0.104 & \cellcolor[rgb]{ .996,  .969,  .945}0.103 \\
    Flowers-102 & 0.807 & 0.647 & 0.703 & \cellcolor[rgb]{ .945,  .973,  .98}0.738 & \cellcolor[rgb]{ .945,  .973,  .98}0.787 & 0.261 & 0.492 & \cellcolor[rgb]{ .945,  .973,  .98}0.449 & \cellcolor[rgb]{ .945,  .973,  .98}0.552 & 0.826 & 0.711 & 0.755 & \cellcolor[rgb]{ .98,  .984,  .957}0.764 & \cellcolor[rgb]{ .98,  .984,  .957}0.774 & 0.377 & 0.538 & \cellcolor[rgb]{ .98,  .984,  .957}0.539 & \cellcolor[rgb]{ .98,  .984,  .957}0.746 & 0.856 & 0.117 & 0.882 & \cellcolor[rgb]{ .996,  .969,  .945}0.830 & \cellcolor[rgb]{ .996,  .969,  .945}0.843 & 0.823 & 0.864 & \cellcolor[rgb]{ .996,  .969,  .945}0.852 & \cellcolor[rgb]{ .996,  .969,  .945}0.842 \\
    Food-101 & 0.777 & 0.562 & 0.683 & \cellcolor[rgb]{ .945,  .973,  .98}0.657 & \cellcolor[rgb]{ .945,  .973,  .98}0.772 & 0.182 & 0.364 & \cellcolor[rgb]{ .945,  .973,  .98}0.150 & \cellcolor[rgb]{ .945,  .973,  .98}0.439 & 0.735 & 0.613 & 0.646 & \cellcolor[rgb]{ .98,  .984,  .957}0.645 & \cellcolor[rgb]{ .98,  .984,  .957}0.749 & 0.047 & 0.483 & \cellcolor[rgb]{ .98,  .984,  .957}0.432 & \cellcolor[rgb]{ .98,  .984,  .957}0.742 & 0.801 & 0.008 & 0.826 & \cellcolor[rgb]{ .996,  .969,  .945}0.791 & \cellcolor[rgb]{ .996,  .969,  .945}0.817 & 0.793 & 0.812 & \cellcolor[rgb]{ .996,  .969,  .945}0.796 & \cellcolor[rgb]{ .996,  .969,  .945}0.811 \\
    GTSRB & 0.383 & 0.276 & 0.283 & \cellcolor[rgb]{ .945,  .973,  .98}0.350 & \cellcolor[rgb]{ .945,  .973,  .98}0.379 & 0.156 & 0.203 & \cellcolor[rgb]{ .945,  .973,  .98}0.275 & \cellcolor[rgb]{ .945,  .973,  .98}0.225 & 0.370 & 0.240 & 0.258 & \cellcolor[rgb]{ .98,  .984,  .957}0.333 & \cellcolor[rgb]{ .98,  .984,  .957}0.353 & 0.186 & 0.240 & \cellcolor[rgb]{ .98,  .984,  .957}0.250 & \cellcolor[rgb]{ .98,  .984,  .957}0.331 & 0.455 & 0.250 & 0.431 & \cellcolor[rgb]{ .996,  .969,  .945}0.245 & \cellcolor[rgb]{ .996,  .969,  .945}0.447 & 0.450 & 0.441 & \cellcolor[rgb]{ .996,  .969,  .945}0.481 & \cellcolor[rgb]{ .996,  .969,  .945}0.439 \\
    PCam & 0.067 & 0.090 & 0.052 & \cellcolor[rgb]{ .945,  .973,  .98}0.088 & \cellcolor[rgb]{ .945,  .973,  .98}0.064 & 0.079 & 0.088 & \cellcolor[rgb]{ .945,  .973,  .98}0.064 & \cellcolor[rgb]{ .945,  .973,  .98}0.090 & 0.058 & 0.086 & 0.064 & \cellcolor[rgb]{ .98,  .984,  .957}0.050 & \cellcolor[rgb]{ .98,  .984,  .957}0.058 & 0.064 & 0.061 & \cellcolor[rgb]{ .98,  .984,  .957}0.065 & \cellcolor[rgb]{ .98,  .984,  .957}0.057 & 0.023 & 0.002 & 0.020 & \cellcolor[rgb]{ .996,  .969,  .945}0.046 & \cellcolor[rgb]{ .996,  .969,  .945}0.023 & 0.025 & 0.023 & \cellcolor[rgb]{ .996,  .969,  .945}0.023 & \cellcolor[rgb]{ .996,  .969,  .945}0.023 \\
    Pets & 0.538 & 0.301 & 0.409 & \cellcolor[rgb]{ .945,  .973,  .98}0.396 & \cellcolor[rgb]{ .945,  .973,  .98}0.518 & 0.098 & 0.138 & \cellcolor[rgb]{ .945,  .973,  .98}0.305 & \cellcolor[rgb]{ .945,  .973,  .98}0.166 & 0.656 & 0.317 & 0.321 & \cellcolor[rgb]{ .98,  .984,  .957}0.373 & \cellcolor[rgb]{ .98,  .984,  .957}0.709 & 0.081 & 0.238 & \cellcolor[rgb]{ .98,  .984,  .957}0.265 & \cellcolor[rgb]{ .98,  .984,  .957}0.445 & 0.802 & 0.064 & 0.805 & \cellcolor[rgb]{ .996,  .969,  .945}0.771 & \cellcolor[rgb]{ .996,  .969,  .945}0.810 & 0.786 & 0.833 & \cellcolor[rgb]{ .996,  .969,  .945}0.796 & \cellcolor[rgb]{ .996,  .969,  .945}0.807 \\
    Rendered SST2 & 0.001 & 0.001 & 0.000 & \cellcolor[rgb]{ .945,  .973,  .98}0.007 & \cellcolor[rgb]{ .945,  .973,  .98}0.001 & 0.001 & 0.002 & \cellcolor[rgb]{ .945,  .973,  .98}0.003 & \cellcolor[rgb]{ .945,  .973,  .98}0.002 & 0.010 & 0.006 & 0.001 & \cellcolor[rgb]{ .98,  .984,  .957}0.003 & \cellcolor[rgb]{ .98,  .984,  .957}0.003 & 0.000 & 0.003 & \cellcolor[rgb]{ .98,  .984,  .957}0.003 & \cellcolor[rgb]{ .98,  .984,  .957}0.005 & 0.004 & 0.001 & 0.007 & \cellcolor[rgb]{ .996,  .969,  .945}0.006 & \cellcolor[rgb]{ .996,  .969,  .945}0.003 & 0.001 & 0.003 & \cellcolor[rgb]{ .996,  .969,  .945}0.003 & \cellcolor[rgb]{ .996,  .969,  .945}0.003 \\
    Resisc45 & 0.626 & 0.547 & 0.595 & \cellcolor[rgb]{ .945,  .973,  .98}0.639 & \cellcolor[rgb]{ .945,  .973,  .98}0.641 & 0.396 & 0.434 & \cellcolor[rgb]{ .945,  .973,  .98}0.355 & \cellcolor[rgb]{ .945,  .973,  .98}0.468 & 0.576 & 0.579 & 0.492 & \cellcolor[rgb]{ .98,  .984,  .957}0.516 & \cellcolor[rgb]{ .98,  .984,  .957}0.603 & 0.284 & 0.578 & \cellcolor[rgb]{ .98,  .984,  .957}0.604 & \cellcolor[rgb]{ .98,  .984,  .957}0.588 & 0.648 & 0.329 & 0.679 & \cellcolor[rgb]{ .996,  .969,  .945}0.643 & \cellcolor[rgb]{ .996,  .969,  .945}0.619 & 0.611 & 0.642 & \cellcolor[rgb]{ .996,  .969,  .945}0.649 & \cellcolor[rgb]{ .996,  .969,  .945}0.633 \\
    STL10 & 0.921 & 0.953 & 0.910 & \cellcolor[rgb]{ .945,  .973,  .98}0.890 & \cellcolor[rgb]{ .945,  .973,  .98}0.921 & 0.410 & 0.746 & \cellcolor[rgb]{ .945,  .973,  .98}0.810 & \cellcolor[rgb]{ .945,  .973,  .98}0.781 & 0.920 & 0.905 & 0.913 & \cellcolor[rgb]{ .98,  .984,  .957}0.917 & \cellcolor[rgb]{ .98,  .984,  .957}0.830 & 0.348 & 0.796 & \cellcolor[rgb]{ .98,  .984,  .957}0.766 & \cellcolor[rgb]{ .98,  .984,  .957}0.920 & 0.903 & 0.788 & 0.900 & \cellcolor[rgb]{ .996,  .969,  .945}0.984 & \cellcolor[rgb]{ .996,  .969,  .945}0.902 & 0.905 & 0.903 & \cellcolor[rgb]{ .996,  .969,  .945}0.818 & \cellcolor[rgb]{ .996,  .969,  .945}0.903 \\
    VOC2007 & 0.229 & 0.224 & 0.216 & \cellcolor[rgb]{ .945,  .973,  .98}0.234 & \cellcolor[rgb]{ .945,  .973,  .98}0.239 & 0.133 & 0.213 & \cellcolor[rgb]{ .945,  .973,  .98}0.158 & \cellcolor[rgb]{ .945,  .973,  .98}0.223 & 0.197 & 0.182 & 0.188 & \cellcolor[rgb]{ .98,  .984,  .957}0.162 & \cellcolor[rgb]{ .98,  .984,  .957}0.176 & 0.095 & 0.129 & \cellcolor[rgb]{ .98,  .984,  .957}0.182 & \cellcolor[rgb]{ .98,  .984,  .957}0.138 & 0.261 & 0.030 & 0.241 & \cellcolor[rgb]{ .996,  .969,  .945}0.248 & \cellcolor[rgb]{ .996,  .969,  .945}0.252 & 0.281 & 0.230 & \cellcolor[rgb]{ .996,  .969,  .945}0.280 & \cellcolor[rgb]{ .996,  .969,  .945}0.232 \\
    ImageNet-1K & 0.418 & 0.246 & 0.331 & \cellcolor[rgb]{ .945,  .973,  .98}0.319 & \cellcolor[rgb]{ .945,  .973,  .98}0.419 & 0.080 & 0.212 & \cellcolor[rgb]{ .945,  .973,  .98}0.166 & \cellcolor[rgb]{ .945,  .973,  .98}0.183 & 0.448 & 0.266 & 0.303 & \cellcolor[rgb]{ .98,  .984,  .957}0.335 & \cellcolor[rgb]{ .98,  .984,  .957}0.443 & 0.053 & 0.279 & \cellcolor[rgb]{ .98,  .984,  .957}0.260 & \cellcolor[rgb]{ .98,  .984,  .957}0.410 & 0.525 & 0.041 & 0.506 & \cellcolor[rgb]{ .996,  .969,  .945}0.469 & \cellcolor[rgb]{ .996,  .969,  .945}0.522 & 0.513 & 0.522 & \cellcolor[rgb]{ .996,  .969,  .945}0.523 & \cellcolor[rgb]{ .996,  .969,  .945}0.524 \\
    ImageNet-V2 & 0.259 & 0.154 & 0.198 & \cellcolor[rgb]{ .945,  .973,  .98}0.212 & \cellcolor[rgb]{ .945,  .973,  .98}0.255 & 0.053 & 0.135 & \cellcolor[rgb]{ .945,  .973,  .98}0.085 & \cellcolor[rgb]{ .945,  .973,  .98}0.114 & 0.271 & 0.171 & 0.197 & \cellcolor[rgb]{ .98,  .984,  .957}0.225 & \cellcolor[rgb]{ .98,  .984,  .957}0.277 & 0.031 & 0.155 & \cellcolor[rgb]{ .98,  .984,  .957}0.150 & \cellcolor[rgb]{ .98,  .984,  .957}0.240 & 0.327 & 0.014 & 0.143 & \cellcolor[rgb]{ .996,  .969,  .945}0.278 & \cellcolor[rgb]{ .996,  .969,  .945}0.325 & 0.317 & 0.327 & \cellcolor[rgb]{ .996,  .969,  .945}0.325 & \cellcolor[rgb]{ .996,  .969,  .945}0.325 \\
    ImageNet-Adv. & 0.274 & 0.074 & 0.244 & \cellcolor[rgb]{ .945,  .973,  .98}0.152 & \cellcolor[rgb]{ .945,  .973,  .98}0.289 & 0.034 & 0.114 & \cellcolor[rgb]{ .945,  .973,  .98}0.045 & \cellcolor[rgb]{ .945,  .973,  .98}0.057 & 0.290 & 0.095 & 0.165 & \cellcolor[rgb]{ .98,  .984,  .957}0.197 & \cellcolor[rgb]{ .98,  .984,  .957}0.293 & 0.025 & 0.093 & \cellcolor[rgb]{ .98,  .984,  .957}0.095 & \cellcolor[rgb]{ .98,  .984,  .957}0.263 & 0.283 & 0.024 & 0.108 & \cellcolor[rgb]{ .996,  .969,  .945}0.175 & \cellcolor[rgb]{ .996,  .969,  .945}0.276 & 0.250 & 0.271 & \cellcolor[rgb]{ .996,  .969,  .945}0.266 & \cellcolor[rgb]{ .996,  .969,  .945}0.279 \\
    ImageNet-Ren. & 0.280 & 0.080 & 0.054 & \cellcolor[rgb]{ .945,  .973,  .98}0.112 & \cellcolor[rgb]{ .945,  .973,  .98}0.293 & 0.018 & 0.059 & \cellcolor[rgb]{ .945,  .973,  .98}0.066 & \cellcolor[rgb]{ .945,  .973,  .98}0.053 & 0.237 & 0.077 & 0.072 & \cellcolor[rgb]{ .98,  .984,  .957}0.089 & \cellcolor[rgb]{ .98,  .984,  .957}0.225 & 0.015 & 0.051 & \cellcolor[rgb]{ .98,  .984,  .957}0.050 & \cellcolor[rgb]{ .98,  .984,  .957}0.185 & 0.287 & 0.006 & 0.266 & \cellcolor[rgb]{ .996,  .969,  .945}0.198 & \cellcolor[rgb]{ .996,  .969,  .945}0.292 & 0.252 & 0.278 & \cellcolor[rgb]{ .996,  .969,  .945}0.266 & \cellcolor[rgb]{ .996,  .969,  .945}0.285 \\
    ImageNet-OOD & 0.104 & 0.116 & 0.094 & \cellcolor[rgb]{ .945,  .973,  .98}0.105 & \cellcolor[rgb]{ .945,  .973,  .98}0.099 & 0.080 & 0.097 & \cellcolor[rgb]{ .945,  .973,  .98}0.140 & \cellcolor[rgb]{ .945,  .973,  .98}0.109 & 0.095 & 0.111 & 0.095 & \cellcolor[rgb]{ .98,  .984,  .957}0.092 & \cellcolor[rgb]{ .98,  .984,  .957}0.094 & 0.083 & 0.095 & \cellcolor[rgb]{ .98,  .984,  .957}0.101 & \cellcolor[rgb]{ .98,  .984,  .957}0.094 & 0.101 & 0.018 & 0.112 & \cellcolor[rgb]{ .996,  .969,  .945}0.111 & \cellcolor[rgb]{ .996,  .969,  .945}0.104 & 0.099 & 0.100 & \cellcolor[rgb]{ .996,  .969,  .945}0.107 & \cellcolor[rgb]{ .996,  .969,  .945}0.106 \\
    ObjectNet & 0.312 & 0.137 & 0.286 & \cellcolor[rgb]{ .945,  .973,  .98}0.261 & \cellcolor[rgb]{ .945,  .973,  .98}0.314 & 0.020 & 0.267 & \cellcolor[rgb]{ .945,  .973,  .98}0.045 & \cellcolor[rgb]{ .945,  .973,  .98}0.092 & 0.379 & 0.265 & 0.316 & \cellcolor[rgb]{ .98,  .984,  .957}0.364 & \cellcolor[rgb]{ .98,  .984,  .957}0.391 & 0.007 & 0.269 & \cellcolor[rgb]{ .98,  .984,  .957}0.239 & \cellcolor[rgb]{ .98,  .984,  .957}0.380 & 0.322 & 0.006 & 0.252 & \cellcolor[rgb]{ .996,  .969,  .945}0.228 & \cellcolor[rgb]{ .996,  .969,  .945}0.328 & 0.300 & 0.205 & \cellcolor[rgb]{ .996,  .969,  .945}0.318 & \cellcolor[rgb]{ .996,  .969,  .945}0.328 \\
    \midrule
    Avg. ARI & 0.386 & 0.290 & 0.312 & \cellcolor[rgb]{ .945,  .973,  .98}0.339 & \cellcolor[rgb]{ .945,  .973,  .98}\textbf{0.388} & 0.152 & 0.239 & \cellcolor[rgb]{ .945,  .973,  .98}0.239 & \cellcolor[rgb]{ .945,  .973,  .98}\textbf{0.243} & 0.399 & 0.301 & 0.321 & \cellcolor[rgb]{ .98,  .984,  .957}0.333 & \cellcolor[rgb]{ .98,  .984,  .957}\textbf{0.395} & 0.156 & 0.285 & \cellcolor[rgb]{ .98,  .984,  .957}0.285 & \cellcolor[rgb]{ .98,  .984,  .957}\textbf{0.375} & 0.450 & 0.136 & 0.422 & \cellcolor[rgb]{ .996,  .969,  .945}0.414 & \cellcolor[rgb]{ .996,  .969,  .945}\textbf{0.451} & 0.446 & 0.448 & \cellcolor[rgb]{ .996,  .969,  .945}\textbf{0.451} & \cellcolor[rgb]{ .996,  .969,  .945}0.450 \\
    \bottomrule
    \end{tabular}%
   }
\label{tab: discriminative ability results ARI}

\end{table}

\newpage

\begin{table}[htbp]

  \centering
  \renewcommand\arraystretch{2}
  \setlength{\tabcolsep}{3pt}
  \caption{\textbf{Performance of \textcolor{DisColor}{Discriminative Ability (D-Ability)} on 26 standard zero-shot clustering benchmarks in terms of \underline{NMI}.} `+DCR' denotes our conference version. This table provides supplementary results for \cref{tab: discriminative ability results ACC}.}
  \resizebox{\linewidth}{!}{
    \begin{tabular}{c|c|cccc|cccc|c|cccc|cccc|c|cccc|cccc}
    \toprule
    \multirow[c]{2}{*}{\textbf{Dataset}} & \multicolumn{9}{c|}{\textbf{OpenAI CLIP ViT-L@224 (O-1)}} & \multicolumn{9}{c|}{\textbf{MetaCLIP ViT-L@224 (M-1)}} & \multicolumn{9}{c}{\textbf{SigLIP ViT-SO@224 (S-1)}} \\
    & \rotatebox[origin=c]{90}{\textbf{Original}} & \rotatebox[origin=c]{90}{\textbf{DIVA}} & \rotatebox[origin=c]{90}{\textbf{+DCR}} & \rotatebox[origin=c]{90}{\textbf{+AUCGR}} & \rotatebox[origin=c]{90}{\textbf{+pAUCGR}} & \rotatebox[origin=c]{90}{\textbf{GenHancer}} & \rotatebox[origin=c]{90}{\textbf{+DCR}} & \rotatebox[origin=c]{90}{\textbf{+AUCGR}} & \rotatebox[origin=c]{90}{\textbf{+pAUCGR}} & \rotatebox[origin=c]{90}{\textbf{Original}} & \rotatebox[origin=c]{90}{\textbf{DIVA}} & \rotatebox[origin=c]{90}{\textbf{+DCR}} & \rotatebox[origin=c]{90}{\textbf{+AUCGR}} & \rotatebox[origin=c]{90}{\textbf{+pAUCGR}} & \rotatebox[origin=c]{90}{\textbf{GenHancer}} & \rotatebox[origin=c]{90}{\textbf{+DCR}} & \rotatebox[origin=c]{90}{\textbf{+AUCGR}} & \rotatebox[origin=c]{90}{\textbf{+pAUCGR}} & \rotatebox[origin=c]{90}{\textbf{Original}} & \rotatebox[origin=c]{90}{\textbf{DIVA}} & \rotatebox[origin=c]{90}{\textbf{+DCR}} & \rotatebox[origin=c]{90}{\textbf{+AUCGR}} & \rotatebox[origin=c]{90}{\textbf{+pAUCGR}} & \rotatebox[origin=c]{90}{\textbf{GenHancer}} & \rotatebox[origin=c]{90}{\textbf{+DCR}} & \rotatebox[origin=c]{90}{\textbf{+AUCGR}} & \rotatebox[origin=c]{90}{\textbf{+pAUCGR}} \\
    \midrule
    CIFAR-10 & 0.841 & 0.745 & 0.716 & \cellcolor[rgb]{ .945,  .973,  .98}0.815 & \cellcolor[rgb]{ .945,  .973,  .98}0.839 & 0.364 & 0.636 & \cellcolor[rgb]{ .945,  .973,  .98}0.677 & \cellcolor[rgb]{ .945,  .973,  .98}0.662 & 0.860 & 0.765 & 0.839 & \cellcolor[rgb]{ .98,  .984,  .957}0.831 & \cellcolor[rgb]{ .98,  .984,  .957}0.865 & 0.508 & 0.811 & \cellcolor[rgb]{ .98,  .984,  .957}0.845 & \cellcolor[rgb]{ .98,  .984,  .957}0.862 & 0.907 & 0.531 & 0.931 & \cellcolor[rgb]{ .996,  .969,  .945}0.863 & \cellcolor[rgb]{ .996,  .969,  .945}0.937 & 0.935 & 0.937 & \cellcolor[rgb]{ .996,  .969,  .945}0.937 & \cellcolor[rgb]{ .996,  .969,  .945}0.938 \\
    CIFAR-100 & 0.679 & 0.611 & 0.592 & \cellcolor[rgb]{ .945,  .973,  .98}0.633 & \cellcolor[rgb]{ .945,  .973,  .98}0.675 & 0.381 & 0.557 & \cellcolor[rgb]{ .945,  .973,  .98}0.541 & \cellcolor[rgb]{ .945,  .973,  .98}0.573 & 0.710 & 0.651 & 0.669 & \cellcolor[rgb]{ .98,  .984,  .957}0.667 & \cellcolor[rgb]{ .98,  .984,  .957}0.722 & 0.498 & 0.683 & \cellcolor[rgb]{ .98,  .984,  .957}0.682 & \cellcolor[rgb]{ .98,  .984,  .957}0.708 & 0.798 & 0.520 & 0.786 & \cellcolor[rgb]{ .996,  .969,  .945}0.744 & \cellcolor[rgb]{ .996,  .969,  .945}0.791 & 0.789 & 0.791 & \cellcolor[rgb]{ .996,  .969,  .945}0.791 & \cellcolor[rgb]{ .996,  .969,  .945}0.788 \\
    MNIST & 0.617 & 0.556 & 0.572 & \cellcolor[rgb]{ .945,  .973,  .98}0.617 & \cellcolor[rgb]{ .945,  .973,  .98}0.647 & 0.525 & 0.538 & \cellcolor[rgb]{ .945,  .973,  .98}0.735 & \cellcolor[rgb]{ .945,  .973,  .98}0.526 & 0.639 & 0.690 & 0.614 & \cellcolor[rgb]{ .98,  .984,  .957}0.627 & \cellcolor[rgb]{ .98,  .984,  .957}0.661 & 0.662 & 0.596 & \cellcolor[rgb]{ .98,  .984,  .957}0.619 & \cellcolor[rgb]{ .98,  .984,  .957}0.624 & 0.696 & 0.680 & 0.746 & \cellcolor[rgb]{ .996,  .969,  .945}0.691 & \cellcolor[rgb]{ .996,  .969,  .945}0.702 & 0.705 & 0.700 & \cellcolor[rgb]{ .996,  .969,  .945}0.685 & \cellcolor[rgb]{ .996,  .969,  .945}0.700 \\
    Caltech-101 & 0.809 & 0.780 & 0.757 & \cellcolor[rgb]{ .945,  .973,  .98}0.793 & \cellcolor[rgb]{ .945,  .973,  .98}0.810 & 0.662 & 0.750 & \cellcolor[rgb]{ .945,  .973,  .98}0.772 & \cellcolor[rgb]{ .945,  .973,  .98}0.759 & 0.823 & 0.807 & 0.818 & \cellcolor[rgb]{ .98,  .984,  .957}0.827 & \cellcolor[rgb]{ .98,  .984,  .957}0.820 & 0.662 & 0.811 & \cellcolor[rgb]{ .98,  .984,  .957}0.816 & \cellcolor[rgb]{ .98,  .984,  .957}0.826 & 0.805 & 0.601 & 0.805 & \cellcolor[rgb]{ .996,  .969,  .945}0.807 & \cellcolor[rgb]{ .996,  .969,  .945}0.809 & 0.809 & 0.807 & \cellcolor[rgb]{ .996,  .969,  .945}0.813 & \cellcolor[rgb]{ .996,  .969,  .945}0.811 \\
    SUN397 & 0.759 & 0.743 & 0.744 & \cellcolor[rgb]{ .945,  .973,  .98}0.740 & \cellcolor[rgb]{ .945,  .973,  .98}0.755 & 0.588 & 0.693 & \cellcolor[rgb]{ .945,  .973,  .98}0.697 & \cellcolor[rgb]{ .945,  .973,  .98}0.716 & 0.764 & 0.715 & 0.715 & \cellcolor[rgb]{ .98,  .984,  .957}0.715 & \cellcolor[rgb]{ .98,  .984,  .957}0.763 & 0.494 & 0.708 & \cellcolor[rgb]{ .98,  .984,  .957}0.703 & \cellcolor[rgb]{ .98,  .984,  .957}0.752 & 0.796 & 0.312 & 0.790 & \cellcolor[rgb]{ .996,  .969,  .945}0.782 & \cellcolor[rgb]{ .996,  .969,  .945}0.795 & 0.796 & 0.795 & \cellcolor[rgb]{ .996,  .969,  .945}0.794 & \cellcolor[rgb]{ .996,  .969,  .945}0.795 \\
    FGVC Aircraft & 0.635 & 0.515 & 0.590 & \cellcolor[rgb]{ .945,  .973,  .98}0.569 & \cellcolor[rgb]{ .945,  .973,  .98}0.627 & 0.393 & 0.430 & \cellcolor[rgb]{ .945,  .973,  .98}0.424 & \cellcolor[rgb]{ .945,  .973,  .98}0.451 & 0.741 & 0.517 & 0.557 & \cellcolor[rgb]{ .98,  .984,  .957}0.596 & \cellcolor[rgb]{ .98,  .984,  .957}0.719 & 0.337 & 0.558 & \cellcolor[rgb]{ .98,  .984,  .957}0.561 & \cellcolor[rgb]{ .98,  .984,  .957}0.710 & 0.841 & 0.333 & 0.838 & \cellcolor[rgb]{ .996,  .969,  .945}0.818 & \cellcolor[rgb]{ .996,  .969,  .945}0.848 & 0.836 & 0.842 & \cellcolor[rgb]{ .996,  .969,  .945}0.842 & \cellcolor[rgb]{ .996,  .969,  .945}0.845 \\
    Country-211 & 0.370 & 0.282 & 0.350 & \cellcolor[rgb]{ .945,  .973,  .98}0.334 & \cellcolor[rgb]{ .945,  .973,  .98}0.367 & 0.273 & 0.269 & \cellcolor[rgb]{ .945,  .973,  .98}0.295 & \cellcolor[rgb]{ .945,  .973,  .98}0.274 & 0.377 & 0.300 & 0.325 & \cellcolor[rgb]{ .98,  .984,  .957}0.333 & \cellcolor[rgb]{ .98,  .984,  .957}0.375 & 0.254 & 0.324 & \cellcolor[rgb]{ .98,  .984,  .957}0.319 & \cellcolor[rgb]{ .98,  .984,  .957}0.369 & 0.357 & 0.228 & 0.353 & \cellcolor[rgb]{ .996,  .969,  .945}0.336 & \cellcolor[rgb]{ .996,  .969,  .945}0.356 & 0.355 & 0.356 & \cellcolor[rgb]{ .996,  .969,  .945}0.355 & \cellcolor[rgb]{ .996,  .969,  .945}0.357 \\
    Stanford Cars & 0.825 & 0.521 & 0.718 & \cellcolor[rgb]{ .945,  .973,  .98}0.689 & \cellcolor[rgb]{ .945,  .973,  .98}0.824 & 0.417 & 0.427 & \cellcolor[rgb]{ .945,  .973,  .98}0.522 & \cellcolor[rgb]{ .945,  .973,  .98}0.445 & 0.856 & 0.618 & 0.646 & \cellcolor[rgb]{ .98,  .984,  .957}0.687 & \cellcolor[rgb]{ .98,  .984,  .957}0.840 & 0.358 & 0.656 & \cellcolor[rgb]{ .98,  .984,  .957}0.639 & \cellcolor[rgb]{ .98,  .984,  .957}0.843 & 0.949 & 0.325 & 0.950 & \cellcolor[rgb]{ .996,  .969,  .945}0.939 & \cellcolor[rgb]{ .996,  .969,  .945}0.952 & 0.951 & 0.950 & \cellcolor[rgb]{ .996,  .969,  .945}0.949 & \cellcolor[rgb]{ .996,  .969,  .945}0.942 \\
    DTD & 0.623 & 0.603 & 0.546 & \cellcolor[rgb]{ .945,  .973,  .98}0.600 & \cellcolor[rgb]{ .945,  .973,  .98}0.644 & 0.440 & 0.562 & \cellcolor[rgb]{ .945,  .973,  .98}0.589 & \cellcolor[rgb]{ .945,  .973,  .98}0.581 & 0.624 & 0.575 & 0.604 & \cellcolor[rgb]{ .98,  .984,  .957}0.601 & \cellcolor[rgb]{ .98,  .984,  .957}0.630 & 0.497 & 0.598 & \cellcolor[rgb]{ .98,  .984,  .957}0.587 & \cellcolor[rgb]{ .98,  .984,  .957}0.631 & 0.681 & 0.337 & 0.663 & \cellcolor[rgb]{ .996,  .969,  .945}0.664 & \cellcolor[rgb]{ .996,  .969,  .945}0.664 & 0.669 & 0.678 & \cellcolor[rgb]{ .996,  .969,  .945}0.677 & \cellcolor[rgb]{ .996,  .969,  .945}0.673 \\
    Eurosat & 0.682 & 0.600 & 0.477 & \cellcolor[rgb]{ .945,  .973,  .98}0.720 & \cellcolor[rgb]{ .945,  .973,  .98}0.681 & 0.610 & 0.549 & \cellcolor[rgb]{ .945,  .973,  .98}0.680 & \cellcolor[rgb]{ .945,  .973,  .98}0.569 & 0.675 & 0.601 & 0.655 & \cellcolor[rgb]{ .98,  .984,  .957}0.633 & \cellcolor[rgb]{ .98,  .984,  .957}0.682 & 0.677 & 0.553 & \cellcolor[rgb]{ .98,  .984,  .957}0.546 & \cellcolor[rgb]{ .98,  .984,  .957}0.645 & 0.637 & 0.444 & 0.649 & \cellcolor[rgb]{ .996,  .969,  .945}0.562 & \cellcolor[rgb]{ .996,  .969,  .945}0.615 & 0.619 & 0.628 & \cellcolor[rgb]{ .996,  .969,  .945}0.650 & \cellcolor[rgb]{ .996,  .969,  .945}0.634 \\
    FER2013 & 0.085 & 0.094 & 0.094 & \cellcolor[rgb]{ .945,  .973,  .98}0.062 & \cellcolor[rgb]{ .945,  .973,  .98}0.086 & 0.131 & 0.088 & \cellcolor[rgb]{ .945,  .973,  .98}0.109 & \cellcolor[rgb]{ .945,  .973,  .98}0.085 & 0.083 & 0.057 & 0.077 & \cellcolor[rgb]{ .98,  .984,  .957}0.079 & \cellcolor[rgb]{ .98,  .984,  .957}0.084 & 0.060 & 0.069 & \cellcolor[rgb]{ .98,  .984,  .957}0.070 & \cellcolor[rgb]{ .98,  .984,  .957}0.076 & 0.138 & 0.043 & 0.098 & \cellcolor[rgb]{ .996,  .969,  .945}0.118 & \cellcolor[rgb]{ .996,  .969,  .945}0.127 & 0.144 & 0.137 & \cellcolor[rgb]{ .996,  .969,  .945}0.137 & \cellcolor[rgb]{ .996,  .969,  .945}0.136 \\
    Flowers-102 & 0.948 & 0.864 & 0.884 & \cellcolor[rgb]{ .945,  .973,  .98}0.919 & \cellcolor[rgb]{ .945,  .973,  .98}0.940 & 0.581 & 0.760 & \cellcolor[rgb]{ .945,  .973,  .98}0.754 & \cellcolor[rgb]{ .945,  .973,  .98}0.798 & 0.950 & 0.894 & 0.907 & \cellcolor[rgb]{ .98,  .984,  .957}0.914 & \cellcolor[rgb]{ .98,  .984,  .957}0.944 & 0.692 & 0.802 & \cellcolor[rgb]{ .98,  .984,  .957}0.803 & \cellcolor[rgb]{ .98,  .984,  .957}0.928 & 0.964 & 0.369 & 0.969 & \cellcolor[rgb]{ .996,  .969,  .945}0.956 & \cellcolor[rgb]{ .996,  .969,  .945}0.962 & 0.959 & 0.964 & \cellcolor[rgb]{ .996,  .969,  .945}0.965 & \cellcolor[rgb]{ .996,  .969,  .945}0.961 \\
    Food-101 & 0.894 & 0.775 & 0.842 & \cellcolor[rgb]{ .945,  .973,  .98}0.824 & \cellcolor[rgb]{ .945,  .973,  .98}0.889 & 0.442 & 0.628 & \cellcolor[rgb]{ .945,  .973,  .98}0.406 & \cellcolor[rgb]{ .945,  .973,  .98}0.685 & 0.882 & 0.805 & 0.822 & \cellcolor[rgb]{ .98,  .984,  .957}0.826 & \cellcolor[rgb]{ .98,  .984,  .957}0.877 & 0.223 & 0.722 & \cellcolor[rgb]{ .98,  .984,  .957}0.707 & \cellcolor[rgb]{ .98,  .984,  .957}0.876 & 0.912 & 0.102 & 0.918 & \cellcolor[rgb]{ .996,  .969,  .945}0.905 & \cellcolor[rgb]{ .996,  .969,  .945}0.916 & 0.905 & 0.916 & \cellcolor[rgb]{ .996,  .969,  .945}0.913 & \cellcolor[rgb]{ .996,  .969,  .945}0.915 \\
    GTSRB & 0.656 & 0.542 & 0.570 & \cellcolor[rgb]{ .945,  .973,  .98}0.594 & \cellcolor[rgb]{ .945,  .973,  .98}0.655 & 0.412 & 0.462 & \cellcolor[rgb]{ .945,  .973,  .98}0.577 & \cellcolor[rgb]{ .945,  .973,  .98}0.488 & 0.618 & 0.515 & 0.542 & \cellcolor[rgb]{ .98,  .984,  .957}0.589 & \cellcolor[rgb]{ .98,  .984,  .957}0.613 & 0.505 & 0.523 & \cellcolor[rgb]{ .98,  .984,  .957}0.543 & \cellcolor[rgb]{ .98,  .984,  .957}0.586 & 0.704 & 0.484 & 0.691 & \cellcolor[rgb]{ .996,  .969,  .945}0.551 & \cellcolor[rgb]{ .996,  .969,  .945}0.702 & 0.695 & 0.693 & \cellcolor[rgb]{ .996,  .969,  .945}0.706 & \cellcolor[rgb]{ .996,  .969,  .945}0.707 \\
    PCam & 0.062 & 0.099 & 0.052 & \cellcolor[rgb]{ .945,  .973,  .98}0.100 & \cellcolor[rgb]{ .945,  .973,  .98}0.060 & 0.076 & 0.091 & \cellcolor[rgb]{ .945,  .973,  .98}0.058 & \cellcolor[rgb]{ .945,  .973,  .98}0.094 & 0.049 & 0.096 & 0.056 & \cellcolor[rgb]{ .98,  .984,  .957}0.041 & \cellcolor[rgb]{ .98,  .984,  .957}0.048 & 0.058 & 0.052 & \cellcolor[rgb]{ .98,  .984,  .957}0.056 & \cellcolor[rgb]{ .98,  .984,  .957}0.048 & 0.018 & 0.002 & 0.016 & \cellcolor[rgb]{ .996,  .969,  .945}0.039 & \cellcolor[rgb]{ .996,  .969,  .945}0.018 & 0.020 & 0.018 & \cellcolor[rgb]{ .996,  .969,  .945}0.018 & \cellcolor[rgb]{ .996,  .969,  .945}0.018 \\
    Pets & 0.759 & 0.579 & 0.654 & \cellcolor[rgb]{ .945,  .973,  .98}0.660 & \cellcolor[rgb]{ .945,  .973,  .98}0.751 & 0.331 & 0.404 & \cellcolor[rgb]{ .945,  .973,  .98}0.572 & \cellcolor[rgb]{ .945,  .973,  .98}0.437 & 0.846 & 0.593 & 0.614 & \cellcolor[rgb]{ .98,  .984,  .957}0.652 & \cellcolor[rgb]{ .98,  .984,  .957}0.855 & 0.289 & 0.520 & \cellcolor[rgb]{ .98,  .984,  .957}0.542 & \cellcolor[rgb]{ .98,  .984,  .957}0.735 & 0.913 & 0.276 & 0.922 & \cellcolor[rgb]{ .996,  .969,  .945}0.891 & \cellcolor[rgb]{ .996,  .969,  .945}0.924 & 0.900 & 0.931 & \cellcolor[rgb]{ .996,  .969,  .945}0.915 & \cellcolor[rgb]{ .996,  .969,  .945}0.921 \\
    Rendered SST2 & 0.001 & 0.002 & 0.000 & \cellcolor[rgb]{ .945,  .973,  .98}0.006 & \cellcolor[rgb]{ .945,  .973,  .98}0.001 & 0.001 & 0.002 & \cellcolor[rgb]{ .945,  .973,  .98}0.003 & \cellcolor[rgb]{ .945,  .973,  .98}0.002 & 0.008 & 0.006 & 0.001 & \cellcolor[rgb]{ .98,  .984,  .957}0.003 & \cellcolor[rgb]{ .98,  .984,  .957}0.002 & 0.001 & 0.003 & \cellcolor[rgb]{ .98,  .984,  .957}0.004 & \cellcolor[rgb]{ .98,  .984,  .957}0.004 & 0.003 & 0.001 & 0.005 & \cellcolor[rgb]{ .996,  .969,  .945}0.005 & \cellcolor[rgb]{ .996,  .969,  .945}0.003 & 0.001 & 0.003 & \cellcolor[rgb]{ .996,  .969,  .945}0.003 & \cellcolor[rgb]{ .996,  .969,  .945}0.003 \\
    Resisc45 & 0.817 & 0.755 & 0.792 & \cellcolor[rgb]{ .945,  .973,  .98}0.814 & \cellcolor[rgb]{ .945,  .973,  .98}0.822 & 0.636 & 0.682 & \cellcolor[rgb]{ .945,  .973,  .98}0.618 & \cellcolor[rgb]{ .945,  .973,  .98}0.704 & 0.776 & 0.776 & 0.729 & \cellcolor[rgb]{ .98,  .984,  .957}0.742 & \cellcolor[rgb]{ .98,  .984,  .957}0.788 & 0.542 & 0.784 & \cellcolor[rgb]{ .98,  .984,  .957}0.793 & \cellcolor[rgb]{ .98,  .984,  .957}0.784 & 0.824 & 0.599 & 0.835 & \cellcolor[rgb]{ .996,  .969,  .945}0.819 & \cellcolor[rgb]{ .996,  .969,  .945}0.810 & 0.810 & 0.821 & \cellcolor[rgb]{ .996,  .969,  .945}0.826 & \cellcolor[rgb]{ .996,  .969,  .945}0.818 \\
    STL10 & 0.951 & 0.949 & 0.941 & \cellcolor[rgb]{ .945,  .973,  .98}0.925 & \cellcolor[rgb]{ .945,  .973,  .98}0.952 & 0.534 & 0.811 & \cellcolor[rgb]{ .945,  .973,  .98}0.863 & \cellcolor[rgb]{ .945,  .973,  .98}0.842 & 0.953 & 0.932 & 0.943 & \cellcolor[rgb]{ .98,  .984,  .957}0.948 & \cellcolor[rgb]{ .98,  .984,  .957}0.912 & 0.513 & 0.859 & \cellcolor[rgb]{ .98,  .984,  .957}0.846 & \cellcolor[rgb]{ .98,  .984,  .957}0.953 & 0.944 & 0.844 & 0.941 & \cellcolor[rgb]{ .996,  .969,  .945}0.980 & \cellcolor[rgb]{ .996,  .969,  .945}0.943 & 0.946 & 0.944 & \cellcolor[rgb]{ .996,  .969,  .945}0.906 & \cellcolor[rgb]{ .996,  .969,  .945}0.944 \\
    VOC2007 & 0.547 & 0.530 & 0.519 & \cellcolor[rgb]{ .945,  .973,  .98}0.530 & \cellcolor[rgb]{ .945,  .973,  .98}0.568 & 0.342 & 0.516 & \cellcolor[rgb]{ .945,  .973,  .98}0.410 & \cellcolor[rgb]{ .945,  .973,  .98}0.516 & 0.461 & 0.479 & 0.461 & \cellcolor[rgb]{ .98,  .984,  .957}0.416 & \cellcolor[rgb]{ .98,  .984,  .957}0.434 & 0.292 & 0.370 & \cellcolor[rgb]{ .98,  .984,  .957}0.424 & \cellcolor[rgb]{ .98,  .984,  .957}0.411 & 0.609 & 0.102 & 0.576 & \cellcolor[rgb]{ .996,  .969,  .945}0.563 & \cellcolor[rgb]{ .996,  .969,  .945}0.607 & 0.615 & 0.580 & \cellcolor[rgb]{ .996,  .969,  .945}0.602 & \cellcolor[rgb]{ .996,  .969,  .945}0.591 \\
    ImageNet-1K & 0.802 & 0.714 & 0.758 & \cellcolor[rgb]{ .945,  .973,  .98}0.750 & \cellcolor[rgb]{ .945,  .973,  .98}0.803 & 0.574 & 0.652 & \cellcolor[rgb]{ .945,  .973,  .98}0.655 & \cellcolor[rgb]{ .945,  .973,  .98}0.673 & 0.820 & 0.726 & 0.744 & \cellcolor[rgb]{ .98,  .984,  .957}0.764 & \cellcolor[rgb]{ .98,  .984,  .957}0.818 & 0.540 & 0.729 & \cellcolor[rgb]{ .98,  .984,  .957}0.721 & \cellcolor[rgb]{ .98,  .984,  .957}0.800 & 0.853 & 0.512 & 0.719 & \cellcolor[rgb]{ .996,  .969,  .945}0.827 & \cellcolor[rgb]{ .996,  .969,  .945}0.852 & 0.848 & 0.850 & \cellcolor[rgb]{ .996,  .969,  .945}0.852 & \cellcolor[rgb]{ .996,  .969,  .945}0.852 \\
    ImageNet-V2 & 0.797 & 0.753 & 0.769 & \cellcolor[rgb]{ .945,  .973,  .98}0.775 & \cellcolor[rgb]{ .945,  .973,  .98}0.797 & 0.695 & 0.722 & \cellcolor[rgb]{ .945,  .973,  .98}0.712 & \cellcolor[rgb]{ .945,  .973,  .98}0.731 & 0.803 & 0.760 & 0.770 & \cellcolor[rgb]{ .98,  .984,  .957}0.783 & \cellcolor[rgb]{ .98,  .984,  .957}0.804 & 0.678 & 0.751 & \cellcolor[rgb]{ .98,  .984,  .957}0.749 & \cellcolor[rgb]{ .98,  .984,  .957}0.789 & 0.824 & 0.628 & 0.745 & \cellcolor[rgb]{ .996,  .969,  .945}0.805 & \cellcolor[rgb]{ .996,  .969,  .945}0.824 & 0.821 & 0.823 & \cellcolor[rgb]{ .996,  .969,  .945}0.822 & \cellcolor[rgb]{ .996,  .969,  .945}0.822 \\
    ImageNet-Adv. & 0.645 & 0.472 & 0.622 & \cellcolor[rgb]{ .945,  .973,  .98}0.552 & \cellcolor[rgb]{ .945,  .973,  .98}0.654 & 0.409 & 0.441 & \cellcolor[rgb]{ .945,  .973,  .98}0.428 & \cellcolor[rgb]{ .945,  .973,  .98}0.449 & 0.647 & 0.497 & 0.561 & \cellcolor[rgb]{ .98,  .984,  .957}0.583 & \cellcolor[rgb]{ .98,  .984,  .957}0.645 & 0.388 & 0.498 & \cellcolor[rgb]{ .98,  .984,  .957}0.497 & \cellcolor[rgb]{ .98,  .984,  .957}0.621 & 0.641 & 0.367 & 0.513 & \cellcolor[rgb]{ .996,  .969,  .945}0.577 & \cellcolor[rgb]{ .996,  .969,  .945}0.634 & 0.620 & 0.633 & \cellcolor[rgb]{ .996,  .969,  .945}0.630 & \cellcolor[rgb]{ .996,  .969,  .945}0.636 \\
    ImageNet-Ren. & 0.606 & 0.401 & 0.340 & \cellcolor[rgb]{ .945,  .973,  .98}0.446 & \cellcolor[rgb]{ .945,  .973,  .98}0.609 & 0.244 & 0.338 & \cellcolor[rgb]{ .945,  .973,  .98}0.360 & \cellcolor[rgb]{ .945,  .973,  .98}0.356 & 0.568 & 0.412 & 0.399 & \cellcolor[rgb]{ .98,  .984,  .957}0.423 & \cellcolor[rgb]{ .98,  .984,  .957}0.557 & 0.235 & 0.339 & \cellcolor[rgb]{ .98,  .984,  .957}0.341 & \cellcolor[rgb]{ .98,  .984,  .957}0.514 & 0.602 & 0.180 & 0.318 & \cellcolor[rgb]{ .996,  .969,  .945}0.527 & \cellcolor[rgb]{ .996,  .969,  .945}0.615 & 0.580 & 0.587 & \cellcolor[rgb]{ .996,  .969,  .945}0.595 & \cellcolor[rgb]{ .996,  .969,  .945}0.605 \\
    ImageNet-OOD & 0.639 & 0.645 & 0.628 & \cellcolor[rgb]{ .945,  .973,  .98}0.642 & \cellcolor[rgb]{ .945,  .973,  .98}0.634 & 0.617 & 0.645 & \cellcolor[rgb]{ .945,  .973,  .98}0.662 & \cellcolor[rgb]{ .945,  .973,  .98}0.645 & 0.628 & 0.647 & 0.630 & \cellcolor[rgb]{ .98,  .984,  .957}0.630 & \cellcolor[rgb]{ .98,  .984,  .957}0.628 & 0.626 & 0.631 & \cellcolor[rgb]{ .98,  .984,  .957}0.635 & \cellcolor[rgb]{ .98,  .984,  .957}0.631 & 0.635 & 0.507 & 0.635 & \cellcolor[rgb]{ .996,  .969,  .945}0.645 & \cellcolor[rgb]{ .996,  .969,  .945}0.639 & 0.641 & 0.637 & \cellcolor[rgb]{ .996,  .969,  .945}0.642 & \cellcolor[rgb]{ .996,  .969,  .945}0.642 \\
    ObjectNet & 0.583 & 0.397 & 0.557 & \cellcolor[rgb]{ .945,  .973,  .98}0.511 & \cellcolor[rgb]{ .945,  .973,  .98}0.594 & 0.183 & 0.310 & \cellcolor[rgb]{ .945,  .973,  .98}0.246 & \cellcolor[rgb]{ .945,  .973,  .98}0.335 & 0.636 & 0.538 & 0.593 & \cellcolor[rgb]{ .98,  .984,  .957}0.628 & \cellcolor[rgb]{ .98,  .984,  .957}0.644 & 0.137 & 0.523 & \cellcolor[rgb]{ .98,  .984,  .957}0.508 & \cellcolor[rgb]{ .98,  .984,  .957}0.633 & 0.588 & 0.127 & 0.464 & \cellcolor[rgb]{ .996,  .969,  .945}0.490 & \cellcolor[rgb]{ .996,  .969,  .945}0.599 & 0.570 & 0.596 & \cellcolor[rgb]{ .996,  .969,  .945}0.596 & \cellcolor[rgb]{ .996,  .969,  .945}0.595 \\
    \midrule
    Avg. NMI & 0.640 & 0.559 & 0.580 & \cellcolor[rgb]{ .945,  .973,  .98}0.601 & \cellcolor[rgb]{ .945,  .973,  .98}\textbf{0.642} & 0.418 & 0.499 & \cellcolor[rgb]{ .945,  .973,  .98}0.514 & \cellcolor[rgb]{ .945,  .973,  .98}\textbf{0.515} & 0.646 & 0.576 & 0.588 & \cellcolor[rgb]{ .98,  .984,  .957}0.598 & \cellcolor[rgb]{ .98,  .984,  .957}\textbf{0.644} & 0.412 & 0.557 & \cellcolor[rgb]{ .98,  .984,  .957}0.560 & \cellcolor[rgb]{ .98,  .984,  .957}\textbf{0.629} & 0.678 & 0.364 & 0.649 & \cellcolor[rgb]{ .996,  .969,  .945}0.650 & \cellcolor[rgb]{ .996,  .969,  .945}\textbf{0.679} & 0.675 & 0.678 & \cellcolor[rgb]{ .996,  .969,  .945}0.678 & \cellcolor[rgb]{ .996,  .969,  .945}\textbf{0.679} \\
    \bottomrule
    \end{tabular}%
   }
\label{tab: discriminative ability results NMI}

\end{table}

\newpage

\subsubsection{Zero-shot Classification}
\label{subsubsec: supp_Zero-shot Classification}
The main paper summarizes average zero-shot classification performance in \cref{fig: classification results}. The complete per-dataset results are reported in \cref{tab: classification acc1}, \cref{tab: classification acc5}, and \cref{tab: classification mrecall}. \cref{tab: classification acc1} reports ACC@1, which measures whether the top-ranked class-name prompt matches the ground-truth label. \cref{tab: classification acc5} reports ACC@5, which measures whether the correct class appears among the top $5$ predictions. \cref{tab: classification mrecall} reports mean recall, which gives a class-balanced view of recognition performance. These metrics evaluate different aspects of prompt-based recognition.

The results show that AUCGR restores the discriminative information weakened by plain generative reconstruction supervision. pAUCGR further improves the results by reducing ambiguous negative comparisons.

\begin{table}[htbp]
  \centering
  \renewcommand\arraystretch{1.7}
  \setlength{\tabcolsep}{3pt}
  \caption{\textbf{Performance on 26 standard \underline{zero-shot classification} benchmarks in terms of \underline{ACC@1}.} This table presents the detailed results corresponding to \cref{fig: classification results}.}
  \resizebox{\linewidth}{!}{
    \begin{tabular}{c|ccc|ccc|ccc|ccc|ccc|ccc}
    \toprule
    \multirow[c]{2}{*}{\textbf{Dataset}} & \multicolumn{6}{c|}{\textbf{OpenAI CLIP ViT-L@224 (O-1)}} & \multicolumn{6}{c|}{\textbf{MetaCLIP ViT-L@224 (M-1)}} & \multicolumn{6}{c}{\textbf{SigLIP ViT-SO@224 (S-1)}} \\
    & \rotatebox[origin=c]{90}{\textbf{DIVA}} & \rotatebox[origin=c]{90}{\textbf{+AUCGR}} & \rotatebox[origin=c]{90}{\textbf{+pAUCGR}} & \rotatebox[origin=c]{90}{\textbf{GenHancer}} & \rotatebox[origin=c]{90}{\textbf{+AUCGR}} & \rotatebox[origin=c]{90}{\textbf{+pAUCGR}} & \rotatebox[origin=c]{90}{\textbf{DIVA}} & \rotatebox[origin=c]{90}{\textbf{+AUCGR}} & \rotatebox[origin=c]{90}{\textbf{+pAUCGR}} & \rotatebox[origin=c]{90}{\textbf{GenHancer}} & \rotatebox[origin=c]{90}{\textbf{+AUCGR}} & \rotatebox[origin=c]{90}{\textbf{+pAUCGR}} & \rotatebox[origin=c]{90}{\textbf{DIVA}} & \rotatebox[origin=c]{90}{\textbf{+AUCGR}} & \rotatebox[origin=c]{90}{\textbf{+pAUCGR}} & \rotatebox[origin=c]{90}{\textbf{GenHancer}} & \rotatebox[origin=c]{90}{\textbf{+AUCGR}} & \rotatebox[origin=c]{90}{\textbf{+pAUCGR}} \\
    \midrule
    CIFAR-10 & 86.93 & \cellcolor[rgb]{ .945,  .973,  .98}92.22 & \cellcolor[rgb]{ .945,  .973,  .98}95.60 & 76.36 & \cellcolor[rgb]{ .945,  .973,  .98}75.76 & \cellcolor[rgb]{ .945,  .973,  .98}79.45 & 96.66 & \cellcolor[rgb]{ .98,  .984,  .957}97.45 & \cellcolor[rgb]{ .98,  .984,  .957}97.56 & 58.19 & \cellcolor[rgb]{ .98,  .984,  .957}96.53 & \cellcolor[rgb]{ .98,  .984,  .957}97.33 & 95.13 & \cellcolor[rgb]{ .996,  .969,  .945}92.30 & \cellcolor[rgb]{ .996,  .969,  .945}96.68 & 96.71 & \cellcolor[rgb]{ .996,  .969,  .945}96.69 & \cellcolor[rgb]{ .996,  .969,  .945}96.70 \\
    CIFAR-100 & 58.44 & \cellcolor[rgb]{ .945,  .973,  .98}70.89 & \cellcolor[rgb]{ .945,  .973,  .98}75.69 & 36.34 & \cellcolor[rgb]{ .945,  .973,  .98}43.81 & \cellcolor[rgb]{ .945,  .973,  .98}51.89 & 79.83 & \cellcolor[rgb]{ .98,  .984,  .957}81.11 & \cellcolor[rgb]{ .98,  .984,  .957}84.12 & 28.63 & \cellcolor[rgb]{ .98,  .984,  .957}81.32 & \cellcolor[rgb]{ .98,  .984,  .957}84.18 & 76.86 & \cellcolor[rgb]{ .996,  .969,  .945}78.41 & \cellcolor[rgb]{ .996,  .969,  .945}84.11 & 83.70 & \cellcolor[rgb]{ .996,  .969,  .945}84.21 & \cellcolor[rgb]{ .996,  .969,  .945}84.23 \\
    MNIST & 53.63 & \cellcolor[rgb]{ .945,  .973,  .98}67.28 & \cellcolor[rgb]{ .945,  .973,  .98}76.65 & 49.02 & \cellcolor[rgb]{ .945,  .973,  .98}66.28 & \cellcolor[rgb]{ .945,  .973,  .98}48.99 & 62.90 & \cellcolor[rgb]{ .98,  .984,  .957}62.21 & \cellcolor[rgb]{ .98,  .984,  .957}63.13 & 29.85 & \cellcolor[rgb]{ .98,  .984,  .957}63.99 & \cellcolor[rgb]{ .98,  .984,  .957}64.71 & 74.84 & \cellcolor[rgb]{ .996,  .969,  .945}81.77 & \cellcolor[rgb]{ .996,  .969,  .945}88.12 & 88.33 & \cellcolor[rgb]{ .996,  .969,  .945}89.11 & \cellcolor[rgb]{ .996,  .969,  .945}89.13 \\
    Caltech-101 & 79.97 & \cellcolor[rgb]{ .945,  .973,  .98}81.35 & \cellcolor[rgb]{ .945,  .973,  .98}83.35 & 69.50 & \cellcolor[rgb]{ .945,  .973,  .98}80.38 & \cellcolor[rgb]{ .945,  .973,  .98}78.54 & 83.81 & \cellcolor[rgb]{ .98,  .984,  .957}89.47 & \cellcolor[rgb]{ .98,  .984,  .957}87.07 & 34.91 & \cellcolor[rgb]{ .98,  .984,  .957}84.77 & \cellcolor[rgb]{ .98,  .984,  .957}86.52 & 71.80 & \cellcolor[rgb]{ .996,  .969,  .945}85.83 & \cellcolor[rgb]{ .996,  .969,  .945}85.98 & 86.06 & \cellcolor[rgb]{ .996,  .969,  .945}86.01 & \cellcolor[rgb]{ .996,  .969,  .945}86.03 \\
    SUN397 & 55.47 & \cellcolor[rgb]{ .945,  .973,  .98}63.43 & \cellcolor[rgb]{ .945,  .973,  .98}67.80 & 30.30 & \cellcolor[rgb]{ .945,  .973,  .98}52.73 & \cellcolor[rgb]{ .945,  .973,  .98}48.99 & 63.63 & \cellcolor[rgb]{ .98,  .984,  .957}70.62 & \cellcolor[rgb]{ .98,  .984,  .957}73.36 & 12.27 & \cellcolor[rgb]{ .98,  .984,  .957}71.79 & \cellcolor[rgb]{ .98,  .984,  .957}74.07 & 26.38 & \cellcolor[rgb]{ .996,  .969,  .945}72.55 & \cellcolor[rgb]{ .996,  .969,  .945}74.43 & 74.42 & \cellcolor[rgb]{ .996,  .969,  .945}74.45 & \cellcolor[rgb]{ .996,  .969,  .945}74.42 \\
    FGVC Aircraft & 19.17 & \cellcolor[rgb]{ .945,  .973,  .98}24.00 & \cellcolor[rgb]{ .945,  .973,  .98}31.98 & 13.02 & \cellcolor[rgb]{ .945,  .973,  .98}12.30 & \cellcolor[rgb]{ .945,  .973,  .98}15.15 & 29.01 & \cellcolor[rgb]{ .98,  .984,  .957}30.99 & \cellcolor[rgb]{ .98,  .984,  .957}43.47 & 2.37 & \cellcolor[rgb]{ .98,  .984,  .957}33.78 & \cellcolor[rgb]{ .98,  .984,  .957}44.64 & 7.92 & \cellcolor[rgb]{ .996,  .969,  .945}55.75 & \cellcolor[rgb]{ .996,  .969,  .945}60.46 & 60.19 & \cellcolor[rgb]{ .996,  .969,  .945}60.46 & \cellcolor[rgb]{ .996,  .969,  .945}60.49 \\
    Country-211 & 12.42 & \cellcolor[rgb]{ .945,  .973,  .98}23.78 & \cellcolor[rgb]{ .945,  .973,  .98}31.75 & 10.38 & \cellcolor[rgb]{ .945,  .973,  .98}16.30 & \cellcolor[rgb]{ .945,  .973,  .98}7.83 & 16.67 & \cellcolor[rgb]{ .98,  .984,  .957}26.57 & \cellcolor[rgb]{ .98,  .984,  .957}33.51 & 1.78 & \cellcolor[rgb]{ .98,  .984,  .957}27.73 & \cellcolor[rgb]{ .98,  .984,  .957}33.81 & 0.50 & \cellcolor[rgb]{ .996,  .969,  .945}28.61 & \cellcolor[rgb]{ .996,  .969,  .945}32.13 & 32.27 & \cellcolor[rgb]{ .996,  .969,  .945}32.48 & \cellcolor[rgb]{ .996,  .969,  .945}32.46 \\
    Stanford Cars & 46.13 & \cellcolor[rgb]{ .945,  .973,  .98}68.20 & \cellcolor[rgb]{ .945,  .973,  .98}77.78 & 38.74 & \cellcolor[rgb]{ .945,  .973,  .98}37.84 & \cellcolor[rgb]{ .945,  .973,  .98}32.99 & 65.87 & \cellcolor[rgb]{ .98,  .984,  .957}81.77 & \cellcolor[rgb]{ .98,  .984,  .957}88.60 & 1.59 & \cellcolor[rgb]{ .98,  .984,  .957}82.81 & \cellcolor[rgb]{ .98,  .984,  .957}88.70 & 15.64 & \cellcolor[rgb]{ .996,  .969,  .945}90.09 & \cellcolor[rgb]{ .996,  .969,  .945}90.31 & 90.25 & \cellcolor[rgb]{ .996,  .969,  .945}90.30 & \cellcolor[rgb]{ .996,  .969,  .945}90.09 \\
    DTD & 46.06 & \cellcolor[rgb]{ .945,  .973,  .98}49.57 & \cellcolor[rgb]{ .945,  .973,  .98}55.32 & 34.10 & \cellcolor[rgb]{ .945,  .973,  .98}39.36 & \cellcolor[rgb]{ .945,  .973,  .98}42.71 & 58.19 & \cellcolor[rgb]{ .98,  .984,  .957}63.94 & \cellcolor[rgb]{ .98,  .984,  .957}68.62 & 23.88 & \cellcolor[rgb]{ .98,  .984,  .957}65.74 & \cellcolor[rgb]{ .98,  .984,  .957}67.93 & 43.30 & \cellcolor[rgb]{ .996,  .969,  .945}69.84 & \cellcolor[rgb]{ .996,  .969,  .945}71.33 & 71.12 & \cellcolor[rgb]{ .996,  .969,  .945}71.22 & \cellcolor[rgb]{ .996,  .969,  .945}71.17 \\
    Eurosat & 43.65 & \cellcolor[rgb]{ .945,  .973,  .98}52.19 & \cellcolor[rgb]{ .945,  .973,  .98}62.30 & 22.07 & \cellcolor[rgb]{ .945,  .973,  .98}23.89 & \cellcolor[rgb]{ .945,  .973,  .98}41.20 & 53.96 & \cellcolor[rgb]{ .98,  .984,  .957}67.48 & \cellcolor[rgb]{ .98,  .984,  .957}61.15 & 21.09 & \cellcolor[rgb]{ .98,  .984,  .957}56.15 & \cellcolor[rgb]{ .98,  .984,  .957}58.67 & 42.85 & \cellcolor[rgb]{ .996,  .969,  .945}46.78 & \cellcolor[rgb]{ .996,  .969,  .945}65.50 & 63.28 & \cellcolor[rgb]{ .996,  .969,  .945}63.80 & \cellcolor[rgb]{ .996,  .969,  .945}63.83 \\
    FER2013 & 26.37 & \cellcolor[rgb]{ .945,  .973,  .98}38.45 & \cellcolor[rgb]{ .945,  .973,  .98}49.51 & 50.88 & \cellcolor[rgb]{ .945,  .973,  .98}43.38 & \cellcolor[rgb]{ .945,  .973,  .98}23.81 & 53.58 & \cellcolor[rgb]{ .98,  .984,  .957}55.57 & \cellcolor[rgb]{ .98,  .984,  .957}51.64 & 24.12 & \cellcolor[rgb]{ .98,  .984,  .957}52.03 & \cellcolor[rgb]{ .98,  .984,  .957}52.70 & 49.29 & \cellcolor[rgb]{ .996,  .969,  .945}48.29 & \cellcolor[rgb]{ .996,  .969,  .945}50.10 & 49.86 & \cellcolor[rgb]{ .996,  .969,  .945}49.58 & \cellcolor[rgb]{ .996,  .969,  .945}49.55 \\
    Flowers-102 & 48.41 & \cellcolor[rgb]{ .945,  .973,  .98}68.06 & \cellcolor[rgb]{ .945,  .973,  .98}79.20 & 19.27 & \cellcolor[rgb]{ .945,  .973,  .98}36.30 & \cellcolor[rgb]{ .945,  .973,  .98}36.04 & 61.51 & \cellcolor[rgb]{ .98,  .984,  .957}79.00 & \cellcolor[rgb]{ .98,  .984,  .957}82.31 & 9.90 & \cellcolor[rgb]{ .98,  .984,  .957}77.72 & \cellcolor[rgb]{ .98,  .984,  .957}82.81 & 12.20 & \cellcolor[rgb]{ .996,  .969,  .945}91.27 & \cellcolor[rgb]{ .996,  .969,  .945}92.65 & 92.44 & \cellcolor[rgb]{ .996,  .969,  .945}92.36 & \cellcolor[rgb]{ .996,  .969,  .945}92.37 \\
    Food-101 & 75.51 & \cellcolor[rgb]{ .945,  .973,  .98}89.31 & \cellcolor[rgb]{ .945,  .973,  .98}93.01 & 59.15 & \cellcolor[rgb]{ .945,  .973,  .98}36.32 & \cellcolor[rgb]{ .945,  .973,  .98}65.31 & 81.73 & \cellcolor[rgb]{ .98,  .984,  .957}89.66 & \cellcolor[rgb]{ .98,  .984,  .957}93.40 & 4.73 & \cellcolor[rgb]{ .98,  .984,  .957}89.76 & \cellcolor[rgb]{ .98,  .984,  .957}93.42 & 19.00 & \cellcolor[rgb]{ .996,  .969,  .945}94.66 & \cellcolor[rgb]{ .996,  .969,  .945}95.56 & 95.46 & \cellcolor[rgb]{ .996,  .969,  .945}95.55 & \cellcolor[rgb]{ .996,  .969,  .945}95.54 \\
    GTSRB & 36.37 & \cellcolor[rgb]{ .945,  .973,  .98}49.26 & \cellcolor[rgb]{ .945,  .973,  .98}50.74 & 27.47 & \cellcolor[rgb]{ .945,  .973,  .98}36.05 & \cellcolor[rgb]{ .945,  .973,  .98}32.99 & 46.63 & \cellcolor[rgb]{ .98,  .984,  .957}50.33 & \cellcolor[rgb]{ .98,  .984,  .957}56.30 & 18.31 & \cellcolor[rgb]{ .98,  .984,  .957}58.64 & \cellcolor[rgb]{ .98,  .984,  .957}56.15 & 54.50 & \cellcolor[rgb]{ .996,  .969,  .945}52.25 & \cellcolor[rgb]{ .996,  .969,  .945}64.27 & 63.67 & \cellcolor[rgb]{ .996,  .969,  .945}64.00 & \cellcolor[rgb]{ .996,  .969,  .945}64.01 \\
    PCam & 53.56 & \cellcolor[rgb]{ .945,  .973,  .98}51.85 & \cellcolor[rgb]{ .945,  .973,  .98}52.44 & 49.98 & \cellcolor[rgb]{ .945,  .973,  .98}50.02 & \cellcolor[rgb]{ .945,  .973,  .98}52.29 & 49.04 & \cellcolor[rgb]{ .98,  .984,  .957}52.23 & \cellcolor[rgb]{ .98,  .984,  .957}63.32 & 51.44 & \cellcolor[rgb]{ .98,  .984,  .957}57.68 & \cellcolor[rgb]{ .98,  .984,  .957}61.30 & 54.01 & \cellcolor[rgb]{ .996,  .969,  .945}50.11 & \cellcolor[rgb]{ .996,  .969,  .945}50.21 & 50.28 & \cellcolor[rgb]{ .996,  .969,  .945}50.65 & \cellcolor[rgb]{ .996,  .969,  .945}50.62 \\
    Pets & 83.18 & \cellcolor[rgb]{ .945,  .973,  .98}89.86 & \cellcolor[rgb]{ .945,  .973,  .98}93.13 & 75.20 & \cellcolor[rgb]{ .945,  .973,  .98}79.20 & \cellcolor[rgb]{ .945,  .973,  .98}76.56 & 82.80 & \cellcolor[rgb]{ .98,  .984,  .957}87.54 & \cellcolor[rgb]{ .98,  .984,  .957}94.11 & 18.89 & \cellcolor[rgb]{ .98,  .984,  .957}92.34 & \cellcolor[rgb]{ .98,  .984,  .957}94.49 & 31.26 & \cellcolor[rgb]{ .996,  .969,  .945}93.79 & \cellcolor[rgb]{ .996,  .969,  .945}95.20 & 95.04 & \cellcolor[rgb]{ .996,  .969,  .945}95.34 & \cellcolor[rgb]{ .996,  .969,  .945}95.34 \\
    Rendered SST2 & 69.63 & \cellcolor[rgb]{ .945,  .973,  .98}56.56 & \cellcolor[rgb]{ .945,  .973,  .98}69.08 & 51.51 & \cellcolor[rgb]{ .945,  .973,  .98}51.89 & \cellcolor[rgb]{ .945,  .973,  .98}64.80 & 59.86 & \cellcolor[rgb]{ .98,  .984,  .957}64.47 & \cellcolor[rgb]{ .98,  .984,  .957}65.95 & 49.92 & \cellcolor[rgb]{ .98,  .984,  .957}68.42 & \cellcolor[rgb]{ .98,  .984,  .957}68.09 & 64.36 & \cellcolor[rgb]{ .996,  .969,  .945}69.69 & \cellcolor[rgb]{ .996,  .969,  .945}69.19 & 70.13 & \cellcolor[rgb]{ .996,  .969,  .945}70.02 & \cellcolor[rgb]{ .996,  .969,  .945}70.02 \\
    Resisc45 & 44.41 & \cellcolor[rgb]{ .945,  .973,  .98}54.81 & \cellcolor[rgb]{ .945,  .973,  .98}63.57 & 25.35 & \cellcolor[rgb]{ .945,  .973,  .98}27.41 & \cellcolor[rgb]{ .945,  .973,  .98}37.52 & 61.98 & \cellcolor[rgb]{ .98,  .984,  .957}69.05 & \cellcolor[rgb]{ .98,  .984,  .957}74.06 & 9.41 & \cellcolor[rgb]{ .98,  .984,  .957}66.70 & \cellcolor[rgb]{ .98,  .984,  .957}73.59 & 53.49 & \cellcolor[rgb]{ .996,  .969,  .945}69.17 & \cellcolor[rgb]{ .996,  .969,  .945}69.65 & 69.78 & \cellcolor[rgb]{ .996,  .969,  .945}69.76 & \cellcolor[rgb]{ .996,  .969,  .945}69.75 \\
    STL10 & 98.35 & \cellcolor[rgb]{ .945,  .973,  .98}99.03 & \cellcolor[rgb]{ .945,  .973,  .98}99.43 & 93.04 & \cellcolor[rgb]{ .945,  .973,  .98}89.56 & \cellcolor[rgb]{ .945,  .973,  .98}96.33 & 98.78 & \cellcolor[rgb]{ .98,  .984,  .957}99.11 & \cellcolor[rgb]{ .98,  .984,  .957}99.29 & 55.61 & \cellcolor[rgb]{ .98,  .984,  .957}99.16 & \cellcolor[rgb]{ .98,  .984,  .957}99.36 & 94.49 & \cellcolor[rgb]{ .996,  .969,  .945}98.64 & \cellcolor[rgb]{ .996,  .969,  .945}99.33 & 99.43 & \cellcolor[rgb]{ .996,  .969,  .945}99.39 & \cellcolor[rgb]{ .996,  .969,  .945}99.39 \\
    VOC2007 & 70.73 & \cellcolor[rgb]{ .945,  .973,  .98}80.41 & \cellcolor[rgb]{ .945,  .973,  .98}78.62 & 67.33 & \cellcolor[rgb]{ .945,  .973,  .98}66.30 & \cellcolor[rgb]{ .945,  .973,  .98}68.38 & 81.32 & \cellcolor[rgb]{ .98,  .984,  .957}81.46 & \cellcolor[rgb]{ .98,  .984,  .957}80.30 & 42.98 & \cellcolor[rgb]{ .98,  .984,  .957}80.50 & \cellcolor[rgb]{ .98,  .984,  .957}80.62 & 64.74 & \cellcolor[rgb]{ .996,  .969,  .945}61.84 & \cellcolor[rgb]{ .996,  .969,  .945}69.89 & 70.04 & \cellcolor[rgb]{ .996,  .969,  .945}70.89 & \cellcolor[rgb]{ .996,  .969,  .945}70.98 \\
    ImageNet-1K & 57.55 & \cellcolor[rgb]{ .945,  .973,  .98}69.45 & \cellcolor[rgb]{ .945,  .973,  .98}75.55 & 34.04 & \cellcolor[rgb]{ .945,  .973,  .98}46.74 & \cellcolor[rgb]{ .945,  .973,  .98}48.58 & 63.52 & \cellcolor[rgb]{ .98,  .984,  .957}73.51 & \cellcolor[rgb]{ .98,  .984,  .957}78.85 & 6.19 & \cellcolor[rgb]{ .98,  .984,  .957}73.63 & \cellcolor[rgb]{ .98,  .984,  .957}78.96 & 18.52 & \cellcolor[rgb]{ .996,  .969,  .945}80.51 & \cellcolor[rgb]{ .996,  .969,  .945}82.01 & 81.94 & \cellcolor[rgb]{ .996,  .969,  .945}82.04 & \cellcolor[rgb]{ .996,  .969,  .945}82.04 \\
    ImageNet-V2 & 50.25 & \cellcolor[rgb]{ .945,  .973,  .98}63.27 & \cellcolor[rgb]{ .945,  .973,  .98}69.77 & 29.11 & \cellcolor[rgb]{ .945,  .973,  .98}40.25 & \cellcolor[rgb]{ .945,  .973,  .98}41.98 & 56.07 & \cellcolor[rgb]{ .98,  .984,  .957}66.51 & \cellcolor[rgb]{ .98,  .984,  .957}72.32 & 5.29 & \cellcolor[rgb]{ .98,  .984,  .957}66.34 & \cellcolor[rgb]{ .98,  .984,  .957}72.28 & 15.53 & \cellcolor[rgb]{ .996,  .969,  .945}74.30 & \cellcolor[rgb]{ .996,  .969,  .945}76.32 & 75.94 & \cellcolor[rgb]{ .996,  .969,  .945}76.14 & \cellcolor[rgb]{ .996,  .969,  .945}76.12 \\
    ImageNet-Adv. & 35.63 & \cellcolor[rgb]{ .945,  .973,  .98}61.84 & \cellcolor[rgb]{ .945,  .973,  .98}70.63 & 20.09 & \cellcolor[rgb]{ .945,  .973,  .98}17.40 & \cellcolor[rgb]{ .945,  .973,  .98}24.17 & 44.73 & \cellcolor[rgb]{ .98,  .984,  .957}63.36 & \cellcolor[rgb]{ .98,  .984,  .957}71.60 & 1.89 & \cellcolor[rgb]{ .98,  .984,  .957}62.43 & \cellcolor[rgb]{ .98,  .984,  .957}71.79 & 17.21 & \cellcolor[rgb]{ .996,  .969,  .945}66.08 & \cellcolor[rgb]{ .996,  .969,  .945}71.73 & 70.93 & \cellcolor[rgb]{ .996,  .969,  .945}72.03 & \cellcolor[rgb]{ .996,  .969,  .945}71.96 \\
    ImageNet-Ren. & 71.26 & \cellcolor[rgb]{ .945,  .973,  .98}81.79 & \cellcolor[rgb]{ .945,  .973,  .98}87.79 & 42.16 & \cellcolor[rgb]{ .945,  .973,  .98}58.18 & \cellcolor[rgb]{ .945,  .973,  .98}61.84 & 79.86 & \cellcolor[rgb]{ .98,  .984,  .957}87.76 & \cellcolor[rgb]{ .98,  .984,  .957}91.88 & 10.47 & \cellcolor[rgb]{ .98,  .984,  .957}88.15 & \cellcolor[rgb]{ .98,  .984,  .957}91.95 & 35.06 & \cellcolor[rgb]{ .996,  .969,  .945}93.58 & \cellcolor[rgb]{ .996,  .969,  .945}95.00 & 94.91 & \cellcolor[rgb]{ .996,  .969,  .945}95.06 & \cellcolor[rgb]{ .996,  .969,  .945}95.07 \\
    ImageNet-OOD & 27.75 & \cellcolor[rgb]{ .945,  .973,  .98}31.25 & \cellcolor[rgb]{ .945,  .973,  .98}32.25 & 21.75 & \cellcolor[rgb]{ .945,  .973,  .98}34.50 & \cellcolor[rgb]{ .945,  .973,  .98}26.05 & 30.10 & \cellcolor[rgb]{ .98,  .984,  .957}29.00 & \cellcolor[rgb]{ .98,  .984,  .957}29.70 & 11.20 & \cellcolor[rgb]{ .98,  .984,  .957}34.60 & \cellcolor[rgb]{ .98,  .984,  .957}30.70 & 11.15 & \cellcolor[rgb]{ .996,  .969,  .945}31.00 & \cellcolor[rgb]{ .996,  .969,  .945}29.90 & 30.20 & \cellcolor[rgb]{ .996,  .969,  .945}29.65 & \cellcolor[rgb]{ .996,  .969,  .945}29.65 \\
    ObjectNet & 46.51 & \cellcolor[rgb]{ .945,  .973,  .98}62.97 & \cellcolor[rgb]{ .945,  .973,  .98}69.13 & 21.15 & \cellcolor[rgb]{ .945,  .973,  .98}27.21 & \cellcolor[rgb]{ .945,  .973,  .98}39.25 & 59.11 & \cellcolor[rgb]{ .98,  .984,  .957}69.81 & \cellcolor[rgb]{ .98,  .984,  .957}74.49 & 2.97 & \cellcolor[rgb]{ .98,  .984,  .957}68.16 & \cellcolor[rgb]{ .98,  .984,  .957}74.22 & 15.30 & \cellcolor[rgb]{ .996,  .969,  .945}65.17 & \cellcolor[rgb]{ .996,  .969,  .945}71.21 & 69.77 & \cellcolor[rgb]{ .996,  .969,  .945}70.53 & \cellcolor[rgb]{ .996,  .969,  .945}70.59 \\
    \midrule
    Avg. ACC@1 & 53.74 & \cellcolor[rgb]{ .945,  .973,  .98}63.12 & \cellcolor[rgb]{ .945,  .973,  .98}\textbf{68.93} & 41.05 & \cellcolor[rgb]{ .945,  .973,  .98}45.75 & \cellcolor[rgb]{ .945,  .973,  .98}\textbf{47.83} & 62.35 & \cellcolor[rgb]{ .98,  .984,  .957}68.85 & \cellcolor[rgb]{ .98,  .984,  .957}\textbf{72.31} & 20.69 & \cellcolor[rgb]{ .98,  .984,  .957}69.26 & \cellcolor[rgb]{ .98,  .984,  .957}\textbf{72.42} & 40.97 & \cellcolor[rgb]{ .996,  .969,  .945}70.86 & \cellcolor[rgb]{ .996,  .969,  .945}\textbf{74.28} & 74.08 & \cellcolor[rgb]{ .996,  .969,  .945}\textbf{74.30} & \cellcolor[rgb]{ .996,  .969,  .945}74.29 \\
    \bottomrule
    \end{tabular}%
   }
\label{tab: classification acc1}
\end{table}

\newpage

\begin{table}[htbp]
  \centering
  \renewcommand\arraystretch{1.7}
  \setlength{\tabcolsep}{3pt}
  \caption{\textbf{Performance on 26 standard \underline{zero-shot classification} benchmarks in terms of \underline{ACC@5}.} This table presents the detailed results corresponding to \cref{fig: classification results}. Since PCam and Rendered SST2 contain only two categories, ACC@5 cannot be computed on these two datasets.}
  \resizebox{\linewidth}{!}{
    \begin{tabular}{c|ccc|ccc|ccc|ccc|ccc|ccc}
    \toprule
    \multirow[c]{2}{*}{\textbf{Dataset}} & \multicolumn{6}{c|}{\textbf{OpenAI CLIP ViT-L@224 (O-1)}} & \multicolumn{6}{c|}{\textbf{MetaCLIP ViT-L@224 (M-1)}} & \multicolumn{6}{c}{\textbf{SigLIP ViT-SO@224 (S-1)}} \\
    & \rotatebox[origin=c]{90}{\textbf{DIVA}} & \rotatebox[origin=c]{90}{\textbf{+AUCGR}} & \rotatebox[origin=c]{90}{\textbf{+pAUCGR}} & \rotatebox[origin=c]{90}{\textbf{GenHancer}} & \rotatebox[origin=c]{90}{\textbf{+AUCGR}} & \rotatebox[origin=c]{90}{\textbf{+pAUCGR}} & \rotatebox[origin=c]{90}{\textbf{DIVA}} & \rotatebox[origin=c]{90}{\textbf{+AUCGR}} & \rotatebox[origin=c]{90}{\textbf{+pAUCGR}} & \rotatebox[origin=c]{90}{\textbf{GenHancer}} & \rotatebox[origin=c]{90}{\textbf{+AUCGR}} & \rotatebox[origin=c]{90}{\textbf{+pAUCGR}} & \rotatebox[origin=c]{90}{\textbf{DIVA}} & \rotatebox[origin=c]{90}{\textbf{+AUCGR}} & \rotatebox[origin=c]{90}{\textbf{+pAUCGR}} & \rotatebox[origin=c]{90}{\textbf{GenHancer}} & \rotatebox[origin=c]{90}{\textbf{+AUCGR}} & \rotatebox[origin=c]{90}{\textbf{+pAUCGR}} \\
    \midrule
    CIFAR-10 & 99.13 & \cellcolor[rgb]{ .945,  .973,  .98}99.41 & \cellcolor[rgb]{ .945,  .973,  .98}99.64 & 96.78 & \cellcolor[rgb]{ .945,  .973,  .98}97.61 & \cellcolor[rgb]{ .945,  .973,  .98}98.48 & 99.94 & \cellcolor[rgb]{ .98,  .984,  .957}99.95 & \cellcolor[rgb]{ .98,  .984,  .957}99.93 & 93.39 & \cellcolor[rgb]{ .98,  .984,  .957}99.90 & \cellcolor[rgb]{ .98,  .984,  .957}99.95 & 99.43 & \cellcolor[rgb]{ .996,  .969,  .945}99.37 & \cellcolor[rgb]{ .996,  .969,  .945}99.81 & 99.83 & \cellcolor[rgb]{ .996,  .969,  .945}99.86 & \cellcolor[rgb]{ .996,  .969,  .945}99.86 \\
    CIFAR-100 & 81.95 & \cellcolor[rgb]{ .945,  .973,  .98}90.58 & \cellcolor[rgb]{ .945,  .973,  .98}92.74 & 64.19 & \cellcolor[rgb]{ .945,  .973,  .98}70.01 & \cellcolor[rgb]{ .945,  .973,  .98}76.23 & 96.10 & \cellcolor[rgb]{ .98,  .984,  .957}95.92 & \cellcolor[rgb]{ .98,  .984,  .957}96.93 & 58.32 & \cellcolor[rgb]{ .98,  .984,  .957}95.85 & \cellcolor[rgb]{ .98,  .984,  .957}96.66 & 91.60 & \cellcolor[rgb]{ .996,  .969,  .945}93.11 & \cellcolor[rgb]{ .996,  .969,  .945}96.07 & 95.83 & \cellcolor[rgb]{ .996,  .969,  .945}96.03 & \cellcolor[rgb]{ .996,  .969,  .945}96.03 \\
    MNIST & 87.91 & \cellcolor[rgb]{ .945,  .973,  .98}89.14 & \cellcolor[rgb]{ .945,  .973,  .98}93.52 & 84.49 & \cellcolor[rgb]{ .945,  .973,  .98}88.97 & \cellcolor[rgb]{ .945,  .973,  .98}88.08 & 92.54 & \cellcolor[rgb]{ .98,  .984,  .957}95.52 & \cellcolor[rgb]{ .98,  .984,  .957}94.38 & 81.94 & \cellcolor[rgb]{ .98,  .984,  .957}93.63 & \cellcolor[rgb]{ .98,  .984,  .957}94.67 & 94.14 & \cellcolor[rgb]{ .996,  .969,  .945}97.16 & \cellcolor[rgb]{ .996,  .969,  .945}98.81 & 98.99 & \cellcolor[rgb]{ .996,  .969,  .945}99.00 & \cellcolor[rgb]{ .996,  .969,  .945}99.00 \\
    Caltech-101 & 92.23 & \cellcolor[rgb]{ .945,  .973,  .98}94.23 & \cellcolor[rgb]{ .945,  .973,  .98}95.43 & 87.87 & \cellcolor[rgb]{ .945,  .973,  .98}96.34 & \cellcolor[rgb]{ .945,  .973,  .98}92.36 & 97.73 & \cellcolor[rgb]{ .98,  .984,  .957}96.06 & \cellcolor[rgb]{ .98,  .984,  .957}96.42 & 64.39 & \cellcolor[rgb]{ .98,  .984,  .957}94.53 & \cellcolor[rgb]{ .98,  .984,  .957}95.27 & 94.95 & \cellcolor[rgb]{ .996,  .969,  .945}96.17 & \cellcolor[rgb]{ .996,  .969,  .945}95.94 & 96.04 & \cellcolor[rgb]{ .996,  .969,  .945}95.79 & \cellcolor[rgb]{ .996,  .969,  .945}95.76 \\
    SUN397 & 84.77 & \cellcolor[rgb]{ .945,  .973,  .98}91.52 & \cellcolor[rgb]{ .945,  .973,  .98}93.70 & 65.04 & \cellcolor[rgb]{ .945,  .973,  .98}84.52 & \cellcolor[rgb]{ .945,  .973,  .98}78.28 & 90.78 & \cellcolor[rgb]{ .98,  .984,  .957}94.66 & \cellcolor[rgb]{ .98,  .984,  .957}95.85 & 32.82 & \cellcolor[rgb]{ .98,  .984,  .957}95.02 & \cellcolor[rgb]{ .98,  .984,  .957}96.06 & 54.22 & \cellcolor[rgb]{ .996,  .969,  .945}95.19 & \cellcolor[rgb]{ .996,  .969,  .945}95.81 & 95.88 & \cellcolor[rgb]{ .996,  .969,  .945}95.83 & \cellcolor[rgb]{ .996,  .969,  .945}95.83 \\
    FGVC Aircraft & 51.10 & \cellcolor[rgb]{ .945,  .973,  .98}60.28 & \cellcolor[rgb]{ .945,  .973,  .98}77.98 & 40.44 & \cellcolor[rgb]{ .945,  .973,  .98}41.04 & \cellcolor[rgb]{ .945,  .973,  .98}40.74 & 69.79 & \cellcolor[rgb]{ .98,  .984,  .957}73.48 & \cellcolor[rgb]{ .98,  .984,  .957}88.78 & 8.58 & \cellcolor[rgb]{ .98,  .984,  .957}77.86 & \cellcolor[rgb]{ .98,  .984,  .957}89.02 & 22.71 & \cellcolor[rgb]{ .996,  .969,  .945}89.98 & \cellcolor[rgb]{ .996,  .969,  .945}91.33 & 91.48 & \cellcolor[rgb]{ .996,  .969,  .945}91.51 & \cellcolor[rgb]{ .996,  .969,  .945}91.48 \\
    Country-211 & 32.69 & \cellcolor[rgb]{ .945,  .973,  .98}50.42 & \cellcolor[rgb]{ .945,  .973,  .98}59.27 & 27.64 & \cellcolor[rgb]{ .945,  .973,  .98}37.44 & \cellcolor[rgb]{ .945,  .973,  .98}24.46 & 37.45 & \cellcolor[rgb]{ .98,  .984,  .957}51.45 & \cellcolor[rgb]{ .98,  .984,  .957}60.10 & 6.62 & \cellcolor[rgb]{ .98,  .984,  .957}52.86 & \cellcolor[rgb]{ .98,  .984,  .957}60.29 & 6.26 & \cellcolor[rgb]{ .996,  .969,  .945}52.40 & \cellcolor[rgb]{ .996,  .969,  .945}56.46 & 56.41 & \cellcolor[rgb]{ .996,  .969,  .945}56.86 & \cellcolor[rgb]{ .996,  .969,  .945}56.82 \\
    Stanford Cars & 82.39 & \cellcolor[rgb]{ .945,  .973,  .98}95.78 & \cellcolor[rgb]{ .945,  .973,  .98}98.36 & 75.80 & \cellcolor[rgb]{ .945,  .973,  .98}76.47 & \cellcolor[rgb]{ .945,  .973,  .98}68.49 & 94.01 & \cellcolor[rgb]{ .98,  .984,  .957}99.14 & \cellcolor[rgb]{ .98,  .984,  .957}99.79 & 6.32 & \cellcolor[rgb]{ .98,  .984,  .957}99.29 & \cellcolor[rgb]{ .98,  .984,  .957}99.78 & 34.26 & \cellcolor[rgb]{ .996,  .969,  .945}99.88 & \cellcolor[rgb]{ .996,  .969,  .945}99.88 & 99.88 & \cellcolor[rgb]{ .996,  .969,  .945}99.89 & \cellcolor[rgb]{ .996,  .969,  .945}99.89 \\
    DTD & 76.91 & \cellcolor[rgb]{ .945,  .973,  .98}77.93 & \cellcolor[rgb]{ .945,  .973,  .98}83.72 & 65.53 & \cellcolor[rgb]{ .945,  .973,  .98}72.93 & \cellcolor[rgb]{ .945,  .973,  .98}73.67 & 86.97 & \cellcolor[rgb]{ .98,  .984,  .957}89.47 & \cellcolor[rgb]{ .98,  .984,  .957}93.14 & 54.73 & \cellcolor[rgb]{ .98,  .984,  .957}91.28 & \cellcolor[rgb]{ .98,  .984,  .957}93.09 & 76.91 & \cellcolor[rgb]{ .996,  .969,  .945}92.82 & \cellcolor[rgb]{ .996,  .969,  .945}94.57 & 94.73 & \cellcolor[rgb]{ .996,  .969,  .945}94.73 & \cellcolor[rgb]{ .996,  .969,  .945}94.73 \\
    Eurosat & 89.43 & \cellcolor[rgb]{ .945,  .973,  .98}89.96 & \cellcolor[rgb]{ .945,  .973,  .98}95.85 & 75.89 & \cellcolor[rgb]{ .945,  .973,  .98}82.22 & \cellcolor[rgb]{ .945,  .973,  .98}86.56 & 91.19 & \cellcolor[rgb]{ .98,  .984,  .957}95.76 & \cellcolor[rgb]{ .98,  .984,  .957}95.89 & 61.26 & \cellcolor[rgb]{ .98,  .984,  .957}95.52 & \cellcolor[rgb]{ .98,  .984,  .957}94.41 & 84.37 & \cellcolor[rgb]{ .996,  .969,  .945}90.13 & \cellcolor[rgb]{ .996,  .969,  .945}97.04 & 96.65 & \cellcolor[rgb]{ .996,  .969,  .945}96.96 & \cellcolor[rgb]{ .996,  .969,  .945}96.96 \\
    FER2013 & 94.89 & \cellcolor[rgb]{ .945,  .973,  .98}94.64 & \cellcolor[rgb]{ .945,  .973,  .98}97.21 & 97.21 & \cellcolor[rgb]{ .945,  .973,  .98}96.75 & \cellcolor[rgb]{ .945,  .973,  .98}95.90 & 97.17 & \cellcolor[rgb]{ .98,  .984,  .957}96.20 & \cellcolor[rgb]{ .98,  .984,  .957}95.83 & 81.74 & \cellcolor[rgb]{ .98,  .984,  .957}95.85 & \cellcolor[rgb]{ .98,  .984,  .957}95.97 & 92.46 & \cellcolor[rgb]{ .996,  .969,  .945}97.12 & \cellcolor[rgb]{ .996,  .969,  .945}97.60 & 97.48 & \cellcolor[rgb]{ .996,  .969,  .945}97.41 & \cellcolor[rgb]{ .996,  .969,  .945}97.41 \\
    Flowers-102 & 73.36 & \cellcolor[rgb]{ .945,  .973,  .98}88.47 & \cellcolor[rgb]{ .945,  .973,  .98}92.01 & 41.31 & \cellcolor[rgb]{ .945,  .973,  .98}64.82 & \cellcolor[rgb]{ .945,  .973,  .98}64.60 & 85.95 & \cellcolor[rgb]{ .98,  .984,  .957}92.49 & \cellcolor[rgb]{ .98,  .984,  .957}93.98 & 28.88 & \cellcolor[rgb]{ .98,  .984,  .957}92.89 & \cellcolor[rgb]{ .98,  .984,  .957}94.18 & 29.61 & \cellcolor[rgb]{ .996,  .969,  .945}97.95 & \cellcolor[rgb]{ .996,  .969,  .945}98.36 & 98.23 & \cellcolor[rgb]{ .996,  .969,  .945}98.29 & \cellcolor[rgb]{ .996,  .969,  .945}98.28 \\
    Food-101 & 94.36 & \cellcolor[rgb]{ .945,  .973,  .98}98.49 & \cellcolor[rgb]{ .945,  .973,  .98}99.31 & 84.41 & \cellcolor[rgb]{ .945,  .973,  .98}61.83 & \cellcolor[rgb]{ .945,  .973,  .98}87.71 & 96.15 & \cellcolor[rgb]{ .98,  .984,  .957}98.60 & \cellcolor[rgb]{ .98,  .984,  .957}99.35 & 17.15 & \cellcolor[rgb]{ .98,  .984,  .957}98.72 & \cellcolor[rgb]{ .98,  .984,  .957}99.41 & 38.34 & \cellcolor[rgb]{ .996,  .969,  .945}99.52 & \cellcolor[rgb]{ .996,  .969,  .945}99.62 & 99.62 & \cellcolor[rgb]{ .996,  .969,  .945}99.63 & \cellcolor[rgb]{ .996,  .969,  .945}99.63 \\
    GTSRB & 60.25 & \cellcolor[rgb]{ .945,  .973,  .98}71.73 & \cellcolor[rgb]{ .945,  .973,  .98}75.95 & 55.45 & \cellcolor[rgb]{ .945,  .973,  .98}69.06 & \cellcolor[rgb]{ .945,  .973,  .98}57.96 & 81.43 & \cellcolor[rgb]{ .98,  .984,  .957}75.71 & \cellcolor[rgb]{ .98,  .984,  .957}82.19 & 46.37 & \cellcolor[rgb]{ .98,  .984,  .957}77.17 & \cellcolor[rgb]{ .98,  .984,  .957}81.26 & 81.69 & \cellcolor[rgb]{ .996,  .969,  .945}77.43 & \cellcolor[rgb]{ .996,  .969,  .945}85.28 & 84.91 & \cellcolor[rgb]{ .996,  .969,  .945}85.41 & \cellcolor[rgb]{ .996,  .969,  .945}85.39 \\
    PCam & - & \cellcolor[rgb]{ .945,  .973,  .98}- & \cellcolor[rgb]{ .945,  .973,  .98}- & - & \cellcolor[rgb]{ .945,  .973,  .98}- & \cellcolor[rgb]{ .945,  .973,  .98}- & - & \cellcolor[rgb]{ .98,  .984,  .957}- & \cellcolor[rgb]{ .98,  .984,  .957}- & - & \cellcolor[rgb]{ .98,  .984,  .957}- & \cellcolor[rgb]{ .98,  .984,  .957}- & - & \cellcolor[rgb]{ .996,  .969,  .945}- & \cellcolor[rgb]{ .996,  .969,  .945}- & - & \cellcolor[rgb]{ .996,  .969,  .945}- & \cellcolor[rgb]{ .996,  .969,  .945}- \\
    Pets & 98.83 & \cellcolor[rgb]{ .945,  .973,  .98}99.67 & \cellcolor[rgb]{ .945,  .973,  .98}99.81 & 96.27 & \cellcolor[rgb]{ .945,  .973,  .98}97.57 & \cellcolor[rgb]{ .945,  .973,  .98}97.49 & 98.69 & \cellcolor[rgb]{ .98,  .984,  .957}99.59 & \cellcolor[rgb]{ .98,  .984,  .957}99.81 & 54.18 & \cellcolor[rgb]{ .98,  .984,  .957}99.81 & \cellcolor[rgb]{ .98,  .984,  .957}99.86 & 57.67 & \cellcolor[rgb]{ .996,  .969,  .945}99.81 & \cellcolor[rgb]{ .996,  .969,  .945}99.86 & 99.92 & \cellcolor[rgb]{ .996,  .969,  .945}99.92 & \cellcolor[rgb]{ .996,  .969,  .945}99.92 \\
    Rendered SST2 & - & \cellcolor[rgb]{ .945,  .973,  .98}- & \cellcolor[rgb]{ .945,  .973,  .98}- & - & \cellcolor[rgb]{ .945,  .973,  .98}- & \cellcolor[rgb]{ .945,  .973,  .98}- & - & \cellcolor[rgb]{ .98,  .984,  .957}- & \cellcolor[rgb]{ .98,  .984,  .957}- & - & \cellcolor[rgb]{ .98,  .984,  .957}- & \cellcolor[rgb]{ .98,  .984,  .957}- & - & \cellcolor[rgb]{ .996,  .969,  .945}- & \cellcolor[rgb]{ .996,  .969,  .945}- & - & \cellcolor[rgb]{ .996,  .969,  .945}- & \cellcolor[rgb]{ .996,  .969,  .945}- \\
    Resisc45 & 80.78 & \cellcolor[rgb]{ .945,  .973,  .98}87.97 & \cellcolor[rgb]{ .945,  .973,  .98}92.35 & 55.52 & \cellcolor[rgb]{ .945,  .973,  .98}59.19 & \cellcolor[rgb]{ .945,  .973,  .98}75.46 & 92.00 & \cellcolor[rgb]{ .98,  .984,  .957}94.19 & \cellcolor[rgb]{ .98,  .984,  .957}96.67 & 32.68 & \cellcolor[rgb]{ .98,  .984,  .957}94.33 & \cellcolor[rgb]{ .98,  .984,  .957}96.43 & 82.19 & \cellcolor[rgb]{ .996,  .969,  .945}94.71 & \cellcolor[rgb]{ .996,  .969,  .945}96.24 & 96.38 & \cellcolor[rgb]{ .996,  .969,  .945}96.37 & \cellcolor[rgb]{ .996,  .969,  .945}96.35 \\
    STL10 & 99.98 & \cellcolor[rgb]{ .945,  .973,  .98}100.00 & \cellcolor[rgb]{ .945,  .973,  .98}100.00 & 99.46 & \cellcolor[rgb]{ .945,  .973,  .98}99.44 & \cellcolor[rgb]{ .945,  .973,  .98}99.94 & 100.00 & \cellcolor[rgb]{ .98,  .984,  .957}100.00 & \cellcolor[rgb]{ .98,  .984,  .957}100.00 & 94.21 & \cellcolor[rgb]{ .98,  .984,  .957}100.00 & \cellcolor[rgb]{ .98,  .984,  .957}100.00 & 99.79 & \cellcolor[rgb]{ .996,  .969,  .945}99.99 & \cellcolor[rgb]{ .996,  .969,  .945}100.00 & 99.99 & \cellcolor[rgb]{ .996,  .969,  .945}100.00 & \cellcolor[rgb]{ .996,  .969,  .945}100.00 \\
    VOC2007 & 94.00 & \cellcolor[rgb]{ .945,  .973,  .98}95.99 & \cellcolor[rgb]{ .945,  .973,  .98}97.01 & 89.00 & \cellcolor[rgb]{ .945,  .973,  .98}92.31 & \cellcolor[rgb]{ .945,  .973,  .98}93.17 & 96.77 & \cellcolor[rgb]{ .98,  .984,  .957}96.71 & \cellcolor[rgb]{ .98,  .984,  .957}97.24 & 78.83 & \cellcolor[rgb]{ .98,  .984,  .957}96.55 & \cellcolor[rgb]{ .98,  .984,  .957}97.18 & 87.35 & \cellcolor[rgb]{ .996,  .969,  .945}78.97 & \cellcolor[rgb]{ .996,  .969,  .945}85.90 & 85.97 & \cellcolor[rgb]{ .996,  .969,  .945}86.87 & \cellcolor[rgb]{ .996,  .969,  .945}86.95 \\
    ImageNet-1K & 85.15 & \cellcolor[rgb]{ .945,  .973,  .98}92.18 & \cellcolor[rgb]{ .945,  .973,  .98}94.60 & 62.05 & \cellcolor[rgb]{ .945,  .973,  .98}75.09 & \cellcolor[rgb]{ .945,  .973,  .98}77.57 & 89.33 & \cellcolor[rgb]{ .98,  .984,  .957}93.81 & \cellcolor[rgb]{ .98,  .984,  .957}95.75 & 16.81 & \cellcolor[rgb]{ .98,  .984,  .957}93.82 & \cellcolor[rgb]{ .98,  .984,  .957}95.82 & 35.10 & \cellcolor[rgb]{ .996,  .969,  .945}96.11 & \cellcolor[rgb]{ .996,  .969,  .945}96.68 & 96.67 & \cellcolor[rgb]{ .996,  .969,  .945}96.69 & \cellcolor[rgb]{ .996,  .969,  .945}96.70 \\
    ImageNet-V2 & 78.25 & \cellcolor[rgb]{ .945,  .973,  .98}87.60 & \cellcolor[rgb]{ .945,  .973,  .98}90.90 & 55.57 & \cellcolor[rgb]{ .945,  .973,  .98}68.64 & \cellcolor[rgb]{ .945,  .973,  .98}70.60 & 83.97 & \cellcolor[rgb]{ .98,  .984,  .957}89.92 & \cellcolor[rgb]{ .98,  .984,  .957}92.52 & 14.46 & \cellcolor[rgb]{ .98,  .984,  .957}89.77 & \cellcolor[rgb]{ .98,  .984,  .957}92.64 & 29.20 & \cellcolor[rgb]{ .996,  .969,  .945}93.27 & \cellcolor[rgb]{ .996,  .969,  .945}94.09 & 93.92 & \cellcolor[rgb]{ .996,  .969,  .945}94.06 & \cellcolor[rgb]{ .996,  .969,  .945}94.01 \\
    ImageNet-Adv. & 64.21 & \cellcolor[rgb]{ .945,  .973,  .98}86.28 & \cellcolor[rgb]{ .945,  .973,  .98}90.76 & 45.11 & \cellcolor[rgb]{ .945,  .973,  .98}41.61 & \cellcolor[rgb]{ .945,  .973,  .98}51.29 & 74.33 & \cellcolor[rgb]{ .98,  .984,  .957}86.67 & \cellcolor[rgb]{ .98,  .984,  .957}90.84 & 7.81 & \cellcolor[rgb]{ .98,  .984,  .957}86.75 & \cellcolor[rgb]{ .98,  .984,  .957}91.24 & 33.11 & \cellcolor[rgb]{ .996,  .969,  .945}86.45 & \cellcolor[rgb]{ .996,  .969,  .945}89.55 & 89.03 & \cellcolor[rgb]{ .996,  .969,  .945}89.68 & \cellcolor[rgb]{ .996,  .969,  .945}89.67 \\
    ImageNet-Ren. & 88.58 & \cellcolor[rgb]{ .945,  .973,  .98}94.55 & \cellcolor[rgb]{ .945,  .973,  .98}97.14 & 70.20 & \cellcolor[rgb]{ .945,  .973,  .98}79.45 & \cellcolor[rgb]{ .945,  .973,  .98}83.01 & 93.46 & \cellcolor[rgb]{ .98,  .984,  .957}96.68 & \cellcolor[rgb]{ .98,  .984,  .957}98.09 & 26.31 & \cellcolor[rgb]{ .98,  .984,  .957}96.81 & \cellcolor[rgb]{ .98,  .984,  .957}98.13 & 58.71 & \cellcolor[rgb]{ .996,  .969,  .945}98.38 & \cellcolor[rgb]{ .996,  .969,  .945}99.09 & 99.03 & \cellcolor[rgb]{ .996,  .969,  .945}99.13 & \cellcolor[rgb]{ .996,  .969,  .945}99.13 \\
    ImageNet-OOD & 54.85 & \cellcolor[rgb]{ .945,  .973,  .98}58.35 & \cellcolor[rgb]{ .945,  .973,  .98}59.45 & 47.60 & \cellcolor[rgb]{ .945,  .973,  .98}66.80 & \cellcolor[rgb]{ .945,  .973,  .98}52.45 & 62.65 & \cellcolor[rgb]{ .98,  .984,  .957}58.15 & \cellcolor[rgb]{ .98,  .984,  .957}60.10 & 28.00 & \cellcolor[rgb]{ .98,  .984,  .957}65.40 & \cellcolor[rgb]{ .98,  .984,  .957}61.50 & 26.50 & \cellcolor[rgb]{ .996,  .969,  .945}61.90 & \cellcolor[rgb]{ .996,  .969,  .945}59.70 & 61.45 & \cellcolor[rgb]{ .996,  .969,  .945}60.35 & \cellcolor[rgb]{ .996,  .969,  .945}60.35 \\
    ObjectNet & 71.28 & \cellcolor[rgb]{ .945,  .973,  .98}83.61 & \cellcolor[rgb]{ .945,  .973,  .98}88.12 & 46.25 & \cellcolor[rgb]{ .945,  .973,  .98}50.65 & \cellcolor[rgb]{ .945,  .973,  .98}65.05 & 82.11 & \cellcolor[rgb]{ .98,  .984,  .957}88.37 & \cellcolor[rgb]{ .98,  .984,  .957}91.25 & 11.49 & \cellcolor[rgb]{ .98,  .984,  .957}87.46 & \cellcolor[rgb]{ .98,  .984,  .957}90.96 & 47.02 & \cellcolor[rgb]{ .996,  .969,  .945}83.85 & \cellcolor[rgb]{ .996,  .969,  .945}88.02 & 87.24 & \cellcolor[rgb]{ .996,  .969,  .945}87.74 & \cellcolor[rgb]{ .996,  .969,  .945}87.78 \\
    \midrule
    Avg. ACC@5 & 79.89 & \cellcolor[rgb]{ .945,  .973,  .98}86.62 & \cellcolor[rgb]{ .945,  .973,  .98}\textbf{90.20} & 67.88 & \cellcolor[rgb]{ .945,  .973,  .98}73.78 & \cellcolor[rgb]{ .945,  .973,  .98}\textbf{74.98} & 87.10 & \cellcolor[rgb]{ .98,  .984,  .957}89.94 & \cellcolor[rgb]{ .98,  .984,  .957}\textbf{92.28} & 41.97 & \cellcolor[rgb]{ .98,  .984,  .957}90.46 & \cellcolor[rgb]{ .98,  .984,  .957}\textbf{92.24} & 60.73 & \cellcolor[rgb]{ .996,  .969,  .945}90.49 & \cellcolor[rgb]{ .996,  .969,  .945}\textbf{92.32} & 92.31 & \cellcolor[rgb]{ .996,  .969,  .945}\textbf{92.42} & \cellcolor[rgb]{ .996,  .969,  .945}92.41 \\
    \bottomrule
    \end{tabular}%
   }
\label{tab: classification acc5}
\end{table}

\newpage

\begin{table}[htbp]
  \centering
  \renewcommand\arraystretch{1.7}
  \setlength{\tabcolsep}{3pt}
  \caption{\textbf{Performance on 26 standard \underline{zero-shot classification} benchmarks in terms of \underline{mRecall}.} This table presents the detailed results corresponding to \cref{fig: classification results}.}
  \resizebox{\linewidth}{!}{
    \begin{tabular}{c|ccc|ccc|ccc|ccc|ccc|ccc}
    \toprule
    \multirow[c]{2}{*}{\textbf{Dataset}} & \multicolumn{6}{c|}{\textbf{OpenAI CLIP ViT-L@224 (O-1)}} & \multicolumn{6}{c|}{\textbf{MetaCLIP ViT-L@224 (M-1)}} & \multicolumn{6}{c}{\textbf{SigLIP ViT-SO@224 (S-1)}} \\
    & \rotatebox[origin=c]{90}{\textbf{DIVA}} & \rotatebox[origin=c]{90}{\textbf{+AUCGR}} & \rotatebox[origin=c]{90}{\textbf{+pAUCGR}} & \rotatebox[origin=c]{90}{\textbf{GenHancer}} & \rotatebox[origin=c]{90}{\textbf{+AUCGR}} & \rotatebox[origin=c]{90}{\textbf{+pAUCGR}} & \rotatebox[origin=c]{90}{\textbf{DIVA}} & \rotatebox[origin=c]{90}{\textbf{+AUCGR}} & \rotatebox[origin=c]{90}{\textbf{+pAUCGR}} & \rotatebox[origin=c]{90}{\textbf{GenHancer}} & \rotatebox[origin=c]{90}{\textbf{+AUCGR}} & \rotatebox[origin=c]{90}{\textbf{+pAUCGR}} & \rotatebox[origin=c]{90}{\textbf{DIVA}} & \rotatebox[origin=c]{90}{\textbf{+AUCGR}} & \rotatebox[origin=c]{90}{\textbf{+pAUCGR}} & \rotatebox[origin=c]{90}{\textbf{GenHancer}} & \rotatebox[origin=c]{90}{\textbf{+AUCGR}} & \rotatebox[origin=c]{90}{\textbf{+pAUCGR}} \\
    \midrule
    CIFAR-10 & 86.95 & \cellcolor[rgb]{ .945,  .973,  .98}92.22 & \cellcolor[rgb]{ .945,  .973,  .98}95.62 & 76.35 & \cellcolor[rgb]{ .945,  .973,  .98}75.76 & \cellcolor[rgb]{ .945,  .973,  .98}79.44 & 96.67 & \cellcolor[rgb]{ .98,  .984,  .957}97.45 & \cellcolor[rgb]{ .98,  .984,  .957}97.57 & 58.09 & \cellcolor[rgb]{ .98,  .984,  .957}96.58 & \cellcolor[rgb]{ .98,  .984,  .957}97.33 & 95.19 & \cellcolor[rgb]{ .996,  .969,  .945}92.31 & \cellcolor[rgb]{ .996,  .969,  .945}96.68 & 96.71 & \cellcolor[rgb]{ .996,  .969,  .945}96.69 & \cellcolor[rgb]{ .996,  .969,  .945}96.69 \\
    CIFAR-100 & 58.45 & \cellcolor[rgb]{ .945,  .973,  .98}70.88 & \cellcolor[rgb]{ .945,  .973,  .98}75.72 & 36.35 & \cellcolor[rgb]{ .945,  .973,  .98}43.79 & \cellcolor[rgb]{ .945,  .973,  .98}51.88 & 79.82 & \cellcolor[rgb]{ .98,  .984,  .957}81.09 & \cellcolor[rgb]{ .98,  .984,  .957}84.10 & 28.64 & \cellcolor[rgb]{ .98,  .984,  .957}81.32 & \cellcolor[rgb]{ .98,  .984,  .957}84.20 & 76.86 & \cellcolor[rgb]{ .996,  .969,  .945}78.42 & \cellcolor[rgb]{ .996,  .969,  .945}84.12 & 83.71 & \cellcolor[rgb]{ .996,  .969,  .945}84.22 & \cellcolor[rgb]{ .996,  .969,  .945}84.26 \\
    MNIST & 53.68 & \cellcolor[rgb]{ .945,  .973,  .98}66.76 & \cellcolor[rgb]{ .945,  .973,  .98}76.32 & 49.37 & \cellcolor[rgb]{ .945,  .973,  .98}65.46 & \cellcolor[rgb]{ .945,  .973,  .98}48.97 & 63.20 & \cellcolor[rgb]{ .98,  .984,  .957}62.89 & \cellcolor[rgb]{ .98,  .984,  .957}63.74 & 29.76 & \cellcolor[rgb]{ .98,  .984,  .957}64.30 & \cellcolor[rgb]{ .98,  .984,  .957}65.12 & 74.33 & \cellcolor[rgb]{ .996,  .969,  .945}81.52 & \cellcolor[rgb]{ .996,  .969,  .945}87.98 & 88.21 & \cellcolor[rgb]{ .996,  .969,  .945}89.01 & \cellcolor[rgb]{ .996,  .969,  .945}89.00 \\
    Caltech-101 & 87.27 & \cellcolor[rgb]{ .945,  .973,  .98}90.89 & \cellcolor[rgb]{ .945,  .973,  .98}92.49 & 78.14 & \cellcolor[rgb]{ .945,  .973,  .98}81.66 & \cellcolor[rgb]{ .945,  .973,  .98}83.82 & 93.05 & \cellcolor[rgb]{ .98,  .984,  .957}95.18 & \cellcolor[rgb]{ .98,  .984,  .957}95.29 & 29.23 & \cellcolor[rgb]{ .98,  .984,  .957}93.85 & \cellcolor[rgb]{ .98,  .984,  .957}95.18 & 84.14 & \cellcolor[rgb]{ .996,  .969,  .945}95.40 & \cellcolor[rgb]{ .996,  .969,  .945}95.99 & 96.11 & \cellcolor[rgb]{ .996,  .969,  .945}96.03 & \cellcolor[rgb]{ .996,  .969,  .945}96.03 \\
    SUN397 & 50.66 & \cellcolor[rgb]{ .945,  .973,  .98}63.31 & \cellcolor[rgb]{ .945,  .973,  .98}68.45 & 30.74 & \cellcolor[rgb]{ .945,  .973,  .98}52.44 & \cellcolor[rgb]{ .945,  .973,  .98}43.27 & 62.02 & \cellcolor[rgb]{ .98,  .984,  .957}70.32 & \cellcolor[rgb]{ .98,  .984,  .957}74.71 & 10.59 & \cellcolor[rgb]{ .98,  .984,  .957}70.32 & \cellcolor[rgb]{ .98,  .984,  .957}75.09 & 25.21 & \cellcolor[rgb]{ .996,  .969,  .945}73.79 & \cellcolor[rgb]{ .996,  .969,  .945}75.82 & 76.05 & \cellcolor[rgb]{ .996,  .969,  .945}75.93 & \cellcolor[rgb]{ .996,  .969,  .945}75.92 \\
    FGVC Aircraft & 19.38 & \cellcolor[rgb]{ .945,  .973,  .98}23.85 & \cellcolor[rgb]{ .945,  .973,  .98}31.89 & 13.10 & \cellcolor[rgb]{ .945,  .973,  .98}12.28 & \cellcolor[rgb]{ .945,  .973,  .98}15.14 & 28.83 & \cellcolor[rgb]{ .98,  .984,  .957}31.00 & \cellcolor[rgb]{ .98,  .984,  .957}43.52 & 2.34 & \cellcolor[rgb]{ .98,  .984,  .957}33.81 & \cellcolor[rgb]{ .98,  .984,  .957}44.56 & 7.95 & \cellcolor[rgb]{ .996,  .969,  .945}55.79 & \cellcolor[rgb]{ .996,  .969,  .945}60.50 & 60.11 & \cellcolor[rgb]{ .996,  .969,  .945}60.41 & \cellcolor[rgb]{ .996,  .969,  .945}60.50 \\
    Country-211 & 12.42 & \cellcolor[rgb]{ .945,  .973,  .98}23.77 & \cellcolor[rgb]{ .945,  .973,  .98}31.75 & 10.37 & \cellcolor[rgb]{ .945,  .973,  .98}16.27 & \cellcolor[rgb]{ .945,  .973,  .98}7.81 & 16.71 & \cellcolor[rgb]{ .98,  .984,  .957}26.56 & \cellcolor[rgb]{ .98,  .984,  .957}33.53 & 1.78 & \cellcolor[rgb]{ .98,  .984,  .957}27.71 & \cellcolor[rgb]{ .98,  .984,  .957}33.82 & 0.50 & \cellcolor[rgb]{ .996,  .969,  .945}28.61 & \cellcolor[rgb]{ .996,  .969,  .945}32.12 & 32.27 & \cellcolor[rgb]{ .996,  .969,  .945}32.48 & \cellcolor[rgb]{ .996,  .969,  .945}32.45 \\
    Stanford Cars & 46.16 & \cellcolor[rgb]{ .945,  .973,  .98}68.31 & \cellcolor[rgb]{ .945,  .973,  .98}77.77 & 39.11 & \cellcolor[rgb]{ .945,  .973,  .98}38.22 & \cellcolor[rgb]{ .945,  .973,  .98}33.26 & 65.57 & \cellcolor[rgb]{ .98,  .984,  .957}81.73 & \cellcolor[rgb]{ .98,  .984,  .957}88.60 & 1.63 & \cellcolor[rgb]{ .98,  .984,  .957}82.61 & \cellcolor[rgb]{ .98,  .984,  .957}88.68 & 15.37 & \cellcolor[rgb]{ .996,  .969,  .945}90.21 & \cellcolor[rgb]{ .996,  .969,  .945}90.42 & 90.22 & \cellcolor[rgb]{ .996,  .969,  .945}90.42 & \cellcolor[rgb]{ .996,  .969,  .945}90.39 \\
    DTD & 45.96 & \cellcolor[rgb]{ .945,  .973,  .98}49.52 & \cellcolor[rgb]{ .945,  .973,  .98}55.27 & 34.04 & \cellcolor[rgb]{ .945,  .973,  .98}39.41 & \cellcolor[rgb]{ .945,  .973,  .98}42.82 & 58.14 & \cellcolor[rgb]{ .98,  .984,  .957}63.99 & \cellcolor[rgb]{ .98,  .984,  .957}68.46 & 23.94 & \cellcolor[rgb]{ .98,  .984,  .957}65.74 & \cellcolor[rgb]{ .98,  .984,  .957}67.93 & 43.40 & \cellcolor[rgb]{ .996,  .969,  .945}69.84 & \cellcolor[rgb]{ .996,  .969,  .945}71.33 & 71.12 & \cellcolor[rgb]{ .996,  .969,  .945}71.22 & \cellcolor[rgb]{ .996,  .969,  .945}71.17 \\
    Eurosat & 44.44 & \cellcolor[rgb]{ .945,  .973,  .98}53.60 & \cellcolor[rgb]{ .945,  .973,  .98}63.60 & 21.65 & \cellcolor[rgb]{ .945,  .973,  .98}23.58 & \cellcolor[rgb]{ .945,  .973,  .98}40.88 & 54.94 & \cellcolor[rgb]{ .98,  .984,  .957}67.80 & \cellcolor[rgb]{ .98,  .984,  .957}60.87 & 20.43 & \cellcolor[rgb]{ .98,  .984,  .957}56.05 & \cellcolor[rgb]{ .98,  .984,  .957}58.63 & 44.72 & \cellcolor[rgb]{ .996,  .969,  .945}46.35 & \cellcolor[rgb]{ .996,  .969,  .945}66.48 & 64.50 & \cellcolor[rgb]{ .996,  .969,  .945}65.06 & \cellcolor[rgb]{ .996,  .969,  .945}65.12 \\
    FER2013 & 24.75 & \cellcolor[rgb]{ .945,  .973,  .98}44.30 & \cellcolor[rgb]{ .945,  .973,  .98}49.05 & 50.63 & \cellcolor[rgb]{ .945,  .973,  .98}39.74 & \cellcolor[rgb]{ .945,  .973,  .98}21.78 & 49.06 & \cellcolor[rgb]{ .98,  .984,  .957}52.87 & \cellcolor[rgb]{ .98,  .984,  .957}52.41 & 28.13 & \cellcolor[rgb]{ .98,  .984,  .957}53.28 & \cellcolor[rgb]{ .98,  .984,  .957}52.87 & 48.81 & \cellcolor[rgb]{ .996,  .969,  .945}47.22 & \cellcolor[rgb]{ .996,  .969,  .945}48.77 & 48.54 & \cellcolor[rgb]{ .996,  .969,  .945}48.75 & \cellcolor[rgb]{ .996,  .969,  .945}48.75 \\
    Flowers-102 & 46.34 & \cellcolor[rgb]{ .945,  .973,  .98}67.93 & \cellcolor[rgb]{ .945,  .973,  .98}79.21 & 21.34 & \cellcolor[rgb]{ .945,  .973,  .98}36.86 & \cellcolor[rgb]{ .945,  .973,  .98}36.35 & 61.40 & \cellcolor[rgb]{ .98,  .984,  .957}77.19 & \cellcolor[rgb]{ .98,  .984,  .957}81.40 & 11.47 & \cellcolor[rgb]{ .98,  .984,  .957}75.52 & \cellcolor[rgb]{ .98,  .984,  .957}81.64 & 13.05 & \cellcolor[rgb]{ .996,  .969,  .945}88.14 & \cellcolor[rgb]{ .996,  .969,  .945}89.98 & 89.90 & \cellcolor[rgb]{ .996,  .969,  .945}89.64 & \cellcolor[rgb]{ .996,  .969,  .945}89.70 \\
    Food-101 & 75.49 & \cellcolor[rgb]{ .945,  .973,  .98}89.32 & \cellcolor[rgb]{ .945,  .973,  .98}93.01 & 59.16 & \cellcolor[rgb]{ .945,  .973,  .98}36.27 & \cellcolor[rgb]{ .945,  .973,  .98}65.28 & 81.72 & \cellcolor[rgb]{ .98,  .984,  .957}89.66 & \cellcolor[rgb]{ .98,  .984,  .957}93.41 & 4.72 & \cellcolor[rgb]{ .98,  .984,  .957}89.78 & \cellcolor[rgb]{ .98,  .984,  .957}93.41 & 19.02 & \cellcolor[rgb]{ .996,  .969,  .945}94.65 & \cellcolor[rgb]{ .996,  .969,  .945}95.56 & 95.46 & \cellcolor[rgb]{ .996,  .969,  .945}95.54 & \cellcolor[rgb]{ .996,  .969,  .945}95.54 \\
    GTSRB & 30.51 & \cellcolor[rgb]{ .945,  .973,  .98}38.01 & \cellcolor[rgb]{ .945,  .973,  .98}44.04 & 20.82 & \cellcolor[rgb]{ .945,  .973,  .98}29.64 & \cellcolor[rgb]{ .945,  .973,  .98}27.06 & 42.35 & \cellcolor[rgb]{ .98,  .984,  .957}48.00 & \cellcolor[rgb]{ .98,  .984,  .957}56.37 & 16.02 & \cellcolor[rgb]{ .98,  .984,  .957}53.23 & \cellcolor[rgb]{ .98,  .984,  .957}55.43 & 49.66 & \cellcolor[rgb]{ .996,  .969,  .945}45.01 & \cellcolor[rgb]{ .996,  .969,  .945}56.93 & 56.03 & \cellcolor[rgb]{ .996,  .969,  .945}56.25 & \cellcolor[rgb]{ .996,  .969,  .945}56.18 \\
    PCam & 53.57 & \cellcolor[rgb]{ .945,  .973,  .98}51.83 & \cellcolor[rgb]{ .945,  .973,  .98}52.43 & 50.00 & \cellcolor[rgb]{ .945,  .973,  .98}50.00 & \cellcolor[rgb]{ .945,  .973,  .98}52.30 & 49.02 & \cellcolor[rgb]{ .98,  .984,  .957}52.21 & \cellcolor[rgb]{ .98,  .984,  .957}63.32 & 51.42 & \cellcolor[rgb]{ .98,  .984,  .957}57.68 & \cellcolor[rgb]{ .98,  .984,  .957}61.29 & 53.99 & \cellcolor[rgb]{ .996,  .969,  .945}50.09 & \cellcolor[rgb]{ .996,  .969,  .945}50.19 & 50.26 & \cellcolor[rgb]{ .996,  .969,  .945}50.63 & \cellcolor[rgb]{ .996,  .969,  .945}50.60 \\
    Pets & 83.07 & \cellcolor[rgb]{ .945,  .973,  .98}89.87 & \cellcolor[rgb]{ .945,  .973,  .98}93.17 & 74.89 & \cellcolor[rgb]{ .945,  .973,  .98}79.24 & \cellcolor[rgb]{ .945,  .973,  .98}76.43 & 82.80 & \cellcolor[rgb]{ .98,  .984,  .957}87.45 & \cellcolor[rgb]{ .98,  .984,  .957}94.14 & 19.07 & \cellcolor[rgb]{ .98,  .984,  .957}92.42 & \cellcolor[rgb]{ .98,  .984,  .957}94.46 & 31.22 & \cellcolor[rgb]{ .996,  .969,  .945}93.77 & \cellcolor[rgb]{ .996,  .969,  .945}95.16 & 95.00 & \cellcolor[rgb]{ .996,  .969,  .945}95.30 & \cellcolor[rgb]{ .996,  .969,  .945}95.30 \\
    Rendered SST2 & 69.64 & \cellcolor[rgb]{ .945,  .973,  .98}56.63 & \cellcolor[rgb]{ .945,  .973,  .98}69.10 & 51.59 & \cellcolor[rgb]{ .945,  .973,  .98}51.97 & \cellcolor[rgb]{ .945,  .973,  .98}64.77 & 59.80 & \cellcolor[rgb]{ .98,  .984,  .957}64.43 & \cellcolor[rgb]{ .98,  .984,  .957}65.98 & 50.00 & \cellcolor[rgb]{ .98,  .984,  .957}68.43 & \cellcolor[rgb]{ .98,  .984,  .957}68.10 & 64.34 & \cellcolor[rgb]{ .996,  .969,  .945}69.68 & \cellcolor[rgb]{ .996,  .969,  .945}69.20 & 70.13 & \cellcolor[rgb]{ .996,  .969,  .945}70.03 & \cellcolor[rgb]{ .996,  .969,  .945}70.03 \\
    Resisc45 & 44.89 & \cellcolor[rgb]{ .945,  .973,  .98}55.46 & \cellcolor[rgb]{ .945,  .973,  .98}64.15 & 25.38 & \cellcolor[rgb]{ .945,  .973,  .98}27.15 & \cellcolor[rgb]{ .945,  .973,  .98}37.89 & 62.68 & \cellcolor[rgb]{ .98,  .984,  .957}69.49 & \cellcolor[rgb]{ .98,  .984,  .957}74.63 & 9.60 & \cellcolor[rgb]{ .98,  .984,  .957}67.50 & \cellcolor[rgb]{ .98,  .984,  .957}74.26 & 53.96 & \cellcolor[rgb]{ .996,  .969,  .945}69.90 & \cellcolor[rgb]{ .996,  .969,  .945}70.18 & 70.37 & \cellcolor[rgb]{ .996,  .969,  .945}70.33 & \cellcolor[rgb]{ .996,  .969,  .945}70.29 \\
    STL10 & 98.36 & \cellcolor[rgb]{ .945,  .973,  .98}99.04 & \cellcolor[rgb]{ .945,  .973,  .98}99.43 & 93.05 & \cellcolor[rgb]{ .945,  .973,  .98}89.56 & \cellcolor[rgb]{ .945,  .973,  .98}96.31 & 98.78 & \cellcolor[rgb]{ .98,  .984,  .957}99.13 & \cellcolor[rgb]{ .98,  .984,  .957}99.29 & 55.73 & \cellcolor[rgb]{ .98,  .984,  .957}99.16 & \cellcolor[rgb]{ .98,  .984,  .957}99.36 & 94.53 & \cellcolor[rgb]{ .996,  .969,  .945}98.64 & \cellcolor[rgb]{ .996,  .969,  .945}99.34 & 99.43 & \cellcolor[rgb]{ .996,  .969,  .945}99.39 & \cellcolor[rgb]{ .996,  .969,  .945}99.39 \\
    VOC2007 & 81.62 & \cellcolor[rgb]{ .945,  .973,  .98}83.36 & \cellcolor[rgb]{ .945,  .973,  .98}86.56 & 74.52 & \cellcolor[rgb]{ .945,  .973,  .98}70.64 & \cellcolor[rgb]{ .945,  .973,  .98}79.16 & 85.16 & \cellcolor[rgb]{ .98,  .984,  .957}87.40 & \cellcolor[rgb]{ .98,  .984,  .957}87.17 & 40.49 & \cellcolor[rgb]{ .98,  .984,  .957}87.23 & \cellcolor[rgb]{ .98,  .984,  .957}87.77 & 77.57 & \cellcolor[rgb]{ .996,  .969,  .945}84.59 & \cellcolor[rgb]{ .996,  .969,  .945}88.75 & 88.98 & \cellcolor[rgb]{ .996,  .969,  .945}88.95 & \cellcolor[rgb]{ .996,  .969,  .945}88.96 \\
    ImageNet-1K & 57.55 & \cellcolor[rgb]{ .945,  .973,  .98}69.47 & \cellcolor[rgb]{ .945,  .973,  .98}75.54 & 34.04 & \cellcolor[rgb]{ .945,  .973,  .98}46.74 & \cellcolor[rgb]{ .945,  .973,  .98}48.58 & 63.54 & \cellcolor[rgb]{ .98,  .984,  .957}73.52 & \cellcolor[rgb]{ .98,  .984,  .957}78.84 & 6.18 & \cellcolor[rgb]{ .98,  .984,  .957}73.62 & \cellcolor[rgb]{ .98,  .984,  .957}78.95 & 18.54 & \cellcolor[rgb]{ .996,  .969,  .945}80.53 & \cellcolor[rgb]{ .996,  .969,  .945}82.01 & 81.95 & \cellcolor[rgb]{ .996,  .969,  .945}82.04 & \cellcolor[rgb]{ .996,  .969,  .945}82.07 \\
    ImageNet-V2 & 50.26 & \cellcolor[rgb]{ .945,  .973,  .98}63.30 & \cellcolor[rgb]{ .945,  .973,  .98}69.77 & 29.12 & \cellcolor[rgb]{ .945,  .973,  .98}40.29 & \cellcolor[rgb]{ .945,  .973,  .98}41.96 & 56.01 & \cellcolor[rgb]{ .98,  .984,  .957}66.50 & \cellcolor[rgb]{ .98,  .984,  .957}72.36 & 5.31 & \cellcolor[rgb]{ .98,  .984,  .957}66.39 & \cellcolor[rgb]{ .98,  .984,  .957}72.33 & 15.56 & \cellcolor[rgb]{ .996,  .969,  .945}74.29 & \cellcolor[rgb]{ .996,  .969,  .945}76.30 & 75.99 & \cellcolor[rgb]{ .996,  .969,  .945}76.19 & \cellcolor[rgb]{ .996,  .969,  .945}76.14 \\
    ImageNet-Adv. & 36.12 & \cellcolor[rgb]{ .945,  .973,  .98}58.98 & \cellcolor[rgb]{ .945,  .973,  .98}67.46 & 20.73 & \cellcolor[rgb]{ .945,  .973,  .98}19.75 & \cellcolor[rgb]{ .945,  .973,  .98}25.79 & 42.92 & \cellcolor[rgb]{ .98,  .984,  .957}60.56 & \cellcolor[rgb]{ .98,  .984,  .957}68.76 & 2.60 & \cellcolor[rgb]{ .98,  .984,  .957}59.83 & \cellcolor[rgb]{ .98,  .984,  .957}69.28 & 17.65 & \cellcolor[rgb]{ .996,  .969,  .945}66.95 & \cellcolor[rgb]{ .996,  .969,  .945}72.63 & 71.36 & \cellcolor[rgb]{ .996,  .969,  .945}72.38 & \cellcolor[rgb]{ .996,  .969,  .945}72.33 \\
    ImageNet-Ren. & 70.01 & \cellcolor[rgb]{ .945,  .973,  .98}79.70 & \cellcolor[rgb]{ .945,  .973,  .98}86.38 & 40.02 & \cellcolor[rgb]{ .945,  .973,  .98}56.36 & \cellcolor[rgb]{ .945,  .973,  .98}60.75 & 78.37 & \cellcolor[rgb]{ .98,  .984,  .957}86.51 & \cellcolor[rgb]{ .98,  .984,  .957}90.84 & 9.67 & \cellcolor[rgb]{ .98,  .984,  .957}86.68 & \cellcolor[rgb]{ .98,  .984,  .957}90.95 & 34.04 & \cellcolor[rgb]{ .996,  .969,  .945}92.45 & \cellcolor[rgb]{ .996,  .969,  .945}93.98 & 93.83 & \cellcolor[rgb]{ .996,  .969,  .945}94.04 & \cellcolor[rgb]{ .996,  .969,  .945}94.06 \\
    ImageNet-OOD & 28.24 & \cellcolor[rgb]{ .945,  .973,  .98}32.20 & \cellcolor[rgb]{ .945,  .973,  .98}32.55 & 23.62 & \cellcolor[rgb]{ .945,  .973,  .98}35.63 & \cellcolor[rgb]{ .945,  .973,  .98}26.66 & 32.57 & \cellcolor[rgb]{ .98,  .984,  .957}31.09 & \cellcolor[rgb]{ .98,  .984,  .957}31.60 & 12.13 & \cellcolor[rgb]{ .98,  .984,  .957}36.07 & \cellcolor[rgb]{ .98,  .984,  .957}32.36 & 11.53 & \cellcolor[rgb]{ .996,  .969,  .945}32.56 & \cellcolor[rgb]{ .996,  .969,  .945}31.40 & 31.57 & \cellcolor[rgb]{ .996,  .969,  .945}31.03 & \cellcolor[rgb]{ .996,  .969,  .945}31.03 \\
    ObjectNet & 44.75 & \cellcolor[rgb]{ .945,  .973,  .98}61.21 & \cellcolor[rgb]{ .945,  .973,  .98}67.56 & 21.36 & \cellcolor[rgb]{ .945,  .973,  .98}26.29 & \cellcolor[rgb]{ .945,  .973,  .98}37.62 & 58.04 & \cellcolor[rgb]{ .98,  .984,  .957}68.14 & \cellcolor[rgb]{ .98,  .984,  .957}73.02 & 3.15 & \cellcolor[rgb]{ .98,  .984,  .957}66.69 & \cellcolor[rgb]{ .98,  .984,  .957}72.85 & 13.70 & \cellcolor[rgb]{ .996,  .969,  .945}63.69 & \cellcolor[rgb]{ .996,  .969,  .945}70.05 & 68.59 & \cellcolor[rgb]{ .996,  .969,  .945}69.45 & \cellcolor[rgb]{ .996,  .969,  .945}69.49 \\
    \midrule
    Avg. mRecall & 53.87 & \cellcolor[rgb]{ .945,  .973,  .98}63.22 & \cellcolor[rgb]{ .945,  .973,  .98}\textbf{69.16} & 41.52 & \cellcolor[rgb]{ .945,  .973,  .98}45.58 & \cellcolor[rgb]{ .945,  .973,  .98}\textbf{47.92} & 62.43 & \cellcolor[rgb]{ .98,  .984,  .957}68.93 & \cellcolor[rgb]{ .98,  .984,  .957}\textbf{72.84} & 20.47 & \cellcolor[rgb]{ .98,  .984,  .957}69.45 & \cellcolor[rgb]{ .98,  .984,  .957}\textbf{72.92} & 41.72 & \cellcolor[rgb]{ .996,  .969,  .945}71.71 & \cellcolor[rgb]{ .996,  .969,  .945}\textbf{75.07} & 74.86 & \cellcolor[rgb]{ .996,  .969,  .945}\textbf{75.05} & \cellcolor[rgb]{ .996,  .969,  .945}\textbf{75.05} \\
    \bottomrule
    \end{tabular}%
   }
\label{tab: classification mrecall}
\end{table}

\newpage

\subsubsection{Zero-shot Retrieval}
\label{subsubsec: supp_Zero-shot Retrieval}
\begin{figure}[htbp]
  \centering
  \subfloat[Image@5 - Flickr30k\label{fig: M1_Image_Flickr30k}]{
    \includegraphics[width=0.23\linewidth]{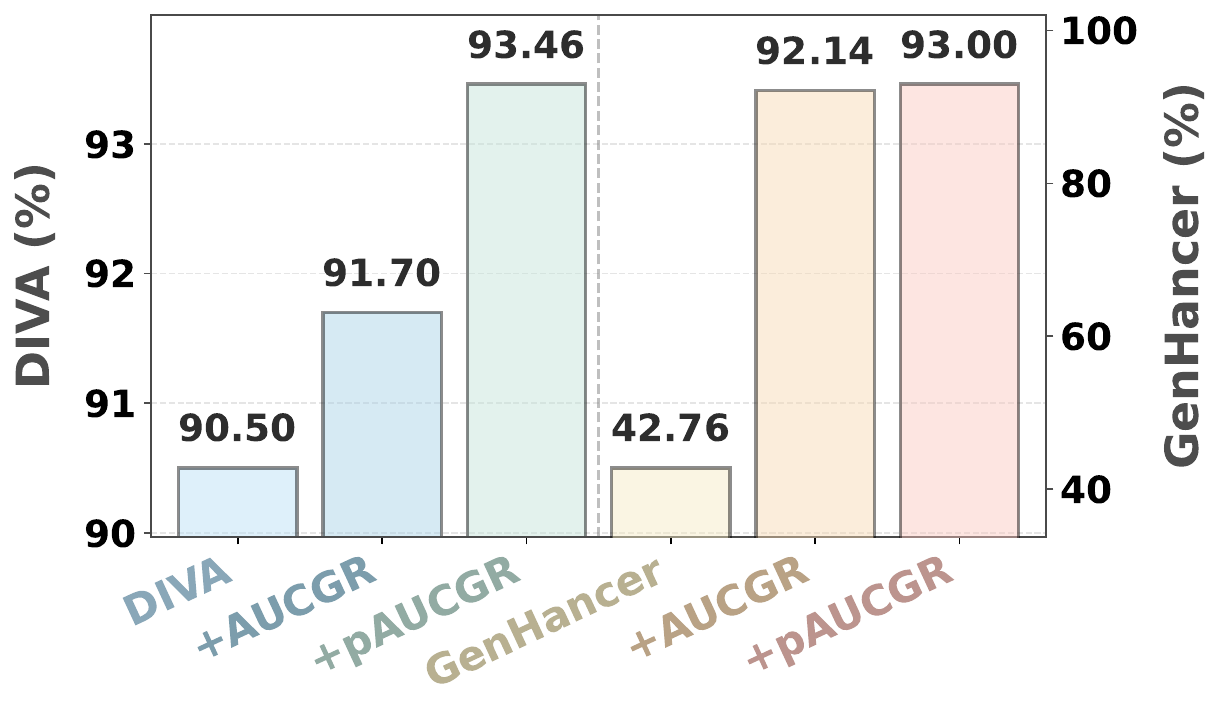}}
  \hfill
  \subfloat[Image@5 - COCO\label{fig: M1_Image_COCO}]{
    \includegraphics[width=0.23\linewidth]{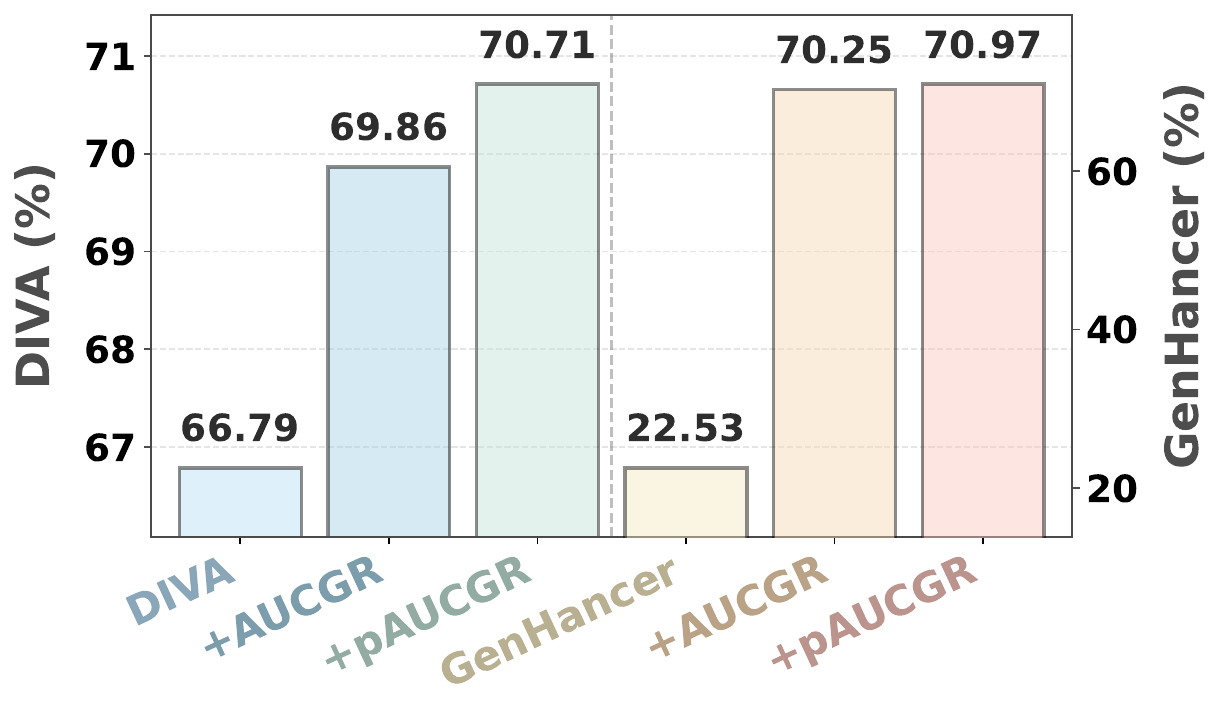}}
  \hfill
  \subfloat[Text@5 - Flickr30k\label{fig: M1_Text_Flickr30k}]{
    \includegraphics[width=0.23\linewidth]{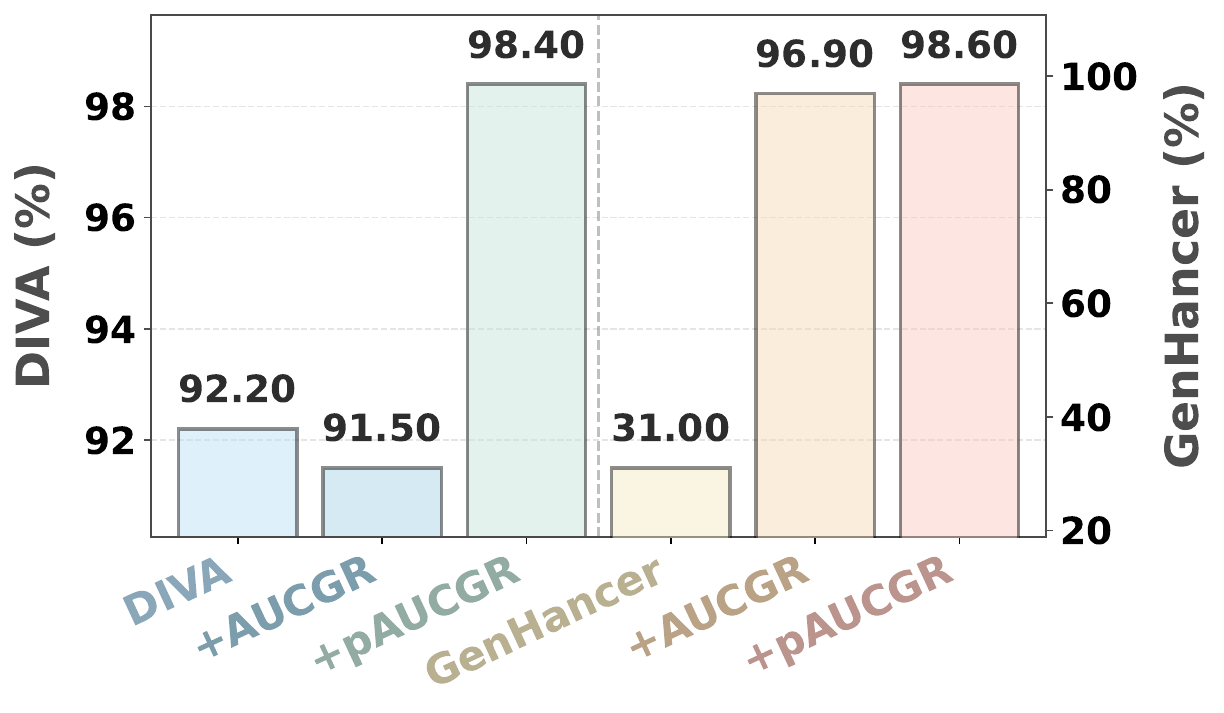}}
  \hfill
  \subfloat[Text@5 - COCO\label{fig: M1_Text_COCO}]{
    \includegraphics[width=0.23\linewidth]{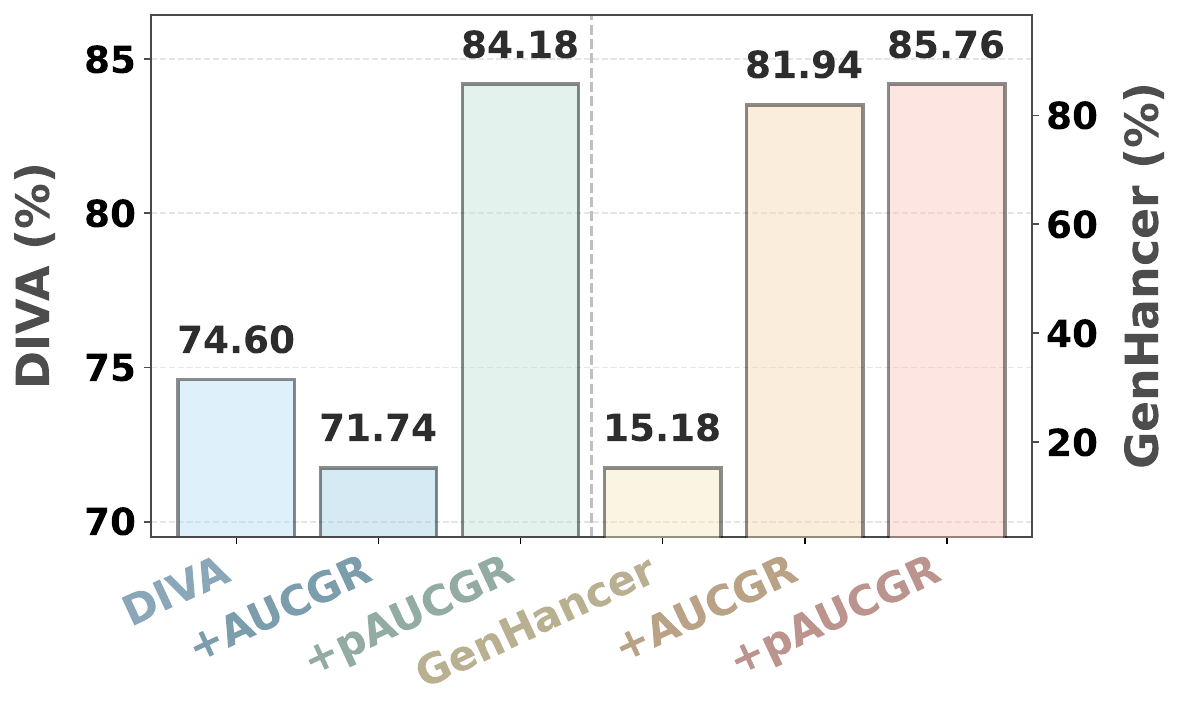}}
  \caption{\textbf{Performance on zero-shot retrieval on MetaCLIP ViT-L@224.} Compared with the original DIVA/GenHancer model, our method consistently improves both image-to-text and text-to-image retrieval on Flickr30k and COCO.}
  \label{fig: M1_retrieval}
\end{figure}

\begin{figure}[htbp]
  \centering
  \subfloat[Image@5 - Flickr30k\label{fig: S1_Image_Flickr30k}]{
    \includegraphics[width=0.23\linewidth]{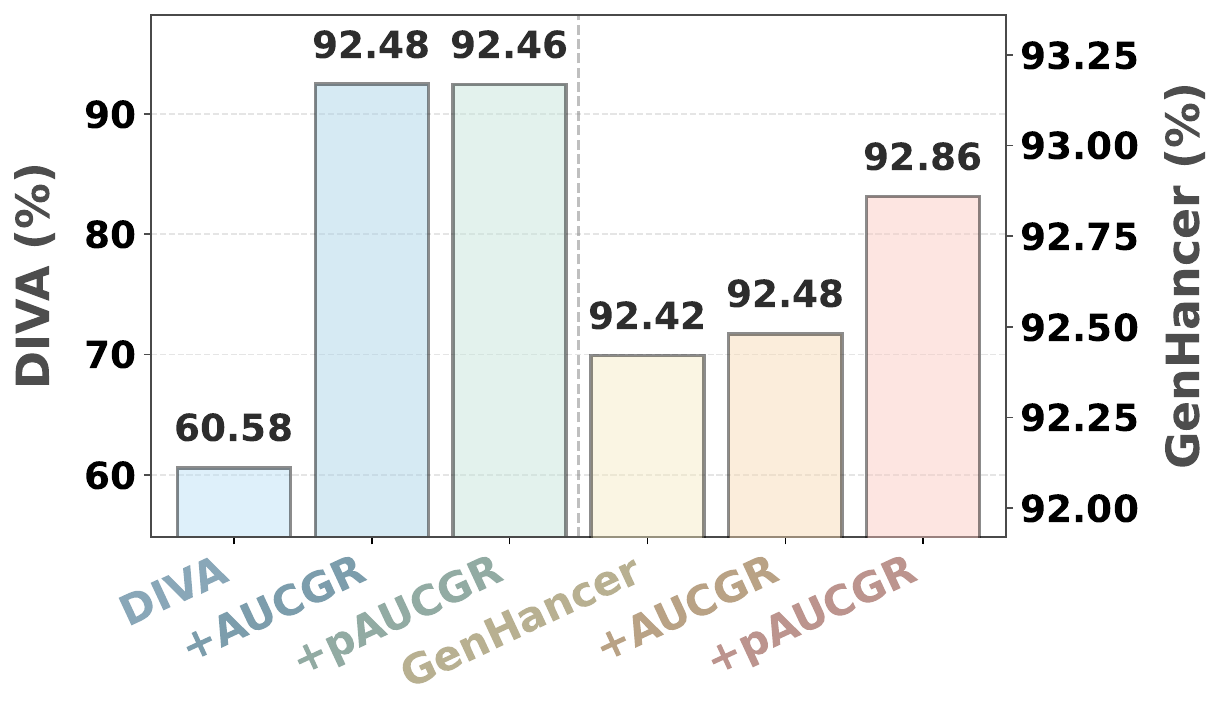}}
  \hfill
  \subfloat[Image@5 - COCO\label{fig: S1_Image_COCO}]{
    \includegraphics[width=0.23\linewidth]{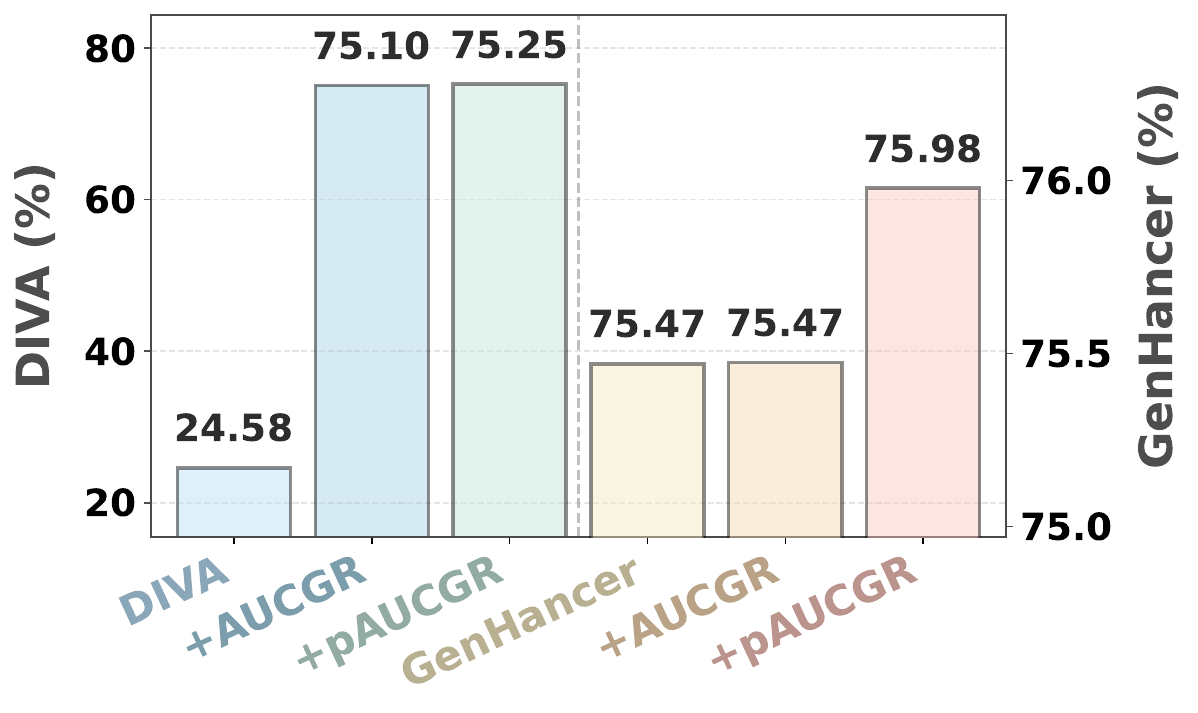}}
  \hfill
  \subfloat[Text@5 - Flickr30k\label{fig: S1_Text_Flickr30k}]{
    \includegraphics[width=0.23\linewidth]{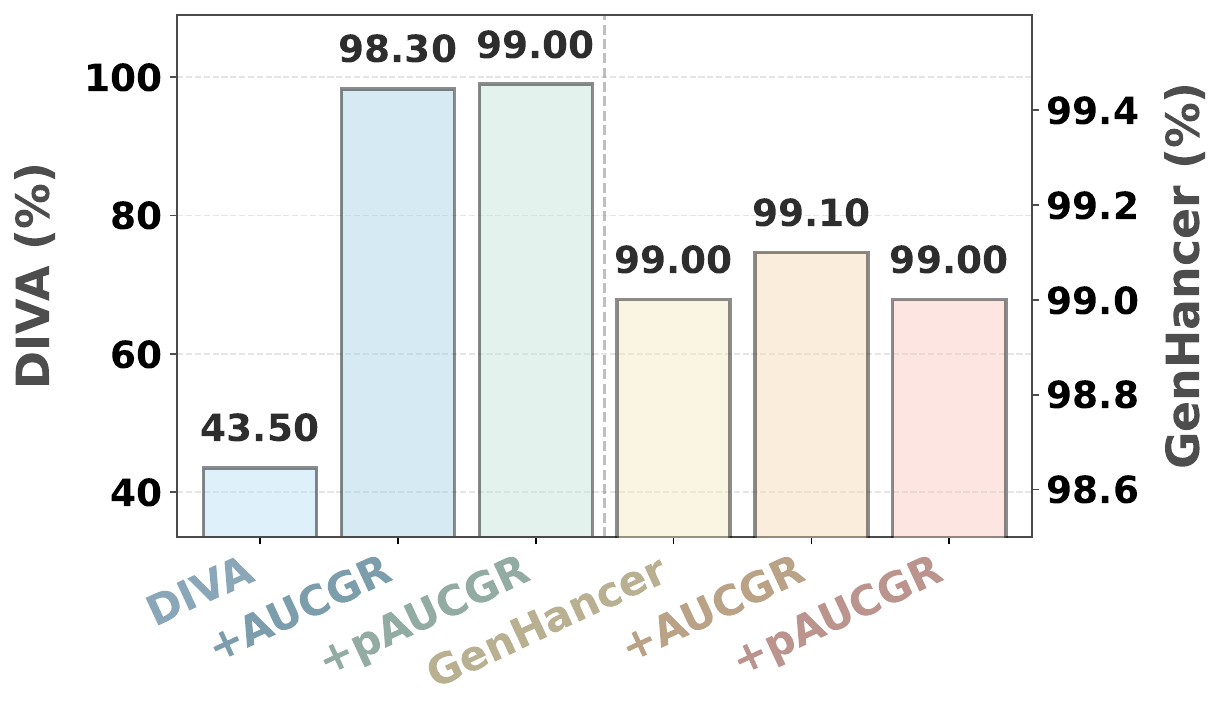}}
  \hfill
  \subfloat[Text@5 - COCO\label{fig: S1_Text_COCO}]{
    \includegraphics[width=0.23\linewidth]{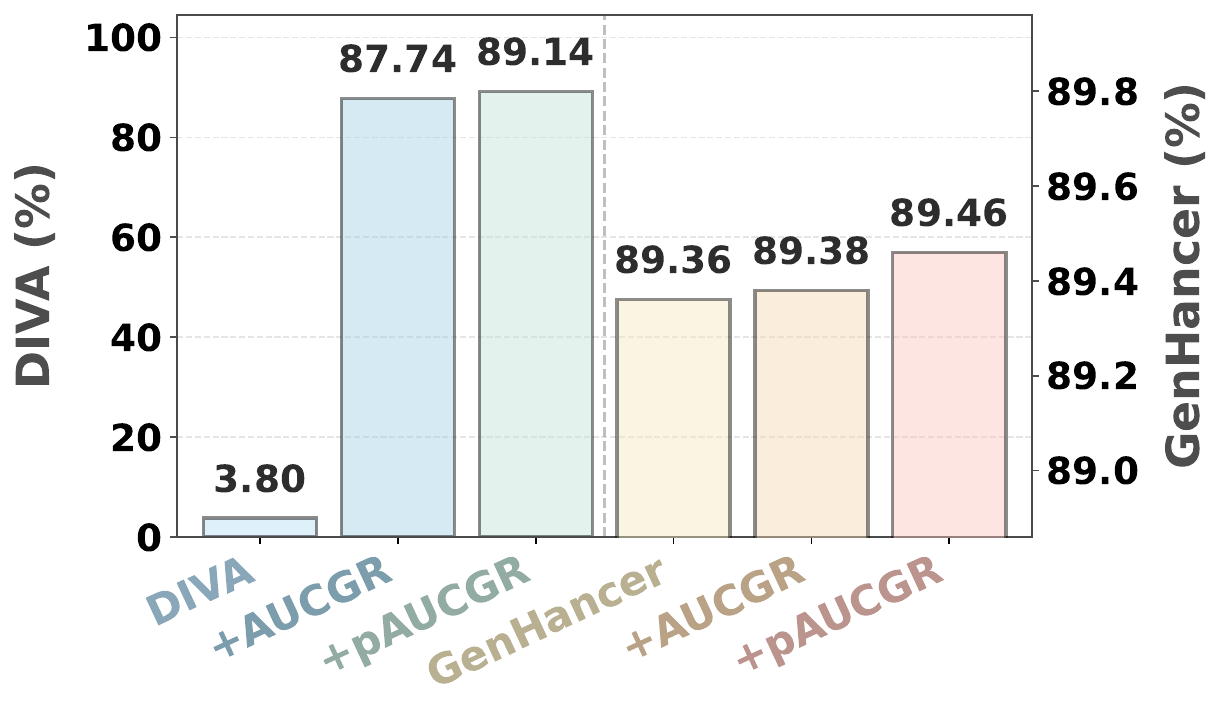}}
  \caption{\textbf{Performance on zero-shot retrieval on SigLIP ViT-SO@224.} Compared with the original DIVA/GenHancer model, our method consistently improves both image-to-text and text-to-image retrieval on Flickr30k and COCO.}
  \label{fig: S1_retrieval}
\end{figure}

The retrieval results evaluate whether the enhanced visual representations remain effective when the model ranks images and natural language descriptions. \cref{fig: O1_retrieval} reports the results on OpenAI CLIP ViT-L@224. \cref{fig: M1_retrieval} and \cref{fig: S1_retrieval} provide the corresponding results on MetaCLIP ViT-L@224 and SigLIP ViT-SO@224. Each figure includes image-to-text retrieval and text-to-image retrieval on Flickr30k and COCO. The improvements over the original DIVA and GenHancer baselines show that the proposed optimization benefits not only image-only discrimination, but also retrieval-based discrimination under cross-modal candidate ranking. This further supports the effectiveness of AUCGR and pAUCGR as plug-and-play optimization modules for strengthening \textbf{\textcolor{DisColor}{D-Ability}}.

\subsection{Expanded Qualitative Results on MLLM Performance}
\label{subsec: supp_Expanded Qualitative Results on MLLM Performance}
\begin{figure}[htbp]
\centering
\subfloat[MMVP-MLLM]{
\begin{minipage}{0.7\linewidth}
\includegraphics[width=\linewidth]{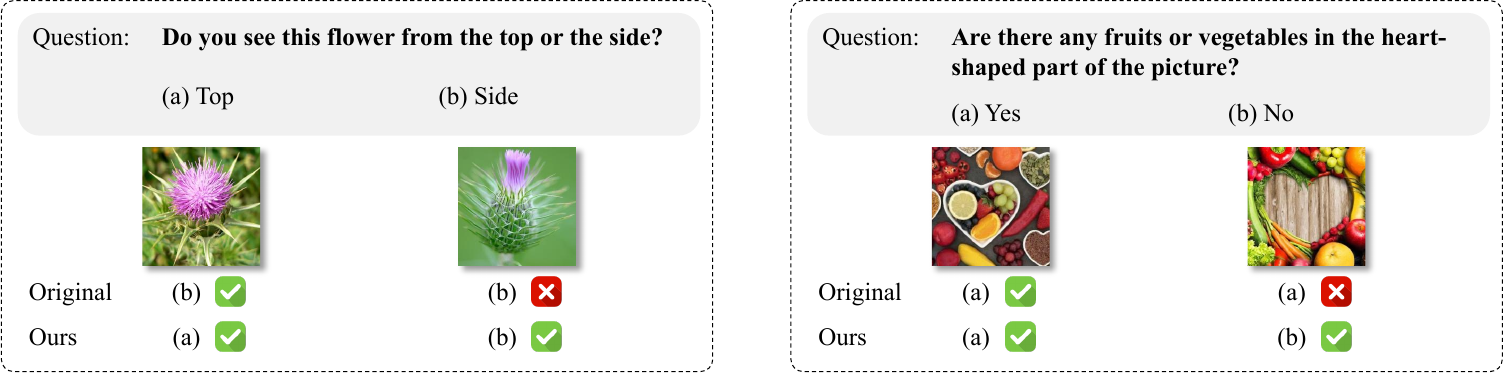}  
\label{fig: expanded_mmvp_mllm}
\end{minipage}
}
\hfil
\subfloat[NaturalBench]{   
\begin{minipage}{0.7\linewidth}
\includegraphics[width=\linewidth]{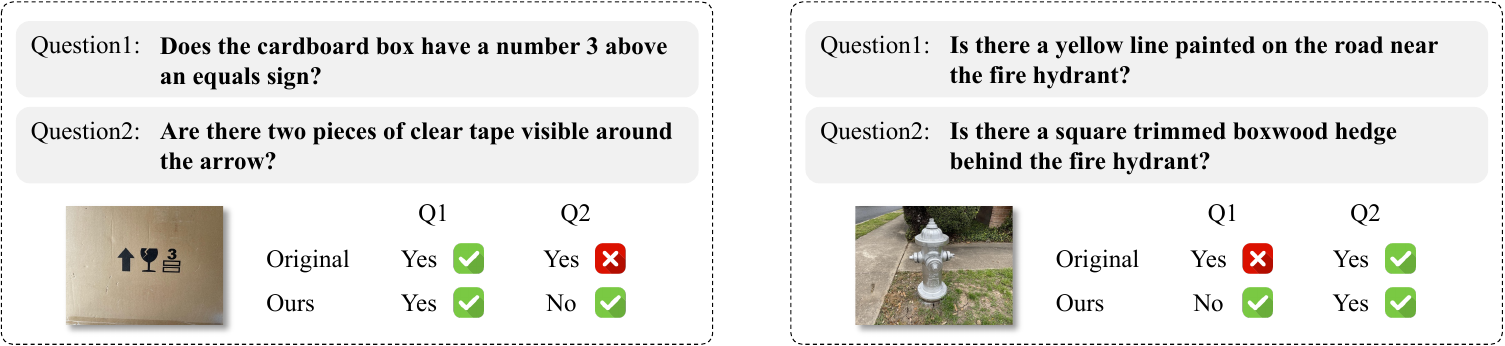}  
\label{fig: expanded_naturalbench}
\end{minipage}
}
\caption{\textbf{Expanded qualitative results on the MLLM benchmarks.} Ours denotes the model that replaces the original CLIP vision encoder with the pAUCGR-enhanced CLIP encoder.}
\label{fig: expanded_mllm}
\vspace{-10pt}
\end{figure}

\cref{fig: expanded_mllm} shows several qualitative examples on the vision-centric benchmarks from MLLMs. The results demonstrate that our enhanced CLIP can be seamlessly integrated into MLLMs to improve their visual capabilities.

\subsection{Expanded Ablation Studies}
\label{subsec: supp_Expanded Ablation Studies}

\subsubsection{Results on Noisy Negative Filtering}
\label{subsubsec: supp_Results on Noisy Negative Filtering}

We conduct two complementary studies to determine whether pAUCGR suppresses semantically noisy negatives rather than merely discarding difficult but valid comparisons.

\noindent\textbf{Semantic validity of the removed negatives.} Following the sampling protocol used in \cref{fig: empirical verification of noisy negatives}, we independently sample $40$ mini-batches with a batch size of $128$. Our training procedure is self-supervised and uses only images. For this diagnostic, we use the similarity between the corresponding texts as a proxy ground truth for semantic relationships. These text-based judgments are used only to evaluate the image-based filter and are not used during training. Among all negatives removed by the image-based filter, $98.75\%$ are also identified as noisy by the text-similarity criterion. Thus, only $1.25\%$ of the removed pairs are not flagged as semantically compatible, indicating that the filter accurately targets likely false negatives while rarely eliminating valid hard negatives. Moreover, filtering produces a $12.73\%$ relative reduction in the false-negative rate compared with the unfiltered negative set. The strong agreement between the image- and text-based criteria confirms that image similarity is an effective surrogate for identifying noisy negatives and substantially reduces false-negative contamination.

\noindent\textbf{Ablation Study on Filtering Strategies.} We further disentangle the effects of AUC optimization and negative filtering in \cref{tab: Noisy Negative Filtering}. For the naive joint objective, we compare retaining all negatives (Naive), random negative removal (Naive+Random), and image-similarity-based removal without the AUC objective (Naive+Partial). We conduct the analogous comparison for AUC-based optimization using AUCGR, AUCGR+Random, and pAUCGR. Random removal consistently fails to improve the two abilities and often degrades both, showing that the gain does not result from simply reducing the number of negative comparisons. Similarity-based filtering alone is also insufficient: Naive+Partial does not yield a balanced improvement over Naive. In contrast, pAUCGR achieves the best P-Ability and D-Ability for both DIVA and GenHancer. These results demonstrate that reliable negative selection and AUC-oriented comparison are complementary: the former suppresses noisy supervision, while the latter converts the retained comparisons into an effective ranking objective.

\begin{table}[htbp]
  
  \centering
  \renewcommand\arraystretch{1}
  \caption{\textbf{Comparison of noisy-negative filtering strategies.} All experiments use OpenAI CLIP ViT-L@224. P-Ability is measured by ACC on MMVP-VLM, and D-Ability is measured by the average ACC over $26$ zero-shot clustering benchmarks. Naive+Random and AUCGR+Random apply random negative removal to the corresponding objectives; Naive+Partial applies image-similarity-based filtering without AUC-oriented optimization; and pAUCGR combines the same filtering principle with AUCGR. The best result for each is highlighted in \textbf{bold}.}
  \resizebox{0.4\linewidth}{!}{
    \begin{tabular}{ccc}
    \toprule
    \textbf{Method} & \textbf{P-Ability} & \textbf{D-Ability} \\
    \midrule
    DIVA  & 21.48  & 44.75  \\
    \midrule
    Naive & 22.22  & 47.23  \\
    Naive+Random & 20.74  & 43.32  \\
    Naive+Partial & 17.78  & 42.75  \\
    \midrule
    AUCGR & 22.22  & 49.59  \\
    AUCGR+Random & 21.48  & 48.88  \\
    pAUCGR & \textbf{24.44} & \textbf{53.88} \\
    \midrule
    \midrule
    GenHancer & 25.19  & 31.13  \\
    \midrule
    Naive & 25.93  & 35.34  \\
    Naive+Random & 25.19  & 32.58  \\
    Naive+Partial & 21.48  & 35.90  \\
    \midrule
    AUCGR & 26.67  & 40.01  \\
    AUCGR+Random & 25.19  & 28.72  \\
    pAUCGR & \textbf{27.41} & \textbf{40.25} \\
    \bottomrule
    \end{tabular}%
  }
\label{tab: Noisy Negative Filtering}

\end{table}

\subsubsection{Results on Different Surrogate Losses}
\label{subsubsec: supp_Results on Different Surrogate Losses}

We evaluate the four surrogate losses introduced in \cref{subsec: Surrogate Loss Instantiation of AUCGR} using OpenAI CLIP ViT-L@224 with DIVA and GenHancer. As reported in \cref{tab: Surrogate Losses}, different surrogate choices generally improve the learned representations over the corresponding generative reconstruction baselines. In particular, the square and hinge surrogates consistently improve both P-Ability and D-Ability on the two backbones, while the logistic surrogate also yields clear gains in most cases. These results demonstrate that the effectiveness of the proposed framework is not tied to a particular surrogate formulation. The exponential surrogate is comparatively less effective: although it improves D-Ability, it reduces P-Ability on both backbones. A likely reason is that its rapidly increasing weighting assigns disproportionately large gradients to a small number of high-loss hard pairs, leading to overly aggressive updates and a less balanced preservation of the two abilities.

\begin{figure}[htbp]

\centering
\subfloat[Hinge/DIVA]{
\begin{minipage}{0.2\linewidth}
\includegraphics[width=\linewidth]{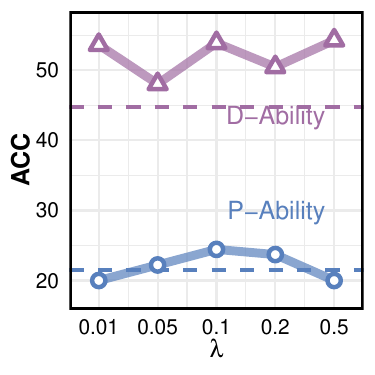}  
\end{minipage}
}
\hfil
\subfloat[Square/DIVA]{   
\begin{minipage}{0.2\linewidth}
\includegraphics[width=\linewidth]{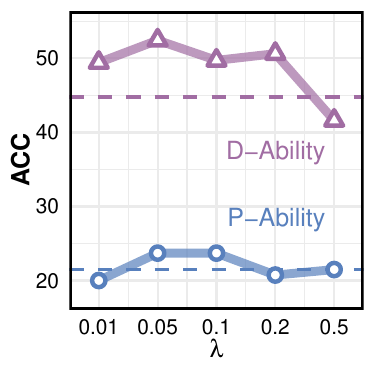} 
\end{minipage}
}
\hfil
\subfloat[Logistic/DIVA]{   
\begin{minipage}{0.2\linewidth}
\includegraphics[width=\linewidth]{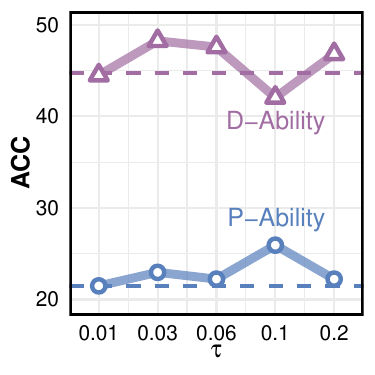}
\end{minipage}
}
\par
\subfloat[Hinge/GenHancer]{
\begin{minipage}{0.2\linewidth}
\includegraphics[width=\linewidth]{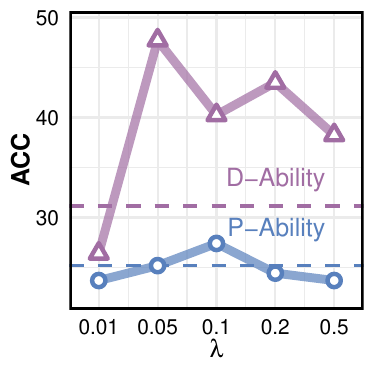}  
\end{minipage}
}
\hfil
\subfloat[Square/GenHancer]{   
\begin{minipage}{0.2\linewidth}
\includegraphics[width=\linewidth]{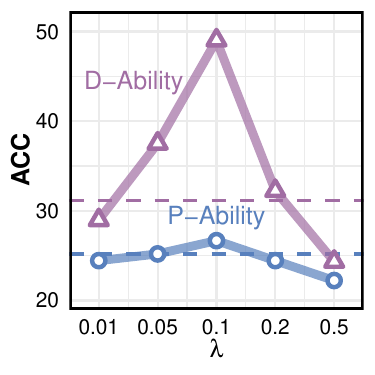} 
\end{minipage}
}
\hfil
\subfloat[Logistic/GenHancer]{   
\begin{minipage}{0.2\linewidth}
\includegraphics[width=\linewidth]{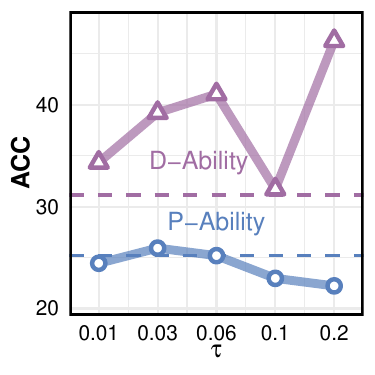}
\end{minipage}
}
\caption{\textbf{Sensitivity to surrogate loss hyperparameters.} We evaluate the trade-off between \textbf{\textcolor{PerColor}{P-Ability}} and \textbf{\textcolor{DisColor}{D-Ability}} of pAUCGR on DIVA and GenHancer using OpenAI CLIP ViT-L@224. Each point corresponds to a different margin $\lambda$ for the hinge or square surrogate, or a different temperature $\tau$ for the logistic surrogate. The dashed lines indicate the original performance of DIVA or GenHancer, respectively.}
\label{fig: surrogate loss}

\end{figure}

We further study the margin $\lambda$ of the hinge and square losses and the temperature $\tau$ of the logistic loss. As shown in \cref{fig: surrogate loss}, the proposed framework achieves consistent improvements over the corresponding reconstruction baselines across a broad range of reasonable hyperparameter values. Performance deteriorates mainly when the hyperparameters are set excessively small or large, because such extreme settings either provide insufficient separation between positive and negative pairs or place excessive emphasis on their differences. Within the practical range, P-Ability remains relatively stable and the improvements in D-Ability are consistently observed, despite its larger variation. These results show that the framework does not rely on a narrowly tuned optimum and is robust to the choice of surrogate-loss hyperparameters.

\begin{table}[htbp]
  
  \centering
  \renewcommand\arraystretch{1}
  \caption{\textbf{Performance with different surrogate losses.} We instantiate pAUCGR with the exponential, square, hinge, and logistic pairwise surrogates using OpenAI CLIP ViT-L@224. P-Ability is measured by ACC on MMVP-VLM, and D-Ability is measured by the average ACC over $26$ zero-shot clustering benchmarks.}
  \label{tab: Surrogate Losses}
  \captionsetup[subfloat]{position=top}
  \begin{minipage}[t]{0.48\linewidth}
    \centering
    \subfloat[DIVA.]{
      \label{tab: DIVA Surrogate Losses}
      \resizebox{0.8\linewidth}{!}{
        \begin{tabular}{>{\centering\arraybackslash}p{1.5cm}cc}
        \toprule
        \textbf{Method} & \textbf{P-Ability} & \textbf{D-Ability} \\
        \midrule
        DIVA  & 21.48  & 44.75  \\
        \midrule
        Exponential   & 18.52  & 53.35  \\
        Square & 23.70  & 52.44  \\
        Hinge & 24.44  & 53.88  \\
        Logistic & 25.93  & 42.06  \\
        \bottomrule
        \end{tabular}
      }
    }
  \end{minipage}
  \hfill
  \begin{minipage}[t]{0.48\linewidth}
    \centering
    \subfloat[GenHancer.]{
      \label{tab: GenHancer Surrogate Losses}
      \resizebox{0.8\linewidth}{!}{
        \begin{tabular}{>{\centering\arraybackslash}p{1.5cm}cc}
        \toprule
        \textbf{Method} & \textbf{P-Ability} & \textbf{D-Ability} \\
        \midrule
        GenHancer & 25.19  & 31.13  \\
        \midrule
        Exponential   & 22.96  & 39.14  \\
        Square & 26.67  & 49.03  \\
        Hinge & 27.41  & 40.25  \\
        Logistic & 25.93  & 39.22  \\
        \bottomrule
        \end{tabular}
      }
    }
  \end{minipage}
  
\end{table}

\subsubsection{Results on Training Stability}
\label{subsubsec: supp_Results on Training Stability}
\begin{figure}[htbp]

\centering
\subfloat[P Ability.]{
\begin{minipage}{0.35\linewidth}
\includegraphics[width=\linewidth]{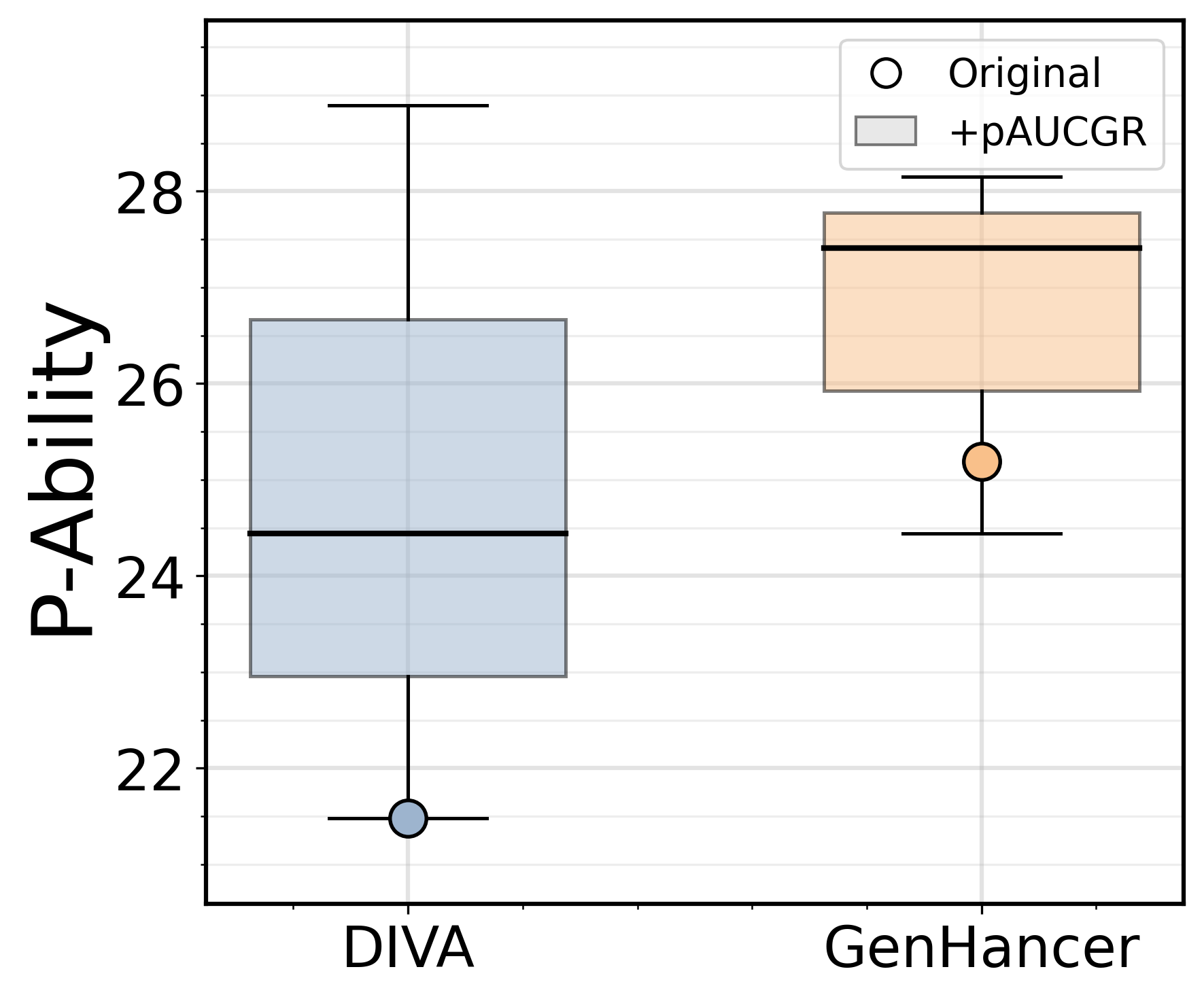}  
\end{minipage}
}
\hfil
\subfloat[D Ability.]{   
\begin{minipage}{0.35\linewidth}
\includegraphics[width=\linewidth]{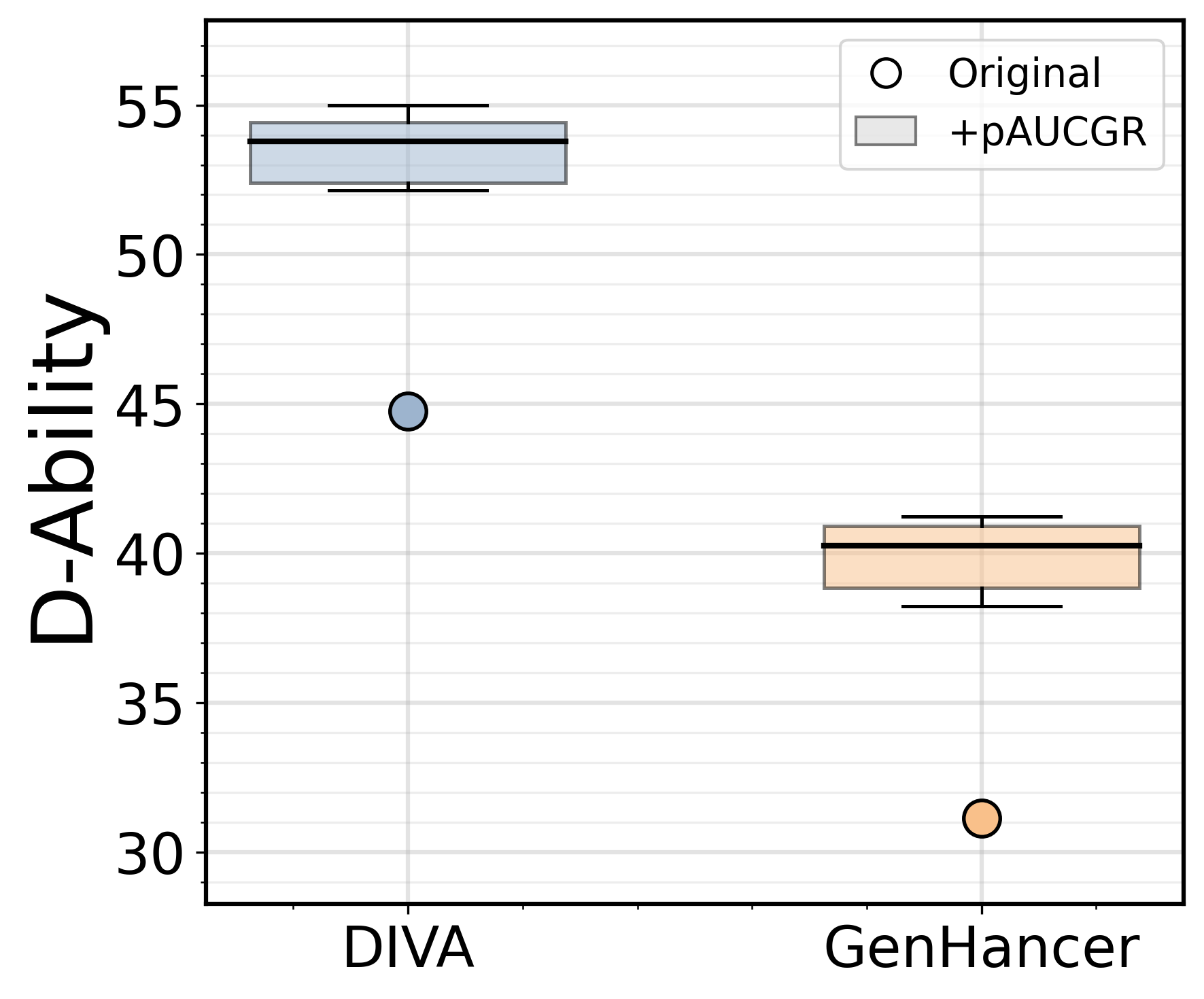} 
\end{minipage}
}
\caption{\textbf{Training stability across five random seeds.} We evaluate pAUCGR on DIVA and GenHancer using OpenAI CLIP ViT-L@224. P-Ability is measured by ACC on MMVP-VLM, and D-Ability is measured by the average ACC over $26$ zero-shot clustering benchmarks. The boxplots summarize the pAUCGR results across five runs. Boxes indicate the interquartile ranges, and the horizontal lines inside them indicate the medians. Circular markers denote the corresponding original DIVA or GenHancer results. The mean P-Ability and D-Ability of pAUCGR over the five seeds both exceed the corresponding baseline values.}
\label{fig: Training Stability}

\end{figure}

To assess the stability of pAUCGR under training randomness, we repeat the experiments with five random seeds on both DIVA and GenHancer using OpenAI CLIP ViT-L@224. All other experimental settings remain unchanged. \cref{fig: Training Stability} presents the results as boxplots to show the variation across runs. Averaged over the five seeds, pAUCGR improves both \textbf{\textcolor{PerColor}{P-Ability}} and \textbf{\textcolor{DisColor}{D-Ability}} over the corresponding original DIVA or GenHancer baseline. In particular, the D-Ability results remain above the corresponding baseline across all five runs on both reconstruction methods. P-Ability exhibits greater variation across seeds, but its average improvement is maintained on both methods. These results support the stability of pAUCGR and show that its gains in balancing the two abilities are reproducible across different random seeds.

\subsection{Computational Costs}
\label{subsec: supp_Computational Costs}
\cref{tab: Training costs} reports the computational costs of our method. Since AUCGR and pAUCGR are optimization methods, they do not introduce extra trainable modules. Therefore, the number of trainable parameters remains the same as that of the original generative reconstruction backbone. The additional training cost mainly comes from pairwise comparison and negative filtering. Relative to DIVA, AUCGR increases the training time by $25.70\%$--$54.79\%$ and GPU memory by $35.53\%$--$74.34\%$. pAUCGR increases the training time by $58.88\%$--$77.11\%$ and GPU memory by $39.50\%$--$85.38\%$. Nevertheless, the overall computation remains within a practical range for the NVIDIA A100 80GB GPU.

\begin{figure}[htbp]

\centering
\subfloat[Time (h)]{
\begin{minipage}{0.35\linewidth}
\includegraphics[width=\linewidth]{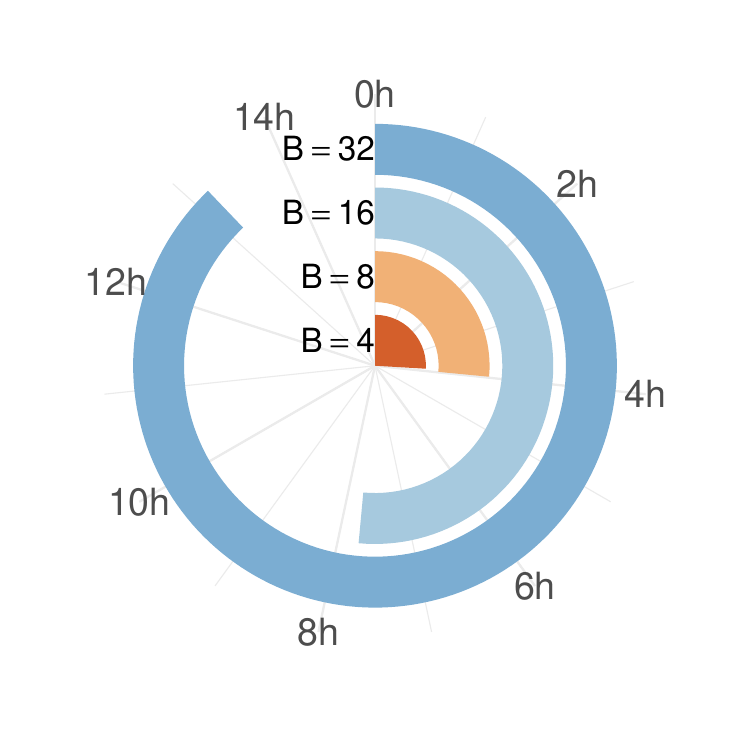}  
\end{minipage}
}
\hfil
\subfloat[GPU Memory (MB).]{   
\begin{minipage}{0.35\linewidth}
\includegraphics[width=\linewidth]{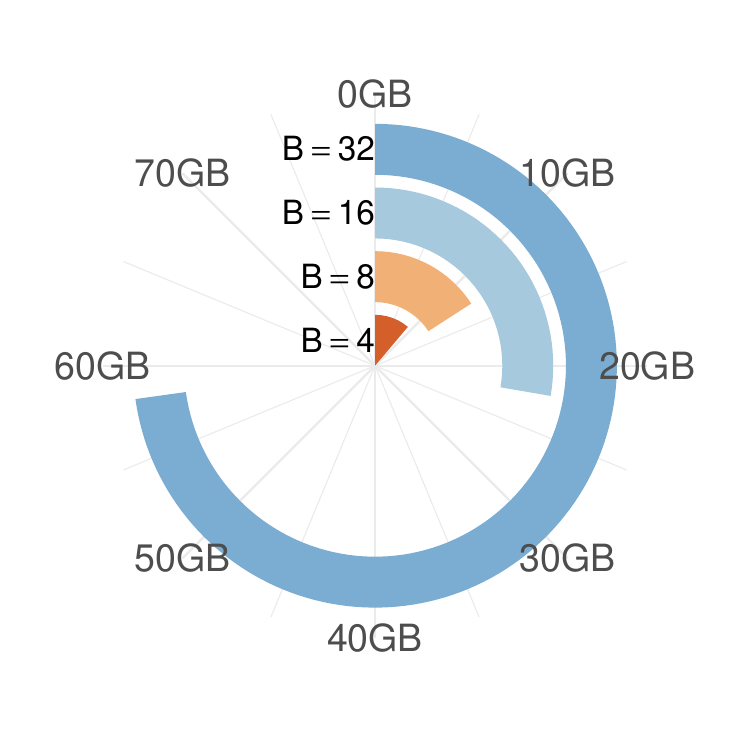} 
\end{minipage}
}
\caption{\textbf{Computational costs under different batch sizes.} Training time and GPU memory are measured for our method on DIVA with OpenAI CLIP ViT-L@224. The pairwise objective constructs a $B\times B$ score matrix, so its per-batch computation and intermediate storage scale as $O(B^2)$. We use $B=16$ in all experiments because it provides a favorable balance between performance and resource consumption.}
\label{fig: Computational Costs Batch Size}

\end{figure}

For a batch size $B$, the pairwise loss constructs a $B\times B$ score matrix. Its per-batch comparison and intermediate-storage complexities therefore scale as $O(B^2)$. As shown in \cref{fig: Computational Costs Batch Size}, training time and GPU memory increase accordingly with the batch size. A batch size of $16$ provides a favorable balance between performance and resource consumption. We therefore use $B=16$ in all experiments. Moreover, pairwise comparison and negative filtering are used only to compute the training objective. They do not change the encoder architecture or inference path and thus introduce no additional inference cost. Overall, our method incurs moderate and manageable training-only overhead while keeping the model size and inference cost unchanged.

\begin{table}[htbp]
  
  \centering
  \renewcommand\arraystretch{1}
  \caption{\textbf{Training costs of our method on DIVA.} Time is measured in hours (h), and GPU memory is measured in gigabytes (GB). The percentages in parentheses denote the relative increases over DIVA with the same backbone and cost metric.}
  \resizebox{0.95\linewidth}{!}{
    \begin{tabular}{ccccc}
    \toprule
    \textbf{Backbone} & \textbf{Training Costs} & \textbf{DIVA} & \textbf{+AUCGR} & \textbf{+pAUCGR} \\
    \midrule
    \multirow{2}[2]{*}{OpenAI CLIP ViT-L@224} & Time & 3.48  & 4.97 ($\uparrow 42.82\%$)  & 5.72 ($\uparrow 64.37\%$)  \\
          & GPU Memory & 13.72 & 21.17 ($\uparrow 54.28\%$) & 22.17 ($\uparrow 61.55\%$) \\
    \midrule
    \multirow{2}[2]{*}{OpenAI CLIP ViT-L@336} & Time & 4.80  & 7.43 ($\uparrow 54.79\%$)  & 8.23 ($\uparrow 71.46\%$)  \\
          & GPU Memory & 22.27 & 38.82 ($\uparrow 74.34\%$) & 41.28 ($\uparrow 85.38\%$) \\
    \midrule
    \multirow{2}[2]{*}{MetaCLIP ViT-L@224} & Time & 3.60  & 4.54 ($\uparrow 26.11\%$)  & 5.73 ($\uparrow 59.17\%$)  \\
          & GPU Memory & 13.75 & 20.94 ($\uparrow 52.27\%$) & 22.17 ($\uparrow 61.25\%$) \\
    \midrule
    \multirow{2}[2]{*}{MetaCLIP ViT-H@224} & Time & 4.28  & 5.38 ($\uparrow 25.70\%$)  & 6.80 ($\uparrow 58.88\%$)  \\
          & GPU Memory & 21.38 & 28.98 ($\uparrow 35.53\%$) & 29.83 ($\uparrow 39.50\%$) \\
    \midrule
    \multirow{2}[2]{*}{SigLIP ViT-SO@224} & Time & 3.80  & 5.46 ($\uparrow 43.68\%$)  & 6.73 ($\uparrow 77.11\%$)  \\
          & GPU Memory & 16.57 & 24.08 ($\uparrow 45.29\%$) & 25.08 ($\uparrow 51.36\%$) \\
    \midrule
    \multirow{2}[2]{*}{SigLIP ViT-SO@384} & Time & 7.00  & 10.02 ($\uparrow 43.14\%$)  & 12.20 ($\uparrow 74.29\%$)  \\
          & GPU Memory & 30.39 & 52.01 ($\uparrow 71.15\%$) & 53.66 ($\uparrow 76.58\%$) \\
    \bottomrule
    \end{tabular}%
  }
\label{tab: Training costs}

\end{table}

\section{Relationship Between the Truncation Ratio $\alpha$ and Threshold $\gamma(\alpha)$ in Expectation}
\label{sec: alpha_gamma_equivalence}

Our goal is to apply the same similarity criterion across mini-batches when removing potentially noisy negatives. Fixing $\alpha$ in each batch controls the removed fraction but allows the threshold to vary. Fixing $\gamma_0$ instead keeps the threshold unchanged and allows the removed fraction to vary. We establish their correspondence below for a fixed negative-score distribution.

Let $n\sim\mathcal{N}_{-}$ be drawn from the negative-sample distribution and define its score as $Q=q_i(n)$. The cumulative distribution function of $Q$ is
\begin{equation}
F_-(s)=\mathbb{P}(Q\le s).
\end{equation}
For a fixed truncation ratio $\alpha$, the corresponding quantile threshold is
\begin{equation}
\gamma(\alpha)=F_-^{-1}(1-\alpha).
\end{equation}
For a fixed threshold $\gamma_0$, the induced truncation ratio is
\begin{equation}
\alpha(\gamma_0)
=
\mathbb{P}(Q>\gamma_0)
=
1-F_-(\gamma_0).
\label{eq: alpha_induced_by_gamma}
\end{equation}

Assume that $F_-$ is continuous and strictly increasing over the relevant score range. Substituting \cref{eq: alpha_induced_by_gamma} into the quantile definition gives
\begin{equation}
\gamma\bigl(\alpha(\gamma_0)\bigr)
=
F_-^{-1}\bigl(F_-(\gamma_0)\bigr)
=
\gamma_0.
\end{equation}
Therefore, the two parameterizations retain the same negative set when $\alpha=\alpha(\gamma_0)$:
\begin{equation}
\{n:q_i(n)\le\gamma_0\}
=
\mathcal{N}_i^{(\alpha(\gamma_0))}.
\end{equation}
Moreover,
\begin{equation}
\alpha(\gamma_0)
=
\mathbb{P}(Q>\gamma_0)
=
\mathrm{FPR}(\gamma_0).
\end{equation}
Thus, the fixed threshold $\gamma_0$ corresponds to the partial AUC interval $[\alpha(\gamma_0),1]$. For a fixed score distribution and independent batch sampling, the average empirical truncation ratio converges to $\alpha(\gamma_0)$. This population-level correspondence does not require every batch to remove the same fraction of negatives.
\end{document}